%% file: DPCP-RANSAC.tex
\documentclass[10pt,twocolumn]{article} 
\usepackage{simpleConference}
\usepackage{times}
\usepackage{epsfig}
\usepackage{graphicx}
\usepackage{amsmath}
\usepackage{amssymb}
\usepackage{algorithmic}
\usepackage[ruled,vlined,linesnumbered]{algorithm2e}
\usepackage{lipsum}
\usepackage{multirow}
\usepackage{subfig}
\usepackage{booktabs}
\usepackage{stackengine}
\usepackage{url}
\usepackage{tikz}
\usepackage{moredefs}

\makeatletter
\@namedef{ver@everyshi.sty}{}
\makeatother
\usepackage{tikz}
\usepackage{moredefs}

\newcommand{\edit}[1]{#1}
\long\def\rene#1{}
\long\def\dpr#1{}

\newcommand{\myparagraph}[1]{\noindent\textbf{#1.}}

\newcommand{\Hcal}{\mathcal{H}}
\newcommand{\Ocal}{\mathcal{O}}
\newcommand{\bF}{\mathbf{F}}
\newcommand{\bG}{\mathbf{G}}
\newcommand{\bH}{\mathbf{H}}

\newcommand{\bY}{\mathbf{Y}}
\newcommand{\bb}{\mathbf{b}}

\newcommand{\f}{\mathbf{f}}

\newcommand{\h}{\mathbf{h}}

\newcommand{\bh}{\boldsymbol{h}}

\newcommand{\x}{\mathbf{x}}

\newcommand{\y}{\mathbf{y}}

\newcommand{\error}{\mathcal{E}}
\newcommand{\R}{\mathbb{R}}

\newcommand{\st}{\operatorname{s.t.}}

\newcommand{\ovec}{\operatorname{vec}}

\usepackage[pagebackref=true,breaklinks=true,colorlinks,bookmarks=false]{hyperref}

\usepackage{cleveref}

\begin{document}

\title{Boosting RANSAC via Dual Principal Component Pursuit}

\author{\normalsize Yunchen Yang$^1$ \,Xinyue Zhang$^1$ \, Tianjiao Ding$^1$ \, Daniel P. Robinson$^2$ \, Ren{\'e} Vidal$^3$ \, Manolis C. Tsakiris$^1$}

\maketitle
\thispagestyle{empty}

\begin{abstract}
In this paper, we revisit the problem of local optimization in RANSAC. Once a so-far-the-best model has been found, we refine it via Dual Principal Component Pursuit (DPCP), a robust subspace learning method with strong theoretical support and efficient algorithms. The proposed DPCP-RANSAC has far fewer parameters than existing methods and is scalable. Experiments on estimating two-view homographies, fundamental and essential matrices, and three-view homographic tensors using large-scale datasets show that our approach consistently has higher accuracy than state-of-the-art alternatives.
\end{abstract}

\section{Introduction} \label{sec:intro}

An important problem in $3$D vision is to fit subspaces to observations corrupted by noise and outliers, e.g. when fitting planes to point clouds, or estimating homographies, fundamental matrices or homographic tensors from feature correspondences detected and matched from images. One of the most widely used robust subspace estimation schemes in the past few decades is Random Sample Consensus (RANSAC) \cite{fischler1981random}, which iteratively computes a model from a random sample of data points, scores this candidate model by measuring its consensus with all data points, and keeps it if it has the highest consensus score. The iterations stop once the algorithm reaches a confidence $p$ that an outlier-free sample has been found. In the noiseless case, the expected number of iterations $T$ to reach such confidence $0 \leq p \leq 1$ can be approximated by
\begin{equation}
    T \approx \frac{\log(1-p)}{\log(1-(1-e)^{N_s})}, \label{eq:T}
\end{equation} where $e$ is the fraction of outliers and $N_s$ is the number of points in each sample needed to compute a model.
Here, the assumption is that the model computed from an outlier-free sample is close to the ground-truth model. 

In practice, when the data are noisy, it has been shown that RANSAC may need many more than $T$ iterations to obtain a subspace model that is close to the ground-truth one~\cite{Chum2003LocallyRANSAC}. The reason is that due to noise not all outlier-free samples give an equally accurate estimate, thus violating the aforementioned assumption. Inspired by this, \cite{Chum2003LocallyRANSAC} proposed to refine the model each time RANSAC finds a model that has a consensus score higher than all previous iterations, termed \textit{so-far-the-best} model. This procedure, known as \textit{local optimization} or LO-RANSAC, includes an extra inner RANSAC step that further samples from inliers to the so-far-the-best model, and recursively applies direct linear transform (DLT) with decreasing thresholds.
LO-RANSAC has been shown to be more accurate and require fewer iterations, albeit it is computationally more demanding \cite{Lebeda2012FixingRANSAC}. To lighten the computational burden, \cite{Lebeda2012FixingRANSAC} proposed to use a fraction of the inliers to the so-far-the-best model in the local optimization step. Nevertheless, the whole local optimization pipeline still requires many parameters, such as the number of inner RANSAC iterations and the range and step-size for the  threshold decrease, and is therefore ad-hoc and complex. Recently, \cite{Barath2018} proposed an alternative local optimization procedure that uses tools from graph theory. It defines an energy function based on spatial coherence and distance to the model, and applies the graph-cut algorithm to obtain a separation of inliers and outliers. This is an alternative to the usual way of labeling points by hard-thresholding the distance to the model. However, the refitting tends not to be robust since the graph-cut often gives false positives.

On the other hand, many robust subspace learning methods based on sparse and low-rank representation have appeared in the last decade. While such methods benefit from both strong theoretical guarantees and efficient algorithms, most of them are based on low-rank assumptions, hence they cannot be applied to robust estimation tasks in geometric vision, which require estimating a hyperplane. One exception is Dual Principal Component Pursuit (DPCP) \cite{Tsakiris:DPCPICCV15,Tsakiris2018DualPursuit,Zhu2018DualAlgorithms,Ding2019NoisyPursuit,Ding2020RobustPursuit}, which is suitable for learning subspaces of dimension close to the ambient dimension, e.g. hyperplanes. Remarkably, DPCP has been shown to provably tolerate $M = \Ocal(N^2)$ outliers, where $N$ and $M$ are the numbers of inliers and outliers respectively, under some uniformity assumptions on the data distribution. This is in sharp contrast to the typical guarantee $M = \Ocal(N)$ of alternative methods. Moreover, many algorithms have been proposed for DPCP, including recursive linear programming \cite{Tsakiris:DPCPICCV15,Tsakiris2018DualPursuit}, alternating minimization \cite{Qu2014,Tsakiris2018DualPursuit}, iteratively reweighed least squares (DPCP-IRLS) \cite{Tsakiris2017HyperplanePursuit,Tsakiris2018DualPursuit,Lerman2018}, projected subgradient (DPCP-PSGM) \cite{Zhu2018DualAlgorithms,Ding2019NoisyPursuit} or Riemannian subgradient \cite{Zhu2019,Li2019} methods, and geodesic gradient descent \cite{Maunu2019}. Notably, DPCP-IRLS and DPCP-PSGM have been shown to outperform the vanilla RANSAC in comparative studies that employ the same running time budget for the task of estimating $3$D road planes \cite{Zhu2018DualAlgorithms,Ding2019NoisyPursuit}, homography matrices and tensors \cite{Ding2020RobustPursuit}, as well as trifocal tensors \cite{Tsakiris2018DualPursuit}.

In this paper we use tools from robust subspace learning to address noise and outliers simultaneously in the local optimization step of RANSAC. In particular, we propose to refine the so-far-the-best model via DPCP. Our local optimization scheme has fewer parameters than existing methods and is efficient. Experiments on estimating two-view homographies, fundamental matrices, and three-view homographic tensors using state-of-the-art datasets show that our approach consistently achieves higher accuracy than alternatives such as Graph-Cut RANSAC \cite{Barath2018}, with comparable running time.

\section{A review of local optimization in RANSAC}
As mentioned above, even if a model is estimated from an all-inlier sample, it may significantly deviate from the true model due to the presence of noise. This is particularly the case when one uses minimal samples, e.g. $3$ points for estimating a plane or $7$ for a fundamental matrix. Thus, the key idea of local optimization is to find more inliers based on the current model, and use them to compute a better estimate. Since this procedure induces additional computational cost, it is typically done only when a so-far-the-best model has been found\footnote{Hence, local optimization is expected to execute at most $log(T)+1$ times \cite{Chum2003LocallyRANSAC}, where $T$ is given in \eqref{eq:T}.}. We summarize \edit{the locally optimized RANSAC} in \Cref{alg:LO-RANSAC}, then review two important lines of work in proposing algorithms for \cref{line:LO-RANSAC-LO} as follows.

\begin{algorithm}
    \SetAlgoNoLine
	\textbf{Input}: data points $P$; threshold $\epsilon$; max iteration $T_{max}$; confidence $p$\;
	\textbf{Output}: best model $M^*$\;
	\textbf{Initialization}: $k \gets 0; r^* \gets -\infty; T \gets T_{max}$\;
	\While{$k<T $ {\rm \textbf{and}} $k < T_{max}$}
	{$S_k \gets$ draw a minimal sample from $P$\; \label{line:minimal-sampler}
	$M_k \gets$ estimate a model from $S_k$\;  \label{line:minimal-estimate}
	$r_k \gets$ score $M_k$ on $P$ with $\epsilon$\; \label{line:score-function}
	\If{$r_k>r^*$}{
	$M^*,r^* \gets$ local optimization$(M_k, P,\epsilon)$\; \label{line:LO-RANSAC-LO}
	$T\gets$ expected number of iterations needed to reach confidence $p$\; \label{update-iterations}
	}
	$k\gets k+1$\;
	}
	\Return $M^*$
	\caption{Locally optimized RANSAC}
	\label{alg:LO-RANSAC}
\end{algorithm}

\subsection{The classical \edit{local optimization}}

One way to find more inliers is to look at points that are close to the so-far-the-best model $M$. Indeed, \cite{Chum2003LocallyRANSAC} proposes to take all points that are within $\epsilon$ distance to $M$, and run DLT on them. However, there may not be a single $\epsilon$ that is always optimal. The reason is that a good $\epsilon$ depends not only on the noise distribution, but possibly also on how much $M$ deviates from the true model. For example, when $M$ is mildly off from the true model, one would expect to use a larger $\epsilon$ to include more inliers. On the other hand, a too-large $\epsilon$ would result in outliers being included, which DLT is not robust to. To resolve those challenges, \cite{Chum2003LocallyRANSAC} further proposes a two-fold approach. It first enlarges the threshold to $K\epsilon$ with $K>1$, and runs DLT. Then, it shrinks $K$ and repeats the process until $K$ drops to $1$. To be robust to outliers, this process is further embedded into an \edit{extra} inner RANSAC step,
which takes non-minimal samples from inliers to $M$. Overall, this method gives more accurate estimates than RANSAC, but it is computationally demanding. For example, \cite{Lebeda2012FixingRANSAC} shows that it could reduce the error in pixels by $78\%$ compared to
plain RANSAC, while it could use an order of magnitude larger running time\footnote{Note that  \cite{Lebeda2012FixingRANSAC} proposed to use a fraction of the inliers with the so-far-the-best model in the local optimization step to reduce computation.}. \edit{Moreover, this outlier rejection process requires many parameters (\S \ref{sec:intro}), and the iterative threshold decreasing process is adhoc.}
 For a summary of the above approach, see \cite[Algorithm 2, 3]{Lebeda2012FixingRANSAC}.

\subsection{The graph-cut \edit{local optimization}}
\begin{algorithm}
    \SetAlgoNoLine
	\textbf{Input}: so-far-the-best model $M$; data points $P$; threshold $\epsilon$; max LO iterations $k_{max}$\;
	\textbf{Output}: best model $M^*$; best score $r^*$\;
	\textbf{Initialization}: $k \gets 0; r^* \gets -\infty$\;
	\While{$k < k_{max}$}
	{$I \gets$ classify inliers to $M$ from $P$ with \edit{$\epsilon$}\; \label{line:find-inliers}
	$M \gets$ estimate a model from $I$\; \label{line:estimate}
	$r \gets$ score $M$ on $P$ with $\epsilon$\;
	\textbf{break if} $r\leq r^*$\; \label{line:stop}
	$r^* \gets r; M^* \gets M; k\gets k+1$\;

	}
	\Return $M^*,r^*$
	\caption{Local Optimization Pipeline}
	\label{alg:LO}
\end{algorithm}
Recently, \cite{Barath2018} argues that the classical approach is ad-hoc and complex, and proposes a new local optimization pipeline (\Cref{alg:LO}). Based on that pipeline, it further proposes a graph-theoretical approach for \cref{line:find-inliers}. It designs an energy function that considers both the spatial coherence of data points $P$ and distances from $P$ to the model $M$, which is optimized using graph-cut to obtain a separation of inliers and outliers. Then, a DLT is performed on the inliers $I$ to $M$ in order to fit a new model $M$. Experimentally, \cite{Barath2018} shows that the composition of the pipeline and graph-cut, termed Graph-Cut-RANSAC, achieves higher accuracy and comparable running time to \cite{Chum2003LocallyRANSAC,Lebeda2012FixingRANSAC}.
\section{DPCP for geometric vision} \label{sec:dpdp-tasks}
We have seen that the challenge in refining a model is to be robust to both noise and outliers. In this section, we review \textbf{Dual Principal Component Pursuit} (DPCP) \cite{Tsakiris:DPCPICCV15,Tsakiris2018DualPursuit,Zhu2018DualAlgorithms,Ding2019NoisyPursuit,Ding2020RobustPursuit}, a robust subspace learning method that can be applied to various geometric vision problems.
Given a dataset $\bY=[\y_1, \dots, \y_N]$, where $\y_i\in \R^{D}$ either lies close to a hyperplane $S$ or is an arbitrary outlier in $\R^{D}$, DPCP estimates a normal vector $\bb$ to $S$ by minimizing the sum of the distances from the data points to the hyperplane:
\begin{equation}
    \min_{\bb} \sum_i^{N} |\bb^\top \y_i| \quad \st \quad \|\bb\|_2 = 1. \label{eq:dpcp}
\end{equation}
It is shown in \cite{Tsakiris:DPCPICCV15,Tsakiris2018DualPursuit,Zhu2018DualAlgorithms,Ding2019NoisyPursuit,Ding2020RobustPursuit} that, under certain conditions, global minima of \eqref{eq:dpcp} are guaranteed to be orthogonal to $S$ (noiseless data) or close to orthogonal to $S$ (noisy data).

\subsection{Robust fundamental matrix estimation} \label{sec:fund}

It is well known that a two-view point-point correspondence $(\x,\x')\in \R^3\times \R^3$ associated with fundamental matrix $\bF\in\R^{3\times3}$ must satisfy the epipolar constraint $\x'^\top \bF \x = 0$, or equivalently $(\x\otimes \x')^\top \f = 0$, where $\f:=\ovec(\bF)$ is the vectorized fundamental matrix. We refer to $(\x\otimes \x')$ as the epipolar embedding of $(\x,\x')$. Given a set of correspondences $\{(\x_i,\x'_i)\}_{i=1}^N$ that contains sufficiently many (noisy) inliers
and possibly a large fraction of mismatches, we propose to robustly estimate an $\bF$ via 
\begin{equation}
    \mathop{\arg\min}_{\f} \sum_i^{N} |\f^\top (\x_i\otimes \x'_i)| \,\,\,\, \st \,\,\,\, \|\f\|_2 = 1. \label{eq:fund-l1}
\end{equation}

On the other hand, when the $3$D points are lying on a single plane or the translation between two views is zero, it has been shown in \cite{Ding2020RobustPursuit} that there is a three-dimensional vector space of fundamental matrices that explain the correspondences, and the true fundamental matrix is one of them. In this case, one could recover an essential matrix as follows.  First, use DPCP to robustly estimate a $3$-dimensional nullspace by computing
\begin{equation}
    \small\hat{\mathbf{B}} := \mathop{\arg\min}_{\mathbf{B}} \sum_i^{N} \|\mathbf{B}^\top (\x_i\otimes \x'_i)\|_2 \,\,\,\, \st \,\,\,\, \mathbf{B}^\top \mathbf{B} = \mathbf{I}_3 \label{eq:fund-c3-l1},
\end{equation} 
and then calculate the subspace distance between the embeddings and the nullspace, $w_i := \|\hat{\mathbf{B}}^\top (\x_i\otimes \x'_i)\|_2$. Second, weight each epipolar embedding by $w_i$, and extract a $5$-dimensional principal subspace from the weighted data. This in turn gives $5$ linear constraints, which one can use to solve for an essential matrix by further considering the polynomial constraints from the essential variety.

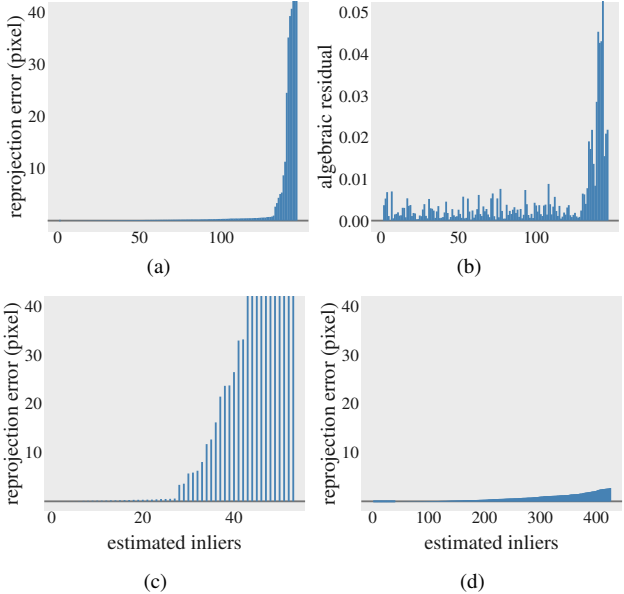
\begin{figure}[hbt!]
\centering
\vspace{-0.47cm}
\hspace{-0.2cm}\subfloat[]{\scalebox{1.17}{
\input{Rfigures/stema_b}}}
\hspace{-0.15cm}\subfloat[]{\scalebox{1.17}{
\input{Rfigures/stemb_b}}}\\
\vspace{-0.4cm}
\hspace{-0.2cm}\subfloat[]{\scalebox{1.17}{
\input{Rfigures/stemc_b}}}
\hspace{-0.15cm}
\subfloat[]{\scalebox{1.17}{
\input{Rfigures/stemd_b}}}
\caption{(a) shows the reprojection errors of the inliers estimated by the 5pt algorithm \cite{stewenius2006recent} with respect to the ground-truth camera pose. (b) shows the subspace distances from the epipolar embeddings to the model estimated by DPCP \eqref{eq:fund-c3-l1} corresponding to the ordering in (a). (c) shows the reprojection errors of inliers estimated from another minimal sample with respect to the ground-truth and (d) shows the reprojection errors of the inlier set refined by DPCP based on the model in (c). The local optimization is done by weighting the embeddings by DPCP \eqref{eq:fund-c3-l1}, extracting the $5$th principle subspace of the weighted data, and solving for the polynomial constraints of the essential matrix.}
\label{fig:motivation}
\end{figure}

This weighting procedure is inspired by the observation demonstrated in \Cref{fig:motivation}(a), where the inlier set identified by a minimal estimate via the 5pt algorithm \cite{stewenius2006recent} (\cref{line:minimal-estimate}, \Cref{alg:LO-RANSAC}) with a threshold on Sampson distance is usually corrupted by outliers. We see that the algebraic residual given by the model estimated by DPCP precisely characterizes the destribution of the reprojection errors in Figure \ref{fig:motivation}(b). Moreover, \Cref{fig:motivation}(c) and (b) compare the model before (c) and after (d) the local optimization via DPCP.

\subsection{Robust homography estimation}
A recent work \cite{Ding2020RobustPursuit} proposed to robustly estimate homographies via DPCP, and shows that it performs on par with USAC \cite{Raguram2008AConsensus}, while running an order of magnitude faster. We briefly review the formulation for completeness.

Let $(\x,\x')$ be a two-view point-point correspondence related by homography $\bH$, i.e., $\x'\sim \bH \x$.  With $\h:=\ovec(\bH)$, this relation is equivalent to $\psi_j(\x,\x')^\top \h=0$, $j=1,2$, where $\psi_j$ are the so-called homographic embeddings \cite{Ding2020RobustPursuit}\cite[\S 4.1]{Hartley2004TheViews}. Similar to the case of fundamental matrix estimation, if $\{(\x_i,\x'_i)\}_{i=1}^N$ contains sufficiently many correspondences that are approximately related by $\bH$ as well as some outliers, one can robustly estimate $\bH$ via DPCP:
\begin{equation}
\!\!\!\min_{\h} \sum_i^{N} \|\h^\top [\psi_1(\x_i,\x'_i), \psi_2(\x_i,\x'_i)]\|_2 \,\, \st \,\, \|\h\|_2 \!=\! 1.\! \label{eq:homo-l1}
\end{equation}

Consider now the case of having \textit{three} views, where a point-point-point correspondence $(\x,\x',\x'')$ is related by two homographies $\bH,\bG$, i.e., $\x \sim \bH\x' \sim \bG\x''$. This is captured by what is known as a homographic tensor $\Hcal\in \R^{3\times 3\times 3}$  and the relation $\phi_j(\x,\x',\x'')^\top \bh=0$, $j=1,\dots,7$, where $\bh:=\ovec(\Hcal)$ \cite{Ding2020RobustPursuit,Shashua1995TrilinearityTensor,Shashua2000HomographyPoints}. The $\phi_j$'s are referred to as three-view homographic embeddings. Again, given $\{(\x_i,\x'_i,\x''_i)\}_{i=1}^N$ corrupted by noise and outliers, the homographic tensor can be robustly estimated via DPCP:
\begin{align}
    & \min_{\bh} \sum_i^{N} \|\bh^\top [\phi_1(\x_i,\x'_i,\x''_i),\dots, \phi_7(\x_i,\x'_i,\x''_i)]\|_2 \nonumber\\
    & \st \,\,\,\, \|\bh\|_2 = 1. \label{eq:homotensor-l1}
\end{align}

\section{Local optimization with DPCP} \label{sec:lo}
In this paper, we propose to apply DPCP in local optimization (\Cref{alg:LO}). \edit{More precisely, in \cref{line:find-inliers}, we select inliers $I$ to the current model $M$ by taking the points from $P$ that are within $\epsilon$ distance to $M$. Then in \cref{line:estimate}, we robustly estimate a model via DPCP from the set $I$, as detailed in \S \ref{sec:dpdp-tasks}.\rene{Why can't DPCP be used on both \cref{line:find-inliers} and \cref{line:estimate}? That is, why can't DPCP be applied to the entire dataset to determine new model and also classify inliers/outliers?}}  We refer to this method as DPCP-RANSAC.

One notable advantage of our method is having fewer parameters. Since we use the local optimization pipeline \Cref{alg:LO}, there is no need for parameters such as a range and step-size for the  threshold decrease as in \cite{Chum2003LocallyRANSAC,Lebeda2012FixingRANSAC}. Furthermore, while both our method and Graph-Cut-RANSAC~\cite{Barath2018} are based on \Cref{alg:LO}, the latter requires a radius parameter for building a neighborhood graph, and a parameter in the energy for balancing the unary and pairwise terms, which in principle require careful tuning. In contrast, our method does not have parameters in the objective. Therefore, DPCP-RANSAC can be more conveniently applied to various tasks and datasets.

\section{Experiments} \label{sec:exp}
In this section, an experimental study is conducted using realistic datasets on the tasks of estimating two-view homography matrices (\S \ref{sec:exp-homo}), fundamental (\S \ref{sec:exp-fund}) and essential matrices (\S \ref{sec:exp-essen}), and three-view homographic tensors (\S \ref{sec:exp-HT}).

\myparagraph{Baselines}
We compare the proposed DPCP-RANSAC (R-DPCP) with Graph-Cut-RANSAC (R-GC) \cite{Barath2018} and Huber-RANSAC (R-Huber) \cite{Serych2016FastEstimation}. For R-GC, we use an off-the-shelf C++ implementation provided by its authors\footnote{https://github.com/danini/graph-cut-ransac, commit: 0321885.}. For a fair comparison we implement R-Huber and R-DPCP in the same C++ framework as R-GC. We also implement a DLT based LO-RANSAC (R-DLT) as a baseline for our comparative study. Although this R-DLT pipeline seems simple, we note that it has not been reported by \cite{Chum2003LocallyRANSAC,Lebeda2012FixingRANSAC,Serych2016FastEstimation,Barath2018}; we will discuss the differences in \S\ref{sec:exp-fund}. On the other hand, for the task of homographic tensor estimation, similar C++ implementations as above are not available. Hence, we restrict the comparison for that task in a Matlab environment where we implement R-DLT, R-Huber and R-DPCP. 
\Cref{tab:algo_summary} summarizes the outlier rejection and non-minimal model refitting steps for the evaluated variants. The confidence $p$ of these LO-RANSAC variants for all tasks is set to $0.95$ with the SPRT test \cite{Matas2005RandomizedTestb,Chum2008OptimalRANSAC} de-activated.

\myparagraph{Experimental setup}
The data points are uniformly sampled during the minimal sampling step (\Cref{alg:LO-RANSAC}, \cref{line:minimal-sampler}). \edit{For fairness, we use the same sampling sequence for all RANSAC variants, and vary the sequences for repeated experiments.} We employ the truncated quadratic cost \cite{lebeda2012fixing,Barath2018} as the scoring criterion (\Cref{alg:LO-RANSAC}, \cref{line:score-function}).
For the embedded subspace learning methods, we use the IRLS algorithm to solve for Huber and DPCP, where we set maximum iteration $\tau_{max} = 100$ and convergence accuracy $\delta = 10^{-5}$. Huber has a parameter that controls the combination of quadratic and linear parts, for which we use $0.01$ for homography matrix estimation as suggested by \cite{Serych2016FastEstimation}. For the remaining tasks, we use $\{0.1,0.01,0.001\}$ and report the mean performance\footnote{We note that R-Huber is similar in spirit with R-DPCP, except that R-Huber uses a Huber loss in-lieu of the $\ell_1$ loss of DPCP. The main disadvantage of Huber is the presence of a parameter to be tuned for different tasks and data. Therefore, for a fair comparison, we report the average over three different values of the Huber parameter. Even though one could tune the parameter for a specific dataset, we empirically observe that the performance of R-Huber always lies between R-DLT and R-DPCP.}.\dpr{is reporting the mean rather than the best performing unfair?}
For Graph-Cut, we use the provided default parameters for each task. In particular, the coherence weight is set to $0.975$ and the neighborhood graph is constructed by dividing an image into $8\times 8$ grid cells. We note that the description in \cite{Barath2018} differs from that in the latest implementation by the authors of \cite{Barath2018}, and we follow the latest implementation\footnote{Earlier versions would not be suitable for experiments, as later ones contain many improvements and bug fixes.}. 

\myparagraph{Metrics}
We use geometric errors as metrics for accuracy, and running time for efficiency.
For fundamental and homography matrix estimation, a validation set of inlier correspondences is manually labelled, on which we report the mean Sampson or reprojection error of the estimated matrix. For homographic tensors, we make a comparison in the context of pose estimation. Namely, we take the estimated homographic tensor, decompose it into two pairwise homography matrices, use the provided camera calibration to further reconstruct rotations and translations, and show the angular rotation and angular translation errors with respect to the ground-truth. 
\edit{Since the threshold $\epsilon$ is often not known a priori and requires tuning in practice, we report geometric errors and running time over different threshold choices to demonstrate the sensitivity of methods.} Meanwhile, in each task and dataset, we also tune an optimal threshold for each method, and show the median and interquartile range (IQR) of errors over \textit{all} frames and trials in box plots. 
The threshold is chosen to correspond to the smallest running time among all thresholds whose error is within $1\%$ of the minimum error.

\begin{table}
\caption{Evaluated LO-RANSAC variants. Outlier rejection and model refitting refer to \cref{line:find-inliers} and \ref{line:estimate} in \Cref{alg:LO}.}
\centering
\begin{tabular}{lcc} \toprule
Algorithm      & outlier rejection & model refitting \\ \cmidrule(r){1-1} \cmidrule(rl){2-2} \cmidrule(l){3-3}
R-DLT  & hard thresholding        & DLT-SVD      \\
R-GC & graph-cut          & DLT-SVD                  \\
R-Huber  & hard thresholding       &     Huber-IRLS       \\
R-DPCP  & hard thresholding        & DPCP-IRLS  \\ \bottomrule
\end{tabular}
\label{tab:algo_summary}
\end{table}

\subsection{Fundamental matrix estimation} \label{sec:exp-fund}

\myparagraph{Data} We take $16$ image pairs from the Kusvod2\footnote{\url{http://cmp.felk.cvut.cz/data/geometry2view/index.xhtml}} dataset and $45$ from the EPFL dataset \cite{strecha2008benchmarking}. Kusvod2 contains images from various indoor and outdoor scenes, such as bookshelves, corridors, and buildings, captured by distinct devices. For each pair of images, we use the provided feature correspondences, and the annotated ground-truth inliers as the validation set (see the beginning of \S\ref{sec:exp}). Beside this standard benchmark dataset, we also include EPFL, which contains outdoor scenes such as city halls and statues. For each image pair, we detect and match features using SURF \cite{bay2006surf} with a threshold of $30$ and a ratio test of $0.7$. Notably, due to the presence of many repetitive patterns and high-resolution images, there is a large number of correspondences and mismatches. EPFL provides ground truth camera matrices, which we use to annotate a set of inliers as the validation set. The correspondences are normalized in each view \cite{466816}, and the epipolar embeddings (\S \ref{sec:fund}) are scaled to have unit $\ell_2$ norm.

\begin{figure}[hbt!]
\centering
\subfloat[Kyoto]{
\includegraphics[height=3.7cm]{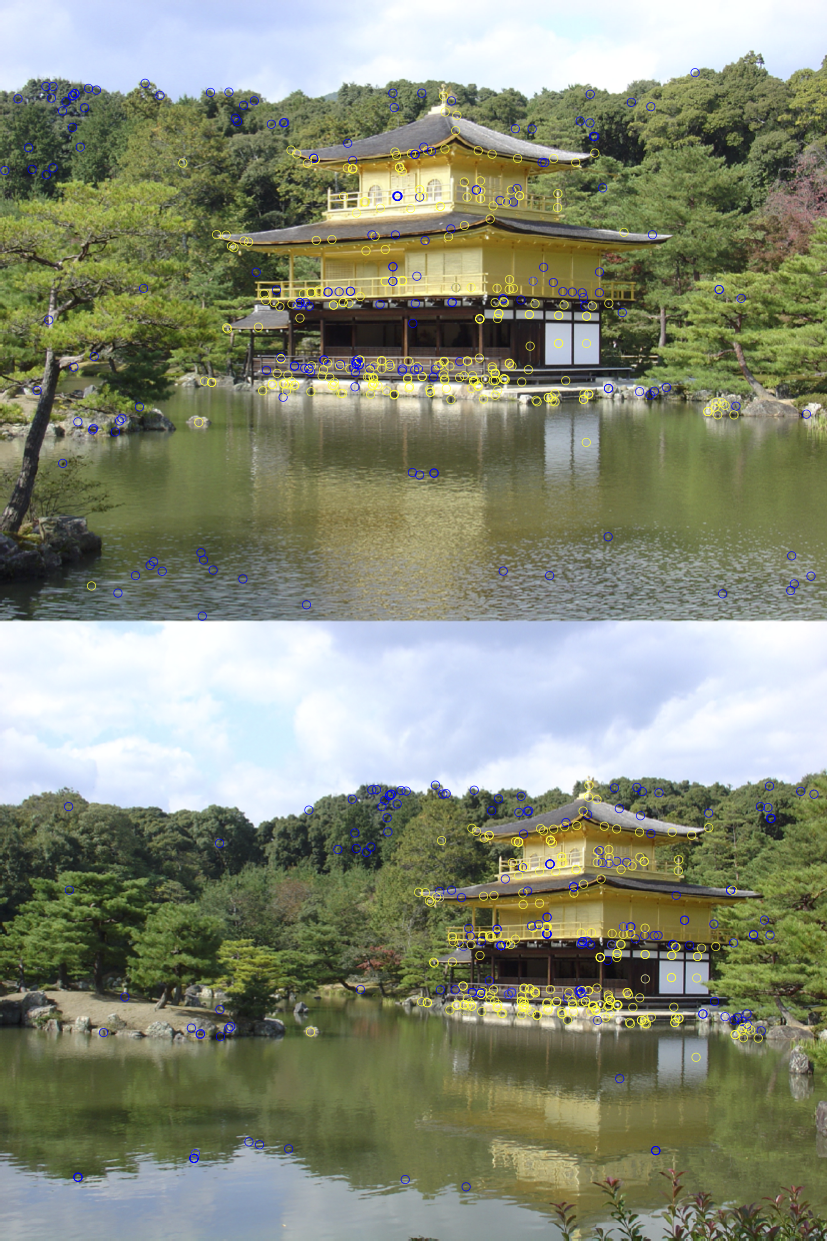}}
\subfloat[City-Hall-Leuven]{
\includegraphics[height=3.7cm]{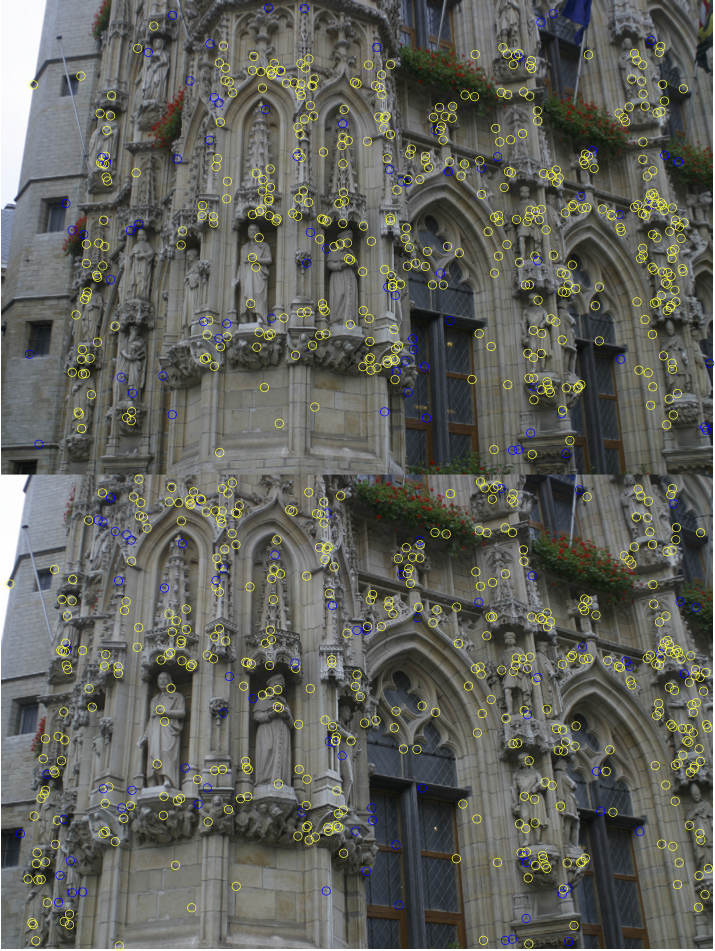}}
\subfloat[Castle]{
\includegraphics[height=3.7cm]{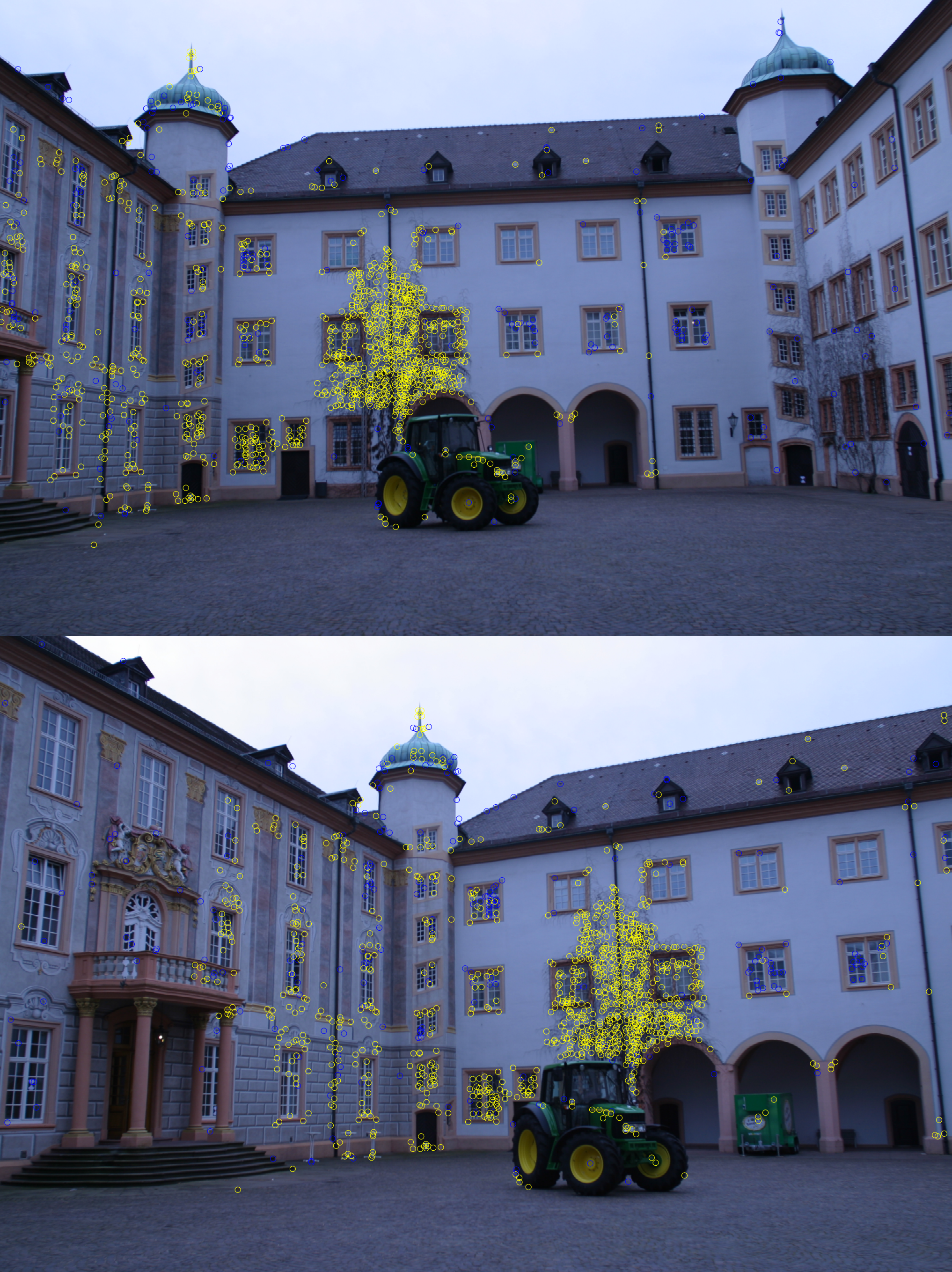}}
\caption{View pairs from Kusvod2 (a) and EPFL (b,c). Correspondences with Sampson distance with respect to the ground-truth less than $1$ pixel are marked in yellow (inliers), otherwise in blue (outliers). }
\end{figure}

\myparagraph{Methods}
The 7-point algorithm \cite[11.1.2]{hartley2003multiple} is used as the minimal solver for RANSAC (\Cref{alg:LO-RANSAC}, \cref{line:minimal-estimate}).
Since the images are of different sizes, the threshold $\epsilon$ is set to be a scalar $\sigma$ multiplied by the diagonal of the image \cite{Lebeda2012FixingRANSAC}, and we run the methods with varying $\sigma$.

\begin{figure}[hbt!]
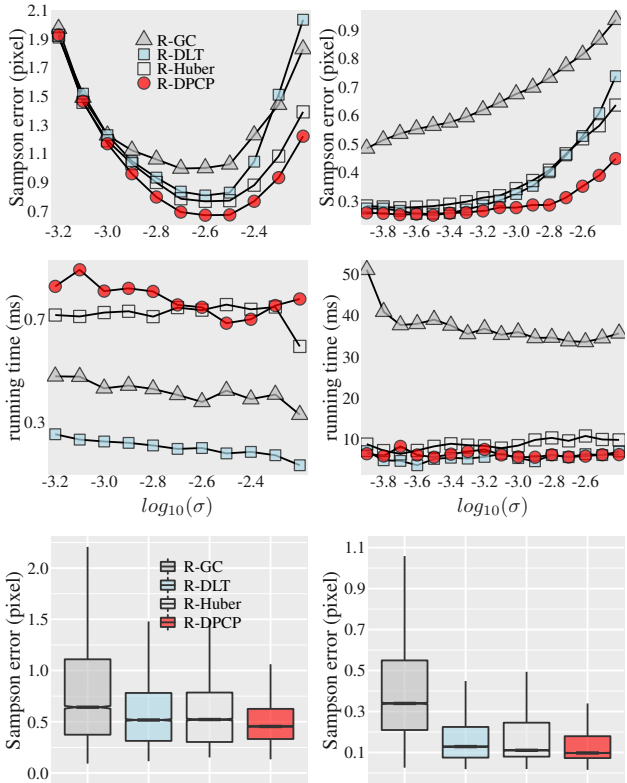

\centering
\vspace{-0.47cm}
\hspace{-0.2cm}\subfloat{\scalebox{1.17}{
\input{Rfigures/KUSVOD2Trans_mean}}}
\hspace{-0.15cm}\subfloat{\scalebox{1.17}{
\input{Rfigures/EPFLTrans_mean}}}
\\
\vspace{-0.8cm}
\hspace{-0.2cm}\subfloat{\scalebox{1.17}{
\input{Rfigures/KUSVOD2Time_mean}}}
\hspace{-0.15cm}
\subfloat{\scalebox{1.17}{
\input{Rfigures/EPFLTime_mean}}}
\\
\vspace{-0.45cm}
\hspace{-0.2cm}\subfloat{\scalebox{1.17}{
\input{Rfigures/KUSVOD2box}}}
\hspace{-0.15cm}
\subfloat{\scalebox{1.17}{
\input{Rfigures/EPFLbox}}}
\setlength{\belowcaptionskip}{-10pt}
\caption{Fundamental matrix estimation using $16$ image pairs from Kusvod2 (left) and $45$ from EPFL (right). All methods terminate when a confidence of $p = 95\%$ has been reached. The top two rows show Sampson error in pixels and running time in milliseconds with varying threshold multiplier $\sigma$, while the bottom row shows Sampson error with threshold optimally tuned for each method. Experiments are averaged over $500$ trials.}
\label{fig:Fundamental}
\end{figure}

\myparagraph{Results} \Cref{fig:Fundamental} presents the mean Sampson error $\error_{samp}$ in pixels and running time $t$ in milliseconds with respect to different threshold multipliers $\sigma$, as well as the median and IQR of $\error_{samp}$ for an optimal $\sigma$ for each method and dataset.
As a first observation, R-DPCP is the method with the smallest Sampson error consistently over all threshold choices, as evidenced by the red curves being the bottom ones in \Cref{fig:Fundamental}. Remarkably, R-DPCP surpasses R-GC \cite{Barath2018}, a recently proposed LO-RANSAC. For instance, on Kusvod2, R-DPCP reaches a Sampson error of $0.67$px with $\sigma=10^{-2.6}$, while R-GC reaches that of $1.00$px with the same $\sigma$. Interestingly, R-DPCP seems to exhibit better robustness to the choice of threshold multiplier $\sigma$, e.g. on EPFL, the Sampson error of R-DPCP stays below $0.29$px for $\sigma \leq 10^{-2.8}$ and increases only gradually after that, in sharp contrast with all other methods. Finally, R-DPCP appears to be scalable to the number of correspondences: On Kusvod2 which has $120$ correspondences on average, all the methods have similar running times which are less than $1$msec; on EPFL which contain about $2$k correspondences, R-DPCP uses significantly smaller running time than R-GC, e.g. $5.55$msec compared to $51.16$msec with optimally chosen $\sigma$. For more detailed efficiency evaluation, please refer to the supplementary material.

It may seem surprising that R-DLT \edit{typically} reaches a smaller error than R-GC. We note that while the locally optimized RANSAC in \cite{Chum2003LocallyRANSAC,Lebeda2012FixingRANSAC} uses DLT to \edit{fit models from non-minimal samples}, it is different from the R-DLT here. Rather than doing iterative DLT with a decreasing threshold, R-DLT uses a fixed threshold and stops the iteration if the model does not give a higher score, as in line \ref{line:find-inliers} and \ref{line:stop} of \Cref{alg:LO}. As such, R-DLT differs from the LO, LO$^+$ or LO' reported in \cite{Barath2018}.

\subsection{Essential matrix estimation} \label{sec:exp-essen}

\begin{figure}[hbt!]
\centering
\subfloat[westminster]{
\includegraphics[height=5.3cm]{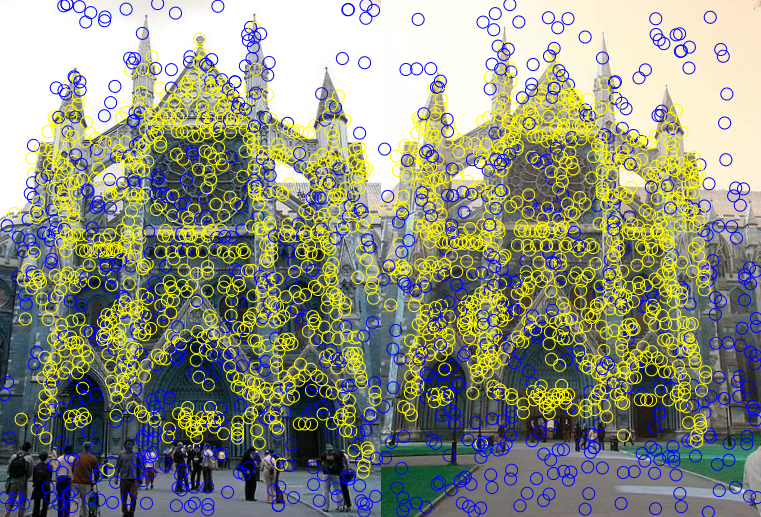}}\\
\subfloat[palace]{
\includegraphics[height=2.9cm]{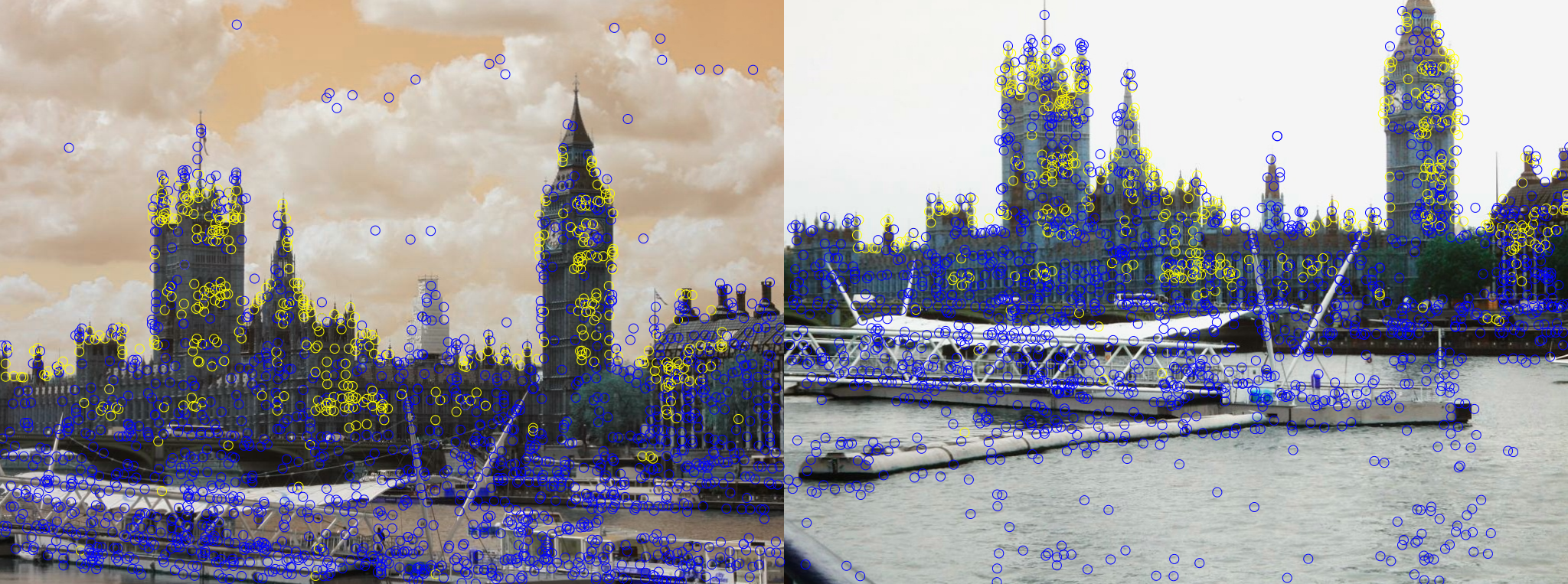}}
\caption{View pairs from PhotoTourism. Correspondences with reprojection error smaller than $5$ pixels (inliers) are marked in yellow, otherwise in blue (outliers).}
\label{fig:hard_ess}
\end{figure}

\myparagraph{Data} For the task of essential matrix estimation, we take $276$ image pairs from the PhotoTourism\footnote{\url{http://phototour.cs.washington.edu/}} dataset, which contains photos of historical sites collected by tourists with very different devices. Each image pair gives around $2000$ correspondences, of which roughly $30\%$ are inliers. Interestingly, the image pairs from PhotoTourism are either capturing planar structures or have small to zero translations, as in \Cref{fig:hard_ess}. 
According to \cite{Ding2020RobustPursuit}, there is a $3$-dimensional vector space of fundamental matrices that approximately explains the inlier correspondences, and the true essential matrix is one of them. 

\myparagraph{Methods} 
Again, we vary the threshold multiplier $\sigma$ to control the threshold of Sampson errors as measured in pixels. The minimal sampler takes $5$ correspondences as linear constraints intersecting the essential variety and solves a polynomial system with $3$ unknown \cite{stewenius2006recent}. During the local optimization, we take the $5$ principle component of all estimated inliers as the linear constraints and solve the polynomial system with $3$ unknowns. However, since the embeddings approximately lie on a lower dimensional subspace, R-DPCP and R-Huber estimates a $3$-dimensional nullspace and take the inverse of cosine subspace angle from the nullspace to the epipolar embeddings as weights for the extraction of principal component. Which serves as a outlier rejection strategy via re-weighting. 

\begin{figure}
\centering
\vspace{-0.55cm}
\hspace{-0.2cm}\subfloat{\scalebox{1.17}{
\input{Rfigures/TutorialTrans_new}}}
\hspace{-0.15cm}
\subfloat{\scalebox{1.17}{
\input{Rfigures/TutorialTime_new}}}
\\
\vspace{-0.4cm}
\hspace{-0.2cm}\subfloat{\scalebox{1.17}{
\input{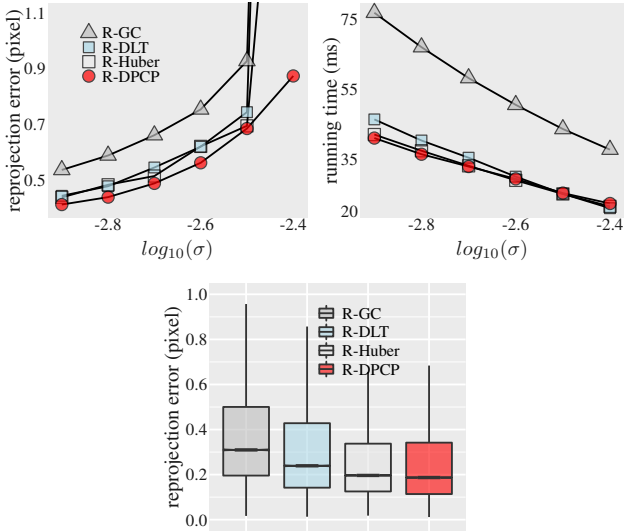}}}
\caption{Essential matrix estimation using $276$ image pairs from the Phototourism dataset. All methods terminate when a confidence of $p=95\%$ has been reached. The top row shows reprojection error and running time with varying threshold multiplier $\sigma$, while the bottom row shows reprojection error with threshold optimally tuned for each method. Experiments are averaged over 300 trials. Reprojection error is in pixels and running time is in milliseconds.}
\label{fig:tutorial}
\end{figure}

\myparagraph{Results} 
In \Cref{fig:tutorial}, we report the mean reprojection error $\error_{repr}$ in pixels and running time $t$ in milliseconds with respect to different $\sigma$, as well as the median and IQR of $\error_{repr}$ with an optimal $\sigma$ tuned for each method. Again, we observe that R-DPCP gives the smallest error among all choices of thresholds compared to other methods. Moreover, we note that with a larger threshold multiplier $\sigma$, the running time of all methods decreases. This is likely due to the fact that with a larger threshold, more inliers tend to be counted, hence less RANSAC iterations are required to reach the confidence $p$.

\subsection{Homography estimation} \label{sec:exp-homo}

\begin{figure}
\subfloat[Adam]{
\includegraphics[height=3.8cm]{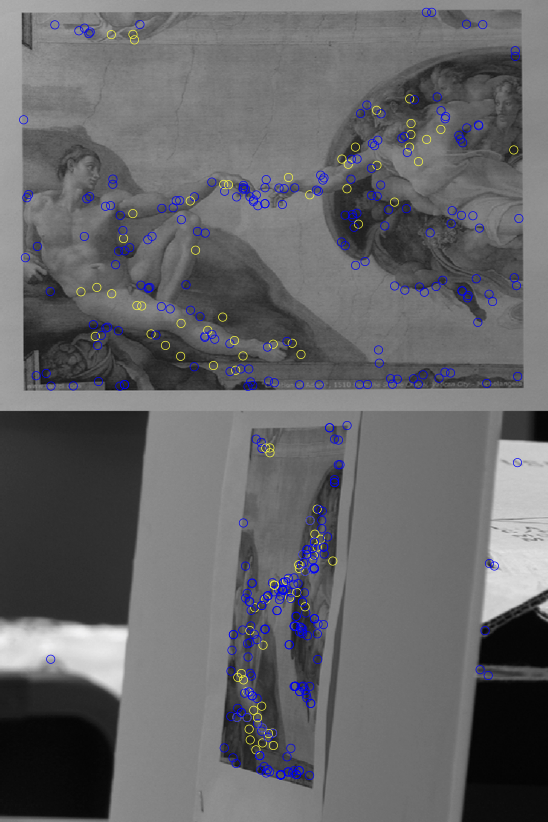}}
\subfloat[Abstract]{
\includegraphics[height=3.8cm]{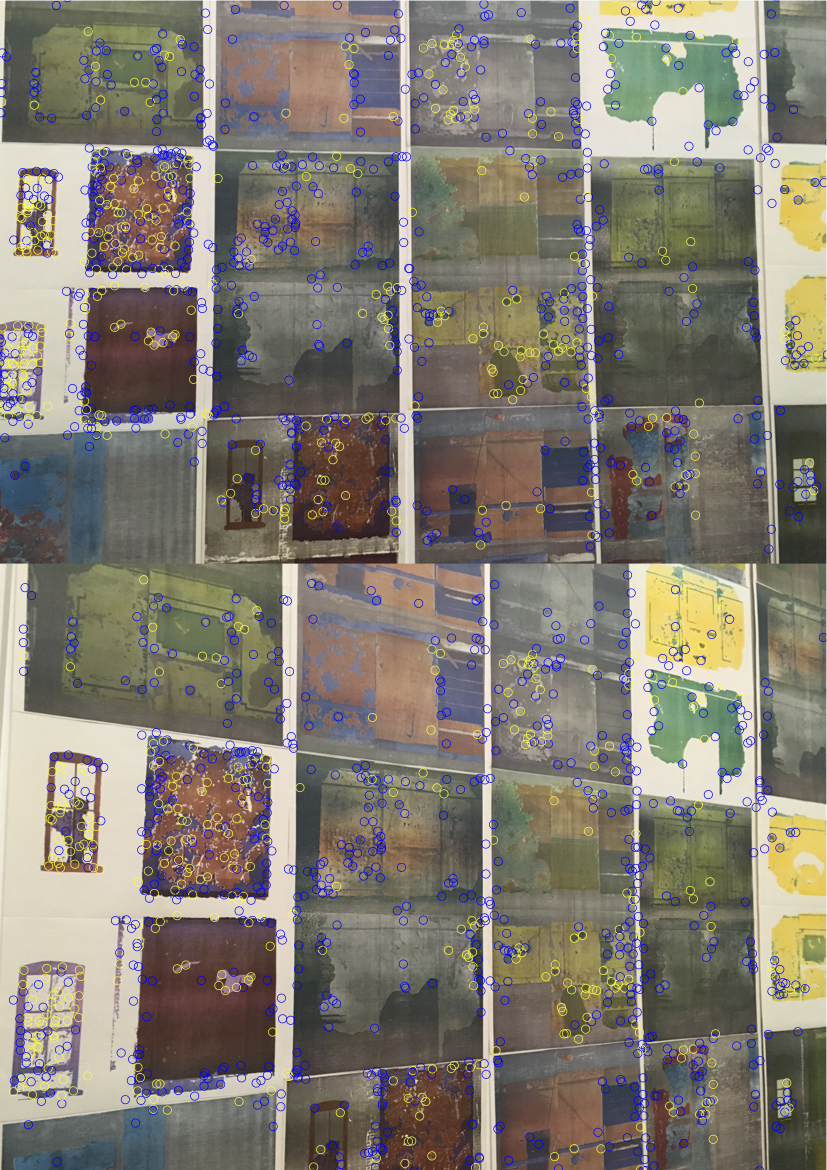}}
\subfloat[Bees]{
\includegraphics[height=3.8cm]{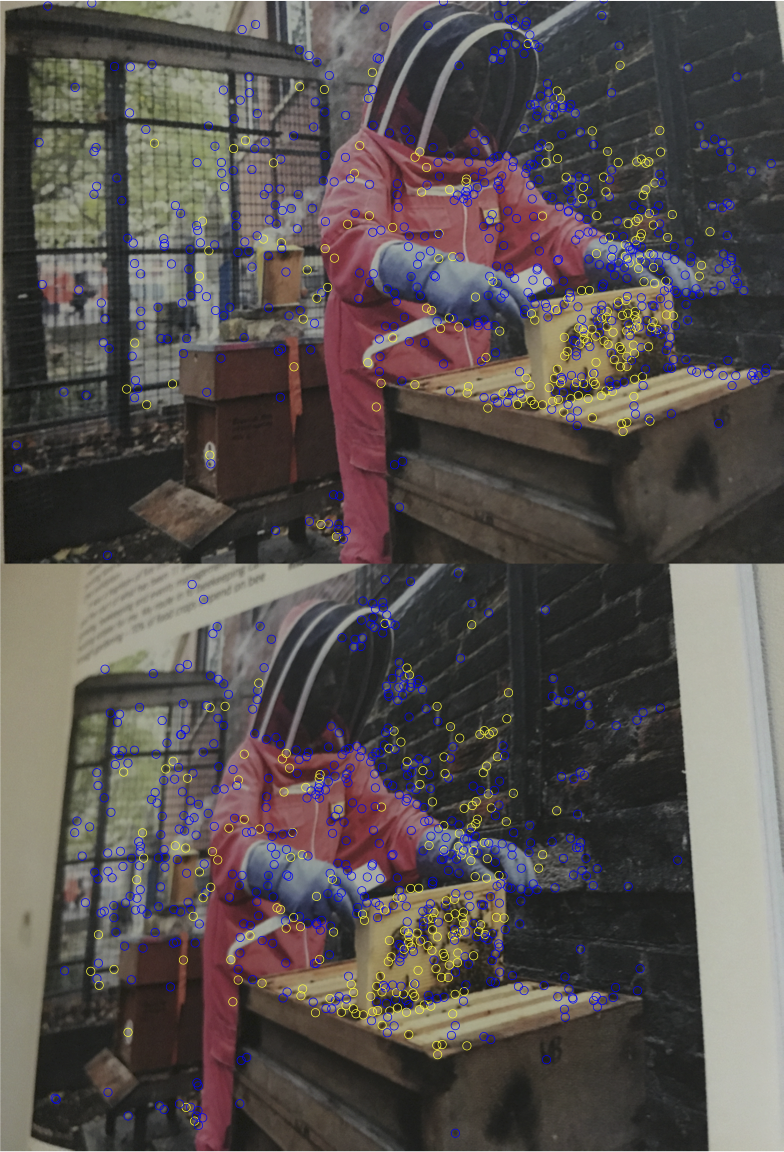}}
\caption{View pairs from EVD (a) and Hpatches (b,c). Correspondences with reprojection error smaller than $5$ pixels (inliers) are marked in yellow, otherwise in blue (outliers).}
\label{fig:hard}
\end{figure}

\myparagraph{Data}
We take $14$ image pairs from the Extreme View Dataset (EVD)\footnote{http://cmp.felk.cvut.cz/wbs. Image pair `cafe' from EVD is not used as the provided correspondences contain less than $5\%$ inliers.} \edit{and $115$ from the Hpatches dataset\footnote{https://hpatches.github.io}}. They collect image pairs of textured planar structures, e.g. graffitis and posters, as well as those with small baselines \cite{Vidal2002StructureBaselines}. For EVD, we use the provided feature correspondences for homography estimation, and the validation set of inliers for evaluation. For Hpatches, we detect and match features using SURF \cite{bay2006surf} with a threshold of $30$ and a ratio test of $0.7$. As shown in \Cref{fig:hard}, image pairs from these datasets are challenging for being highly noisy and having extreme view-point changes. We perform a similar normalization to the correspondences and homographic embeddings as described in \S \ref{sec:exp-fund}. 

\myparagraph{Methods} $4$ pairs of correspondences are drawn as minimal samples, from which a homography matrix can be estimated \cite{hartley2003multiple}. 
We vary $\sigma$ to control the threshold of reprojection errors as measured in pixels.\dpr{reprojection or projection?  What is reprojection? This may be correct, but I just do not know.}

\begin{figure}
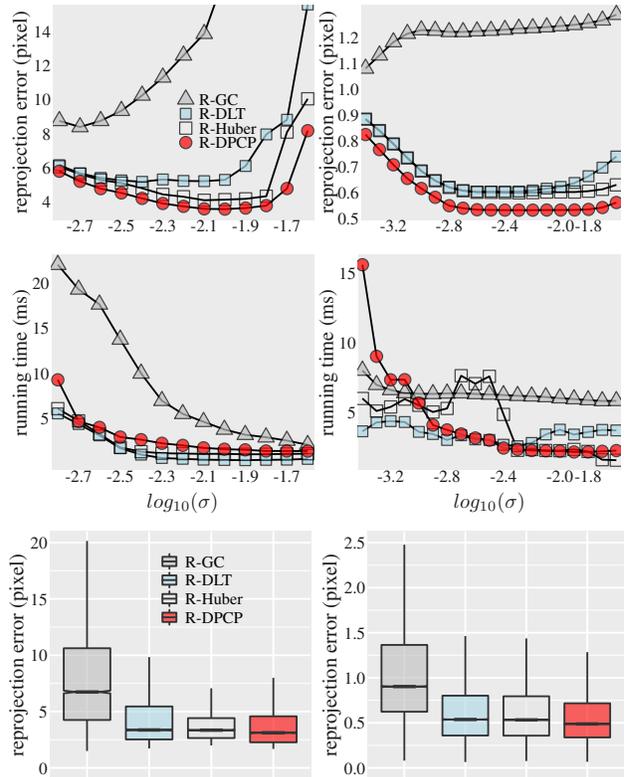

\centering
\vspace{-0.4cm}
\hspace{-0.2cm}\subfloat{\scalebox{1.17}{
\input{Rfigures/EVDsampson_mean}}}
\hspace{-0.2cm}\subfloat{\scalebox{1.17}{
\input{Rfigures/HPatchesTrans_mean}}}
\\
\vspace{-0.8cm}\hspace{-0.2cm}\subfloat{\scalebox{1.17}{
\input{Rfigures/EVDTime_median}}}
\hspace{-0.2cm}\subfloat{\scalebox{1.17}{
\input{Rfigures/HPatchesTime_mean}}}
\\
\vspace{-0.45cm}\hspace{-0.2cm}\subfloat{\scalebox{1.17}{
\input{Rfigures/EVDbox}}}
\hspace{-0.2cm}
\subfloat{\scalebox{1.17}{
\input{Rfigures/Hpatchesbox}}}
\caption{Homography estimation using $14$ image pairs from the EVD dataset (left) and $115$ form the Hpatches (right). All methods terminate when a confidence of $p=95\%$ has been reached. The top two rows show reprojection error and running time with varying threshold multiplier $\sigma$, while the bottom row shows reprojection error with threshold optimally tuned for each method. Experiments are averaged over $500$ trials. Reprojection error is in pixels, running time is in milliseconds.  }
\label{fig:Homography}
\end{figure}

\myparagraph{Results} In \Cref{fig:Homography}, we report the mean reprojection error $\error_{repr}$ in pixels and running time $t$ in milliseconds with respect to different $\sigma$, as well as the median and IQR of $\error_{repr}$ with an optimal $\sigma$ tuned for each method. Again, R-DPCP consistently yields the smallest reprojection error on all datasets. 
Moreover, R-DPCP seems to be less sensitive to the choice of threshold, \edit{as for example in Hpatches, the reprojection error only increases from $0.55$px to $0.56$px as $\sigma$ goes from $10^{-2.4}$ to $10^{-1.6}$ ($\epsilon$ being $\sim 6\times$ larger), while that of R-DLT increases from $0.62$px to $0.74$px.} Finally, R-DPCP has comparably small running time with other methods, e.g. below $5$ms for most threshold choices. 

\subsection{Homographic tensor estimation} \label{sec:exp-HT}

\myparagraph{Data} We use the TUM dataset \cite{sturm12iros}, which consists of video sequences of indoor environments taken by a hand-held RGB-D camera. We take two sequences suitable for this task: `nostructure\_texture\_near\_withloop' (Near) where the scene is planar posters so that rich features can be detected and matched, and
`360\_hemisphere' (Hemi) where the motion is almost pure rotation.
We take a frame gap of $20$ for 'Near' and $10$ for `Hemi' as used in \cite{Ding2020RobustPursuit}. A similar normalization is applied to correspondences in each view and the homographic tensor embeddings as in \S \ref{sec:exp-fund}. As the geometrically related views increase from $2$ to $3$, the dimension of the optimization problem increases from $9$ to $27$. The task of homographic tensor estimation is more challenging than the case of two-views because a hyperplane in $\mathbb{R}^{27}$ is less distinguishable than in $\mathbb{R}^9$.

\begin{figure}
\centering
\subfloat[Near]{
\includegraphics[width=0.9\linewidth]{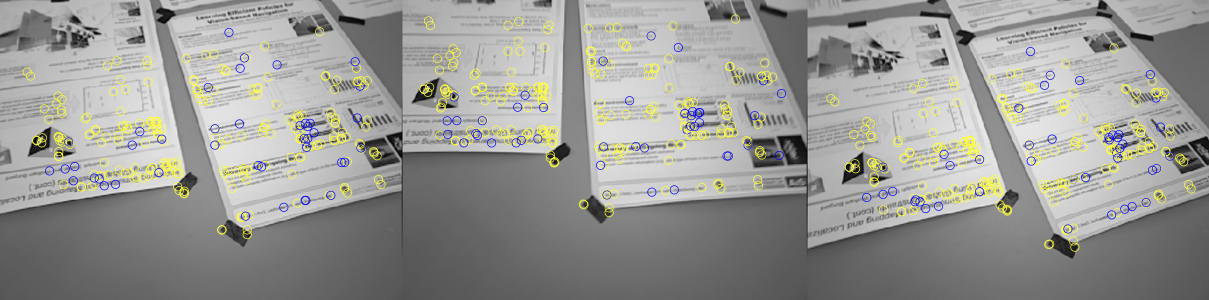}}\\
\subfloat[Hemi]{
\includegraphics[width=0.9\linewidth]{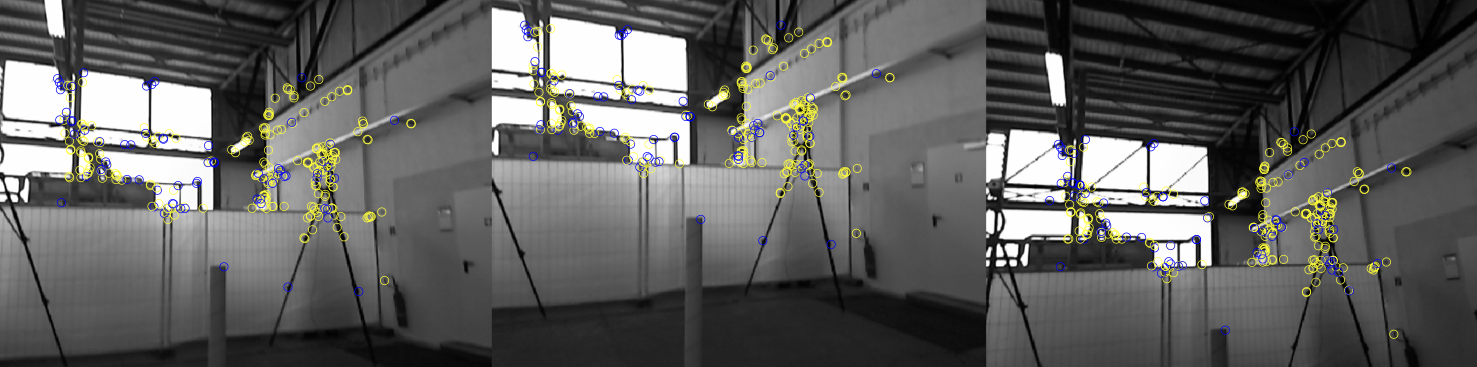}}
\caption{View triplets from TUM: (a) Near (b) Hemi. Correspondences with reprojection error smaller than $5$ pixel are marked in yellow (inliers), otherwise in blue (outliers).}
\label{fig:HT}
\end{figure}

\myparagraph{Methods} The minimal sampler (\Cref{alg:LO-RANSAC}, \cref{line:minimal-sampler}) takes $4$ correspondences to linearly estimate a homographic tensor. Since each homographic tensor can be decomposed into two pairwise homography matrices, the reprojection error is used as a scoring function for RANSAC. 

\begin{figure}
\centering
\vspace{-0.4cm}
\hspace{-0.2cm}\subfloat{\scalebox{1.17}{
\input{Rfigures/NearRot_mean}}}
\hspace{-0.2cm}\subfloat{\scalebox{1.17}{
\input{Rfigures/HemiRot_mean}}}
\\
\vspace{-0.8cm}
\hspace{-0.2cm}\subfloat{\scalebox{1.17}{
\input{Rfigures/NearTrans_mean}}}
\hspace{-0.2cm}\subfloat{\scalebox{1.17}{
\input{Rfigures/HemiTrans_mean}}}

\caption{Homographic tensor estimation using $31$ image triples from `Near' (left)  and $73$ from `Hemi' (right), with varying threshold multiplier $\sigma$. All methods terminate when a confidence of $p=95\%$ has been reached. Experiments are averaged over $100$ trials. Rotation and translation errors are in degrees.}
\label{fig:HomoTensor}
\end{figure}
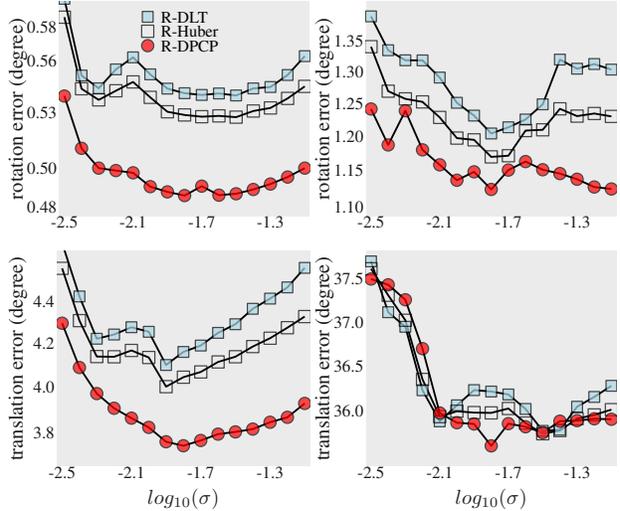

\myparagraph{Results} We report the average angular rotation and angular translation errors ($\error_{Rot}$, $\error_{tran}$) in degrees with respect to different thresholds in \Cref{fig:HomoTensor}. 
As a first observation, in the downstream task of retrieving relative poses, R-DPCP produces the most accurate estimate. In 'Near', it yields $0.49^\circ$ and $3.81^\circ$ of rotation and translation error compared with $0.54^\circ$ and $4.10^\circ$ given by R-Huber.
Interestingly, the translation errors given by all methods are significant ($\geq 30^\circ$) on 'Hemi'. This is because the inter-frame translation (baseline) for the sequence 'Hemi' is small. As such, the signal-to-noise ratio for translation is small, leading to a large translation error.

\section{Conclusion}

We employed the robust subspace learning method Dual Principal Component Pursuit inside the local optimization step of RANSAC. The proposed DPCP-RANSAC has far fewer parameters than existing methods \cite{Chum2003LocallyRANSAC,Lebeda2012FixingRANSAC,Serych2016FastEstimation,Barath2018} and can be extended to various geometric vision tasks conveniently. Our experimental evaluation on large-scale datasets shows that DPCP-RANSAC (1) offers a stable estimation that is more accurate than its recent predecessors, (2) is scalable with time complexity comparable to that of its simplest alternatives, and (3) its sensitivity to the threshold parameter is significantly improved.


{\small
\bibliographystyle{abbrv}
\bibliography{refs}
}

\end{document}

%% file: Rfigures/stema_b.tex
\begin{tikzpicture}[x=1pt,y=1pt]
\clip (5,10) rectangle (104, 93.95);
\definecolor{fillColor}{RGB}{255,255,255}
\path[use as bounding box,fill=fillColor,fill opacity=0.00] (0,0) rectangle (108.41, 93.95);
\begin{scope}
\path[clip] (  0.00,  0.00) rectangle (108.40, 93.95);
\definecolor{drawColor}{RGB}{255,255,255}
\definecolor{fillColor}{RGB}{255,255,255}

\path[draw=drawColor,line width= 0.6pt,line join=round,line cap=round,fill=fillColor] (  0.00,  0.00) rectangle (108.41, 93.95);
\end{scope}
\begin{scope}
\path[clip] ( 18.58, 14.34) rectangle (102.90, 88.45);
\definecolor{fillColor}{gray}{0.92}

\path[fill=fillColor] ( 18.58, 14.34) rectangle (102.91, 88.45);
\definecolor{drawColor}{RGB}{0,0,0}

\path[draw=drawColor,draw opacity=0.50,line width= 0.6pt,line join=round] ( 18.58, 17.71) -- (102.90, 17.71);
\definecolor{drawColor}{RGB}{70,130,180}

\path[draw=drawColor,line width= 0.6pt,line join=round] ( 22.41, 17.71) -- ( 22.41, 17.71);

\path[draw=drawColor,line width= 0.6pt,line join=round] ( 22.94, 17.72) -- ( 22.94, 17.71);

\path[draw=drawColor,line width= 0.6pt,line join=round] ( 23.47, 17.72) -- ( 23.47, 17.71);

\path[draw=drawColor,line width= 0.6pt,line join=round] ( 24.00, 17.73) -- ( 24.00, 17.71);

\path[draw=drawColor,line width= 0.6pt,line join=round] ( 24.53, 17.73) -- ( 24.53, 17.71);

\path[draw=drawColor,line width= 0.6pt,line join=round] ( 25.05, 17.73) -- ( 25.05, 17.71);

\path[draw=drawColor,line width= 0.6pt,line join=round] ( 25.58, 17.73) -- ( 25.58, 17.71);

\path[draw=drawColor,line width= 0.6pt,line join=round] ( 26.11, 17.74) -- ( 26.11, 17.71);

\path[draw=drawColor,line width= 0.6pt,line join=round] ( 26.64, 17.74) -- ( 26.64, 17.71);

\path[draw=drawColor,line width= 0.6pt,line join=round] ( 27.17, 17.75) -- ( 27.17, 17.71);

\path[draw=drawColor,line width= 0.6pt,line join=round] ( 27.70, 17.75) -- ( 27.70, 17.71);

\path[draw=drawColor,line width= 0.6pt,line join=round] ( 28.23, 17.75) -- ( 28.23, 17.71);

\path[draw=drawColor,line width= 0.6pt,line join=round] ( 28.75, 17.76) -- ( 28.75, 17.71);

\path[draw=drawColor,line width= 0.6pt,line join=round] ( 29.28, 17.76) -- ( 29.28, 17.71);

\path[draw=drawColor,line width= 0.6pt,line join=round] ( 29.81, 17.76) -- ( 29.81, 17.71);

\path[draw=drawColor,line width= 0.6pt,line join=round] ( 30.34, 17.77) -- ( 30.34, 17.71);

\path[draw=drawColor,line width= 0.6pt,line join=round] ( 30.87, 17.78) -- ( 30.87, 17.71);

\path[draw=drawColor,line width= 0.6pt,line join=round] ( 31.40, 17.78) -- ( 31.40, 17.71);

\path[draw=drawColor,line width= 0.6pt,line join=round] ( 31.93, 17.78) -- ( 31.93, 17.71);

\path[draw=drawColor,line width= 0.6pt,line join=round] ( 32.46, 17.79) -- ( 32.46, 17.71);

\path[draw=drawColor,line width= 0.6pt,line join=round] ( 32.98, 17.79) -- ( 32.98, 17.71);

\path[draw=drawColor,line width= 0.6pt,line join=round] ( 33.51, 17.79) -- ( 33.51, 17.71);

\path[draw=drawColor,line width= 0.6pt,line join=round] ( 34.04, 17.80) -- ( 34.04, 17.71);

\path[draw=drawColor,line width= 0.6pt,line join=round] ( 34.57, 17.81) -- ( 34.57, 17.71);

\path[draw=drawColor,line width= 0.6pt,line join=round] ( 35.10, 17.81) -- ( 35.10, 17.71);

\path[draw=drawColor,line width= 0.6pt,line join=round] ( 35.63, 17.81) -- ( 35.63, 17.71);

\path[draw=drawColor,line width= 0.6pt,line join=round] ( 36.16, 17.81) -- ( 36.16, 17.71);

\path[draw=drawColor,line width= 0.6pt,line join=round] ( 36.69, 17.81) -- ( 36.69, 17.71);

\path[draw=drawColor,line width= 0.6pt,line join=round] ( 37.21, 17.81) -- ( 37.21, 17.71);

\path[draw=drawColor,line width= 0.6pt,line join=round] ( 37.74, 17.82) -- ( 37.74, 17.71);

\path[draw=drawColor,line width= 0.6pt,line join=round] ( 38.27, 17.82) -- ( 38.27, 17.71);

\path[draw=drawColor,line width= 0.6pt,line join=round] ( 38.80, 17.83) -- ( 38.80, 17.71);

\path[draw=drawColor,line width= 0.6pt,line join=round] ( 39.33, 17.83) -- ( 39.33, 17.71);

\path[draw=drawColor,line width= 0.6pt,line join=round] ( 39.86, 17.84) -- ( 39.86, 17.71);

\path[draw=drawColor,line width= 0.6pt,line join=round] ( 40.39, 17.84) -- ( 40.39, 17.71);

\path[draw=drawColor,line width= 0.6pt,line join=round] ( 40.91, 17.84) -- ( 40.91, 17.71);

\path[draw=drawColor,line width= 0.6pt,line join=round] ( 41.44, 17.85) -- ( 41.44, 17.71);

\path[draw=drawColor,line width= 0.6pt,line join=round] ( 41.97, 17.86) -- ( 41.97, 17.71);

\path[draw=drawColor,line width= 0.6pt,line join=round] ( 42.50, 17.86) -- ( 42.50, 17.71);

\path[draw=drawColor,line width= 0.6pt,line join=round] ( 43.03, 17.86) -- ( 43.03, 17.71);

\path[draw=drawColor,line width= 0.6pt,line join=round] ( 43.56, 17.86) -- ( 43.56, 17.71);

\path[draw=drawColor,line width= 0.6pt,line join=round] ( 44.09, 17.87) -- ( 44.09, 17.71);

\path[draw=drawColor,line width= 0.6pt,line join=round] ( 44.62, 17.88) -- ( 44.62, 17.71);

\path[draw=drawColor,line width= 0.6pt,line join=round] ( 45.14, 17.88) -- ( 45.14, 17.71);

\path[draw=drawColor,line width= 0.6pt,line join=round] ( 45.67, 17.88) -- ( 45.67, 17.71);

\path[draw=drawColor,line width= 0.6pt,line join=round] ( 46.20, 17.88) -- ( 46.20, 17.71);

\path[draw=drawColor,line width= 0.6pt,line join=round] ( 46.73, 17.89) -- ( 46.73, 17.71);

\path[draw=drawColor,line width= 0.6pt,line join=round] ( 47.26, 17.89) -- ( 47.26, 17.71);

\path[draw=drawColor,line width= 0.6pt,line join=round] ( 47.79, 17.90) -- ( 47.79, 17.71);

\path[draw=drawColor,line width= 0.6pt,line join=round] ( 48.32, 17.90) -- ( 48.32, 17.71);

\path[draw=drawColor,line width= 0.6pt,line join=round] ( 48.85, 17.92) -- ( 48.85, 17.71);

\path[draw=drawColor,line width= 0.6pt,line join=round] ( 49.37, 17.92) -- ( 49.37, 17.71);

\path[draw=drawColor,line width= 0.6pt,line join=round] ( 49.90, 17.93) -- ( 49.90, 17.71);

\path[draw=drawColor,line width= 0.6pt,line join=round] ( 50.43, 17.94) -- ( 50.43, 17.71);

\path[draw=drawColor,line width= 0.6pt,line join=round] ( 50.96, 17.94) -- ( 50.96, 17.71);

\path[draw=drawColor,line width= 0.6pt,line join=round] ( 51.49, 17.94) -- ( 51.49, 17.71);

\path[draw=drawColor,line width= 0.6pt,line join=round] ( 52.02, 17.95) -- ( 52.02, 17.71);

\path[draw=drawColor,line width= 0.6pt,line join=round] ( 52.55, 17.95) -- ( 52.55, 17.71);

\path[draw=drawColor,line width= 0.6pt,line join=round] ( 53.07, 17.95) -- ( 53.07, 17.71);

\path[draw=drawColor,line width= 0.6pt,line join=round] ( 53.60, 17.96) -- ( 53.60, 17.71);

\path[draw=drawColor,line width= 0.6pt,line join=round] ( 54.13, 17.96) -- ( 54.13, 17.71);

\path[draw=drawColor,line width= 0.6pt,line join=round] ( 54.66, 17.97) -- ( 54.66, 17.71);

\path[draw=drawColor,line width= 0.6pt,line join=round] ( 55.19, 17.98) -- ( 55.19, 17.71);

\path[draw=drawColor,line width= 0.6pt,line join=round] ( 55.72, 17.98) -- ( 55.72, 17.71);

\path[draw=drawColor,line width= 0.6pt,line join=round] ( 56.25, 17.99) -- ( 56.25, 17.71);

\path[draw=drawColor,line width= 0.6pt,line join=round] ( 56.78, 17.99) -- ( 56.78, 17.71);

\path[draw=drawColor,line width= 0.6pt,line join=round] ( 57.30, 17.99) -- ( 57.30, 17.71);

\path[draw=drawColor,line width= 0.6pt,line join=round] ( 57.83, 18.00) -- ( 57.83, 17.71);

\path[draw=drawColor,line width= 0.6pt,line join=round] ( 58.36, 18.00) -- ( 58.36, 17.71);

\path[draw=drawColor,line width= 0.6pt,line join=round] ( 58.89, 18.01) -- ( 58.89, 17.71);

\path[draw=drawColor,line width= 0.6pt,line join=round] ( 59.42, 18.01) -- ( 59.42, 17.71);

\path[draw=drawColor,line width= 0.6pt,line join=round] ( 59.95, 18.02) -- ( 59.95, 17.71);

\path[draw=drawColor,line width= 0.6pt,line join=round] ( 60.48, 18.02) -- ( 60.48, 17.71);

\path[draw=drawColor,line width= 0.6pt,line join=round] ( 61.01, 18.03) -- ( 61.01, 17.71);

\path[draw=drawColor,line width= 0.6pt,line join=round] ( 61.53, 18.03) -- ( 61.53, 17.71);

\path[draw=drawColor,line width= 0.6pt,line join=round] ( 62.06, 18.04) -- ( 62.06, 17.71);

\path[draw=drawColor,line width= 0.6pt,line join=round] ( 62.59, 18.05) -- ( 62.59, 17.71);

\path[draw=drawColor,line width= 0.6pt,line join=round] ( 63.12, 18.05) -- ( 63.12, 17.71);

\path[draw=drawColor,line width= 0.6pt,line join=round] ( 63.65, 18.06) -- ( 63.65, 17.71);

\path[draw=drawColor,line width= 0.6pt,line join=round] ( 64.18, 18.07) -- ( 64.18, 17.71);

\path[draw=drawColor,line width= 0.6pt,line join=round] ( 64.71, 18.07) -- ( 64.71, 17.71);

\path[draw=drawColor,line width= 0.6pt,line join=round] ( 65.24, 18.08) -- ( 65.24, 17.71);

\path[draw=drawColor,line width= 0.6pt,line join=round] ( 65.76, 18.08) -- ( 65.76, 17.71);

\path[draw=drawColor,line width= 0.6pt,line join=round] ( 66.29, 18.08) -- ( 66.29, 17.71);

\path[draw=drawColor,line width= 0.6pt,line join=round] ( 66.82, 18.08) -- ( 66.82, 17.71);

\path[draw=drawColor,line width= 0.6pt,line join=round] ( 67.35, 18.08) -- ( 67.35, 17.71);

\path[draw=drawColor,line width= 0.6pt,line join=round] ( 67.88, 18.09) -- ( 67.88, 17.71);

\path[draw=drawColor,line width= 0.6pt,line join=round] ( 68.41, 18.13) -- ( 68.41, 17.71);

\path[draw=drawColor,line width= 0.6pt,line join=round] ( 68.94, 18.13) -- ( 68.94, 17.71);

\path[draw=drawColor,line width= 0.6pt,line join=round] ( 69.46, 18.13) -- ( 69.46, 17.71);

\path[draw=drawColor,line width= 0.6pt,line join=round] ( 69.99, 18.13) -- ( 69.99, 17.71);

\path[draw=drawColor,line width= 0.6pt,line join=round] ( 70.52, 18.17) -- ( 70.52, 17.71);

\path[draw=drawColor,line width= 0.6pt,line join=round] ( 71.05, 18.18) -- ( 71.05, 17.71);

\path[draw=drawColor,line width= 0.6pt,line join=round] ( 71.58, 18.18) -- ( 71.58, 17.71);

\path[draw=drawColor,line width= 0.6pt,line join=round] ( 72.11, 18.18) -- ( 72.11, 17.71);

\path[draw=drawColor,line width= 0.6pt,line join=round] ( 72.64, 18.20) -- ( 72.64, 17.71);

\path[draw=drawColor,line width= 0.6pt,line join=round] ( 73.17, 18.20) -- ( 73.17, 17.71);

\path[draw=drawColor,line width= 0.6pt,line join=round] ( 73.69, 18.24) -- ( 73.69, 17.71);

\path[draw=drawColor,line width= 0.6pt,line join=round] ( 74.22, 18.25) -- ( 74.22, 17.71);

\path[draw=drawColor,line width= 0.6pt,line join=round] ( 74.75, 18.26) -- ( 74.75, 17.71);

\path[draw=drawColor,line width= 0.6pt,line join=round] ( 75.28, 18.28) -- ( 75.28, 17.71);

\path[draw=drawColor,line width= 0.6pt,line join=round] ( 75.81, 18.29) -- ( 75.81, 17.71);

\path[draw=drawColor,line width= 0.6pt,line join=round] ( 76.34, 18.30) -- ( 76.34, 17.71);

\path[draw=drawColor,line width= 0.6pt,line join=round] ( 76.87, 18.31) -- ( 76.87, 17.71);

\path[draw=drawColor,line width= 0.6pt,line join=round] ( 77.40, 18.36) -- ( 77.40, 17.71);

\path[draw=drawColor,line width= 0.6pt,line join=round] ( 77.92, 18.39) -- ( 77.92, 17.71);

\path[draw=drawColor,line width= 0.6pt,line join=round] ( 78.45, 18.39) -- ( 78.45, 17.71);

\path[draw=drawColor,line width= 0.6pt,line join=round] ( 78.98, 18.39) -- ( 78.98, 17.71);

\path[draw=drawColor,line width= 0.6pt,line join=round] ( 79.51, 18.40) -- ( 79.51, 17.71);

\path[draw=drawColor,line width= 0.6pt,line join=round] ( 80.04, 18.40) -- ( 80.04, 17.71);

\path[draw=drawColor,line width= 0.6pt,line join=round] ( 80.57, 18.41) -- ( 80.57, 17.71);

\path[draw=drawColor,line width= 0.6pt,line join=round] ( 81.10, 18.43) -- ( 81.10, 17.71);

\path[draw=drawColor,line width= 0.6pt,line join=round] ( 81.62, 18.45) -- ( 81.62, 17.71);

\path[draw=drawColor,line width= 0.6pt,line join=round] ( 82.15, 18.45) -- ( 82.15, 17.71);

\path[draw=drawColor,line width= 0.6pt,line join=round] ( 82.68, 18.46) -- ( 82.68, 17.71);

\path[draw=drawColor,line width= 0.6pt,line join=round] ( 83.21, 18.48) -- ( 83.21, 17.71);

\path[draw=drawColor,line width= 0.6pt,line join=round] ( 83.74, 18.49) -- ( 83.74, 17.71);

\path[draw=drawColor,line width= 0.6pt,line join=round] ( 84.27, 18.49) -- ( 84.27, 17.71);

\path[draw=drawColor,line width= 0.6pt,line join=round] ( 84.80, 18.51) -- ( 84.80, 17.71);

\path[draw=drawColor,line width= 0.6pt,line join=round] ( 85.33, 18.55) -- ( 85.33, 17.71);

\path[draw=drawColor,line width= 0.6pt,line join=round] ( 85.85, 18.56) -- ( 85.85, 17.71);

\path[draw=drawColor,line width= 0.6pt,line join=round] ( 86.38, 18.56) -- ( 86.38, 17.71);

\path[draw=drawColor,line width= 0.6pt,line join=round] ( 86.91, 18.60) -- ( 86.91, 17.71);

\path[draw=drawColor,line width= 0.6pt,line join=round] ( 87.44, 18.65) -- ( 87.44, 17.71);

\path[draw=drawColor,line width= 0.6pt,line join=round] ( 87.97, 18.69) -- ( 87.97, 17.71);

\path[draw=drawColor,line width= 0.6pt,line join=round] ( 88.50, 18.70) -- ( 88.50, 17.71);

\path[draw=drawColor,line width= 0.6pt,line join=round] ( 89.03, 18.70) -- ( 89.03, 17.71);

\path[draw=drawColor,line width= 0.6pt,line join=round] ( 89.56, 18.86) -- ( 89.56, 17.71);

\path[draw=drawColor,line width= 0.6pt,line join=round] ( 90.08, 18.89) -- ( 90.08, 17.71);

\path[draw=drawColor,line width= 0.6pt,line join=round] ( 90.61, 18.91) -- ( 90.61, 17.71);

\path[draw=drawColor,line width= 0.6pt,line join=round] ( 91.14, 19.03) -- ( 91.14, 17.71);

\path[draw=drawColor,line width= 0.6pt,line join=round] ( 91.67, 19.33) -- ( 91.67, 17.71);

\path[draw=drawColor,line width= 0.6pt,line join=round] ( 92.20, 22.23) -- ( 92.20, 17.71);

\path[draw=drawColor,line width= 0.6pt,line join=round] ( 92.73, 23.39) -- ( 92.73, 17.71);

\path[draw=drawColor,line width= 0.6pt,line join=round] ( 93.26, 25.10) -- ( 93.26, 17.71);

\path[draw=drawColor,line width= 0.6pt,line join=round] ( 93.78, 26.32) -- ( 93.78, 17.71);

\path[draw=drawColor,line width= 0.6pt,line join=round] ( 94.31, 26.79) -- ( 94.31, 17.71);

\path[draw=drawColor,line width= 0.6pt,line join=round] ( 94.84, 32.36) -- ( 94.84, 17.71);

\path[draw=drawColor,line width= 0.6pt,line join=round] ( 95.37, 36.77) -- ( 95.37, 17.71);

\path[draw=drawColor,line width= 0.6pt,line join=round] ( 95.90, 59.06) -- ( 95.90, 17.71);

\path[draw=drawColor,line width= 0.6pt,line join=round] ( 96.43, 76.96) -- ( 96.43, 17.71);

\path[draw=drawColor,line width= 0.6pt,line join=round] ( 96.96, 83.90) -- ( 96.96, 17.71);

\path[draw=drawColor,line width= 0.6pt,line join=round] ( 97.49, 86.26) -- ( 97.49, 17.71);

\path[draw=drawColor,line width= 0.6pt,line join=round] ( 98.01,177.93) -- ( 98.01, 17.71);

\path[draw=drawColor,line width= 0.6pt,line join=round] ( 98.54,458.28) -- ( 98.54, 17.71);

\path[draw=drawColor,line width= 0.6pt,line join=round] ( 99.07,527.57) -- ( 99.07, 17.71);
\end{scope}
\begin{scope}
\path[clip] (  0.00,  0.00) rectangle (108.41, 93.95);
\definecolor{drawColor}{gray}{0.10}

\node[text=drawColor,anchor=base east,inner sep=0pt, outer sep=0pt, scale=  0.55] at ( 17.83, 32.66) {10};

\node[text=drawColor,anchor=base east,inner sep=0pt, outer sep=0pt, scale=  0.55] at ( 17.83, 49.50) {20};

\node[text=drawColor,anchor=base east,inner sep=0pt, outer sep=0pt, scale=  0.55] at ( 17.83, 66.34) {30};

\node[text=drawColor,anchor=base east,inner sep=0pt, outer sep=0pt, scale=  0.55] at ( 17.83, 83.19) {40};
\end{scope}
\begin{scope}
\path[clip] (  0.00,  0.00) rectangle (108.41, 93.95);
\definecolor{drawColor}{gray}{0.10}

\node[text=drawColor,anchor=base,inner sep=0pt, outer sep=0pt, scale=  0.55] at ( 21.88, 10.60) {0};

\node[text=drawColor,anchor=base,inner sep=0pt, outer sep=0pt, scale=  0.55] at ( 48.32, 10.60) {50};

\node[text=drawColor,anchor=base,inner sep=0pt, outer sep=0pt, scale=  0.55] at ( 74.75, 10.60) {100};
\end{scope}
\begin{scope}
\path[clip] (  0.00,  0.00) rectangle (108.41, 93.95);
\definecolor{drawColor}{gray}{0.10}

\node[text=drawColor,rotate= 90.00,anchor=base,inner sep=0pt, outer sep=0pt, scale=  0.66] at ( 10.05, 51.40) {reprojection error (pixel)};
\end{scope}
\end{tikzpicture}

%% file: Rfigures/stemb_b.tex
\begin{tikzpicture}[x=1pt,y=1pt]
\clip (5,10) rectangle (104, 93.95);
\definecolor{fillColor}{RGB}{255,255,255}
\path[use as bounding box,fill=fillColor,fill opacity=0.00] (0,0) rectangle (108.41, 93.95);
\begin{scope}
\path[clip] (  0.00,  0.00) rectangle (108.41, 93.95);
\definecolor{drawColor}{RGB}{255,255,255}
\definecolor{fillColor}{RGB}{255,255,255}

\path[draw=drawColor,line width= 0.6pt,line join=round,line cap=round,fill=fillColor] ( -0.00,  0.00) rectangle (108.41, 93.95);
\end{scope}
\begin{scope}
\path[clip] ( 22.85, 14.34) rectangle (102.90, 88.45);
\definecolor{fillColor}{gray}{0.92}

\path[fill=fillColor] ( 22.85, 14.34) rectangle (102.91, 88.45);
\definecolor{drawColor}{RGB}{0,0,0}

\path[draw=drawColor,draw opacity=0.50,line width= 0.6pt,line join=round] ( 22.85, 17.71) -- (102.90, 17.71);
\definecolor{drawColor}{RGB}{70,130,180}

\path[draw=drawColor,line width= 0.6pt,line join=round] ( 26.49, 17.77) -- ( 26.49, 17.71);

\path[draw=drawColor,line width= 0.6pt,line join=round] ( 26.99, 22.70) -- ( 26.99, 17.71);

\path[draw=drawColor,line width= 0.6pt,line join=round] ( 27.50, 24.85) -- ( 27.50, 17.71);

\path[draw=drawColor,line width= 0.6pt,line join=round] ( 28.00, 26.93) -- ( 28.00, 17.71);

\path[draw=drawColor,line width= 0.6pt,line join=round] ( 28.50, 19.20) -- ( 28.50, 17.71);

\path[draw=drawColor,line width= 0.6pt,line join=round] ( 29.00, 17.81) -- ( 29.00, 17.71);

\path[draw=drawColor,line width= 0.6pt,line join=round] ( 29.50, 27.15) -- ( 29.50, 17.71);

\path[draw=drawColor,line width= 0.6pt,line join=round] ( 30.01, 18.46) -- ( 30.01, 17.71);

\path[draw=drawColor,line width= 0.6pt,line join=round] ( 30.51, 19.72) -- ( 30.51, 17.71);

\path[draw=drawColor,line width= 0.6pt,line join=round] ( 31.01, 19.90) -- ( 31.01, 17.71);

\path[draw=drawColor,line width= 0.6pt,line join=round] ( 31.51, 20.39) -- ( 31.51, 17.71);

\path[draw=drawColor,line width= 0.6pt,line join=round] ( 32.01, 19.33) -- ( 32.01, 17.71);

\path[draw=drawColor,line width= 0.6pt,line join=round] ( 32.52, 19.43) -- ( 32.52, 17.71);

\path[draw=drawColor,line width= 0.6pt,line join=round] ( 33.02, 21.86) -- ( 33.02, 17.71);

\path[draw=drawColor,line width= 0.6pt,line join=round] ( 33.52, 21.83) -- ( 33.52, 17.71);

\path[draw=drawColor,line width= 0.6pt,line join=round] ( 34.02, 18.79) -- ( 34.02, 17.71);

\path[draw=drawColor,line width= 0.6pt,line join=round] ( 34.52, 24.93) -- ( 34.52, 17.71);

\path[draw=drawColor,line width= 0.6pt,line join=round] ( 35.02, 22.47) -- ( 35.02, 17.71);

\path[draw=drawColor,line width= 0.6pt,line join=round] ( 35.53, 22.75) -- ( 35.53, 17.71);

\path[draw=drawColor,line width= 0.6pt,line join=round] ( 36.03, 19.20) -- ( 36.03, 17.71);

\path[draw=drawColor,line width= 0.6pt,line join=round] ( 36.53, 20.18) -- ( 36.53, 17.71);

\path[draw=drawColor,line width= 0.6pt,line join=round] ( 37.03, 20.02) -- ( 37.03, 17.71);

\path[draw=drawColor,line width= 0.6pt,line join=round] ( 37.53, 18.15) -- ( 37.53, 17.71);

\path[draw=drawColor,line width= 0.6pt,line join=round] ( 38.04, 18.20) -- ( 38.04, 17.71);

\path[draw=drawColor,line width= 0.6pt,line join=round] ( 38.54, 19.37) -- ( 38.54, 17.71);

\path[draw=drawColor,line width= 0.6pt,line join=round] ( 39.04, 19.15) -- ( 39.04, 17.71);

\path[draw=drawColor,line width= 0.6pt,line join=round] ( 39.54, 21.31) -- ( 39.54, 17.71);

\path[draw=drawColor,line width= 0.6pt,line join=round] ( 40.04, 23.04) -- ( 40.04, 17.71);

\path[draw=drawColor,line width= 0.6pt,line join=round] ( 40.55, 19.25) -- ( 40.55, 17.71);

\path[draw=drawColor,line width= 0.6pt,line join=round] ( 41.05, 18.44) -- ( 41.05, 17.71);

\path[draw=drawColor,line width= 0.6pt,line join=round] ( 41.55, 24.72) -- ( 41.55, 17.71);

\path[draw=drawColor,line width= 0.6pt,line join=round] ( 42.05, 21.71) -- ( 42.05, 17.71);

\path[draw=drawColor,line width= 0.6pt,line join=round] ( 42.55, 21.55) -- ( 42.55, 17.71);

\path[draw=drawColor,line width= 0.6pt,line join=round] ( 43.05, 20.96) -- ( 43.05, 17.71);

\path[draw=drawColor,line width= 0.6pt,line join=round] ( 43.56, 18.60) -- ( 43.56, 17.71);

\path[draw=drawColor,line width= 0.6pt,line join=round] ( 44.06, 20.69) -- ( 44.06, 17.71);

\path[draw=drawColor,line width= 0.6pt,line join=round] ( 44.56, 22.41) -- ( 44.56, 17.71);

\path[draw=drawColor,line width= 0.6pt,line join=round] ( 45.06, 18.52) -- ( 45.06, 17.71);

\path[draw=drawColor,line width= 0.6pt,line join=round] ( 45.56, 18.64) -- ( 45.56, 17.71);

\path[draw=drawColor,line width= 0.6pt,line join=round] ( 46.07, 18.86) -- ( 46.07, 17.71);

\path[draw=drawColor,line width= 0.6pt,line join=round] ( 46.57, 20.36) -- ( 46.57, 17.71);

\path[draw=drawColor,line width= 0.6pt,line join=round] ( 47.07, 23.85) -- ( 47.07, 17.71);

\path[draw=drawColor,line width= 0.6pt,line join=round] ( 47.57, 20.17) -- ( 47.57, 17.71);

\path[draw=drawColor,line width= 0.6pt,line join=round] ( 48.07, 18.06) -- ( 48.07, 17.71);

\path[draw=drawColor,line width= 0.6pt,line join=round] ( 48.58, 18.88) -- ( 48.58, 17.71);

\path[draw=drawColor,line width= 0.6pt,line join=round] ( 49.08, 21.43) -- ( 49.08, 17.71);

\path[draw=drawColor,line width= 0.6pt,line join=round] ( 49.58, 19.44) -- ( 49.58, 17.71);

\path[draw=drawColor,line width= 0.6pt,line join=round] ( 50.08, 18.96) -- ( 50.08, 17.71);

\path[draw=drawColor,line width= 0.6pt,line join=round] ( 50.58, 19.13) -- ( 50.58, 17.71);

\path[draw=drawColor,line width= 0.6pt,line join=round] ( 51.09, 21.24) -- ( 51.09, 17.71);

\path[draw=drawColor,line width= 0.6pt,line join=round] ( 51.59, 19.88) -- ( 51.59, 17.71);

\path[draw=drawColor,line width= 0.6pt,line join=round] ( 52.09, 18.61) -- ( 52.09, 17.71);

\path[draw=drawColor,line width= 0.6pt,line join=round] ( 52.59, 25.46) -- ( 52.59, 17.71);

\path[draw=drawColor,line width= 0.6pt,line join=round] ( 53.09, 18.45) -- ( 53.09, 17.71);

\path[draw=drawColor,line width= 0.6pt,line join=round] ( 53.59, 18.61) -- ( 53.59, 17.71);

\path[draw=drawColor,line width= 0.6pt,line join=round] ( 54.10, 24.85) -- ( 54.10, 17.71);

\path[draw=drawColor,line width= 0.6pt,line join=round] ( 54.60, 20.65) -- ( 54.60, 17.71);

\path[draw=drawColor,line width= 0.6pt,line join=round] ( 55.10, 18.01) -- ( 55.10, 17.71);

\path[draw=drawColor,line width= 0.6pt,line join=round] ( 55.60, 20.99) -- ( 55.60, 17.71);

\path[draw=drawColor,line width= 0.6pt,line join=round] ( 56.10, 18.67) -- ( 56.10, 17.71);

\path[draw=drawColor,line width= 0.6pt,line join=round] ( 56.61, 19.66) -- ( 56.61, 17.71);

\path[draw=drawColor,line width= 0.6pt,line join=round] ( 57.11, 21.42) -- ( 57.11, 17.71);

\path[draw=drawColor,line width= 0.6pt,line join=round] ( 57.61, 26.02) -- ( 57.61, 17.71);

\path[draw=drawColor,line width= 0.6pt,line join=round] ( 58.11, 19.84) -- ( 58.11, 17.71);

\path[draw=drawColor,line width= 0.6pt,line join=round] ( 58.61, 19.20) -- ( 58.61, 17.71);

\path[draw=drawColor,line width= 0.6pt,line join=round] ( 59.12, 22.31) -- ( 59.12, 17.71);

\path[draw=drawColor,line width= 0.6pt,line join=round] ( 59.62, 20.74) -- ( 59.62, 17.71);

\path[draw=drawColor,line width= 0.6pt,line join=round] ( 60.12, 21.47) -- ( 60.12, 17.71);

\path[draw=drawColor,line width= 0.6pt,line join=round] ( 60.62, 19.70) -- ( 60.62, 17.71);

\path[draw=drawColor,line width= 0.6pt,line join=round] ( 61.12, 18.82) -- ( 61.12, 17.71);

\path[draw=drawColor,line width= 0.6pt,line join=round] ( 61.62, 24.92) -- ( 61.62, 17.71);

\path[draw=drawColor,line width= 0.6pt,line join=round] ( 62.13, 26.50) -- ( 62.13, 17.71);

\path[draw=drawColor,line width= 0.6pt,line join=round] ( 62.63, 19.14) -- ( 62.63, 17.71);

\path[draw=drawColor,line width= 0.6pt,line join=round] ( 63.13, 18.04) -- ( 63.13, 17.71);

\path[draw=drawColor,line width= 0.6pt,line join=round] ( 63.63, 24.90) -- ( 63.63, 17.71);

\path[draw=drawColor,line width= 0.6pt,line join=round] ( 64.13, 18.79) -- ( 64.13, 17.71);

\path[draw=drawColor,line width= 0.6pt,line join=round] ( 64.64, 27.98) -- ( 64.64, 17.71);

\path[draw=drawColor,line width= 0.6pt,line join=round] ( 65.14, 20.85) -- ( 65.14, 17.71);

\path[draw=drawColor,line width= 0.6pt,line join=round] ( 65.64, 18.34) -- ( 65.64, 17.71);

\path[draw=drawColor,line width= 0.6pt,line join=round] ( 66.14, 18.37) -- ( 66.14, 17.71);

\path[draw=drawColor,line width= 0.6pt,line join=round] ( 66.64, 22.84) -- ( 66.64, 17.71);

\path[draw=drawColor,line width= 0.6pt,line join=round] ( 67.15, 19.55) -- ( 67.15, 17.71);

\path[draw=drawColor,line width= 0.6pt,line join=round] ( 67.65, 20.86) -- ( 67.65, 17.71);

\path[draw=drawColor,line width= 0.6pt,line join=round] ( 68.15, 19.85) -- ( 68.15, 17.71);

\path[draw=drawColor,line width= 0.6pt,line join=round] ( 68.65, 21.24) -- ( 68.65, 17.71);

\path[draw=drawColor,line width= 0.6pt,line join=round] ( 69.15, 22.05) -- ( 69.15, 17.71);

\path[draw=drawColor,line width= 0.6pt,line join=round] ( 69.65, 20.20) -- ( 69.65, 17.71);

\path[draw=drawColor,line width= 0.6pt,line join=round] ( 70.16, 21.27) -- ( 70.16, 17.71);

\path[draw=drawColor,line width= 0.6pt,line join=round] ( 70.66, 18.93) -- ( 70.66, 17.71);

\path[draw=drawColor,line width= 0.6pt,line join=round] ( 71.16, 21.48) -- ( 71.16, 17.71);

\path[draw=drawColor,line width= 0.6pt,line join=round] ( 71.66, 18.46) -- ( 71.66, 17.71);

\path[draw=drawColor,line width= 0.6pt,line join=round] ( 72.16, 19.60) -- ( 72.16, 17.71);

\path[draw=drawColor,line width= 0.6pt,line join=round] ( 72.67, 27.62) -- ( 72.67, 17.71);

\path[draw=drawColor,line width= 0.6pt,line join=round] ( 73.17, 23.02) -- ( 73.17, 17.71);

\path[draw=drawColor,line width= 0.6pt,line join=round] ( 73.67, 19.53) -- ( 73.67, 17.71);

\path[draw=drawColor,line width= 0.6pt,line join=round] ( 74.17, 18.63) -- ( 74.17, 17.71);

\path[draw=drawColor,line width= 0.6pt,line join=round] ( 74.67, 19.90) -- ( 74.67, 17.71);

\path[draw=drawColor,line width= 0.6pt,line join=round] ( 75.18, 18.56) -- ( 75.18, 17.71);

\path[draw=drawColor,line width= 0.6pt,line join=round] ( 75.68, 23.49) -- ( 75.68, 17.71);

\path[draw=drawColor,line width= 0.6pt,line join=round] ( 76.18, 22.62) -- ( 76.18, 17.71);

\path[draw=drawColor,line width= 0.6pt,line join=round] ( 76.68, 20.64) -- ( 76.68, 17.71);

\path[draw=drawColor,line width= 0.6pt,line join=round] ( 77.18, 22.71) -- ( 77.18, 17.71);

\path[draw=drawColor,line width= 0.6pt,line join=round] ( 77.69, 19.80) -- ( 77.69, 17.71);

\path[draw=drawColor,line width= 0.6pt,line join=round] ( 78.19, 19.21) -- ( 78.19, 17.71);

\path[draw=drawColor,line width= 0.6pt,line join=round] ( 78.69, 22.59) -- ( 78.69, 17.71);

\path[draw=drawColor,line width= 0.6pt,line join=round] ( 79.19, 22.89) -- ( 79.19, 17.71);

\path[draw=drawColor,line width= 0.6pt,line join=round] ( 79.69, 19.67) -- ( 79.69, 17.71);

\path[draw=drawColor,line width= 0.6pt,line join=round] ( 80.19, 29.59) -- ( 80.19, 17.71);

\path[draw=drawColor,line width= 0.6pt,line join=round] ( 80.70, 23.02) -- ( 80.70, 17.71);

\path[draw=drawColor,line width= 0.6pt,line join=round] ( 81.20, 19.73) -- ( 81.20, 17.71);

\path[draw=drawColor,line width= 0.6pt,line join=round] ( 81.70, 22.24) -- ( 81.70, 17.71);

\path[draw=drawColor,line width= 0.6pt,line join=round] ( 82.20, 25.40) -- ( 82.20, 17.71);

\path[draw=drawColor,line width= 0.6pt,line join=round] ( 82.70, 20.82) -- ( 82.70, 17.71);

\path[draw=drawColor,line width= 0.6pt,line join=round] ( 83.21, 19.22) -- ( 83.21, 17.71);

\path[draw=drawColor,line width= 0.6pt,line join=round] ( 83.71, 22.47) -- ( 83.71, 17.71);

\path[draw=drawColor,line width= 0.6pt,line join=round] ( 84.21, 18.37) -- ( 84.21, 17.71);

\path[draw=drawColor,line width= 0.6pt,line join=round] ( 84.71, 19.00) -- ( 84.71, 17.71);

\path[draw=drawColor,line width= 0.6pt,line join=round] ( 85.21, 20.32) -- ( 85.21, 17.71);

\path[draw=drawColor,line width= 0.6pt,line join=round] ( 85.72, 19.66) -- ( 85.72, 17.71);

\path[draw=drawColor,line width= 0.6pt,line join=round] ( 86.22, 21.95) -- ( 86.22, 17.71);

\path[draw=drawColor,line width= 0.6pt,line join=round] ( 86.72, 22.72) -- ( 86.72, 17.71);

\path[draw=drawColor,line width= 0.6pt,line join=round] ( 87.22, 19.45) -- ( 87.22, 17.71);

\path[draw=drawColor,line width= 0.6pt,line join=round] ( 87.72, 18.86) -- ( 87.72, 17.71);

\path[draw=drawColor,line width= 0.6pt,line join=round] ( 88.22, 19.58) -- ( 88.22, 17.71);

\path[draw=drawColor,line width= 0.6pt,line join=round] ( 88.73, 20.23) -- ( 88.73, 17.71);

\path[draw=drawColor,line width= 0.6pt,line join=round] ( 89.23, 19.07) -- ( 89.23, 17.71);

\path[draw=drawColor,line width= 0.6pt,line join=round] ( 89.73, 18.65) -- ( 89.73, 17.71);

\path[draw=drawColor,line width= 0.6pt,line join=round] ( 90.23, 18.92) -- ( 90.23, 17.71);

\path[draw=drawColor,line width= 0.6pt,line join=round] ( 90.73, 20.97) -- ( 90.73, 17.71);

\path[draw=drawColor,line width= 0.6pt,line join=round] ( 91.24, 26.45) -- ( 91.24, 17.71);

\path[draw=drawColor,line width= 0.6pt,line join=round] ( 91.74, 26.20) -- ( 91.74, 17.71);

\path[draw=drawColor,line width= 0.6pt,line join=round] ( 92.24, 23.13) -- ( 92.24, 17.71);

\path[draw=drawColor,line width= 0.6pt,line join=round] ( 92.74, 28.20) -- ( 92.74, 17.71);

\path[draw=drawColor,line width= 0.6pt,line join=round] ( 93.24, 43.35) -- ( 93.24, 17.71);

\path[draw=drawColor,line width= 0.6pt,line join=round] ( 93.75, 40.88) -- ( 93.75, 17.71);

\path[draw=drawColor,line width= 0.6pt,line join=round] ( 94.25, 47.08) -- ( 94.25, 17.71);

\path[draw=drawColor,line width= 0.6pt,line join=round] ( 94.75, 36.08) -- ( 94.75, 17.71);

\path[draw=drawColor,line width= 0.6pt,line join=round] ( 95.25, 29.00) -- ( 95.25, 17.71);

\path[draw=drawColor,line width= 0.6pt,line join=round] ( 95.75, 56.15) -- ( 95.75, 17.71);

\path[draw=drawColor,line width= 0.6pt,line join=round] ( 96.25, 78.82) -- ( 96.25, 17.71);

\path[draw=drawColor,line width= 0.6pt,line join=round] ( 96.76, 75.22) -- ( 96.76, 17.71);

\path[draw=drawColor,line width= 0.6pt,line join=round] ( 97.26, 75.75) -- ( 97.26, 17.71);

\path[draw=drawColor,line width= 0.6pt,line join=round] ( 97.76, 91.81) -- ( 97.76, 17.71);

\path[draw=drawColor,line width= 0.6pt,line join=round] ( 98.26, 38.57) -- ( 98.26, 17.71);

\path[draw=drawColor,line width= 0.6pt,line join=round] ( 98.76, 45.85) -- ( 98.76, 17.71);

\path[draw=drawColor,line width= 0.6pt,line join=round] ( 99.27, 47.12) -- ( 99.27, 17.71);
\end{scope}
\begin{scope}
\path[clip] (  0.00,  0.00) rectangle (108.41, 93.95);
\definecolor{drawColor}{gray}{0.10}

\node[text=drawColor,anchor=base east,inner sep=0pt, outer sep=0pt, scale=  0.55] at ( 22.10, 15.81) {0.00};

\node[text=drawColor,anchor=base east,inner sep=0pt, outer sep=0pt, scale=  0.55] at ( 22.10, 29.29) {0.01};

\node[text=drawColor,anchor=base east,inner sep=0pt, outer sep=0pt, scale=  0.55] at ( 22.10, 42.76) {0.02};

\node[text=drawColor,anchor=base east,inner sep=0pt, outer sep=0pt, scale=  0.55] at ( 22.10, 56.24) {0.03};

\node[text=drawColor,anchor=base east,inner sep=0pt, outer sep=0pt, scale=  0.55] at ( 22.10, 69.71) {0.04};

\node[text=drawColor,anchor=base east,inner sep=0pt, outer sep=0pt, scale=  0.55] at ( 22.10, 83.19) {0.05};
\end{scope}
\begin{scope}
\path[clip] (  0.00,  0.00) rectangle (108.41, 93.95);
\definecolor{drawColor}{gray}{0.10}

\node[text=drawColor,anchor=base,inner sep=0pt, outer sep=0pt, scale=  0.55] at ( 25.99, 10.60) {0};

\node[text=drawColor,anchor=base,inner sep=0pt, outer sep=0pt, scale=  0.55] at ( 51.09, 10.60) {50};

\node[text=drawColor,anchor=base,inner sep=0pt, outer sep=0pt, scale=  0.55] at ( 76.18, 10.60) {100};
\end{scope}
\begin{scope}
\path[clip] (  0.00,  0.00) rectangle (108.41, 93.95);
\definecolor{drawColor}{gray}{0.10}

\node[text=drawColor,rotate= 90.00,anchor=base,inner sep=0pt, outer sep=0pt, scale=  0.66] at ( 10.05, 51.40) {algebraic residual};
\end{scope}
\end{tikzpicture}

%% file: Rfigures/stemc_b.tex
\begin{tikzpicture}[x=1pt,y=1pt]
\clip (5,3) rectangle (104, 93.95);
\definecolor{fillColor}{RGB}{255,255,255}
\path[use as bounding box,fill=fillColor,fill opacity=0.00] (0,0) rectangle (108.41, 93.95);
\begin{scope}
\path[clip] (  0.00,  0.00) rectangle (108.40, 93.95);
\definecolor{drawColor}{RGB}{255,255,255}
\definecolor{fillColor}{RGB}{255,255,255}

\path[draw=drawColor,line width= 0.6pt,line join=round,line cap=round,fill=fillColor] (  0.00,  0.00) rectangle (108.41, 93.95);
\end{scope}
\begin{scope}
\path[clip] ( 18.58, 18.89) rectangle (102.90, 88.45);
\definecolor{fillColor}{gray}{0.92}

\path[fill=fillColor] ( 18.58, 18.89) rectangle (102.91, 88.45);
\definecolor{drawColor}{RGB}{0,0,0}

\path[draw=drawColor,draw opacity=0.50,line width= 0.6pt,line join=round] ( 18.58, 22.05) -- (102.90, 22.05);
\definecolor{drawColor}{RGB}{70,130,180}

\path[draw=drawColor,line width= 0.6pt,line join=round] ( 22.41, 22.07) -- ( 22.41, 22.05);

\path[draw=drawColor,line width= 0.6pt,line join=round] ( 23.88, 22.08) -- ( 23.88, 22.05);

\path[draw=drawColor,line width= 0.6pt,line join=round] ( 25.36, 22.09) -- ( 25.36, 22.05);

\path[draw=drawColor,line width= 0.6pt,line join=round] ( 26.83, 22.09) -- ( 26.83, 22.05);

\path[draw=drawColor,line width= 0.6pt,line join=round] ( 28.31, 22.10) -- ( 28.31, 22.05);

\path[draw=drawColor,line width= 0.6pt,line join=round] ( 29.78, 22.12) -- ( 29.78, 22.05);

\path[draw=drawColor,line width= 0.6pt,line join=round] ( 31.26, 22.19) -- ( 31.26, 22.05);

\path[draw=drawColor,line width= 0.6pt,line join=round] ( 32.73, 22.21) -- ( 32.73, 22.05);

\path[draw=drawColor,line width= 0.6pt,line join=round] ( 34.20, 22.25) -- ( 34.20, 22.05);

\path[draw=drawColor,line width= 0.6pt,line join=round] ( 35.68, 22.27) -- ( 35.68, 22.05);

\path[draw=drawColor,line width= 0.6pt,line join=round] ( 37.15, 22.27) -- ( 37.15, 22.05);

\path[draw=drawColor,line width= 0.6pt,line join=round] ( 38.63, 22.30) -- ( 38.63, 22.05);

\path[draw=drawColor,line width= 0.6pt,line join=round] ( 40.10, 22.36) -- ( 40.10, 22.05);

\path[draw=drawColor,line width= 0.6pt,line join=round] ( 41.58, 22.36) -- ( 41.58, 22.05);

\path[draw=drawColor,line width= 0.6pt,line join=round] ( 43.05, 22.38) -- ( 43.05, 22.05);

\path[draw=drawColor,line width= 0.6pt,line join=round] ( 44.52, 22.44) -- ( 44.52, 22.05);

\path[draw=drawColor,line width= 0.6pt,line join=round] ( 46.00, 22.46) -- ( 46.00, 22.05);

\path[draw=drawColor,line width= 0.6pt,line join=round] ( 47.47, 22.53) -- ( 47.47, 22.05);

\path[draw=drawColor,line width= 0.6pt,line join=round] ( 48.95, 22.55) -- ( 48.95, 22.05);

\path[draw=drawColor,line width= 0.6pt,line join=round] ( 50.42, 22.56) -- ( 50.42, 22.05);

\path[draw=drawColor,line width= 0.6pt,line join=round] ( 51.90, 22.59) -- ( 51.90, 22.05);

\path[draw=drawColor,line width= 0.6pt,line join=round] ( 53.37, 22.61) -- ( 53.37, 22.05);

\path[draw=drawColor,line width= 0.6pt,line join=round] ( 54.84, 22.69) -- ( 54.84, 22.05);

\path[draw=drawColor,line width= 0.6pt,line join=round] ( 56.32, 22.80) -- ( 56.32, 22.05);

\path[draw=drawColor,line width= 0.6pt,line join=round] ( 57.79, 22.81) -- ( 57.79, 22.05);

\path[draw=drawColor,line width= 0.6pt,line join=round] ( 59.27, 22.86) -- ( 59.27, 22.05);

\path[draw=drawColor,line width= 0.6pt,line join=round] ( 60.74, 22.95) -- ( 60.74, 22.05);

\path[draw=drawColor,line width= 0.6pt,line join=round] ( 62.22, 27.38) -- ( 62.22, 22.05);

\path[draw=drawColor,line width= 0.6pt,line join=round] ( 63.69, 27.76) -- ( 63.69, 22.05);

\path[draw=drawColor,line width= 0.6pt,line join=round] ( 65.16, 31.06) -- ( 65.16, 22.05);

\path[draw=drawColor,line width= 0.6pt,line join=round] ( 66.64, 31.35) -- ( 66.64, 22.05);

\path[draw=drawColor,line width= 0.6pt,line join=round] ( 68.11, 31.95) -- ( 68.11, 22.05);

\path[draw=drawColor,line width= 0.6pt,line join=round] ( 69.59, 34.78) -- ( 69.59, 22.05);

\path[draw=drawColor,line width= 0.6pt,line join=round] ( 71.06, 40.56) -- ( 71.06, 22.05);

\path[draw=drawColor,line width= 0.6pt,line join=round] ( 72.54, 42.06) -- ( 72.54, 22.05);

\path[draw=drawColor,line width= 0.6pt,line join=round] ( 74.01, 47.59) -- ( 74.01, 22.05);

\path[draw=drawColor,line width= 0.6pt,line join=round] ( 75.48, 55.93) -- ( 75.48, 22.05);

\path[draw=drawColor,line width= 0.6pt,line join=round] ( 76.96, 59.42) -- ( 76.96, 22.05);

\path[draw=drawColor,line width= 0.6pt,line join=round] ( 78.43, 59.55) -- ( 78.43, 22.05);

\path[draw=drawColor,line width= 0.6pt,line join=round] ( 79.91, 63.85) -- ( 79.91, 22.05);

\path[draw=drawColor,line width= 0.6pt,line join=round] ( 81.38, 74.07) -- ( 81.38, 22.05);

\path[draw=drawColor,line width= 0.6pt,line join=round] ( 82.86, 74.44) -- ( 82.86, 22.05);

\path[draw=drawColor,line width= 0.6pt,line join=round] ( 84.33, 89.22) -- ( 84.33, 22.05);

\path[draw=drawColor,line width= 0.6pt,line join=round] ( 85.80, 93.95) -- ( 85.80, 22.05);

\path[draw=drawColor,line width= 0.6pt,line join=round] ( 87.28, 93.95) -- ( 87.28, 22.05);

\path[draw=drawColor,line width= 0.6pt,line join=round] ( 88.75, 93.95) -- ( 88.75, 22.05);

\path[draw=drawColor,line width= 0.6pt,line join=round] ( 90.23, 93.95) -- ( 90.23, 22.05);

\path[draw=drawColor,line width= 0.6pt,line join=round] ( 91.70, 93.95) -- ( 91.70, 22.05);

\path[draw=drawColor,line width= 0.6pt,line join=round] ( 93.17, 93.95) -- ( 93.17, 22.05);

\path[draw=drawColor,line width= 0.6pt,line join=round] ( 94.65, 93.95) -- ( 94.65, 22.05);

\path[draw=drawColor,line width= 0.6pt,line join=round] ( 96.12, 93.95) -- ( 96.12, 22.05);

\path[draw=drawColor,line width= 0.6pt,line join=round] ( 97.60, 93.95) -- ( 97.60, 22.05);

\path[draw=drawColor,line width= 0.6pt,line join=round] ( 99.07, 93.95) -- ( 99.07, 22.05);
\end{scope}
\begin{scope}
\path[clip] (  0.00,  0.00) rectangle (108.41, 93.95);
\definecolor{drawColor}{gray}{0.10}

\node[text=drawColor,anchor=base east,inner sep=0pt, outer sep=0pt, scale=  0.55] at ( 17.83, 35.96) {10};

\node[text=drawColor,anchor=base east,inner sep=0pt, outer sep=0pt, scale=  0.55] at ( 17.83, 51.77) {20};

\node[text=drawColor,anchor=base east,inner sep=0pt, outer sep=0pt, scale=  0.55] at ( 17.83, 67.58) {30};

\node[text=drawColor,anchor=base east,inner sep=0pt, outer sep=0pt, scale=  0.55] at ( 17.83, 83.39) {40};
\end{scope}
\begin{scope}
\path[clip] (  0.00,  0.00) rectangle (108.41, 93.95);
\definecolor{drawColor}{gray}{0.10}

\node[text=drawColor,anchor=base,inner sep=0pt, outer sep=0pt, scale=  0.55] at ( 20.94, 15.15) {0};

\node[text=drawColor,anchor=base,inner sep=0pt, outer sep=0pt, scale=  0.55] at ( 50.42, 15.15) {20};

\node[text=drawColor,anchor=base,inner sep=0pt, outer sep=0pt, scale=  0.55] at ( 79.91, 15.15) {40};
\end{scope}
\begin{scope}
\path[clip] (  0.00,  0.00) rectangle (108.41, 93.95);
\definecolor{drawColor}{gray}{0.10}

\node[text=drawColor,anchor=base,inner sep=0pt, outer sep=0pt, scale=  0.66] at ( 60.74,  6.78) {estimated inliers};
\end{scope}
\begin{scope}
\path[clip] (  0.00,  0.00) rectangle (108.41, 93.95);
\definecolor{drawColor}{gray}{0.10}

\node[text=drawColor,rotate= 90.00,anchor=base,inner sep=0pt, outer sep=0pt, scale=  0.66] at ( 10.05, 53.67) {reprojection error (pixel)};
\end{scope}
\end{tikzpicture}

%% file: Rfigures/stemd_b.tex
\begin{tikzpicture}[x=1pt,y=1pt]
\clip (5,3) rectangle (104, 93.95);
\definecolor{fillColor}{RGB}{255,255,255}
\path[use as bounding box,fill=fillColor,fill opacity=0.00] (0,0) rectangle (108.41, 93.95);
\begin{scope}
\path[clip] (  0.00,  0.00) rectangle (108.40, 93.95);
\definecolor{drawColor}{RGB}{255,255,255}
\definecolor{fillColor}{RGB}{255,255,255}

\path[draw=drawColor,line width= 0.6pt,line join=round,line cap=round,fill=fillColor] (  0.00,  0.00) rectangle (108.41, 93.95);
\end{scope}
\begin{scope}
\path[clip] ( 18.58, 18.89) rectangle (102.90, 88.45);
\definecolor{fillColor}{gray}{0.92}

\path[fill=fillColor] ( 18.58, 18.89) rectangle (102.91, 88.45);
\definecolor{drawColor}{RGB}{0,0,0}

\path[draw=drawColor,draw opacity=0.50,line width= 0.6pt,line join=round] ( 18.58, 22.05) -- (102.90, 22.05);
\definecolor{drawColor}{RGB}{70,130,180}

\path[draw=drawColor,line width= 0.6pt,line join=round] ( 22.41, 22.05) -- ( 22.41, 22.05);

\path[draw=drawColor,line width= 0.6pt,line join=round] ( 22.59, 22.05) -- ( 22.59, 22.05);

\path[draw=drawColor,line width= 0.6pt,line join=round] ( 22.77, 22.05) -- ( 22.77, 22.05);

\path[draw=drawColor,line width= 0.6pt,line join=round] ( 22.95, 22.05) -- ( 22.95, 22.05);

\path[draw=drawColor,line width= 0.6pt,line join=round] ( 23.13, 22.05) -- ( 23.13, 22.05);

\path[draw=drawColor,line width= 0.6pt,line join=round] ( 23.31, 22.05) -- ( 23.31, 22.05);

\path[draw=drawColor,line width= 0.6pt,line join=round] ( 23.50, 22.05) -- ( 23.50, 22.05);

\path[draw=drawColor,line width= 0.6pt,line join=round] ( 23.68, 22.05) -- ( 23.68, 22.05);

\path[draw=drawColor,line width= 0.6pt,line join=round] ( 23.86, 22.05) -- ( 23.86, 22.05);

\path[draw=drawColor,line width= 0.6pt,line join=round] ( 24.04, 22.05) -- ( 24.04, 22.05);

\path[draw=drawColor,line width= 0.6pt,line join=round] ( 24.22, 22.05) -- ( 24.22, 22.05);

\path[draw=drawColor,line width= 0.6pt,line join=round] ( 24.40, 22.05) -- ( 24.40, 22.05);

\path[draw=drawColor,line width= 0.6pt,line join=round] ( 24.58, 22.05) -- ( 24.58, 22.05);

\path[draw=drawColor,line width= 0.6pt,line join=round] ( 24.76, 22.05) -- ( 24.76, 22.05);

\path[draw=drawColor,line width= 0.6pt,line join=round] ( 24.94, 22.05) -- ( 24.94, 22.05);

\path[draw=drawColor,line width= 0.6pt,line join=round] ( 25.12, 22.05) -- ( 25.12, 22.05);

\path[draw=drawColor,line width= 0.6pt,line join=round] ( 25.30, 22.05) -- ( 25.30, 22.05);

\path[draw=drawColor,line width= 0.6pt,line join=round] ( 25.48, 22.05) -- ( 25.48, 22.05);

\path[draw=drawColor,line width= 0.6pt,line join=round] ( 25.66, 22.05) -- ( 25.66, 22.05);

\path[draw=drawColor,line width= 0.6pt,line join=round] ( 25.85, 22.05) -- ( 25.85, 22.05);

\path[draw=drawColor,line width= 0.6pt,line join=round] ( 26.03, 22.05) -- ( 26.03, 22.05);

\path[draw=drawColor,line width= 0.6pt,line join=round] ( 26.21, 22.05) -- ( 26.21, 22.05);

\path[draw=drawColor,line width= 0.6pt,line join=round] ( 26.39, 22.05) -- ( 26.39, 22.05);

\path[draw=drawColor,line width= 0.6pt,line join=round] ( 26.57, 22.05) -- ( 26.57, 22.05);

\path[draw=drawColor,line width= 0.6pt,line join=round] ( 26.75, 22.05) -- ( 26.75, 22.05);

\path[draw=drawColor,line width= 0.6pt,line join=round] ( 26.93, 22.05) -- ( 26.93, 22.05);

\path[draw=drawColor,line width= 0.6pt,line join=round] ( 27.11, 22.05) -- ( 27.11, 22.05);

\path[draw=drawColor,line width= 0.6pt,line join=round] ( 27.29, 22.05) -- ( 27.29, 22.05);

\path[draw=drawColor,line width= 0.6pt,line join=round] ( 27.47, 22.05) -- ( 27.47, 22.05);

\path[draw=drawColor,line width= 0.6pt,line join=round] ( 27.65, 22.05) -- ( 27.65, 22.05);

\path[draw=drawColor,line width= 0.6pt,line join=round] ( 27.83, 22.05) -- ( 27.83, 22.05);

\path[draw=drawColor,line width= 0.6pt,line join=round] ( 28.02, 22.05) -- ( 28.02, 22.05);

\path[draw=drawColor,line width= 0.6pt,line join=round] ( 28.20, 22.05) -- ( 28.20, 22.05);

\path[draw=drawColor,line width= 0.6pt,line join=round] ( 28.38, 22.05) -- ( 28.38, 22.05);

\path[draw=drawColor,line width= 0.6pt,line join=round] ( 28.56, 22.05) -- ( 28.56, 22.05);

\path[draw=drawColor,line width= 0.6pt,line join=round] ( 28.74, 22.05) -- ( 28.74, 22.05);

\path[draw=drawColor,line width= 0.6pt,line join=round] ( 28.92, 22.05) -- ( 28.92, 22.05);

\path[draw=drawColor,line width= 0.6pt,line join=round] ( 29.10, 22.05) -- ( 29.10, 22.05);

\path[draw=drawColor,line width= 0.6pt,line join=round] ( 29.28, 22.05) -- ( 29.28, 22.05);

\path[draw=drawColor,line width= 0.6pt,line join=round] ( 29.46, 22.06) -- ( 29.46, 22.05);

\path[draw=drawColor,line width= 0.6pt,line join=round] ( 29.64, 22.06) -- ( 29.64, 22.05);

\path[draw=drawColor,line width= 0.6pt,line join=round] ( 29.82, 22.06) -- ( 29.82, 22.05);

\path[draw=drawColor,line width= 0.6pt,line join=round] ( 30.00, 22.06) -- ( 30.00, 22.05);

\path[draw=drawColor,line width= 0.6pt,line join=round] ( 30.18, 22.06) -- ( 30.18, 22.05);

\path[draw=drawColor,line width= 0.6pt,line join=round] ( 30.37, 22.06) -- ( 30.37, 22.05);

\path[draw=drawColor,line width= 0.6pt,line join=round] ( 30.55, 22.06) -- ( 30.55, 22.05);

\path[draw=drawColor,line width= 0.6pt,line join=round] ( 30.73, 22.06) -- ( 30.73, 22.05);

\path[draw=drawColor,line width= 0.6pt,line join=round] ( 30.91, 22.06) -- ( 30.91, 22.05);

\path[draw=drawColor,line width= 0.6pt,line join=round] ( 31.09, 22.06) -- ( 31.09, 22.05);

\path[draw=drawColor,line width= 0.6pt,line join=round] ( 31.27, 22.06) -- ( 31.27, 22.05);

\path[draw=drawColor,line width= 0.6pt,line join=round] ( 31.45, 22.06) -- ( 31.45, 22.05);

\path[draw=drawColor,line width= 0.6pt,line join=round] ( 31.63, 22.06) -- ( 31.63, 22.05);

\path[draw=drawColor,line width= 0.6pt,line join=round] ( 31.81, 22.06) -- ( 31.81, 22.05);

\path[draw=drawColor,line width= 0.6pt,line join=round] ( 31.99, 22.07) -- ( 31.99, 22.05);

\path[draw=drawColor,line width= 0.6pt,line join=round] ( 32.17, 22.07) -- ( 32.17, 22.05);

\path[draw=drawColor,line width= 0.6pt,line join=round] ( 32.35, 22.07) -- ( 32.35, 22.05);

\path[draw=drawColor,line width= 0.6pt,line join=round] ( 32.54, 22.07) -- ( 32.54, 22.05);

\path[draw=drawColor,line width= 0.6pt,line join=round] ( 32.72, 22.07) -- ( 32.72, 22.05);

\path[draw=drawColor,line width= 0.6pt,line join=round] ( 32.90, 22.07) -- ( 32.90, 22.05);

\path[draw=drawColor,line width= 0.6pt,line join=round] ( 33.08, 22.07) -- ( 33.08, 22.05);

\path[draw=drawColor,line width= 0.6pt,line join=round] ( 33.26, 22.07) -- ( 33.26, 22.05);

\path[draw=drawColor,line width= 0.6pt,line join=round] ( 33.44, 22.07) -- ( 33.44, 22.05);

\path[draw=drawColor,line width= 0.6pt,line join=round] ( 33.62, 22.07) -- ( 33.62, 22.05);

\path[draw=drawColor,line width= 0.6pt,line join=round] ( 33.80, 22.08) -- ( 33.80, 22.05);

\path[draw=drawColor,line width= 0.6pt,line join=round] ( 33.98, 22.08) -- ( 33.98, 22.05);

\path[draw=drawColor,line width= 0.6pt,line join=round] ( 34.16, 22.08) -- ( 34.16, 22.05);

\path[draw=drawColor,line width= 0.6pt,line join=round] ( 34.34, 22.08) -- ( 34.34, 22.05);

\path[draw=drawColor,line width= 0.6pt,line join=round] ( 34.52, 22.08) -- ( 34.52, 22.05);

\path[draw=drawColor,line width= 0.6pt,line join=round] ( 34.71, 22.08) -- ( 34.71, 22.05);

\path[draw=drawColor,line width= 0.6pt,line join=round] ( 34.89, 22.08) -- ( 34.89, 22.05);

\path[draw=drawColor,line width= 0.6pt,line join=round] ( 35.07, 22.08) -- ( 35.07, 22.05);

\path[draw=drawColor,line width= 0.6pt,line join=round] ( 35.25, 22.08) -- ( 35.25, 22.05);

\path[draw=drawColor,line width= 0.6pt,line join=round] ( 35.43, 22.08) -- ( 35.43, 22.05);

\path[draw=drawColor,line width= 0.6pt,line join=round] ( 35.61, 22.08) -- ( 35.61, 22.05);

\path[draw=drawColor,line width= 0.6pt,line join=round] ( 35.79, 22.09) -- ( 35.79, 22.05);

\path[draw=drawColor,line width= 0.6pt,line join=round] ( 35.97, 22.09) -- ( 35.97, 22.05);

\path[draw=drawColor,line width= 0.6pt,line join=round] ( 36.15, 22.09) -- ( 36.15, 22.05);

\path[draw=drawColor,line width= 0.6pt,line join=round] ( 36.33, 22.09) -- ( 36.33, 22.05);

\path[draw=drawColor,line width= 0.6pt,line join=round] ( 36.51, 22.09) -- ( 36.51, 22.05);

\path[draw=drawColor,line width= 0.6pt,line join=round] ( 36.69, 22.09) -- ( 36.69, 22.05);

\path[draw=drawColor,line width= 0.6pt,line join=round] ( 36.87, 22.09) -- ( 36.87, 22.05);

\path[draw=drawColor,line width= 0.6pt,line join=round] ( 37.06, 22.09) -- ( 37.06, 22.05);

\path[draw=drawColor,line width= 0.6pt,line join=round] ( 37.24, 22.09) -- ( 37.24, 22.05);

\path[draw=drawColor,line width= 0.6pt,line join=round] ( 37.42, 22.09) -- ( 37.42, 22.05);

\path[draw=drawColor,line width= 0.6pt,line join=round] ( 37.60, 22.10) -- ( 37.60, 22.05);

\path[draw=drawColor,line width= 0.6pt,line join=round] ( 37.78, 22.10) -- ( 37.78, 22.05);

\path[draw=drawColor,line width= 0.6pt,line join=round] ( 37.96, 22.10) -- ( 37.96, 22.05);

\path[draw=drawColor,line width= 0.6pt,line join=round] ( 38.14, 22.10) -- ( 38.14, 22.05);

\path[draw=drawColor,line width= 0.6pt,line join=round] ( 38.32, 22.10) -- ( 38.32, 22.05);

\path[draw=drawColor,line width= 0.6pt,line join=round] ( 38.50, 22.10) -- ( 38.50, 22.05);

\path[draw=drawColor,line width= 0.6pt,line join=round] ( 38.68, 22.10) -- ( 38.68, 22.05);

\path[draw=drawColor,line width= 0.6pt,line join=round] ( 38.86, 22.11) -- ( 38.86, 22.05);

\path[draw=drawColor,line width= 0.6pt,line join=round] ( 39.04, 22.11) -- ( 39.04, 22.05);

\path[draw=drawColor,line width= 0.6pt,line join=round] ( 39.23, 22.11) -- ( 39.23, 22.05);

\path[draw=drawColor,line width= 0.6pt,line join=round] ( 39.41, 22.11) -- ( 39.41, 22.05);

\path[draw=drawColor,line width= 0.6pt,line join=round] ( 39.59, 22.12) -- ( 39.59, 22.05);

\path[draw=drawColor,line width= 0.6pt,line join=round] ( 39.77, 22.12) -- ( 39.77, 22.05);

\path[draw=drawColor,line width= 0.6pt,line join=round] ( 39.95, 22.12) -- ( 39.95, 22.05);

\path[draw=drawColor,line width= 0.6pt,line join=round] ( 40.13, 22.12) -- ( 40.13, 22.05);

\path[draw=drawColor,line width= 0.6pt,line join=round] ( 40.31, 22.13) -- ( 40.31, 22.05);

\path[draw=drawColor,line width= 0.6pt,line join=round] ( 40.49, 22.13) -- ( 40.49, 22.05);

\path[draw=drawColor,line width= 0.6pt,line join=round] ( 40.67, 22.13) -- ( 40.67, 22.05);

\path[draw=drawColor,line width= 0.6pt,line join=round] ( 40.85, 22.13) -- ( 40.85, 22.05);

\path[draw=drawColor,line width= 0.6pt,line join=round] ( 41.03, 22.13) -- ( 41.03, 22.05);

\path[draw=drawColor,line width= 0.6pt,line join=round] ( 41.21, 22.13) -- ( 41.21, 22.05);

\path[draw=drawColor,line width= 0.6pt,line join=round] ( 41.39, 22.13) -- ( 41.39, 22.05);

\path[draw=drawColor,line width= 0.6pt,line join=round] ( 41.58, 22.14) -- ( 41.58, 22.05);

\path[draw=drawColor,line width= 0.6pt,line join=round] ( 41.76, 22.14) -- ( 41.76, 22.05);

\path[draw=drawColor,line width= 0.6pt,line join=round] ( 41.94, 22.14) -- ( 41.94, 22.05);

\path[draw=drawColor,line width= 0.6pt,line join=round] ( 42.12, 22.14) -- ( 42.12, 22.05);

\path[draw=drawColor,line width= 0.6pt,line join=round] ( 42.30, 22.14) -- ( 42.30, 22.05);

\path[draw=drawColor,line width= 0.6pt,line join=round] ( 42.48, 22.15) -- ( 42.48, 22.05);

\path[draw=drawColor,line width= 0.6pt,line join=round] ( 42.66, 22.15) -- ( 42.66, 22.05);

\path[draw=drawColor,line width= 0.6pt,line join=round] ( 42.84, 22.15) -- ( 42.84, 22.05);

\path[draw=drawColor,line width= 0.6pt,line join=round] ( 43.02, 22.15) -- ( 43.02, 22.05);

\path[draw=drawColor,line width= 0.6pt,line join=round] ( 43.20, 22.15) -- ( 43.20, 22.05);

\path[draw=drawColor,line width= 0.6pt,line join=round] ( 43.38, 22.15) -- ( 43.38, 22.05);

\path[draw=drawColor,line width= 0.6pt,line join=round] ( 43.56, 22.16) -- ( 43.56, 22.05);

\path[draw=drawColor,line width= 0.6pt,line join=round] ( 43.75, 22.16) -- ( 43.75, 22.05);

\path[draw=drawColor,line width= 0.6pt,line join=round] ( 43.93, 22.17) -- ( 43.93, 22.05);

\path[draw=drawColor,line width= 0.6pt,line join=round] ( 44.11, 22.17) -- ( 44.11, 22.05);

\path[draw=drawColor,line width= 0.6pt,line join=round] ( 44.29, 22.17) -- ( 44.29, 22.05);

\path[draw=drawColor,line width= 0.6pt,line join=round] ( 44.47, 22.18) -- ( 44.47, 22.05);

\path[draw=drawColor,line width= 0.6pt,line join=round] ( 44.65, 22.18) -- ( 44.65, 22.05);

\path[draw=drawColor,line width= 0.6pt,line join=round] ( 44.83, 22.18) -- ( 44.83, 22.05);

\path[draw=drawColor,line width= 0.6pt,line join=round] ( 45.01, 22.18) -- ( 45.01, 22.05);

\path[draw=drawColor,line width= 0.6pt,line join=round] ( 45.19, 22.19) -- ( 45.19, 22.05);

\path[draw=drawColor,line width= 0.6pt,line join=round] ( 45.37, 22.19) -- ( 45.37, 22.05);

\path[draw=drawColor,line width= 0.6pt,line join=round] ( 45.55, 22.19) -- ( 45.55, 22.05);

\path[draw=drawColor,line width= 0.6pt,line join=round] ( 45.73, 22.19) -- ( 45.73, 22.05);

\path[draw=drawColor,line width= 0.6pt,line join=round] ( 45.92, 22.19) -- ( 45.92, 22.05);

\path[draw=drawColor,line width= 0.6pt,line join=round] ( 46.10, 22.20) -- ( 46.10, 22.05);

\path[draw=drawColor,line width= 0.6pt,line join=round] ( 46.28, 22.20) -- ( 46.28, 22.05);

\path[draw=drawColor,line width= 0.6pt,line join=round] ( 46.46, 22.21) -- ( 46.46, 22.05);

\path[draw=drawColor,line width= 0.6pt,line join=round] ( 46.64, 22.21) -- ( 46.64, 22.05);

\path[draw=drawColor,line width= 0.6pt,line join=round] ( 46.82, 22.22) -- ( 46.82, 22.05);

\path[draw=drawColor,line width= 0.6pt,line join=round] ( 47.00, 22.22) -- ( 47.00, 22.05);

\path[draw=drawColor,line width= 0.6pt,line join=round] ( 47.18, 22.23) -- ( 47.18, 22.05);

\path[draw=drawColor,line width= 0.6pt,line join=round] ( 47.36, 22.23) -- ( 47.36, 22.05);

\path[draw=drawColor,line width= 0.6pt,line join=round] ( 47.54, 22.25) -- ( 47.54, 22.05);

\path[draw=drawColor,line width= 0.6pt,line join=round] ( 47.72, 22.25) -- ( 47.72, 22.05);

\path[draw=drawColor,line width= 0.6pt,line join=round] ( 47.90, 22.25) -- ( 47.90, 22.05);

\path[draw=drawColor,line width= 0.6pt,line join=round] ( 48.08, 22.25) -- ( 48.08, 22.05);

\path[draw=drawColor,line width= 0.6pt,line join=round] ( 48.27, 22.26) -- ( 48.27, 22.05);

\path[draw=drawColor,line width= 0.6pt,line join=round] ( 48.45, 22.26) -- ( 48.45, 22.05);

\path[draw=drawColor,line width= 0.6pt,line join=round] ( 48.63, 22.26) -- ( 48.63, 22.05);

\path[draw=drawColor,line width= 0.6pt,line join=round] ( 48.81, 22.27) -- ( 48.81, 22.05);

\path[draw=drawColor,line width= 0.6pt,line join=round] ( 48.99, 22.27) -- ( 48.99, 22.05);

\path[draw=drawColor,line width= 0.6pt,line join=round] ( 49.17, 22.27) -- ( 49.17, 22.05);

\path[draw=drawColor,line width= 0.6pt,line join=round] ( 49.35, 22.28) -- ( 49.35, 22.05);

\path[draw=drawColor,line width= 0.6pt,line join=round] ( 49.53, 22.29) -- ( 49.53, 22.05);

\path[draw=drawColor,line width= 0.6pt,line join=round] ( 49.71, 22.29) -- ( 49.71, 22.05);

\path[draw=drawColor,line width= 0.6pt,line join=round] ( 49.89, 22.29) -- ( 49.89, 22.05);

\path[draw=drawColor,line width= 0.6pt,line join=round] ( 50.07, 22.29) -- ( 50.07, 22.05);

\path[draw=drawColor,line width= 0.6pt,line join=round] ( 50.25, 22.30) -- ( 50.25, 22.05);

\path[draw=drawColor,line width= 0.6pt,line join=round] ( 50.44, 22.31) -- ( 50.44, 22.05);

\path[draw=drawColor,line width= 0.6pt,line join=round] ( 50.62, 22.31) -- ( 50.62, 22.05);

\path[draw=drawColor,line width= 0.6pt,line join=round] ( 50.80, 22.31) -- ( 50.80, 22.05);

\path[draw=drawColor,line width= 0.6pt,line join=round] ( 50.98, 22.31) -- ( 50.98, 22.05);

\path[draw=drawColor,line width= 0.6pt,line join=round] ( 51.16, 22.32) -- ( 51.16, 22.05);

\path[draw=drawColor,line width= 0.6pt,line join=round] ( 51.34, 22.33) -- ( 51.34, 22.05);

\path[draw=drawColor,line width= 0.6pt,line join=round] ( 51.52, 22.35) -- ( 51.52, 22.05);

\path[draw=drawColor,line width= 0.6pt,line join=round] ( 51.70, 22.35) -- ( 51.70, 22.05);

\path[draw=drawColor,line width= 0.6pt,line join=round] ( 51.88, 22.35) -- ( 51.88, 22.05);

\path[draw=drawColor,line width= 0.6pt,line join=round] ( 52.06, 22.36) -- ( 52.06, 22.05);

\path[draw=drawColor,line width= 0.6pt,line join=round] ( 52.24, 22.36) -- ( 52.24, 22.05);

\path[draw=drawColor,line width= 0.6pt,line join=round] ( 52.42, 22.37) -- ( 52.42, 22.05);

\path[draw=drawColor,line width= 0.6pt,line join=round] ( 52.60, 22.38) -- ( 52.60, 22.05);

\path[draw=drawColor,line width= 0.6pt,line join=round] ( 52.79, 22.39) -- ( 52.79, 22.05);

\path[draw=drawColor,line width= 0.6pt,line join=round] ( 52.97, 22.40) -- ( 52.97, 22.05);

\path[draw=drawColor,line width= 0.6pt,line join=round] ( 53.15, 22.40) -- ( 53.15, 22.05);

\path[draw=drawColor,line width= 0.6pt,line join=round] ( 53.33, 22.41) -- ( 53.33, 22.05);

\path[draw=drawColor,line width= 0.6pt,line join=round] ( 53.51, 22.41) -- ( 53.51, 22.05);

\path[draw=drawColor,line width= 0.6pt,line join=round] ( 53.69, 22.41) -- ( 53.69, 22.05);

\path[draw=drawColor,line width= 0.6pt,line join=round] ( 53.87, 22.42) -- ( 53.87, 22.05);

\path[draw=drawColor,line width= 0.6pt,line join=round] ( 54.05, 22.43) -- ( 54.05, 22.05);

\path[draw=drawColor,line width= 0.6pt,line join=round] ( 54.23, 22.43) -- ( 54.23, 22.05);

\path[draw=drawColor,line width= 0.6pt,line join=round] ( 54.41, 22.44) -- ( 54.41, 22.05);

\path[draw=drawColor,line width= 0.6pt,line join=round] ( 54.59, 22.44) -- ( 54.59, 22.05);

\path[draw=drawColor,line width= 0.6pt,line join=round] ( 54.77, 22.45) -- ( 54.77, 22.05);

\path[draw=drawColor,line width= 0.6pt,line join=round] ( 54.96, 22.46) -- ( 54.96, 22.05);

\path[draw=drawColor,line width= 0.6pt,line join=round] ( 55.14, 22.47) -- ( 55.14, 22.05);

\path[draw=drawColor,line width= 0.6pt,line join=round] ( 55.32, 22.47) -- ( 55.32, 22.05);

\path[draw=drawColor,line width= 0.6pt,line join=round] ( 55.50, 22.48) -- ( 55.50, 22.05);

\path[draw=drawColor,line width= 0.6pt,line join=round] ( 55.68, 22.48) -- ( 55.68, 22.05);

\path[draw=drawColor,line width= 0.6pt,line join=round] ( 55.86, 22.48) -- ( 55.86, 22.05);

\path[draw=drawColor,line width= 0.6pt,line join=round] ( 56.04, 22.49) -- ( 56.04, 22.05);

\path[draw=drawColor,line width= 0.6pt,line join=round] ( 56.22, 22.49) -- ( 56.22, 22.05);

\path[draw=drawColor,line width= 0.6pt,line join=round] ( 56.40, 22.50) -- ( 56.40, 22.05);

\path[draw=drawColor,line width= 0.6pt,line join=round] ( 56.58, 22.50) -- ( 56.58, 22.05);

\path[draw=drawColor,line width= 0.6pt,line join=round] ( 56.76, 22.51) -- ( 56.76, 22.05);

\path[draw=drawColor,line width= 0.6pt,line join=round] ( 56.94, 22.52) -- ( 56.94, 22.05);

\path[draw=drawColor,line width= 0.6pt,line join=round] ( 57.13, 22.52) -- ( 57.13, 22.05);

\path[draw=drawColor,line width= 0.6pt,line join=round] ( 57.31, 22.52) -- ( 57.31, 22.05);

\path[draw=drawColor,line width= 0.6pt,line join=round] ( 57.49, 22.53) -- ( 57.49, 22.05);

\path[draw=drawColor,line width= 0.6pt,line join=round] ( 57.67, 22.53) -- ( 57.67, 22.05);

\path[draw=drawColor,line width= 0.6pt,line join=round] ( 57.85, 22.54) -- ( 57.85, 22.05);

\path[draw=drawColor,line width= 0.6pt,line join=round] ( 58.03, 22.54) -- ( 58.03, 22.05);

\path[draw=drawColor,line width= 0.6pt,line join=round] ( 58.21, 22.55) -- ( 58.21, 22.05);

\path[draw=drawColor,line width= 0.6pt,line join=round] ( 58.39, 22.55) -- ( 58.39, 22.05);

\path[draw=drawColor,line width= 0.6pt,line join=round] ( 58.57, 22.55) -- ( 58.57, 22.05);

\path[draw=drawColor,line width= 0.6pt,line join=round] ( 58.75, 22.56) -- ( 58.75, 22.05);

\path[draw=drawColor,line width= 0.6pt,line join=round] ( 58.93, 22.57) -- ( 58.93, 22.05);

\path[draw=drawColor,line width= 0.6pt,line join=round] ( 59.11, 22.58) -- ( 59.11, 22.05);

\path[draw=drawColor,line width= 0.6pt,line join=round] ( 59.29, 22.58) -- ( 59.29, 22.05);

\path[draw=drawColor,line width= 0.6pt,line join=round] ( 59.48, 22.61) -- ( 59.48, 22.05);

\path[draw=drawColor,line width= 0.6pt,line join=round] ( 59.66, 22.62) -- ( 59.66, 22.05);

\path[draw=drawColor,line width= 0.6pt,line join=round] ( 59.84, 22.64) -- ( 59.84, 22.05);

\path[draw=drawColor,line width= 0.6pt,line join=round] ( 60.02, 22.64) -- ( 60.02, 22.05);

\path[draw=drawColor,line width= 0.6pt,line join=round] ( 60.20, 22.66) -- ( 60.20, 22.05);

\path[draw=drawColor,line width= 0.6pt,line join=round] ( 60.38, 22.66) -- ( 60.38, 22.05);

\path[draw=drawColor,line width= 0.6pt,line join=round] ( 60.56, 22.66) -- ( 60.56, 22.05);

\path[draw=drawColor,line width= 0.6pt,line join=round] ( 60.74, 22.68) -- ( 60.74, 22.05);

\path[draw=drawColor,line width= 0.6pt,line join=round] ( 60.92, 22.71) -- ( 60.92, 22.05);

\path[draw=drawColor,line width= 0.6pt,line join=round] ( 61.10, 22.71) -- ( 61.10, 22.05);

\path[draw=drawColor,line width= 0.6pt,line join=round] ( 61.28, 22.75) -- ( 61.28, 22.05);

\path[draw=drawColor,line width= 0.6pt,line join=round] ( 61.46, 22.76) -- ( 61.46, 22.05);

\path[draw=drawColor,line width= 0.6pt,line join=round] ( 61.65, 22.76) -- ( 61.65, 22.05);

\path[draw=drawColor,line width= 0.6pt,line join=round] ( 61.83, 22.77) -- ( 61.83, 22.05);

\path[draw=drawColor,line width= 0.6pt,line join=round] ( 62.01, 22.78) -- ( 62.01, 22.05);

\path[draw=drawColor,line width= 0.6pt,line join=round] ( 62.19, 22.78) -- ( 62.19, 22.05);

\path[draw=drawColor,line width= 0.6pt,line join=round] ( 62.37, 22.79) -- ( 62.37, 22.05);

\path[draw=drawColor,line width= 0.6pt,line join=round] ( 62.55, 22.80) -- ( 62.55, 22.05);

\path[draw=drawColor,line width= 0.6pt,line join=round] ( 62.73, 22.80) -- ( 62.73, 22.05);

\path[draw=drawColor,line width= 0.6pt,line join=round] ( 62.91, 22.80) -- ( 62.91, 22.05);

\path[draw=drawColor,line width= 0.6pt,line join=round] ( 63.09, 22.82) -- ( 63.09, 22.05);

\path[draw=drawColor,line width= 0.6pt,line join=round] ( 63.27, 22.82) -- ( 63.27, 22.05);

\path[draw=drawColor,line width= 0.6pt,line join=round] ( 63.45, 22.82) -- ( 63.45, 22.05);

\path[draw=drawColor,line width= 0.6pt,line join=round] ( 63.63, 22.82) -- ( 63.63, 22.05);

\path[draw=drawColor,line width= 0.6pt,line join=round] ( 63.81, 22.83) -- ( 63.81, 22.05);

\path[draw=drawColor,line width= 0.6pt,line join=round] ( 64.00, 22.83) -- ( 64.00, 22.05);

\path[draw=drawColor,line width= 0.6pt,line join=round] ( 64.18, 22.84) -- ( 64.18, 22.05);

\path[draw=drawColor,line width= 0.6pt,line join=round] ( 64.36, 22.84) -- ( 64.36, 22.05);

\path[draw=drawColor,line width= 0.6pt,line join=round] ( 64.54, 22.84) -- ( 64.54, 22.05);

\path[draw=drawColor,line width= 0.6pt,line join=round] ( 64.72, 22.86) -- ( 64.72, 22.05);

\path[draw=drawColor,line width= 0.6pt,line join=round] ( 64.90, 22.86) -- ( 64.90, 22.05);

\path[draw=drawColor,line width= 0.6pt,line join=round] ( 65.08, 22.87) -- ( 65.08, 22.05);

\path[draw=drawColor,line width= 0.6pt,line join=round] ( 65.26, 22.88) -- ( 65.26, 22.05);

\path[draw=drawColor,line width= 0.6pt,line join=round] ( 65.44, 22.90) -- ( 65.44, 22.05);

\path[draw=drawColor,line width= 0.6pt,line join=round] ( 65.62, 22.90) -- ( 65.62, 22.05);

\path[draw=drawColor,line width= 0.6pt,line join=round] ( 65.80, 22.90) -- ( 65.80, 22.05);

\path[draw=drawColor,line width= 0.6pt,line join=round] ( 65.98, 22.91) -- ( 65.98, 22.05);

\path[draw=drawColor,line width= 0.6pt,line join=round] ( 66.17, 22.92) -- ( 66.17, 22.05);

\path[draw=drawColor,line width= 0.6pt,line join=round] ( 66.35, 22.92) -- ( 66.35, 22.05);

\path[draw=drawColor,line width= 0.6pt,line join=round] ( 66.53, 22.94) -- ( 66.53, 22.05);

\path[draw=drawColor,line width= 0.6pt,line join=round] ( 66.71, 22.94) -- ( 66.71, 22.05);

\path[draw=drawColor,line width= 0.6pt,line join=round] ( 66.89, 22.95) -- ( 66.89, 22.05);

\path[draw=drawColor,line width= 0.6pt,line join=round] ( 67.07, 22.95) -- ( 67.07, 22.05);

\path[draw=drawColor,line width= 0.6pt,line join=round] ( 67.25, 22.95) -- ( 67.25, 22.05);

\path[draw=drawColor,line width= 0.6pt,line join=round] ( 67.43, 22.95) -- ( 67.43, 22.05);

\path[draw=drawColor,line width= 0.6pt,line join=round] ( 67.61, 22.96) -- ( 67.61, 22.05);

\path[draw=drawColor,line width= 0.6pt,line join=round] ( 67.79, 22.97) -- ( 67.79, 22.05);

\path[draw=drawColor,line width= 0.6pt,line join=round] ( 67.97, 22.99) -- ( 67.97, 22.05);

\path[draw=drawColor,line width= 0.6pt,line join=round] ( 68.15, 22.99) -- ( 68.15, 22.05);

\path[draw=drawColor,line width= 0.6pt,line join=round] ( 68.33, 22.99) -- ( 68.33, 22.05);

\path[draw=drawColor,line width= 0.6pt,line join=round] ( 68.52, 23.00) -- ( 68.52, 22.05);

\path[draw=drawColor,line width= 0.6pt,line join=round] ( 68.70, 23.00) -- ( 68.70, 22.05);

\path[draw=drawColor,line width= 0.6pt,line join=round] ( 68.88, 23.00) -- ( 68.88, 22.05);

\path[draw=drawColor,line width= 0.6pt,line join=round] ( 69.06, 23.01) -- ( 69.06, 22.05);

\path[draw=drawColor,line width= 0.6pt,line join=round] ( 69.24, 23.02) -- ( 69.24, 22.05);

\path[draw=drawColor,line width= 0.6pt,line join=round] ( 69.42, 23.04) -- ( 69.42, 22.05);

\path[draw=drawColor,line width= 0.6pt,line join=round] ( 69.60, 23.07) -- ( 69.60, 22.05);

\path[draw=drawColor,line width= 0.6pt,line join=round] ( 69.78, 23.08) -- ( 69.78, 22.05);

\path[draw=drawColor,line width= 0.6pt,line join=round] ( 69.96, 23.09) -- ( 69.96, 22.05);

\path[draw=drawColor,line width= 0.6pt,line join=round] ( 70.14, 23.10) -- ( 70.14, 22.05);

\path[draw=drawColor,line width= 0.6pt,line join=round] ( 70.32, 23.11) -- ( 70.32, 22.05);

\path[draw=drawColor,line width= 0.6pt,line join=round] ( 70.50, 23.14) -- ( 70.50, 22.05);

\path[draw=drawColor,line width= 0.6pt,line join=round] ( 70.69, 23.15) -- ( 70.69, 22.05);

\path[draw=drawColor,line width= 0.6pt,line join=round] ( 70.87, 23.16) -- ( 70.87, 22.05);

\path[draw=drawColor,line width= 0.6pt,line join=round] ( 71.05, 23.16) -- ( 71.05, 22.05);

\path[draw=drawColor,line width= 0.6pt,line join=round] ( 71.23, 23.16) -- ( 71.23, 22.05);

\path[draw=drawColor,line width= 0.6pt,line join=round] ( 71.41, 23.17) -- ( 71.41, 22.05);

\path[draw=drawColor,line width= 0.6pt,line join=round] ( 71.59, 23.17) -- ( 71.59, 22.05);

\path[draw=drawColor,line width= 0.6pt,line join=round] ( 71.77, 23.18) -- ( 71.77, 22.05);

\path[draw=drawColor,line width= 0.6pt,line join=round] ( 71.95, 23.19) -- ( 71.95, 22.05);

\path[draw=drawColor,line width= 0.6pt,line join=round] ( 72.13, 23.20) -- ( 72.13, 22.05);

\path[draw=drawColor,line width= 0.6pt,line join=round] ( 72.31, 23.22) -- ( 72.31, 22.05);

\path[draw=drawColor,line width= 0.6pt,line join=round] ( 72.49, 23.22) -- ( 72.49, 22.05);

\path[draw=drawColor,line width= 0.6pt,line join=round] ( 72.67, 23.24) -- ( 72.67, 22.05);

\path[draw=drawColor,line width= 0.6pt,line join=round] ( 72.86, 23.24) -- ( 72.86, 22.05);

\path[draw=drawColor,line width= 0.6pt,line join=round] ( 73.04, 23.24) -- ( 73.04, 22.05);

\path[draw=drawColor,line width= 0.6pt,line join=round] ( 73.22, 23.24) -- ( 73.22, 22.05);

\path[draw=drawColor,line width= 0.6pt,line join=round] ( 73.40, 23.24) -- ( 73.40, 22.05);

\path[draw=drawColor,line width= 0.6pt,line join=round] ( 73.58, 23.26) -- ( 73.58, 22.05);

\path[draw=drawColor,line width= 0.6pt,line join=round] ( 73.76, 23.27) -- ( 73.76, 22.05);

\path[draw=drawColor,line width= 0.6pt,line join=round] ( 73.94, 23.29) -- ( 73.94, 22.05);

\path[draw=drawColor,line width= 0.6pt,line join=round] ( 74.12, 23.31) -- ( 74.12, 22.05);

\path[draw=drawColor,line width= 0.6pt,line join=round] ( 74.30, 23.33) -- ( 74.30, 22.05);

\path[draw=drawColor,line width= 0.6pt,line join=round] ( 74.48, 23.35) -- ( 74.48, 22.05);

\path[draw=drawColor,line width= 0.6pt,line join=round] ( 74.66, 23.35) -- ( 74.66, 22.05);

\path[draw=drawColor,line width= 0.6pt,line join=round] ( 74.84, 23.36) -- ( 74.84, 22.05);

\path[draw=drawColor,line width= 0.6pt,line join=round] ( 75.02, 23.37) -- ( 75.02, 22.05);

\path[draw=drawColor,line width= 0.6pt,line join=round] ( 75.21, 23.37) -- ( 75.21, 22.05);

\path[draw=drawColor,line width= 0.6pt,line join=round] ( 75.39, 23.37) -- ( 75.39, 22.05);

\path[draw=drawColor,line width= 0.6pt,line join=round] ( 75.57, 23.40) -- ( 75.57, 22.05);

\path[draw=drawColor,line width= 0.6pt,line join=round] ( 75.75, 23.44) -- ( 75.75, 22.05);

\path[draw=drawColor,line width= 0.6pt,line join=round] ( 75.93, 23.47) -- ( 75.93, 22.05);

\path[draw=drawColor,line width= 0.6pt,line join=round] ( 76.11, 23.51) -- ( 76.11, 22.05);

\path[draw=drawColor,line width= 0.6pt,line join=round] ( 76.29, 23.52) -- ( 76.29, 22.05);

\path[draw=drawColor,line width= 0.6pt,line join=round] ( 76.47, 23.53) -- ( 76.47, 22.05);

\path[draw=drawColor,line width= 0.6pt,line join=round] ( 76.65, 23.54) -- ( 76.65, 22.05);

\path[draw=drawColor,line width= 0.6pt,line join=round] ( 76.83, 23.57) -- ( 76.83, 22.05);

\path[draw=drawColor,line width= 0.6pt,line join=round] ( 77.01, 23.57) -- ( 77.01, 22.05);

\path[draw=drawColor,line width= 0.6pt,line join=round] ( 77.19, 23.57) -- ( 77.19, 22.05);

\path[draw=drawColor,line width= 0.6pt,line join=round] ( 77.38, 23.58) -- ( 77.38, 22.05);

\path[draw=drawColor,line width= 0.6pt,line join=round] ( 77.56, 23.66) -- ( 77.56, 22.05);

\path[draw=drawColor,line width= 0.6pt,line join=round] ( 77.74, 23.67) -- ( 77.74, 22.05);

\path[draw=drawColor,line width= 0.6pt,line join=round] ( 77.92, 23.68) -- ( 77.92, 22.05);

\path[draw=drawColor,line width= 0.6pt,line join=round] ( 78.10, 23.68) -- ( 78.10, 22.05);

\path[draw=drawColor,line width= 0.6pt,line join=round] ( 78.28, 23.69) -- ( 78.28, 22.05);

\path[draw=drawColor,line width= 0.6pt,line join=round] ( 78.46, 23.70) -- ( 78.46, 22.05);

\path[draw=drawColor,line width= 0.6pt,line join=round] ( 78.64, 23.72) -- ( 78.64, 22.05);

\path[draw=drawColor,line width= 0.6pt,line join=round] ( 78.82, 23.72) -- ( 78.82, 22.05);

\path[draw=drawColor,line width= 0.6pt,line join=round] ( 79.00, 23.73) -- ( 79.00, 22.05);

\path[draw=drawColor,line width= 0.6pt,line join=round] ( 79.18, 23.75) -- ( 79.18, 22.05);

\path[draw=drawColor,line width= 0.6pt,line join=round] ( 79.36, 23.76) -- ( 79.36, 22.05);

\path[draw=drawColor,line width= 0.6pt,line join=round] ( 79.54, 23.76) -- ( 79.54, 22.05);

\path[draw=drawColor,line width= 0.6pt,line join=round] ( 79.73, 23.78) -- ( 79.73, 22.05);

\path[draw=drawColor,line width= 0.6pt,line join=round] ( 79.91, 23.78) -- ( 79.91, 22.05);

\path[draw=drawColor,line width= 0.6pt,line join=round] ( 80.09, 23.79) -- ( 80.09, 22.05);

\path[draw=drawColor,line width= 0.6pt,line join=round] ( 80.27, 23.79) -- ( 80.27, 22.05);

\path[draw=drawColor,line width= 0.6pt,line join=round] ( 80.45, 23.80) -- ( 80.45, 22.05);

\path[draw=drawColor,line width= 0.6pt,line join=round] ( 80.63, 23.81) -- ( 80.63, 22.05);

\path[draw=drawColor,line width= 0.6pt,line join=round] ( 80.81, 23.81) -- ( 80.81, 22.05);

\path[draw=drawColor,line width= 0.6pt,line join=round] ( 80.99, 23.82) -- ( 80.99, 22.05);

\path[draw=drawColor,line width= 0.6pt,line join=round] ( 81.17, 23.82) -- ( 81.17, 22.05);

\path[draw=drawColor,line width= 0.6pt,line join=round] ( 81.35, 23.83) -- ( 81.35, 22.05);

\path[draw=drawColor,line width= 0.6pt,line join=round] ( 81.53, 23.83) -- ( 81.53, 22.05);

\path[draw=drawColor,line width= 0.6pt,line join=round] ( 81.71, 23.84) -- ( 81.71, 22.05);

\path[draw=drawColor,line width= 0.6pt,line join=round] ( 81.90, 23.87) -- ( 81.90, 22.05);

\path[draw=drawColor,line width= 0.6pt,line join=round] ( 82.08, 23.88) -- ( 82.08, 22.05);

\path[draw=drawColor,line width= 0.6pt,line join=round] ( 82.26, 23.89) -- ( 82.26, 22.05);

\path[draw=drawColor,line width= 0.6pt,line join=round] ( 82.44, 23.89) -- ( 82.44, 22.05);

\path[draw=drawColor,line width= 0.6pt,line join=round] ( 82.62, 23.90) -- ( 82.62, 22.05);

\path[draw=drawColor,line width= 0.6pt,line join=round] ( 82.80, 23.94) -- ( 82.80, 22.05);

\path[draw=drawColor,line width= 0.6pt,line join=round] ( 82.98, 23.94) -- ( 82.98, 22.05);

\path[draw=drawColor,line width= 0.6pt,line join=round] ( 83.16, 23.96) -- ( 83.16, 22.05);

\path[draw=drawColor,line width= 0.6pt,line join=round] ( 83.34, 23.96) -- ( 83.34, 22.05);

\path[draw=drawColor,line width= 0.6pt,line join=round] ( 83.52, 23.98) -- ( 83.52, 22.05);

\path[draw=drawColor,line width= 0.6pt,line join=round] ( 83.70, 23.98) -- ( 83.70, 22.05);

\path[draw=drawColor,line width= 0.6pt,line join=round] ( 83.88, 23.98) -- ( 83.88, 22.05);

\path[draw=drawColor,line width= 0.6pt,line join=round] ( 84.07, 23.98) -- ( 84.07, 22.05);

\path[draw=drawColor,line width= 0.6pt,line join=round] ( 84.25, 24.01) -- ( 84.25, 22.05);

\path[draw=drawColor,line width= 0.6pt,line join=round] ( 84.43, 24.04) -- ( 84.43, 22.05);

\path[draw=drawColor,line width= 0.6pt,line join=round] ( 84.61, 24.04) -- ( 84.61, 22.05);

\path[draw=drawColor,line width= 0.6pt,line join=round] ( 84.79, 24.04) -- ( 84.79, 22.05);

\path[draw=drawColor,line width= 0.6pt,line join=round] ( 84.97, 24.05) -- ( 84.97, 22.05);

\path[draw=drawColor,line width= 0.6pt,line join=round] ( 85.15, 24.07) -- ( 85.15, 22.05);

\path[draw=drawColor,line width= 0.6pt,line join=round] ( 85.33, 24.08) -- ( 85.33, 22.05);

\path[draw=drawColor,line width= 0.6pt,line join=round] ( 85.51, 24.08) -- ( 85.51, 22.05);

\path[draw=drawColor,line width= 0.6pt,line join=round] ( 85.69, 24.09) -- ( 85.69, 22.05);

\path[draw=drawColor,line width= 0.6pt,line join=round] ( 85.87, 24.16) -- ( 85.87, 22.05);

\path[draw=drawColor,line width= 0.6pt,line join=round] ( 86.05, 24.16) -- ( 86.05, 22.05);

\path[draw=drawColor,line width= 0.6pt,line join=round] ( 86.23, 24.22) -- ( 86.23, 22.05);

\path[draw=drawColor,line width= 0.6pt,line join=round] ( 86.42, 24.24) -- ( 86.42, 22.05);

\path[draw=drawColor,line width= 0.6pt,line join=round] ( 86.60, 24.25) -- ( 86.60, 22.05);

\path[draw=drawColor,line width= 0.6pt,line join=round] ( 86.78, 24.26) -- ( 86.78, 22.05);

\path[draw=drawColor,line width= 0.6pt,line join=round] ( 86.96, 24.26) -- ( 86.96, 22.05);

\path[draw=drawColor,line width= 0.6pt,line join=round] ( 87.14, 24.27) -- ( 87.14, 22.05);

\path[draw=drawColor,line width= 0.6pt,line join=round] ( 87.32, 24.30) -- ( 87.32, 22.05);

\path[draw=drawColor,line width= 0.6pt,line join=round] ( 87.50, 24.30) -- ( 87.50, 22.05);

\path[draw=drawColor,line width= 0.6pt,line join=round] ( 87.68, 24.31) -- ( 87.68, 22.05);

\path[draw=drawColor,line width= 0.6pt,line join=round] ( 87.86, 24.31) -- ( 87.86, 22.05);

\path[draw=drawColor,line width= 0.6pt,line join=round] ( 88.04, 24.33) -- ( 88.04, 22.05);

\path[draw=drawColor,line width= 0.6pt,line join=round] ( 88.22, 24.35) -- ( 88.22, 22.05);

\path[draw=drawColor,line width= 0.6pt,line join=round] ( 88.40, 24.36) -- ( 88.40, 22.05);

\path[draw=drawColor,line width= 0.6pt,line join=round] ( 88.59, 24.38) -- ( 88.59, 22.05);

\path[draw=drawColor,line width= 0.6pt,line join=round] ( 88.77, 24.40) -- ( 88.77, 22.05);

\path[draw=drawColor,line width= 0.6pt,line join=round] ( 88.95, 24.42) -- ( 88.95, 22.05);

\path[draw=drawColor,line width= 0.6pt,line join=round] ( 89.13, 24.48) -- ( 89.13, 22.05);

\path[draw=drawColor,line width= 0.6pt,line join=round] ( 89.31, 24.51) -- ( 89.31, 22.05);

\path[draw=drawColor,line width= 0.6pt,line join=round] ( 89.49, 24.52) -- ( 89.49, 22.05);

\path[draw=drawColor,line width= 0.6pt,line join=round] ( 89.67, 24.52) -- ( 89.67, 22.05);

\path[draw=drawColor,line width= 0.6pt,line join=round] ( 89.85, 24.53) -- ( 89.85, 22.05);

\path[draw=drawColor,line width= 0.6pt,line join=round] ( 90.03, 24.56) -- ( 90.03, 22.05);

\path[draw=drawColor,line width= 0.6pt,line join=round] ( 90.21, 24.66) -- ( 90.21, 22.05);

\path[draw=drawColor,line width= 0.6pt,line join=round] ( 90.39, 24.67) -- ( 90.39, 22.05);

\path[draw=drawColor,line width= 0.6pt,line join=round] ( 90.57, 24.70) -- ( 90.57, 22.05);

\path[draw=drawColor,line width= 0.6pt,line join=round] ( 90.75, 24.79) -- ( 90.75, 22.05);

\path[draw=drawColor,line width= 0.6pt,line join=round] ( 90.94, 24.80) -- ( 90.94, 22.05);

\path[draw=drawColor,line width= 0.6pt,line join=round] ( 91.12, 24.84) -- ( 91.12, 22.05);

\path[draw=drawColor,line width= 0.6pt,line join=round] ( 91.30, 24.87) -- ( 91.30, 22.05);

\path[draw=drawColor,line width= 0.6pt,line join=round] ( 91.48, 24.88) -- ( 91.48, 22.05);

\path[draw=drawColor,line width= 0.6pt,line join=round] ( 91.66, 24.88) -- ( 91.66, 22.05);

\path[draw=drawColor,line width= 0.6pt,line join=round] ( 91.84, 24.95) -- ( 91.84, 22.05);

\path[draw=drawColor,line width= 0.6pt,line join=round] ( 92.02, 24.98) -- ( 92.02, 22.05);

\path[draw=drawColor,line width= 0.6pt,line join=round] ( 92.20, 25.00) -- ( 92.20, 22.05);

\path[draw=drawColor,line width= 0.6pt,line join=round] ( 92.38, 25.05) -- ( 92.38, 22.05);

\path[draw=drawColor,line width= 0.6pt,line join=round] ( 92.56, 25.05) -- ( 92.56, 22.05);

\path[draw=drawColor,line width= 0.6pt,line join=round] ( 92.74, 25.10) -- ( 92.74, 22.05);

\path[draw=drawColor,line width= 0.6pt,line join=round] ( 92.92, 25.10) -- ( 92.92, 22.05);

\path[draw=drawColor,line width= 0.6pt,line join=round] ( 93.11, 25.18) -- ( 93.11, 22.05);

\path[draw=drawColor,line width= 0.6pt,line join=round] ( 93.29, 25.20) -- ( 93.29, 22.05);

\path[draw=drawColor,line width= 0.6pt,line join=round] ( 93.47, 25.20) -- ( 93.47, 22.05);

\path[draw=drawColor,line width= 0.6pt,line join=round] ( 93.65, 25.22) -- ( 93.65, 22.05);

\path[draw=drawColor,line width= 0.6pt,line join=round] ( 93.83, 25.27) -- ( 93.83, 22.05);

\path[draw=drawColor,line width= 0.6pt,line join=round] ( 94.01, 25.30) -- ( 94.01, 22.05);

\path[draw=drawColor,line width= 0.6pt,line join=round] ( 94.19, 25.30) -- ( 94.19, 22.05);

\path[draw=drawColor,line width= 0.6pt,line join=round] ( 94.37, 25.34) -- ( 94.37, 22.05);

\path[draw=drawColor,line width= 0.6pt,line join=round] ( 94.55, 25.37) -- ( 94.55, 22.05);

\path[draw=drawColor,line width= 0.6pt,line join=round] ( 94.73, 25.48) -- ( 94.73, 22.05);

\path[draw=drawColor,line width= 0.6pt,line join=round] ( 94.91, 25.58) -- ( 94.91, 22.05);

\path[draw=drawColor,line width= 0.6pt,line join=round] ( 95.09, 25.62) -- ( 95.09, 22.05);

\path[draw=drawColor,line width= 0.6pt,line join=round] ( 95.28, 25.68) -- ( 95.28, 22.05);

\path[draw=drawColor,line width= 0.6pt,line join=round] ( 95.46, 25.74) -- ( 95.46, 22.05);

\path[draw=drawColor,line width= 0.6pt,line join=round] ( 95.64, 25.79) -- ( 95.64, 22.05);

\path[draw=drawColor,line width= 0.6pt,line join=round] ( 95.82, 25.86) -- ( 95.82, 22.05);

\path[draw=drawColor,line width= 0.6pt,line join=round] ( 96.00, 25.86) -- ( 96.00, 22.05);

\path[draw=drawColor,line width= 0.6pt,line join=round] ( 96.18, 25.89) -- ( 96.18, 22.05);

\path[draw=drawColor,line width= 0.6pt,line join=round] ( 96.36, 25.91) -- ( 96.36, 22.05);

\path[draw=drawColor,line width= 0.6pt,line join=round] ( 96.54, 25.93) -- ( 96.54, 22.05);

\path[draw=drawColor,line width= 0.6pt,line join=round] ( 96.72, 26.02) -- ( 96.72, 22.05);

\path[draw=drawColor,line width= 0.6pt,line join=round] ( 96.90, 26.02) -- ( 96.90, 22.05);

\path[draw=drawColor,line width= 0.6pt,line join=round] ( 97.08, 26.03) -- ( 97.08, 22.05);

\path[draw=drawColor,line width= 0.6pt,line join=round] ( 97.26, 26.04) -- ( 97.26, 22.05);

\path[draw=drawColor,line width= 0.6pt,line join=round] ( 97.44, 26.08) -- ( 97.44, 22.05);

\path[draw=drawColor,line width= 0.6pt,line join=round] ( 97.63, 26.09) -- ( 97.63, 22.05);

\path[draw=drawColor,line width= 0.6pt,line join=round] ( 97.81, 26.11) -- ( 97.81, 22.05);

\path[draw=drawColor,line width= 0.6pt,line join=round] ( 97.99, 26.12) -- ( 97.99, 22.05);

\path[draw=drawColor,line width= 0.6pt,line join=round] ( 98.17, 26.14) -- ( 98.17, 22.05);

\path[draw=drawColor,line width= 0.6pt,line join=round] ( 98.35, 26.16) -- ( 98.35, 22.05);

\path[draw=drawColor,line width= 0.6pt,line join=round] ( 98.53, 26.17) -- ( 98.53, 22.05);

\path[draw=drawColor,line width= 0.6pt,line join=round] ( 98.71, 26.19) -- ( 98.71, 22.05);

\path[draw=drawColor,line width= 0.6pt,line join=round] ( 98.89, 26.27) -- ( 98.89, 22.05);

\path[draw=drawColor,line width= 0.6pt,line join=round] ( 99.07, 26.36) -- ( 99.07, 22.05);
\end{scope}
\begin{scope}
\path[clip] (  0.00,  0.00) rectangle (108.41, 93.95);
\definecolor{drawColor}{gray}{0.10}

\node[text=drawColor,anchor=base east,inner sep=0pt, outer sep=0pt, scale=  0.55] at ( 17.83, 35.96) {10};

\node[text=drawColor,anchor=base east,inner sep=0pt, outer sep=0pt, scale=  0.55] at ( 17.83, 51.77) {20};

\node[text=drawColor,anchor=base east,inner sep=0pt, outer sep=0pt, scale=  0.55] at ( 17.83, 67.58) {30};

\node[text=drawColor,anchor=base east,inner sep=0pt, outer sep=0pt, scale=  0.55] at ( 17.83, 83.39) {40};
\end{scope}
\begin{scope}
\path[clip] (  0.00,  0.00) rectangle (108.41, 93.95);
\definecolor{drawColor}{gray}{0.10}

\node[text=drawColor,anchor=base,inner sep=0pt, outer sep=0pt, scale=  0.55] at ( 22.23, 15.15) {0};

\node[text=drawColor,anchor=base,inner sep=0pt, outer sep=0pt, scale=  0.55] at ( 40.31, 15.15) {100};

\node[text=drawColor,anchor=base,inner sep=0pt, outer sep=0pt, scale=  0.55] at ( 58.39, 15.15) {200};

\node[text=drawColor,anchor=base,inner sep=0pt, outer sep=0pt, scale=  0.55] at ( 76.47, 15.15) {300};

\node[text=drawColor,anchor=base,inner sep=0pt, outer sep=0pt, scale=  0.55] at ( 94.55, 15.15) {400};
\end{scope}
\begin{scope}
\path[clip] (  0.00,  0.00) rectangle (108.41, 93.95);
\definecolor{drawColor}{gray}{0.10}

\node[text=drawColor,anchor=base,inner sep=0pt, outer sep=0pt, scale=  0.66] at ( 60.74,  6.78) {estimated inliers};
\end{scope}
\begin{scope}
\path[clip] (  0.00,  0.00) rectangle (108.41, 93.95);
\definecolor{drawColor}{gray}{0.10}

\node[text=drawColor,rotate= 90.00,anchor=base,inner sep=0pt, outer sep=0pt, scale=  0.66] at ( 10.05, 53.67) {reprojection error (pixel)};
\end{scope}
\end{tikzpicture}

%% file: Rfigures/KUSVOD2Trans_mean.tex
\begin{tikzpicture}[x=1pt,y=1pt]
\clip (5,3) rectangle (104, 93.95);
\definecolor{fillColor}{RGB}{255,255,255}
\path[use as bounding box,fill=fillColor,fill opacity=0.00] (0,0) rectangle (108.41, 93.95);
\begin{scope}
\path[clip] (  0.00,  0.00) rectangle (108.41, 93.95);
\definecolor{drawColor}{RGB}{255,255,255}
\definecolor{fillColor}{RGB}{255,255,255}

\path[draw=drawColor,line width= 0.6pt,line join=round,line cap=round,fill=fillColor] ( -0.00,  0.00) rectangle (108.41, 93.95);
\end{scope}
\begin{scope}
\path[clip] ( 19.10, 18.89) rectangle (102.90, 88.45);
\definecolor{fillColor}{gray}{0.92}

\path[fill=fillColor] ( 19.10, 18.89) rectangle (102.90, 88.45);
\definecolor{drawColor}{RGB}{0,0,0}

\path[draw=drawColor,line width= 0.6pt,line join=round] ( 21.48, 82.41) --
	( 29.38, 60.09) --
	( 37.29, 47.91) --
	( 45.19, 43.03) --
	( 53.10, 40.13) --
	( 61.00, 37.16) --
	( 68.91, 37.33) --
	( 76.82, 38.51) --
	( 84.72, 47.97) --
	( 92.63, 57.81) --
	(100.53, 75.80);

\path[draw=drawColor,line width= 0.6pt,line join=round] ( 21.48, 79.52) --
	( 29.38, 61.37) --
	( 37.29, 47.98) --
	( 45.19, 39.54) --
	( 53.10, 34.21) --
	( 61.00, 29.68) --
	( 68.91, 28.44) --
	( 76.82, 29.35) --
	( 84.72, 39.38) --
	( 92.63, 61.03) --
	(100.53, 85.29);

\path[draw=drawColor,line width= 0.6pt,line join=round] ( 21.48, 80.27) --
	( 29.38, 58.58) --
	( 37.29, 46.14) --
	( 45.19, 38.73) --
	( 53.10, 32.64) --
	( 61.00, 27.38) --
	( 68.91, 26.54) --
	( 76.82, 26.76) --
	( 84.72, 31.76) --
	( 92.63, 41.16) --
	(100.53, 55.49);

\path[draw=drawColor,line width= 0.6pt,line join=round] ( 21.48, 80.34) --
	( 29.38, 58.92) --
	( 37.29, 45.19) --
	( 45.19, 35.49) --
	( 53.10, 27.94) --
	( 61.00, 23.07) --
	( 68.91, 22.05) --
	( 76.82, 22.16) --
	( 84.72, 26.56) --
	( 92.63, 34.20) --
	(100.53, 47.61);
\definecolor{drawColor}{RGB}{0,0,0}
\definecolor{fillColor}{RGB}{190,190,190}

\path[draw=drawColor,draw opacity=0.70,line width= 0.4pt,line join=round,line cap=round,fill=fillColor,fill opacity=0.70] ( 21.48, 85.46) --
	( 24.12, 80.88) --
	( 18.83, 80.88) --
	cycle;

\path[draw=drawColor,draw opacity=0.70,line width= 0.4pt,line join=round,line cap=round,fill=fillColor,fill opacity=0.70] ( 29.38, 63.14) --
	( 32.02, 58.56) --
	( 26.74, 58.56) --
	cycle;

\path[draw=drawColor,draw opacity=0.70,line width= 0.4pt,line join=round,line cap=round,fill=fillColor,fill opacity=0.70] ( 37.29, 50.96) --
	( 39.93, 46.39) --
	( 34.65, 46.39) --
	cycle;

\path[draw=drawColor,draw opacity=0.70,line width= 0.4pt,line join=round,line cap=round,fill=fillColor,fill opacity=0.70] ( 45.19, 46.08) --
	( 47.84, 41.51) --
	( 42.55, 41.51) --
	cycle;

\path[draw=drawColor,draw opacity=0.70,line width= 0.4pt,line join=round,line cap=round,fill=fillColor,fill opacity=0.70] ( 53.10, 43.18) --
	( 55.74, 38.61) --
	( 50.46, 38.61) --
	cycle;

\path[draw=drawColor,draw opacity=0.70,line width= 0.4pt,line join=round,line cap=round,fill=fillColor,fill opacity=0.70] ( 61.00, 40.21) --
	( 63.65, 35.63) --
	( 58.36, 35.63) --
	cycle;

\path[draw=drawColor,draw opacity=0.70,line width= 0.4pt,line join=round,line cap=round,fill=fillColor,fill opacity=0.70] ( 68.91, 40.38) --
	( 71.55, 35.80) --
	( 66.27, 35.80) --
	cycle;

\path[draw=drawColor,draw opacity=0.70,line width= 0.4pt,line join=round,line cap=round,fill=fillColor,fill opacity=0.70] ( 76.82, 41.56) --
	( 79.46, 36.98) --
	( 74.17, 36.98) --
	cycle;

\path[draw=drawColor,draw opacity=0.70,line width= 0.4pt,line join=round,line cap=round,fill=fillColor,fill opacity=0.70] ( 84.72, 51.02) --
	( 87.36, 46.45) --
	( 82.08, 46.45) --
	cycle;

\path[draw=drawColor,draw opacity=0.70,line width= 0.4pt,line join=round,line cap=round,fill=fillColor,fill opacity=0.70] ( 92.63, 60.86) --
	( 95.27, 56.28) --
	( 89.99, 56.28) --
	cycle;

\path[draw=drawColor,draw opacity=0.70,line width= 0.4pt,line join=round,line cap=round,fill=fillColor,fill opacity=0.70] (100.53, 78.85) --
	(103.18, 74.28) --
	( 97.89, 74.28) --
	cycle;
\definecolor{fillColor}{RGB}{173,216,230}

\path[draw=drawColor,draw opacity=0.70,line width= 0.4pt,line join=round,line cap=round,fill=fillColor,fill opacity=0.70] ( 19.74, 77.78) rectangle ( 23.22, 81.26);

\path[draw=drawColor,draw opacity=0.70,line width= 0.4pt,line join=round,line cap=round,fill=fillColor,fill opacity=0.70] ( 27.64, 59.63) rectangle ( 31.12, 63.11);

\path[draw=drawColor,draw opacity=0.70,line width= 0.4pt,line join=round,line cap=round,fill=fillColor,fill opacity=0.70] ( 35.55, 46.24) rectangle ( 39.03, 49.71);

\path[draw=drawColor,draw opacity=0.70,line width= 0.4pt,line join=round,line cap=round,fill=fillColor,fill opacity=0.70] ( 43.45, 37.80) rectangle ( 46.93, 41.28);

\path[draw=drawColor,draw opacity=0.70,line width= 0.4pt,line join=round,line cap=round,fill=fillColor,fill opacity=0.70] ( 51.36, 32.47) rectangle ( 54.84, 35.95);

\path[draw=drawColor,draw opacity=0.70,line width= 0.4pt,line join=round,line cap=round,fill=fillColor,fill opacity=0.70] ( 59.27, 27.95) rectangle ( 62.74, 31.42);

\path[draw=drawColor,draw opacity=0.70,line width= 0.4pt,line join=round,line cap=round,fill=fillColor,fill opacity=0.70] ( 67.17, 26.70) rectangle ( 70.65, 30.18);

\path[draw=drawColor,draw opacity=0.70,line width= 0.4pt,line join=round,line cap=round,fill=fillColor,fill opacity=0.70] ( 75.08, 27.61) rectangle ( 78.56, 31.09);

\path[draw=drawColor,draw opacity=0.70,line width= 0.4pt,line join=round,line cap=round,fill=fillColor,fill opacity=0.70] ( 82.98, 37.64) rectangle ( 86.46, 41.12);

\path[draw=drawColor,draw opacity=0.70,line width= 0.4pt,line join=round,line cap=round,fill=fillColor,fill opacity=0.70] ( 90.89, 59.29) rectangle ( 94.37, 62.77);

\path[draw=drawColor,draw opacity=0.70,line width= 0.4pt,line join=round,line cap=round,fill=fillColor,fill opacity=0.70] ( 98.79, 83.55) rectangle (102.27, 87.03);

\path[draw=drawColor,draw opacity=0.70,line width= 0.4pt,line join=round,line cap=round] ( 19.51, 78.30) rectangle ( 23.44, 82.23);

\path[draw=drawColor,draw opacity=0.70,line width= 0.4pt,line join=round,line cap=round] ( 27.42, 56.62) rectangle ( 31.34, 60.54);

\path[draw=drawColor,draw opacity=0.70,line width= 0.4pt,line join=round,line cap=round] ( 35.33, 44.18) rectangle ( 39.25, 48.11);

\path[draw=drawColor,draw opacity=0.70,line width= 0.4pt,line join=round,line cap=round] ( 43.23, 36.77) rectangle ( 47.16, 40.69);

\path[draw=drawColor,draw opacity=0.70,line width= 0.4pt,line join=round,line cap=round] ( 51.14, 30.68) rectangle ( 55.06, 34.61);

\path[draw=drawColor,draw opacity=0.70,line width= 0.4pt,line join=round,line cap=round] ( 59.04, 25.42) rectangle ( 62.97, 29.35);

\path[draw=drawColor,draw opacity=0.70,line width= 0.4pt,line join=round,line cap=round] ( 66.95, 24.58) rectangle ( 70.87, 28.50);

\path[draw=drawColor,draw opacity=0.70,line width= 0.4pt,line join=round,line cap=round] ( 74.85, 24.80) rectangle ( 78.78, 28.72);

\path[draw=drawColor,draw opacity=0.70,line width= 0.4pt,line join=round,line cap=round] ( 82.76, 29.80) rectangle ( 86.68, 33.72);

\path[draw=drawColor,draw opacity=0.70,line width= 0.4pt,line join=round,line cap=round] ( 90.67, 39.20) rectangle ( 94.59, 43.12);

\path[draw=drawColor,draw opacity=0.70,line width= 0.4pt,line join=round,line cap=round] ( 98.57, 53.53) rectangle (102.50, 57.45);
\definecolor{fillColor}{RGB}{255,0,0}

\path[draw=drawColor,draw opacity=0.70,line width= 0.4pt,line join=round,line cap=round,fill=fillColor,fill opacity=0.70] ( 21.48, 80.34) circle (  1.96);

\path[draw=drawColor,draw opacity=0.70,line width= 0.4pt,line join=round,line cap=round,fill=fillColor,fill opacity=0.70] ( 29.38, 58.92) circle (  1.96);

\path[draw=drawColor,draw opacity=0.70,line width= 0.4pt,line join=round,line cap=round,fill=fillColor,fill opacity=0.70] ( 37.29, 45.19) circle (  1.96);

\path[draw=drawColor,draw opacity=0.70,line width= 0.4pt,line join=round,line cap=round,fill=fillColor,fill opacity=0.70] ( 45.19, 35.49) circle (  1.96);

\path[draw=drawColor,draw opacity=0.70,line width= 0.4pt,line join=round,line cap=round,fill=fillColor,fill opacity=0.70] ( 53.10, 27.94) circle (  1.96);

\path[draw=drawColor,draw opacity=0.70,line width= 0.4pt,line join=round,line cap=round,fill=fillColor,fill opacity=0.70] ( 61.00, 23.07) circle (  1.96);

\path[draw=drawColor,draw opacity=0.70,line width= 0.4pt,line join=round,line cap=round,fill=fillColor,fill opacity=0.70] ( 68.91, 22.05) circle (  1.96);

\path[draw=drawColor,draw opacity=0.70,line width= 0.4pt,line join=round,line cap=round,fill=fillColor,fill opacity=0.70] ( 76.82, 22.16) circle (  1.96);

\path[draw=drawColor,draw opacity=0.70,line width= 0.4pt,line join=round,line cap=round,fill=fillColor,fill opacity=0.70] ( 84.72, 26.56) circle (  1.96);

\path[draw=drawColor,draw opacity=0.70,line width= 0.4pt,line join=round,line cap=round,fill=fillColor,fill opacity=0.70] ( 92.63, 34.20) circle (  1.96);

\path[draw=drawColor,draw opacity=0.70,line width= 0.4pt,line join=round,line cap=round,fill=fillColor,fill opacity=0.70] (100.53, 47.61) circle (  1.96);
\end{scope}
\begin{scope}
\path[clip] (  0.00,  0.00) rectangle (108.41, 93.95);
\definecolor{drawColor}{gray}{0.10}

\node[text=drawColor,anchor=base east,inner sep=0pt, outer sep=0pt, scale=  0.55] at ( 18.35, 21.52) {0.7};

\node[text=drawColor,anchor=base east,inner sep=0pt, outer sep=0pt, scale=  0.55] at ( 18.35, 30.81) {0.9};

\node[text=drawColor,anchor=base east,inner sep=0pt, outer sep=0pt, scale=  0.55] at ( 18.35, 40.10) {1.1};

\node[text=drawColor,anchor=base east,inner sep=0pt, outer sep=0pt, scale=  0.55] at ( 18.35, 49.39) {1.3};

\node[text=drawColor,anchor=base east,inner sep=0pt, outer sep=0pt, scale=  0.55] at ( 18.35, 58.67) {1.5};

\node[text=drawColor,anchor=base east,inner sep=0pt, outer sep=0pt, scale=  0.55] at ( 18.35, 67.96) {1.7};

\node[text=drawColor,anchor=base east,inner sep=0pt, outer sep=0pt, scale=  0.55] at ( 18.35, 77.25) {1.9};

\node[text=drawColor,anchor=base east,inner sep=0pt, outer sep=0pt, scale=  0.55] at ( 18.35, 86.54) {2.1};
\end{scope}
\begin{scope}
\path[clip] (  0.00,  0.00) rectangle (108.41, 93.95);
\definecolor{drawColor}{gray}{0.10}

\node[text=drawColor,anchor=base,inner sep=0pt, outer sep=0pt, scale=  0.55] at ( 21.48, 15.15) {-3.2};

\node[text=drawColor,anchor=base,inner sep=0pt, outer sep=0pt, scale=  0.55] at ( 37.29, 15.15) {-3.0};

\node[text=drawColor,anchor=base,inner sep=0pt, outer sep=0pt, scale=  0.55] at ( 53.10, 15.15) {-2.8};

\node[text=drawColor,anchor=base,inner sep=0pt, outer sep=0pt, scale=  0.55] at ( 68.91, 15.15) {-2.6};

\node[text=drawColor,anchor=base,inner sep=0pt, outer sep=0pt, scale=  0.55] at ( 84.72, 15.15) {-2.4};
\end{scope}
\begin{scope}
\path[clip] (  0.00,  0.00) rectangle (108.41, 93.95);
\definecolor{drawColor}{gray}{0.10}

\node[text=drawColor,anchor=base,inner sep=0pt, outer sep=0pt, scale=  0.66] at ( 61.00,  6.78) {$log_{10}(\sigma)$};
\end{scope}
\begin{scope}
\path[clip] (  0.00,  0.00) rectangle (108.41, 93.95);
\definecolor{drawColor}{gray}{0.10}

\node[text=drawColor,rotate= 90.00,anchor=base,inner sep=0pt, outer sep=0pt, scale=  0.66] at ( 10.05, 53.67) {Sampson error (pixel)};
\end{scope}
\begin{scope}
\path[clip] (  0.00,  0.00) rectangle (108.41, 93.95);
\definecolor{drawColor}{RGB}{0,0,0}
\definecolor{fillColor}{RGB}{190,190,190}

\path[draw=drawColor,draw opacity=0.70,line width= 0.4pt,line join=round,line cap=round,fill=fillColor,fill opacity=0.70] ( 49.08, 81.83) --
	( 51.73, 77.26) --
	( 46.44, 77.26) --
	cycle;
\end{scope}
\begin{scope}
\path[clip] (  0.00,  0.00) rectangle (108.41, 93.95);
\definecolor{drawColor}{RGB}{0,0,0}
\definecolor{fillColor}{RGB}{173,216,230}

\path[draw=drawColor,draw opacity=0.70,line width= 0.4pt,line join=round,line cap=round,fill=fillColor,fill opacity=0.70] ( 47.34, 72.38) rectangle ( 50.82, 75.86);
\end{scope}
\begin{scope}
\path[clip] (  0.00,  0.00) rectangle (108.41, 93.95);
\definecolor{drawColor}{RGB}{0,0,0}

\path[draw=drawColor,draw opacity=0.70,line width= 0.4pt,line join=round,line cap=round] ( 47.12, 67.49) rectangle ( 51.05, 71.42);
\end{scope}
\begin{scope}
\path[clip] (  0.00,  0.00) rectangle (108.41, 93.95);
\definecolor{drawColor}{RGB}{0,0,0}
\definecolor{fillColor}{RGB}{255,0,0}

\path[draw=drawColor,draw opacity=0.70,line width= 0.4pt,line join=round,line cap=round,fill=fillColor,fill opacity=0.70] ( 49.08, 64.79) circle (  1.96);
\end{scope}
\begin{scope}
\path[clip] (  0.00,  0.00) rectangle (108.41, 93.95);
\definecolor{drawColor}{RGB}{0,0,0}

\node[text=drawColor,anchor=base west,inner sep=0pt, outer sep=0pt, scale=  0.53] at ( 53.39, 76.96) {R-GC};
\end{scope}
\begin{scope}
\path[clip] (  0.00,  0.00) rectangle (108.41, 93.95);
\definecolor{drawColor}{RGB}{0,0,0}

\node[text=drawColor,anchor=base west,inner sep=0pt, outer sep=0pt, scale=  0.53] at ( 53.39, 72.30) {R-DLT};
\end{scope}
\begin{scope}
\path[clip] (  0.00,  0.00) rectangle (108.41, 93.95);
\definecolor{drawColor}{RGB}{0,0,0}

\node[text=drawColor,anchor=base west,inner sep=0pt, outer sep=0pt, scale=  0.53] at ( 53.39, 67.64) {R-Huber};
\end{scope}
\begin{scope}
\path[clip] (  0.00,  0.00) rectangle (108.41, 93.95);
\definecolor{drawColor}{RGB}{0,0,0}

\node[text=drawColor,anchor=base west,inner sep=0pt, outer sep=0pt, scale=  0.53] at ( 53.39, 62.98) {R-DPCP};
\end{scope}
\end{tikzpicture}

%% file: Rfigures/EPFLTrans_mean.tex
\begin{tikzpicture}[x=1pt,y=1pt]
\clip (5,3) rectangle (104, 93.95);
\definecolor{fillColor}{RGB}{255,255,255}
\path[use as bounding box,fill=fillColor,fill opacity=0.00] (0,0) rectangle (108.41, 93.95);
\begin{scope}
\path[clip] (  0.00,  0.00) rectangle (108.41, 93.95);
\definecolor{drawColor}{RGB}{255,255,255}
\definecolor{fillColor}{RGB}{255,255,255}

\path[draw=drawColor,line width= 0.6pt,line join=round,line cap=round,fill=fillColor] ( -0.00,  0.00) rectangle (108.41, 93.95);
\end{scope}
\begin{scope}
\path[clip] ( 19.10, 18.89) rectangle (102.90, 88.45);
\definecolor{fillColor}{gray}{0.92}

\path[fill=fillColor] ( 19.10, 18.89) rectangle (102.90, 88.45);
\definecolor{drawColor}{RGB}{0,0,0}

\path[draw=drawColor,line width= 0.6pt,line join=round] ( 20.72, 43.65) --
	( 26.09, 46.38) --
	( 31.46, 48.42) --
	( 36.83, 49.93) --
	( 42.20, 51.01) --
	( 47.58, 52.08) --
	( 52.95, 53.83) --
	( 58.32, 56.12) --
	( 63.69, 58.57) --
	( 69.06, 61.26) --
	( 74.43, 63.39) --
	( 79.81, 66.55) --
	( 85.18, 70.33) --
	( 90.55, 74.06) --
	( 95.92, 78.93) --
	(101.29, 85.29);

\path[draw=drawColor,line width= 0.6pt,line join=round] ( 20.72, 24.23) --
	( 26.09, 24.20) --
	( 31.46, 23.47) --
	( 36.83, 22.47) --
	( 42.20, 22.56) --
	( 47.58, 22.73) --
	( 52.95, 23.91) --
	( 58.32, 25.22) --
	( 63.69, 26.83) --
	( 69.06, 28.91) --
	( 74.43, 31.55) --
	( 79.81, 35.92) --
	( 85.18, 41.41) --
	( 90.55, 47.67) --
	( 95.92, 55.07) --
	(101.29, 67.04);

\path[draw=drawColor,line width= 0.6pt,line join=round] ( 20.72, 25.04) --
	( 26.09, 24.73) --
	( 31.46, 24.46) --
	( 36.83, 24.66) --
	( 42.20, 25.00) --
	( 47.58, 25.62) --
	( 52.95, 26.43) --
	( 58.32, 27.73) --
	( 63.69, 28.35) --
	( 69.06, 30.66) --
	( 74.43, 33.28) --
	( 79.81, 36.61) --
	( 85.18, 41.94) --
	( 90.55, 46.78) --
	( 95.92, 50.99) --
	(101.29, 57.69);

\path[draw=drawColor,line width= 0.6pt,line join=round] ( 20.72, 22.82) --
	( 26.09, 22.65) --
	( 31.46, 22.23) --
	( 36.83, 22.41) --
	( 42.20, 22.05) --
	( 47.58, 22.71) --
	( 52.95, 22.95) --
	( 58.32, 23.44) --
	( 63.69, 24.53) --
	( 69.06, 24.51) --
	( 74.43, 25.37) --
	( 79.81, 25.35) --
	( 85.18, 27.74) --
	( 90.55, 31.42) --
	( 95.92, 34.93) --
	(101.29, 40.36);
\definecolor{drawColor}{RGB}{0,0,0}
\definecolor{fillColor}{RGB}{190,190,190}

\path[draw=drawColor,draw opacity=0.70,line width= 0.4pt,line join=round,line cap=round,fill=fillColor,fill opacity=0.70] ( 20.72, 46.70) --
	( 23.36, 42.13) --
	( 18.07, 42.13) --
	cycle;

\path[draw=drawColor,draw opacity=0.70,line width= 0.4pt,line join=round,line cap=round,fill=fillColor,fill opacity=0.70] ( 26.09, 49.43) --
	( 28.73, 44.86) --
	( 23.45, 44.86) --
	cycle;

\path[draw=drawColor,draw opacity=0.70,line width= 0.4pt,line join=round,line cap=round,fill=fillColor,fill opacity=0.70] ( 31.46, 51.47) --
	( 34.10, 46.89) --
	( 28.82, 46.89) --
	cycle;

\path[draw=drawColor,draw opacity=0.70,line width= 0.4pt,line join=round,line cap=round,fill=fillColor,fill opacity=0.70] ( 36.83, 52.98) --
	( 39.47, 48.40) --
	( 34.19, 48.40) --
	cycle;

\path[draw=drawColor,draw opacity=0.70,line width= 0.4pt,line join=round,line cap=round,fill=fillColor,fill opacity=0.70] ( 42.20, 54.06) --
	( 44.85, 49.48) --
	( 39.56, 49.48) --
	cycle;

\path[draw=drawColor,draw opacity=0.70,line width= 0.4pt,line join=round,line cap=round,fill=fillColor,fill opacity=0.70] ( 47.58, 55.13) --
	( 50.22, 50.56) --
	( 44.93, 50.56) --
	cycle;

\path[draw=drawColor,draw opacity=0.70,line width= 0.4pt,line join=round,line cap=round,fill=fillColor,fill opacity=0.70] ( 52.95, 56.88) --
	( 55.59, 52.30) --
	( 50.30, 52.30) --
	cycle;

\path[draw=drawColor,draw opacity=0.70,line width= 0.4pt,line join=round,line cap=round,fill=fillColor,fill opacity=0.70] ( 58.32, 59.17) --
	( 60.96, 54.60) --
	( 55.68, 54.60) --
	cycle;

\path[draw=drawColor,draw opacity=0.70,line width= 0.4pt,line join=round,line cap=round,fill=fillColor,fill opacity=0.70] ( 63.69, 61.62) --
	( 66.33, 57.04) --
	( 61.05, 57.04) --
	cycle;

\path[draw=drawColor,draw opacity=0.70,line width= 0.4pt,line join=round,line cap=round,fill=fillColor,fill opacity=0.70] ( 69.06, 64.32) --
	( 71.71, 59.74) --
	( 66.42, 59.74) --
	cycle;

\path[draw=drawColor,draw opacity=0.70,line width= 0.4pt,line join=round,line cap=round,fill=fillColor,fill opacity=0.70] ( 74.43, 66.44) --
	( 77.08, 61.86) --
	( 71.79, 61.86) --
	cycle;

\path[draw=drawColor,draw opacity=0.70,line width= 0.4pt,line join=round,line cap=round,fill=fillColor,fill opacity=0.70] ( 79.81, 69.60) --
	( 82.45, 65.02) --
	( 77.16, 65.02) --
	cycle;

\path[draw=drawColor,draw opacity=0.70,line width= 0.4pt,line join=round,line cap=round,fill=fillColor,fill opacity=0.70] ( 85.18, 73.38) --
	( 87.82, 68.80) --
	( 82.54, 68.80) --
	cycle;

\path[draw=drawColor,draw opacity=0.70,line width= 0.4pt,line join=round,line cap=round,fill=fillColor,fill opacity=0.70] ( 90.55, 77.11) --
	( 93.19, 72.53) --
	( 87.91, 72.53) --
	cycle;

\path[draw=drawColor,draw opacity=0.70,line width= 0.4pt,line join=round,line cap=round,fill=fillColor,fill opacity=0.70] ( 95.92, 81.98) --
	( 98.56, 77.40) --
	( 93.28, 77.40) --
	cycle;

\path[draw=drawColor,draw opacity=0.70,line width= 0.4pt,line join=round,line cap=round,fill=fillColor,fill opacity=0.70] (101.29, 88.34) --
	(103.94, 83.76) --
	( 98.65, 83.76) --
	cycle;
\definecolor{fillColor}{RGB}{173,216,230}

\path[draw=drawColor,draw opacity=0.70,line width= 0.4pt,line join=round,line cap=round,fill=fillColor,fill opacity=0.70] ( 18.98, 22.49) rectangle ( 22.46, 25.97);

\path[draw=drawColor,draw opacity=0.70,line width= 0.4pt,line join=round,line cap=round,fill=fillColor,fill opacity=0.70] ( 24.35, 22.46) rectangle ( 27.83, 25.94);

\path[draw=drawColor,draw opacity=0.70,line width= 0.4pt,line join=round,line cap=round,fill=fillColor,fill opacity=0.70] ( 29.72, 21.73) rectangle ( 33.20, 25.21);

\path[draw=drawColor,draw opacity=0.70,line width= 0.4pt,line join=round,line cap=round,fill=fillColor,fill opacity=0.70] ( 35.09, 20.73) rectangle ( 38.57, 24.21);

\path[draw=drawColor,draw opacity=0.70,line width= 0.4pt,line join=round,line cap=round,fill=fillColor,fill opacity=0.70] ( 40.46, 20.82) rectangle ( 43.94, 24.30);

\path[draw=drawColor,draw opacity=0.70,line width= 0.4pt,line join=round,line cap=round,fill=fillColor,fill opacity=0.70] ( 45.84, 20.99) rectangle ( 49.31, 24.47);

\path[draw=drawColor,draw opacity=0.70,line width= 0.4pt,line join=round,line cap=round,fill=fillColor,fill opacity=0.70] ( 51.21, 22.17) rectangle ( 54.69, 25.65);

\path[draw=drawColor,draw opacity=0.70,line width= 0.4pt,line join=round,line cap=round,fill=fillColor,fill opacity=0.70] ( 56.58, 23.48) rectangle ( 60.06, 26.95);

\path[draw=drawColor,draw opacity=0.70,line width= 0.4pt,line join=round,line cap=round,fill=fillColor,fill opacity=0.70] ( 61.95, 25.09) rectangle ( 65.43, 28.57);

\path[draw=drawColor,draw opacity=0.70,line width= 0.4pt,line join=round,line cap=round,fill=fillColor,fill opacity=0.70] ( 67.32, 27.17) rectangle ( 70.80, 30.65);

\path[draw=drawColor,draw opacity=0.70,line width= 0.4pt,line join=round,line cap=round,fill=fillColor,fill opacity=0.70] ( 72.70, 29.81) rectangle ( 76.17, 33.29);

\path[draw=drawColor,draw opacity=0.70,line width= 0.4pt,line join=round,line cap=round,fill=fillColor,fill opacity=0.70] ( 78.07, 34.19) rectangle ( 81.55, 37.66);

\path[draw=drawColor,draw opacity=0.70,line width= 0.4pt,line join=round,line cap=round,fill=fillColor,fill opacity=0.70] ( 83.44, 39.67) rectangle ( 86.92, 43.15);

\path[draw=drawColor,draw opacity=0.70,line width= 0.4pt,line join=round,line cap=round,fill=fillColor,fill opacity=0.70] ( 88.81, 45.93) rectangle ( 92.29, 49.41);

\path[draw=drawColor,draw opacity=0.70,line width= 0.4pt,line join=round,line cap=round,fill=fillColor,fill opacity=0.70] ( 94.18, 53.33) rectangle ( 97.66, 56.81);

\path[draw=drawColor,draw opacity=0.70,line width= 0.4pt,line join=round,line cap=round,fill=fillColor,fill opacity=0.70] ( 99.55, 65.30) rectangle (103.03, 68.78);

\path[draw=drawColor,draw opacity=0.70,line width= 0.4pt,line join=round,line cap=round] ( 18.75, 23.08) rectangle ( 22.68, 27.01);

\path[draw=drawColor,draw opacity=0.70,line width= 0.4pt,line join=round,line cap=round] ( 24.13, 22.76) rectangle ( 28.05, 26.69);

\path[draw=drawColor,draw opacity=0.70,line width= 0.4pt,line join=round,line cap=round] ( 29.50, 22.50) rectangle ( 33.42, 26.42);

\path[draw=drawColor,draw opacity=0.70,line width= 0.4pt,line join=round,line cap=round] ( 34.87, 22.70) rectangle ( 38.79, 26.62);

\path[draw=drawColor,draw opacity=0.70,line width= 0.4pt,line join=round,line cap=round] ( 40.24, 23.04) rectangle ( 44.17, 26.97);

\path[draw=drawColor,draw opacity=0.70,line width= 0.4pt,line join=round,line cap=round] ( 45.61, 23.66) rectangle ( 49.54, 27.58);

\path[draw=drawColor,draw opacity=0.70,line width= 0.4pt,line join=round,line cap=round] ( 50.99, 24.46) rectangle ( 54.91, 28.39);

\path[draw=drawColor,draw opacity=0.70,line width= 0.4pt,line join=round,line cap=round] ( 56.36, 25.77) rectangle ( 60.28, 29.69);

\path[draw=drawColor,draw opacity=0.70,line width= 0.4pt,line join=round,line cap=round] ( 61.73, 26.39) rectangle ( 65.65, 30.32);

\path[draw=drawColor,draw opacity=0.70,line width= 0.4pt,line join=round,line cap=round] ( 67.10, 28.70) rectangle ( 71.02, 32.63);

\path[draw=drawColor,draw opacity=0.70,line width= 0.4pt,line join=round,line cap=round] ( 72.47, 31.32) rectangle ( 76.40, 35.24);

\path[draw=drawColor,draw opacity=0.70,line width= 0.4pt,line join=round,line cap=round] ( 77.84, 34.65) rectangle ( 81.77, 38.57);

\path[draw=drawColor,draw opacity=0.70,line width= 0.4pt,line join=round,line cap=round] ( 83.22, 39.98) rectangle ( 87.14, 43.90);

\path[draw=drawColor,draw opacity=0.70,line width= 0.4pt,line join=round,line cap=round] ( 88.59, 44.81) rectangle ( 92.51, 48.74);

\path[draw=drawColor,draw opacity=0.70,line width= 0.4pt,line join=round,line cap=round] ( 93.96, 49.03) rectangle ( 97.88, 52.96);

\path[draw=drawColor,draw opacity=0.70,line width= 0.4pt,line join=round,line cap=round] ( 99.33, 55.73) rectangle (103.26, 59.65);
\definecolor{fillColor}{RGB}{255,0,0}

\path[draw=drawColor,draw opacity=0.70,line width= 0.4pt,line join=round,line cap=round,fill=fillColor,fill opacity=0.70] ( 20.72, 22.82) circle (  1.96);

\path[draw=drawColor,draw opacity=0.70,line width= 0.4pt,line join=round,line cap=round,fill=fillColor,fill opacity=0.70] ( 26.09, 22.65) circle (  1.96);

\path[draw=drawColor,draw opacity=0.70,line width= 0.4pt,line join=round,line cap=round,fill=fillColor,fill opacity=0.70] ( 31.46, 22.23) circle (  1.96);

\path[draw=drawColor,draw opacity=0.70,line width= 0.4pt,line join=round,line cap=round,fill=fillColor,fill opacity=0.70] ( 36.83, 22.41) circle (  1.96);

\path[draw=drawColor,draw opacity=0.70,line width= 0.4pt,line join=round,line cap=round,fill=fillColor,fill opacity=0.70] ( 42.20, 22.05) circle (  1.96);

\path[draw=drawColor,draw opacity=0.70,line width= 0.4pt,line join=round,line cap=round,fill=fillColor,fill opacity=0.70] ( 47.58, 22.71) circle (  1.96);

\path[draw=drawColor,draw opacity=0.70,line width= 0.4pt,line join=round,line cap=round,fill=fillColor,fill opacity=0.70] ( 52.95, 22.95) circle (  1.96);

\path[draw=drawColor,draw opacity=0.70,line width= 0.4pt,line join=round,line cap=round,fill=fillColor,fill opacity=0.70] ( 58.32, 23.44) circle (  1.96);

\path[draw=drawColor,draw opacity=0.70,line width= 0.4pt,line join=round,line cap=round,fill=fillColor,fill opacity=0.70] ( 63.69, 24.53) circle (  1.96);

\path[draw=drawColor,draw opacity=0.70,line width= 0.4pt,line join=round,line cap=round,fill=fillColor,fill opacity=0.70] ( 69.06, 24.51) circle (  1.96);

\path[draw=drawColor,draw opacity=0.70,line width= 0.4pt,line join=round,line cap=round,fill=fillColor,fill opacity=0.70] ( 74.43, 25.37) circle (  1.96);

\path[draw=drawColor,draw opacity=0.70,line width= 0.4pt,line join=round,line cap=round,fill=fillColor,fill opacity=0.70] ( 79.81, 25.35) circle (  1.96);

\path[draw=drawColor,draw opacity=0.70,line width= 0.4pt,line join=round,line cap=round,fill=fillColor,fill opacity=0.70] ( 85.18, 27.74) circle (  1.96);

\path[draw=drawColor,draw opacity=0.70,line width= 0.4pt,line join=round,line cap=round,fill=fillColor,fill opacity=0.70] ( 90.55, 31.42) circle (  1.96);

\path[draw=drawColor,draw opacity=0.70,line width= 0.4pt,line join=round,line cap=round,fill=fillColor,fill opacity=0.70] ( 95.92, 34.93) circle (  1.96);

\path[draw=drawColor,draw opacity=0.70,line width= 0.4pt,line join=round,line cap=round,fill=fillColor,fill opacity=0.70] (101.29, 40.36) circle (  1.96);
\end{scope}
\begin{scope}
\path[clip] (  0.00,  0.00) rectangle (108.41, 93.95);
\definecolor{drawColor}{gray}{0.10}

\node[text=drawColor,anchor=base east,inner sep=0pt, outer sep=0pt, scale=  0.55] at ( 18.35, 24.79) {0.3};

\node[text=drawColor,anchor=base east,inner sep=0pt, outer sep=0pt, scale=  0.55] at ( 18.35, 34.02) {0.4};

\node[text=drawColor,anchor=base east,inner sep=0pt, outer sep=0pt, scale=  0.55] at ( 18.35, 43.24) {0.5};

\node[text=drawColor,anchor=base east,inner sep=0pt, outer sep=0pt, scale=  0.55] at ( 18.35, 52.47) {0.6};

\node[text=drawColor,anchor=base east,inner sep=0pt, outer sep=0pt, scale=  0.55] at ( 18.35, 61.69) {0.7};

\node[text=drawColor,anchor=base east,inner sep=0pt, outer sep=0pt, scale=  0.55] at ( 18.35, 70.92) {0.8};

\node[text=drawColor,anchor=base east,inner sep=0pt, outer sep=0pt, scale=  0.55] at ( 18.35, 80.14) {0.9};
\end{scope}
\begin{scope}
\path[clip] (  0.00,  0.00) rectangle (108.41, 93.95);
\definecolor{drawColor}{gray}{0.10}

\node[text=drawColor,anchor=base,inner sep=0pt, outer sep=0pt, scale=  0.55] at ( 26.09, 15.15) {-3.8};

\node[text=drawColor,anchor=base,inner sep=0pt, outer sep=0pt, scale=  0.55] at ( 36.83, 15.15) {-3.6};

\node[text=drawColor,anchor=base,inner sep=0pt, outer sep=0pt, scale=  0.55] at ( 47.58, 15.15) {-3.4};

\node[text=drawColor,anchor=base,inner sep=0pt, outer sep=0pt, scale=  0.55] at ( 58.32, 15.15) {-3.2};

\node[text=drawColor,anchor=base,inner sep=0pt, outer sep=0pt, scale=  0.55] at ( 69.06, 15.15) {-3.0};

\node[text=drawColor,anchor=base,inner sep=0pt, outer sep=0pt, scale=  0.55] at ( 79.81, 15.15) {-2.8};

\node[text=drawColor,anchor=base,inner sep=0pt, outer sep=0pt, scale=  0.55] at ( 90.55, 15.15) {-2.6};
\end{scope}
\begin{scope}
\path[clip] (  0.00,  0.00) rectangle (108.41, 93.95);
\definecolor{drawColor}{gray}{0.10}

\node[text=drawColor,anchor=base,inner sep=0pt, outer sep=0pt, scale=  0.66] at ( 61.00,  6.78) {$log_{10}(\sigma)$};
\end{scope}
\begin{scope}
\path[clip] (  0.00,  0.00) rectangle (108.41, 93.95);
\definecolor{drawColor}{gray}{0.10}

\node[text=drawColor,rotate= 90.00,anchor=base,inner sep=0pt, outer sep=0pt, scale=  0.66] at ( 10.05, 53.67) {Sampson error (pixel)};
\end{scope}
\end{tikzpicture}

%% file: Rfigures/KUSVOD2Time_mean.tex
\begin{tikzpicture}[x=1pt,y=1pt]
\clip (5,3) rectangle (104, 93.95);
\definecolor{fillColor}{RGB}{255,255,255}
\path[use as bounding box,fill=fillColor,fill opacity=0.00] (0,0) rectangle (108.41, 93.95);
\begin{scope}
\path[clip] (  0.00,  0.00) rectangle (108.41, 93.95);
\definecolor{drawColor}{RGB}{255,255,255}
\definecolor{fillColor}{RGB}{255,255,255}

\path[draw=drawColor,line width= 0.6pt,line join=round,line cap=round,fill=fillColor] ( -0.00,  0.00) rectangle (108.41, 93.95);
\end{scope}
\begin{scope}
\path[clip] ( 19.10, 18.89) rectangle (102.90, 88.45);
\definecolor{fillColor}{gray}{0.92}

\path[fill=fillColor] ( 19.10, 18.89) rectangle (102.90, 88.45);
\definecolor{drawColor}{RGB}{0,0,0}

\path[draw=drawColor,line width= 0.6pt,line join=round] ( 21.48, 50.81) --
	( 29.38, 50.70) --
	( 37.29, 46.94) --
	( 45.19, 47.82) --
	( 53.10, 46.73) --
	( 61.00, 44.82) --
	( 68.91, 42.54) --
	( 76.82, 46.19) --
	( 84.72, 43.45) --
	( 92.63, 44.92) --
	(100.53, 38.40);

\path[draw=drawColor,line width= 0.6pt,line join=round] ( 21.48, 32.08) --
	( 29.38, 30.34) --
	( 37.29, 29.73) --
	( 45.19, 29.29) --
	( 53.10, 28.45) --
	( 61.00, 27.34) --
	( 68.91, 27.62) --
	( 76.82, 25.88) --
	( 84.72, 26.37) --
	( 92.63, 25.29) --
	(100.53, 22.05);

\path[draw=drawColor,line width= 0.6pt,line join=round] ( 21.48, 70.65) --
	( 29.38, 70.19) --
	( 37.29, 71.45) --
	( 45.19, 71.86) --
	( 53.10, 70.14) --
	( 61.00, 73.06) --
	( 68.91, 72.23) --
	( 76.82, 74.02) --
	( 84.72, 72.50) --
	( 92.63, 73.24) --
	(100.53, 60.50);

\path[draw=drawColor,line width= 0.6pt,line join=round] ( 21.48, 79.92) --
	( 29.38, 85.29) --
	( 37.29, 78.39) --
	( 45.19, 79.28) --
	( 53.10, 78.28) --
	( 61.00, 73.93) --
	( 68.91, 73.15) --
	( 76.82, 68.01) --
	( 84.72, 69.28) --
	( 92.63, 73.72) --
	(100.53, 75.84);
\definecolor{drawColor}{RGB}{0,0,0}
\definecolor{fillColor}{RGB}{190,190,190}

\path[draw=drawColor,draw opacity=0.70,line width= 0.4pt,line join=round,line cap=round,fill=fillColor,fill opacity=0.70] ( 21.48, 53.86) --
	( 24.12, 49.29) --
	( 18.83, 49.29) --
	cycle;

\path[draw=drawColor,draw opacity=0.70,line width= 0.4pt,line join=round,line cap=round,fill=fillColor,fill opacity=0.70] ( 29.38, 53.75) --
	( 32.02, 49.17) --
	( 26.74, 49.17) --
	cycle;

\path[draw=drawColor,draw opacity=0.70,line width= 0.4pt,line join=round,line cap=round,fill=fillColor,fill opacity=0.70] ( 37.29, 49.99) --
	( 39.93, 45.42) --
	( 34.65, 45.42) --
	cycle;

\path[draw=drawColor,draw opacity=0.70,line width= 0.4pt,line join=round,line cap=round,fill=fillColor,fill opacity=0.70] ( 45.19, 50.87) --
	( 47.84, 46.29) --
	( 42.55, 46.29) --
	cycle;

\path[draw=drawColor,draw opacity=0.70,line width= 0.4pt,line join=round,line cap=round,fill=fillColor,fill opacity=0.70] ( 53.10, 49.78) --
	( 55.74, 45.20) --
	( 50.46, 45.20) --
	cycle;

\path[draw=drawColor,draw opacity=0.70,line width= 0.4pt,line join=round,line cap=round,fill=fillColor,fill opacity=0.70] ( 61.00, 47.87) --
	( 63.65, 43.29) --
	( 58.36, 43.29) --
	cycle;

\path[draw=drawColor,draw opacity=0.70,line width= 0.4pt,line join=round,line cap=round,fill=fillColor,fill opacity=0.70] ( 68.91, 45.60) --
	( 71.55, 41.02) --
	( 66.27, 41.02) --
	cycle;

\path[draw=drawColor,draw opacity=0.70,line width= 0.4pt,line join=round,line cap=round,fill=fillColor,fill opacity=0.70] ( 76.82, 49.24) --
	( 79.46, 44.67) --
	( 74.17, 44.67) --
	cycle;

\path[draw=drawColor,draw opacity=0.70,line width= 0.4pt,line join=round,line cap=round,fill=fillColor,fill opacity=0.70] ( 84.72, 46.50) --
	( 87.36, 41.92) --
	( 82.08, 41.92) --
	cycle;

\path[draw=drawColor,draw opacity=0.70,line width= 0.4pt,line join=round,line cap=round,fill=fillColor,fill opacity=0.70] ( 92.63, 47.97) --
	( 95.27, 43.39) --
	( 89.99, 43.39) --
	cycle;

\path[draw=drawColor,draw opacity=0.70,line width= 0.4pt,line join=round,line cap=round,fill=fillColor,fill opacity=0.70] (100.53, 41.45) --
	(103.18, 36.87) --
	( 97.89, 36.87) --
	cycle;
\definecolor{fillColor}{RGB}{173,216,230}

\path[draw=drawColor,draw opacity=0.70,line width= 0.4pt,line join=round,line cap=round,fill=fillColor,fill opacity=0.70] ( 19.74, 30.34) rectangle ( 23.22, 33.82);

\path[draw=drawColor,draw opacity=0.70,line width= 0.4pt,line join=round,line cap=round,fill=fillColor,fill opacity=0.70] ( 27.64, 28.60) rectangle ( 31.12, 32.08);

\path[draw=drawColor,draw opacity=0.70,line width= 0.4pt,line join=round,line cap=round,fill=fillColor,fill opacity=0.70] ( 35.55, 28.00) rectangle ( 39.03, 31.47);

\path[draw=drawColor,draw opacity=0.70,line width= 0.4pt,line join=round,line cap=round,fill=fillColor,fill opacity=0.70] ( 43.45, 27.55) rectangle ( 46.93, 31.02);

\path[draw=drawColor,draw opacity=0.70,line width= 0.4pt,line join=round,line cap=round,fill=fillColor,fill opacity=0.70] ( 51.36, 26.72) rectangle ( 54.84, 30.19);

\path[draw=drawColor,draw opacity=0.70,line width= 0.4pt,line join=round,line cap=round,fill=fillColor,fill opacity=0.70] ( 59.27, 25.60) rectangle ( 62.74, 29.08);

\path[draw=drawColor,draw opacity=0.70,line width= 0.4pt,line join=round,line cap=round,fill=fillColor,fill opacity=0.70] ( 67.17, 25.88) rectangle ( 70.65, 29.36);

\path[draw=drawColor,draw opacity=0.70,line width= 0.4pt,line join=round,line cap=round,fill=fillColor,fill opacity=0.70] ( 75.08, 24.14) rectangle ( 78.56, 27.62);

\path[draw=drawColor,draw opacity=0.70,line width= 0.4pt,line join=round,line cap=round,fill=fillColor,fill opacity=0.70] ( 82.98, 24.63) rectangle ( 86.46, 28.11);

\path[draw=drawColor,draw opacity=0.70,line width= 0.4pt,line join=round,line cap=round,fill=fillColor,fill opacity=0.70] ( 90.89, 23.55) rectangle ( 94.37, 27.03);

\path[draw=drawColor,draw opacity=0.70,line width= 0.4pt,line join=round,line cap=round,fill=fillColor,fill opacity=0.70] ( 98.79, 20.31) rectangle (102.27, 23.79);

\path[draw=drawColor,draw opacity=0.70,line width= 0.4pt,line join=round,line cap=round] ( 19.51, 68.69) rectangle ( 23.44, 72.61);

\path[draw=drawColor,draw opacity=0.70,line width= 0.4pt,line join=round,line cap=round] ( 27.42, 68.23) rectangle ( 31.34, 72.15);

\path[draw=drawColor,draw opacity=0.70,line width= 0.4pt,line join=round,line cap=round] ( 35.33, 69.49) rectangle ( 39.25, 73.42);

\path[draw=drawColor,draw opacity=0.70,line width= 0.4pt,line join=round,line cap=round] ( 43.23, 69.90) rectangle ( 47.16, 73.82);

\path[draw=drawColor,draw opacity=0.70,line width= 0.4pt,line join=round,line cap=round] ( 51.14, 68.18) rectangle ( 55.06, 72.10);

\path[draw=drawColor,draw opacity=0.70,line width= 0.4pt,line join=round,line cap=round] ( 59.04, 71.10) rectangle ( 62.97, 75.02);

\path[draw=drawColor,draw opacity=0.70,line width= 0.4pt,line join=round,line cap=round] ( 66.95, 70.27) rectangle ( 70.87, 74.20);

\path[draw=drawColor,draw opacity=0.70,line width= 0.4pt,line join=round,line cap=round] ( 74.85, 72.06) rectangle ( 78.78, 75.99);

\path[draw=drawColor,draw opacity=0.70,line width= 0.4pt,line join=round,line cap=round] ( 82.76, 70.54) rectangle ( 86.68, 74.46);

\path[draw=drawColor,draw opacity=0.70,line width= 0.4pt,line join=round,line cap=round] ( 90.67, 71.28) rectangle ( 94.59, 75.20);

\path[draw=drawColor,draw opacity=0.70,line width= 0.4pt,line join=round,line cap=round] ( 98.57, 58.53) rectangle (102.50, 62.46);
\definecolor{fillColor}{RGB}{255,0,0}

\path[draw=drawColor,draw opacity=0.70,line width= 0.4pt,line join=round,line cap=round,fill=fillColor,fill opacity=0.70] ( 21.48, 79.92) circle (  1.96);

\path[draw=drawColor,draw opacity=0.70,line width= 0.4pt,line join=round,line cap=round,fill=fillColor,fill opacity=0.70] ( 29.38, 85.29) circle (  1.96);

\path[draw=drawColor,draw opacity=0.70,line width= 0.4pt,line join=round,line cap=round,fill=fillColor,fill opacity=0.70] ( 37.29, 78.39) circle (  1.96);

\path[draw=drawColor,draw opacity=0.70,line width= 0.4pt,line join=round,line cap=round,fill=fillColor,fill opacity=0.70] ( 45.19, 79.28) circle (  1.96);

\path[draw=drawColor,draw opacity=0.70,line width= 0.4pt,line join=round,line cap=round,fill=fillColor,fill opacity=0.70] ( 53.10, 78.28) circle (  1.96);

\path[draw=drawColor,draw opacity=0.70,line width= 0.4pt,line join=round,line cap=round,fill=fillColor,fill opacity=0.70] ( 61.00, 73.93) circle (  1.96);

\path[draw=drawColor,draw opacity=0.70,line width= 0.4pt,line join=round,line cap=round,fill=fillColor,fill opacity=0.70] ( 68.91, 73.15) circle (  1.96);

\path[draw=drawColor,draw opacity=0.70,line width= 0.4pt,line join=round,line cap=round,fill=fillColor,fill opacity=0.70] ( 76.82, 68.01) circle (  1.96);

\path[draw=drawColor,draw opacity=0.70,line width= 0.4pt,line join=round,line cap=round,fill=fillColor,fill opacity=0.70] ( 84.72, 69.28) circle (  1.96);

\path[draw=drawColor,draw opacity=0.70,line width= 0.4pt,line join=round,line cap=round,fill=fillColor,fill opacity=0.70] ( 92.63, 73.72) circle (  1.96);

\path[draw=drawColor,draw opacity=0.70,line width= 0.4pt,line join=round,line cap=round,fill=fillColor,fill opacity=0.70] (100.53, 75.84) circle (  1.96);
\end{scope}
\begin{scope}
\path[clip] (  0.00,  0.00) rectangle (108.41, 93.95);
\definecolor{drawColor}{gray}{0.10}

\node[text=drawColor,anchor=base east,inner sep=0pt, outer sep=0pt, scale=  0.55] at ( 18.35, 34.06) {0.3};

\node[text=drawColor,anchor=base east,inner sep=0pt, outer sep=0pt, scale=  0.55] at ( 18.35, 67.42) {0.7};
\end{scope}
\begin{scope}
\path[clip] (  0.00,  0.00) rectangle (108.41, 93.95);
\definecolor{drawColor}{gray}{0.10}

\node[text=drawColor,anchor=base,inner sep=0pt, outer sep=0pt, scale=  0.55] at ( 21.48, 15.15) {-3.2};

\node[text=drawColor,anchor=base,inner sep=0pt, outer sep=0pt, scale=  0.55] at ( 37.29, 15.15) {-3.0};

\node[text=drawColor,anchor=base,inner sep=0pt, outer sep=0pt, scale=  0.55] at ( 53.10, 15.15) {-2.8};

\node[text=drawColor,anchor=base,inner sep=0pt, outer sep=0pt, scale=  0.55] at ( 68.91, 15.15) {-2.6};

\node[text=drawColor,anchor=base,inner sep=0pt, outer sep=0pt, scale=  0.55] at ( 84.72, 15.15) {-2.4};
\end{scope}
\begin{scope}
\path[clip] (  0.00,  0.00) rectangle (108.41, 93.95);
\definecolor{drawColor}{gray}{0.10}

\node[text=drawColor,anchor=base,inner sep=0pt, outer sep=0pt, scale=  0.66] at ( 61.00,  6.78) {$log_{10}(\sigma)$};
\end{scope}
\begin{scope}
\path[clip] (  0.00,  0.00) rectangle (108.41, 93.95);
\definecolor{drawColor}{gray}{0.10}

\node[text=drawColor,rotate= 90.00,anchor=base,inner sep=0pt, outer sep=0pt, scale=  0.66] at ( 10.05, 53.67) {running time (ms)};
\end{scope}
\end{tikzpicture}

%% file: Rfigures/EPFLTime_mean.tex
\begin{tikzpicture}[x=1pt,y=1pt]
\clip (5,3) rectangle (104, 93.95);
\definecolor{fillColor}{RGB}{255,255,255}
\path[use as bounding box,fill=fillColor,fill opacity=0.00] (0,0) rectangle (108.41, 93.95);
\begin{scope}
\path[clip] (  0.00,  0.00) rectangle (108.41, 93.95);
\definecolor{drawColor}{RGB}{255,255,255}
\definecolor{fillColor}{RGB}{255,255,255}

\path[draw=drawColor,line width= 0.6pt,line join=round,line cap=round,fill=fillColor] (  0.00,  0.00) rectangle (108.41, 93.95);
\end{scope}
\begin{scope}
\path[clip] ( 18.08, 18.89) rectangle (102.91, 88.45);
\definecolor{fillColor}{gray}{0.92}

\path[fill=fillColor] ( 18.08, 18.89) rectangle (102.91, 88.45);
\definecolor{drawColor}{RGB}{0,0,0}

\path[draw=drawColor,line width= 0.6pt,line join=round] ( 19.71, 85.29) --
	( 25.15, 71.73) --
	( 30.58, 67.42) --
	( 36.02, 67.80) --
	( 41.46, 69.15) --
	( 46.90, 67.27) --
	( 52.33, 64.53) --
	( 57.77, 66.23) --
	( 63.21, 64.27) --
	( 68.65, 65.12) --
	( 74.09, 63.24) --
	( 79.52, 63.34) --
	( 84.96, 62.20) --
	( 90.40, 61.95) --
	( 95.84, 63.17) --
	(101.27, 64.70);

\path[draw=drawColor,line width= 0.6pt,line join=round] ( 19.71, 26.62) --
	( 25.15, 23.66) --
	( 30.58, 23.46) --
	( 36.02, 22.05) --
	( 41.46, 24.10) --
	( 46.90, 24.49) --
	( 52.33, 24.20) --
	( 57.77, 24.74) --
	( 63.21, 25.73) --
	( 68.65, 24.37) --
	( 74.09, 23.34) --
	( 79.52, 25.34) --
	( 84.96, 24.67) --
	( 90.40, 25.64) --
	( 95.84, 25.33) --
	(101.27, 26.50);

\path[draw=drawColor,line width= 0.6pt,line join=round] ( 19.71, 28.88) --
	( 25.15, 26.79) --
	( 30.58, 25.72) --
	( 36.02, 26.84) --
	( 41.46, 28.16) --
	( 46.90, 29.00) --
	( 52.33, 28.55) --
	( 57.77, 28.30) --
	( 63.21, 27.59) --
	( 68.65, 28.40) --
	( 74.09, 30.24) --
	( 79.52, 30.90) --
	( 84.96, 29.94) --
	( 90.40, 31.52) --
	( 95.84, 30.38) --
	(101.27, 30.25);

\path[draw=drawColor,line width= 0.6pt,line join=round] ( 19.71, 25.78) --
	( 25.15, 25.05) --
	( 30.58, 28.14) --
	( 36.02, 25.38) --
	( 41.46, 24.64) --
	( 46.90, 25.63) --
	( 52.33, 26.46) --
	( 57.77, 27.15) --
	( 63.21, 25.39) --
	( 68.65, 24.74) --
	( 74.09, 24.64) --
	( 79.52, 25.43) --
	( 84.96, 24.71) --
	( 90.40, 25.00) --
	( 95.84, 25.46) --
	(101.27, 25.48);
\definecolor{drawColor}{RGB}{0,0,0}
\definecolor{fillColor}{RGB}{190,190,190}

\path[draw=drawColor,draw opacity=0.70,line width= 0.4pt,line join=round,line cap=round,fill=fillColor,fill opacity=0.70] ( 19.71, 88.34) --
	( 22.35, 83.76) --
	( 17.07, 83.76) --
	cycle;

\path[draw=drawColor,draw opacity=0.70,line width= 0.4pt,line join=round,line cap=round,fill=fillColor,fill opacity=0.70] ( 25.15, 74.78) --
	( 27.79, 70.20) --
	( 22.50, 70.20) --
	cycle;

\path[draw=drawColor,draw opacity=0.70,line width= 0.4pt,line join=round,line cap=round,fill=fillColor,fill opacity=0.70] ( 30.58, 70.47) --
	( 33.23, 65.89) --
	( 27.94, 65.89) --
	cycle;

\path[draw=drawColor,draw opacity=0.70,line width= 0.4pt,line join=round,line cap=round,fill=fillColor,fill opacity=0.70] ( 36.02, 70.85) --
	( 38.66, 66.27) --
	( 33.38, 66.27) --
	cycle;

\path[draw=drawColor,draw opacity=0.70,line width= 0.4pt,line join=round,line cap=round,fill=fillColor,fill opacity=0.70] ( 41.46, 72.20) --
	( 44.10, 67.62) --
	( 38.82, 67.62) --
	cycle;

\path[draw=drawColor,draw opacity=0.70,line width= 0.4pt,line join=round,line cap=round,fill=fillColor,fill opacity=0.70] ( 46.90, 70.32) --
	( 49.54, 65.74) --
	( 44.25, 65.74) --
	cycle;

\path[draw=drawColor,draw opacity=0.70,line width= 0.4pt,line join=round,line cap=round,fill=fillColor,fill opacity=0.70] ( 52.33, 67.58) --
	( 54.98, 63.00) --
	( 49.69, 63.00) --
	cycle;

\path[draw=drawColor,draw opacity=0.70,line width= 0.4pt,line join=round,line cap=round,fill=fillColor,fill opacity=0.70] ( 57.77, 69.28) --
	( 60.41, 64.70) --
	( 55.13, 64.70) --
	cycle;

\path[draw=drawColor,draw opacity=0.70,line width= 0.4pt,line join=round,line cap=round,fill=fillColor,fill opacity=0.70] ( 63.21, 67.32) --
	( 65.85, 62.74) --
	( 60.57, 62.74) --
	cycle;

\path[draw=drawColor,draw opacity=0.70,line width= 0.4pt,line join=round,line cap=round,fill=fillColor,fill opacity=0.70] ( 68.65, 68.17) --
	( 71.29, 63.60) --
	( 66.01, 63.60) --
	cycle;

\path[draw=drawColor,draw opacity=0.70,line width= 0.4pt,line join=round,line cap=round,fill=fillColor,fill opacity=0.70] ( 74.09, 66.29) --
	( 76.73, 61.71) --
	( 71.44, 61.71) --
	cycle;

\path[draw=drawColor,draw opacity=0.70,line width= 0.4pt,line join=round,line cap=round,fill=fillColor,fill opacity=0.70] ( 79.52, 66.39) --
	( 82.17, 61.82) --
	( 76.88, 61.82) --
	cycle;

\path[draw=drawColor,draw opacity=0.70,line width= 0.4pt,line join=round,line cap=round,fill=fillColor,fill opacity=0.70] ( 84.96, 65.25) --
	( 87.60, 60.67) --
	( 82.32, 60.67) --
	cycle;

\path[draw=drawColor,draw opacity=0.70,line width= 0.4pt,line join=round,line cap=round,fill=fillColor,fill opacity=0.70] ( 90.40, 65.00) --
	( 93.04, 60.42) --
	( 87.76, 60.42) --
	cycle;

\path[draw=drawColor,draw opacity=0.70,line width= 0.4pt,line join=round,line cap=round,fill=fillColor,fill opacity=0.70] ( 95.84, 66.22) --
	( 98.48, 61.64) --
	( 93.19, 61.64) --
	cycle;

\path[draw=drawColor,draw opacity=0.70,line width= 0.4pt,line join=round,line cap=round,fill=fillColor,fill opacity=0.70] (101.27, 67.76) --
	(103.92, 63.18) --
	( 98.63, 63.18) --
	cycle;
\definecolor{fillColor}{RGB}{173,216,230}

\path[draw=drawColor,draw opacity=0.70,line width= 0.4pt,line join=round,line cap=round,fill=fillColor,fill opacity=0.70] ( 17.97, 24.88) rectangle ( 21.45, 28.36);

\path[draw=drawColor,draw opacity=0.70,line width= 0.4pt,line join=round,line cap=round,fill=fillColor,fill opacity=0.70] ( 23.41, 21.93) rectangle ( 26.89, 25.40);

\path[draw=drawColor,draw opacity=0.70,line width= 0.4pt,line join=round,line cap=round,fill=fillColor,fill opacity=0.70] ( 28.85, 21.72) rectangle ( 32.32, 25.20);

\path[draw=drawColor,draw opacity=0.70,line width= 0.4pt,line join=round,line cap=round,fill=fillColor,fill opacity=0.70] ( 34.28, 20.31) rectangle ( 37.76, 23.79);

\path[draw=drawColor,draw opacity=0.70,line width= 0.4pt,line join=round,line cap=round,fill=fillColor,fill opacity=0.70] ( 39.72, 22.36) rectangle ( 43.20, 25.84);

\path[draw=drawColor,draw opacity=0.70,line width= 0.4pt,line join=round,line cap=round,fill=fillColor,fill opacity=0.70] ( 45.16, 22.75) rectangle ( 48.64, 26.23);

\path[draw=drawColor,draw opacity=0.70,line width= 0.4pt,line join=round,line cap=round,fill=fillColor,fill opacity=0.70] ( 50.60, 22.46) rectangle ( 54.07, 25.94);

\path[draw=drawColor,draw opacity=0.70,line width= 0.4pt,line join=round,line cap=round,fill=fillColor,fill opacity=0.70] ( 56.03, 23.01) rectangle ( 59.51, 26.48);

\path[draw=drawColor,draw opacity=0.70,line width= 0.4pt,line join=round,line cap=round,fill=fillColor,fill opacity=0.70] ( 61.47, 23.99) rectangle ( 64.95, 27.46);

\path[draw=drawColor,draw opacity=0.70,line width= 0.4pt,line join=round,line cap=round,fill=fillColor,fill opacity=0.70] ( 66.91, 22.63) rectangle ( 70.39, 26.11);

\path[draw=drawColor,draw opacity=0.70,line width= 0.4pt,line join=round,line cap=round,fill=fillColor,fill opacity=0.70] ( 72.35, 21.60) rectangle ( 75.82, 25.08);

\path[draw=drawColor,draw opacity=0.70,line width= 0.4pt,line join=round,line cap=round,fill=fillColor,fill opacity=0.70] ( 77.78, 23.60) rectangle ( 81.26, 27.07);

\path[draw=drawColor,draw opacity=0.70,line width= 0.4pt,line join=round,line cap=round,fill=fillColor,fill opacity=0.70] ( 83.22, 22.93) rectangle ( 86.70, 26.41);

\path[draw=drawColor,draw opacity=0.70,line width= 0.4pt,line join=round,line cap=round,fill=fillColor,fill opacity=0.70] ( 88.66, 23.90) rectangle ( 92.14, 27.38);

\path[draw=drawColor,draw opacity=0.70,line width= 0.4pt,line join=round,line cap=round,fill=fillColor,fill opacity=0.70] ( 94.10, 23.59) rectangle ( 97.57, 27.07);

\path[draw=drawColor,draw opacity=0.70,line width= 0.4pt,line join=round,line cap=round,fill=fillColor,fill opacity=0.70] ( 99.53, 24.76) rectangle (103.01, 28.24);

\path[draw=drawColor,draw opacity=0.70,line width= 0.4pt,line join=round,line cap=round] ( 17.75, 26.92) rectangle ( 21.67, 30.85);

\path[draw=drawColor,draw opacity=0.70,line width= 0.4pt,line join=round,line cap=round] ( 23.18, 24.83) rectangle ( 27.11, 28.75);

\path[draw=drawColor,draw opacity=0.70,line width= 0.4pt,line join=round,line cap=round] ( 28.62, 23.76) rectangle ( 32.55, 27.68);

\path[draw=drawColor,draw opacity=0.70,line width= 0.4pt,line join=round,line cap=round] ( 34.06, 24.87) rectangle ( 37.98, 28.80);

\path[draw=drawColor,draw opacity=0.70,line width= 0.4pt,line join=round,line cap=round] ( 39.50, 26.20) rectangle ( 43.42, 30.13);

\path[draw=drawColor,draw opacity=0.70,line width= 0.4pt,line join=round,line cap=round] ( 44.93, 27.03) rectangle ( 48.86, 30.96);

\path[draw=drawColor,draw opacity=0.70,line width= 0.4pt,line join=round,line cap=round] ( 50.37, 26.59) rectangle ( 54.30, 30.51);

\path[draw=drawColor,draw opacity=0.70,line width= 0.4pt,line join=round,line cap=round] ( 55.81, 26.34) rectangle ( 59.73, 30.27);

\path[draw=drawColor,draw opacity=0.70,line width= 0.4pt,line join=round,line cap=round] ( 61.25, 25.63) rectangle ( 65.17, 29.55);

\path[draw=drawColor,draw opacity=0.70,line width= 0.4pt,line join=round,line cap=round] ( 66.69, 26.44) rectangle ( 70.61, 30.36);

\path[draw=drawColor,draw opacity=0.70,line width= 0.4pt,line join=round,line cap=round] ( 72.12, 28.28) rectangle ( 76.05, 32.20);

\path[draw=drawColor,draw opacity=0.70,line width= 0.4pt,line join=round,line cap=round] ( 77.56, 28.94) rectangle ( 81.49, 32.86);

\path[draw=drawColor,draw opacity=0.70,line width= 0.4pt,line join=round,line cap=round] ( 83.00, 27.97) rectangle ( 86.92, 31.90);

\path[draw=drawColor,draw opacity=0.70,line width= 0.4pt,line join=round,line cap=round] ( 88.44, 29.55) rectangle ( 92.36, 33.48);

\path[draw=drawColor,draw opacity=0.70,line width= 0.4pt,line join=round,line cap=round] ( 93.87, 28.42) rectangle ( 97.80, 32.34);

\path[draw=drawColor,draw opacity=0.70,line width= 0.4pt,line join=round,line cap=round] ( 99.31, 28.29) rectangle (103.24, 32.21);
\definecolor{fillColor}{RGB}{255,0,0}

\path[draw=drawColor,draw opacity=0.70,line width= 0.4pt,line join=round,line cap=round,fill=fillColor,fill opacity=0.70] ( 19.71, 25.78) circle (  1.96);

\path[draw=drawColor,draw opacity=0.70,line width= 0.4pt,line join=round,line cap=round,fill=fillColor,fill opacity=0.70] ( 25.15, 25.05) circle (  1.96);

\path[draw=drawColor,draw opacity=0.70,line width= 0.4pt,line join=round,line cap=round,fill=fillColor,fill opacity=0.70] ( 30.58, 28.14) circle (  1.96);

\path[draw=drawColor,draw opacity=0.70,line width= 0.4pt,line join=round,line cap=round,fill=fillColor,fill opacity=0.70] ( 36.02, 25.38) circle (  1.96);

\path[draw=drawColor,draw opacity=0.70,line width= 0.4pt,line join=round,line cap=round,fill=fillColor,fill opacity=0.70] ( 41.46, 24.64) circle (  1.96);

\path[draw=drawColor,draw opacity=0.70,line width= 0.4pt,line join=round,line cap=round,fill=fillColor,fill opacity=0.70] ( 46.90, 25.63) circle (  1.96);

\path[draw=drawColor,draw opacity=0.70,line width= 0.4pt,line join=round,line cap=round,fill=fillColor,fill opacity=0.70] ( 52.33, 26.46) circle (  1.96);

\path[draw=drawColor,draw opacity=0.70,line width= 0.4pt,line join=round,line cap=round,fill=fillColor,fill opacity=0.70] ( 57.77, 27.15) circle (  1.96);

\path[draw=drawColor,draw opacity=0.70,line width= 0.4pt,line join=round,line cap=round,fill=fillColor,fill opacity=0.70] ( 63.21, 25.39) circle (  1.96);

\path[draw=drawColor,draw opacity=0.70,line width= 0.4pt,line join=round,line cap=round,fill=fillColor,fill opacity=0.70] ( 68.65, 24.74) circle (  1.96);

\path[draw=drawColor,draw opacity=0.70,line width= 0.4pt,line join=round,line cap=round,fill=fillColor,fill opacity=0.70] ( 74.09, 24.64) circle (  1.96);

\path[draw=drawColor,draw opacity=0.70,line width= 0.4pt,line join=round,line cap=round,fill=fillColor,fill opacity=0.70] ( 79.52, 25.43) circle (  1.96);

\path[draw=drawColor,draw opacity=0.70,line width= 0.4pt,line join=round,line cap=round,fill=fillColor,fill opacity=0.70] ( 84.96, 24.71) circle (  1.96);

\path[draw=drawColor,draw opacity=0.70,line width= 0.4pt,line join=round,line cap=round,fill=fillColor,fill opacity=0.70] ( 90.40, 25.00) circle (  1.96);

\path[draw=drawColor,draw opacity=0.70,line width= 0.4pt,line join=round,line cap=round,fill=fillColor,fill opacity=0.70] ( 95.84, 25.46) circle (  1.96);

\path[draw=drawColor,draw opacity=0.70,line width= 0.4pt,line join=round,line cap=round,fill=fillColor,fill opacity=0.70] (101.27, 25.48) circle (  1.96);
\end{scope}
\begin{scope}
\path[clip] (  0.00,  0.00) rectangle (108.41, 93.95);
\definecolor{drawColor}{gray}{0.10}

\node[text=drawColor,anchor=base east,inner sep=0pt, outer sep=0pt, scale=  0.55] at ( 17.33, 28.72) {10};

\node[text=drawColor,anchor=base east,inner sep=0pt, outer sep=0pt, scale=  0.55] at ( 17.33, 42.00) {20};

\node[text=drawColor,anchor=base east,inner sep=0pt, outer sep=0pt, scale=  0.55] at ( 17.33, 55.28) {30};

\node[text=drawColor,anchor=base east,inner sep=0pt, outer sep=0pt, scale=  0.55] at ( 17.33, 68.57) {40};

\node[text=drawColor,anchor=base east,inner sep=0pt, outer sep=0pt, scale=  0.55] at ( 17.33, 81.85) {50};
\end{scope}
\begin{scope}
\path[clip] (  0.00,  0.00) rectangle (108.41, 93.95);
\definecolor{drawColor}{gray}{0.10}

\node[text=drawColor,anchor=base,inner sep=0pt, outer sep=0pt, scale=  0.55] at ( 25.15, 15.15) {-3.8};

\node[text=drawColor,anchor=base,inner sep=0pt, outer sep=0pt, scale=  0.55] at ( 36.02, 15.15) {-3.6};

\node[text=drawColor,anchor=base,inner sep=0pt, outer sep=0pt, scale=  0.55] at ( 46.90, 15.15) {-3.4};

\node[text=drawColor,anchor=base,inner sep=0pt, outer sep=0pt, scale=  0.55] at ( 57.77, 15.15) {-3.2};

\node[text=drawColor,anchor=base,inner sep=0pt, outer sep=0pt, scale=  0.55] at ( 68.65, 15.15) {-3.0};

\node[text=drawColor,anchor=base,inner sep=0pt, outer sep=0pt, scale=  0.55] at ( 79.52, 15.15) {-2.8};

\node[text=drawColor,anchor=base,inner sep=0pt, outer sep=0pt, scale=  0.55] at ( 90.40, 15.15) {-2.6};
\end{scope}
\begin{scope}
\path[clip] (  0.00,  0.00) rectangle (108.41, 93.95);
\definecolor{drawColor}{gray}{0.10}

\node[text=drawColor,anchor=base,inner sep=0pt, outer sep=0pt, scale=  0.66] at ( 60.49,  6.78) {$log_{10}(\sigma)$};
\end{scope}
\begin{scope}
\path[clip] (  0.00,  0.00) rectangle (108.41, 93.95);
\definecolor{drawColor}{gray}{0.10}

\node[text=drawColor,rotate= 90.00,anchor=base,inner sep=0pt, outer sep=0pt, scale=  0.66] at ( 10.05, 53.67) {running time (ms)};
\end{scope}
\end{tikzpicture}

%% file: Rfigures/TutorialTrans_new.tex
\begin{tikzpicture}[x=1pt,y=1pt]
\clip (5,3) rectangle (104, 93.95);
\definecolor{fillColor}{RGB}{255,255,255}
\path[use as bounding box,fill=fillColor,fill opacity=0.00] (0,0) rectangle (108.41, 93.95);
\begin{scope}
\path[clip] (  0.00,  0.00) rectangle (108.41, 93.95);
\definecolor{drawColor}{RGB}{255,255,255}
\definecolor{fillColor}{RGB}{255,255,255}

\path[draw=drawColor,line width= 0.6pt,line join=round,line cap=round,fill=fillColor] ( -0.00,  0.00) rectangle (108.41, 93.95);
\end{scope}
\begin{scope}
\path[clip] ( 19.10, 18.89) rectangle (102.90, 88.45);
\definecolor{fillColor}{gray}{0.92}

\path[fill=fillColor] ( 19.10, 18.89) rectangle (102.90, 88.45);
\definecolor{drawColor}{RGB}{0,0,0}

\path[draw=drawColor,line width= 0.6pt,line join=round] ( 23.59, 34.43) --
	( 38.56, 39.11) --
	( 53.52, 45.68) --
	( 68.49, 54.01) --
	( 83.45, 69.67) --
	( 84.22, 93.95);

\path[draw=drawColor,line width= 0.6pt,line join=round] ( 23.59, 26.06) --
	( 38.56, 29.16) --
	( 53.52, 35.22) --
	( 68.49, 41.95) --
	( 83.45, 53.13) --
	( 84.76, 93.95);

\path[draw=drawColor,line width= 0.6pt,line join=round] ( 23.59, 25.56) --
	( 38.56, 29.63) --
	( 53.52, 32.46) --
	( 68.49, 42.05) --
	( 83.45, 48.70) --
	( 87.45, 93.95);

\path[draw=drawColor,line width= 0.6pt,line join=round] ( 23.59, 23.23) --
	( 38.56, 25.63) --
	( 53.52, 30.08) --
	( 68.49, 36.74) --
	( 83.45, 47.83) --
	( 98.42, 64.84);
\definecolor{drawColor}{RGB}{0,0,0}
\definecolor{fillColor}{RGB}{190,190,190}

\path[draw=drawColor,draw opacity=0.70,line width= 0.4pt,line join=round,line cap=round,fill=fillColor,fill opacity=0.70] ( 23.59, 37.48) --
	( 26.24, 32.90) --
	( 20.95, 32.90) --
	cycle;

\path[draw=drawColor,draw opacity=0.70,line width= 0.4pt,line join=round,line cap=round,fill=fillColor,fill opacity=0.70] ( 38.56, 42.17) --
	( 41.20, 37.59) --
	( 35.92, 37.59) --
	cycle;

\path[draw=drawColor,draw opacity=0.70,line width= 0.4pt,line join=round,line cap=round,fill=fillColor,fill opacity=0.70] ( 53.52, 48.73) --
	( 56.17, 44.15) --
	( 50.88, 44.15) --
	cycle;

\path[draw=drawColor,draw opacity=0.70,line width= 0.4pt,line join=round,line cap=round,fill=fillColor,fill opacity=0.70] ( 68.49, 57.07) --
	( 71.13, 52.49) --
	( 65.84, 52.49) --
	cycle;

\path[draw=drawColor,draw opacity=0.70,line width= 0.4pt,line join=round,line cap=round,fill=fillColor,fill opacity=0.70] ( 83.45, 72.72) --
	( 86.09, 68.14) --
	( 80.81, 68.14) --
	cycle;
\definecolor{fillColor}{RGB}{173,216,230}

\path[draw=drawColor,draw opacity=0.70,line width= 0.4pt,line join=round,line cap=round,fill=fillColor,fill opacity=0.70] ( 21.86, 24.32) rectangle ( 25.33, 27.79);

\path[draw=drawColor,draw opacity=0.70,line width= 0.4pt,line join=round,line cap=round,fill=fillColor,fill opacity=0.70] ( 36.82, 27.42) rectangle ( 40.30, 30.90);

\path[draw=drawColor,draw opacity=0.70,line width= 0.4pt,line join=round,line cap=round,fill=fillColor,fill opacity=0.70] ( 51.78, 33.49) rectangle ( 55.26, 36.96);

\path[draw=drawColor,draw opacity=0.70,line width= 0.4pt,line join=round,line cap=round,fill=fillColor,fill opacity=0.70] ( 66.75, 40.21) rectangle ( 70.23, 43.68);

\path[draw=drawColor,draw opacity=0.70,line width= 0.4pt,line join=round,line cap=round,fill=fillColor,fill opacity=0.70] ( 81.71, 51.39) rectangle ( 85.19, 54.87);

\path[draw=drawColor,draw opacity=0.70,line width= 0.4pt,line join=round,line cap=round] ( 21.63, 23.60) rectangle ( 25.56, 27.53);

\path[draw=drawColor,draw opacity=0.70,line width= 0.4pt,line join=round,line cap=round] ( 36.60, 27.67) rectangle ( 40.52, 31.59);

\path[draw=drawColor,draw opacity=0.70,line width= 0.4pt,line join=round,line cap=round] ( 51.56, 30.50) rectangle ( 55.48, 34.42);

\path[draw=drawColor,draw opacity=0.70,line width= 0.4pt,line join=round,line cap=round] ( 66.52, 40.09) rectangle ( 70.45, 44.01);

\path[draw=drawColor,draw opacity=0.70,line width= 0.4pt,line join=round,line cap=round] ( 81.49, 46.74) rectangle ( 85.41, 50.66);
\definecolor{fillColor}{RGB}{255,0,0}

\path[draw=drawColor,draw opacity=0.70,line width= 0.4pt,line join=round,line cap=round,fill=fillColor,fill opacity=0.70] ( 23.59, 23.23) circle (  1.96);

\path[draw=drawColor,draw opacity=0.70,line width= 0.4pt,line join=round,line cap=round,fill=fillColor,fill opacity=0.70] ( 38.56, 25.63) circle (  1.96);

\path[draw=drawColor,draw opacity=0.70,line width= 0.4pt,line join=round,line cap=round,fill=fillColor,fill opacity=0.70] ( 53.52, 30.08) circle (  1.96);

\path[draw=drawColor,draw opacity=0.70,line width= 0.4pt,line join=round,line cap=round,fill=fillColor,fill opacity=0.70] ( 68.49, 36.74) circle (  1.96);

\path[draw=drawColor,draw opacity=0.70,line width= 0.4pt,line join=round,line cap=round,fill=fillColor,fill opacity=0.70] ( 83.45, 47.83) circle (  1.96);

\path[draw=drawColor,draw opacity=0.70,line width= 0.4pt,line join=round,line cap=round,fill=fillColor,fill opacity=0.70] ( 98.42, 64.84) circle (  1.96);
\end{scope}
\begin{scope}
\path[clip] (  0.00,  0.00) rectangle (108.41, 93.95);
\definecolor{drawColor}{gray}{0.10}

\node[text=drawColor,anchor=base east,inner sep=0pt, outer sep=0pt, scale=  0.55] at ( 18.35, 29.19) {0.5};

\node[text=drawColor,anchor=base east,inner sep=0pt, outer sep=0pt, scale=  0.55] at ( 18.35, 47.26) {0.7};

\node[text=drawColor,anchor=base east,inner sep=0pt, outer sep=0pt, scale=  0.55] at ( 18.35, 65.33) {0.9};

\node[text=drawColor,anchor=base east,inner sep=0pt, outer sep=0pt, scale=  0.55] at ( 18.35, 83.39) {1.1};
\end{scope}
\begin{scope}
\path[clip] (  0.00,  0.00) rectangle (108.41, 93.95);
\definecolor{drawColor}{gray}{0.10}

\node[text=drawColor,anchor=base,inner sep=0pt, outer sep=0pt, scale=  0.55] at ( 38.56, 15.15) {-2.8};

\node[text=drawColor,anchor=base,inner sep=0pt, outer sep=0pt, scale=  0.55] at ( 68.49, 15.15) {-2.6};

\node[text=drawColor,anchor=base,inner sep=0pt, outer sep=0pt, scale=  0.55] at ( 98.42, 15.15) {-2.4};
\end{scope}
\begin{scope}
\path[clip] (  0.00,  0.00) rectangle (108.41, 93.95);
\definecolor{drawColor}{gray}{0.10}

\node[text=drawColor,anchor=base,inner sep=0pt, outer sep=0pt, scale=  0.66] at ( 61.00,  6.78) {$log_{10}(\sigma)$};
\end{scope}
\begin{scope}
\path[clip] (  0.00,  0.00) rectangle (108.41, 93.95);
\definecolor{drawColor}{gray}{0.10}

\node[text=drawColor,rotate= 90.00,anchor=base,inner sep=0pt, outer sep=0pt, scale=  0.66] at ( 10.05, 53.67) {reprojection error (pixel)};
\end{scope}
\begin{scope}
\path[clip] (  0.00,  0.00) rectangle (108.41, 93.95);
\definecolor{drawColor}{RGB}{0,0,0}
\definecolor{fillColor}{RGB}{190,190,190}

\path[draw=drawColor,draw opacity=0.70,line width= 0.4pt,line join=round,line cap=round,fill=fillColor,fill opacity=0.70] ( 32.32, 81.83) --
	( 34.97, 77.26) --
	( 29.68, 77.26) --
	cycle;
\end{scope}
\begin{scope}
\path[clip] (  0.00,  0.00) rectangle (108.41, 93.95);
\definecolor{drawColor}{RGB}{0,0,0}
\definecolor{fillColor}{RGB}{173,216,230}

\path[draw=drawColor,draw opacity=0.70,line width= 0.4pt,line join=round,line cap=round,fill=fillColor,fill opacity=0.70] ( 30.58, 72.38) rectangle ( 34.06, 75.86);
\end{scope}
\begin{scope}
\path[clip] (  0.00,  0.00) rectangle (108.41, 93.95);
\definecolor{drawColor}{RGB}{0,0,0}

\path[draw=drawColor,draw opacity=0.70,line width= 0.4pt,line join=round,line cap=round] ( 30.36, 67.49) rectangle ( 34.29, 71.42);
\end{scope}
\begin{scope}
\path[clip] (  0.00,  0.00) rectangle (108.41, 93.95);
\definecolor{drawColor}{RGB}{0,0,0}
\definecolor{fillColor}{RGB}{255,0,0}

\path[draw=drawColor,draw opacity=0.70,line width= 0.4pt,line join=round,line cap=round,fill=fillColor,fill opacity=0.70] ( 32.32, 64.79) circle (  1.96);
\end{scope}
\begin{scope}
\path[clip] (  0.00,  0.00) rectangle (108.41, 93.95);
\definecolor{drawColor}{RGB}{0,0,0}

\node[text=drawColor,anchor=base west,inner sep=0pt, outer sep=0pt, scale=  0.53] at ( 36.63, 76.96) {R-GC};
\end{scope}
\begin{scope}
\path[clip] (  0.00,  0.00) rectangle (108.41, 93.95);
\definecolor{drawColor}{RGB}{0,0,0}

\node[text=drawColor,anchor=base west,inner sep=0pt, outer sep=0pt, scale=  0.53] at ( 36.63, 72.30) {R-DLT};
\end{scope}
\begin{scope}
\path[clip] (  0.00,  0.00) rectangle (108.41, 93.95);
\definecolor{drawColor}{RGB}{0,0,0}

\node[text=drawColor,anchor=base west,inner sep=0pt, outer sep=0pt, scale=  0.53] at ( 36.63, 67.64) {R-Huber};
\end{scope}
\begin{scope}
\path[clip] (  0.00,  0.00) rectangle (108.41, 93.95);
\definecolor{drawColor}{RGB}{0,0,0}

\node[text=drawColor,anchor=base west,inner sep=0pt, outer sep=0pt, scale=  0.53] at ( 36.63, 62.98) {R-DPCP};
\end{scope}
\end{tikzpicture}

%% file: Rfigures/TutorialTime_new.tex
\begin{tikzpicture}[x=1pt,y=1pt]
\clip (5,3) rectangle (104, 93.95);
\definecolor{fillColor}{RGB}{255,255,255}
\path[use as bounding box,fill=fillColor,fill opacity=0.00] (0,0) rectangle (108.41, 93.95);
\begin{scope}
\path[clip] (  0.00,  0.00) rectangle (108.41, 93.95);
\definecolor{drawColor}{RGB}{255,255,255}
\definecolor{fillColor}{RGB}{255,255,255}

\path[draw=drawColor,line width= 0.6pt,line join=round,line cap=round,fill=fillColor] ( -0.00,  0.00) rectangle (108.41, 93.95);
\end{scope}
\begin{scope}
\path[clip] ( 17.58, 18.89) rectangle (102.91, 88.45);
\definecolor{fillColor}{gray}{0.92}

\path[fill=fillColor] ( 17.58, 18.89) rectangle (102.91, 88.45);
\definecolor{drawColor}{RGB}{0,0,0}

\path[draw=drawColor,line width= 0.6pt,line join=round] ( 22.15, 85.29) --
	( 37.39, 74.15) --
	( 52.62, 64.16) --
	( 67.86, 55.49) --
	( 83.10, 47.67) --
	( 98.33, 41.00);

\path[draw=drawColor,line width= 0.6pt,line join=round] ( 22.15, 50.78) --
	( 37.39, 43.95) --
	( 52.62, 38.39) --
	( 67.86, 32.26) --
	( 83.10, 26.71) --
	( 98.33, 22.05);

\path[draw=drawColor,line width= 0.6pt,line join=round] ( 22.15, 45.88) --
	( 37.39, 40.64) --
	( 52.62, 35.77) --
	( 67.86, 31.06) --
	( 83.10, 26.66) --
	( 98.33, 22.64);

\path[draw=drawColor,line width= 0.6pt,line join=round] ( 22.15, 44.74) --
	( 37.39, 39.51) --
	( 52.62, 35.53) --
	( 67.86, 31.54) --
	( 83.10, 26.94) --
	( 98.33, 23.57);
\definecolor{drawColor}{RGB}{0,0,0}
\definecolor{fillColor}{RGB}{190,190,190}

\path[draw=drawColor,draw opacity=0.70,line width= 0.4pt,line join=round,line cap=round,fill=fillColor,fill opacity=0.70] ( 22.15, 88.34) --
	( 24.79, 83.76) --
	( 19.51, 83.76) --
	cycle;

\path[draw=drawColor,draw opacity=0.70,line width= 0.4pt,line join=round,line cap=round,fill=fillColor,fill opacity=0.70] ( 37.39, 77.21) --
	( 40.03, 72.63) --
	( 34.74, 72.63) --
	cycle;

\path[draw=drawColor,draw opacity=0.70,line width= 0.4pt,line join=round,line cap=round,fill=fillColor,fill opacity=0.70] ( 52.62, 67.21) --
	( 55.27, 62.63) --
	( 49.98, 62.63) --
	cycle;

\path[draw=drawColor,draw opacity=0.70,line width= 0.4pt,line join=round,line cap=round,fill=fillColor,fill opacity=0.70] ( 67.86, 58.55) --
	( 70.50, 53.97) --
	( 65.22, 53.97) --
	cycle;

\path[draw=drawColor,draw opacity=0.70,line width= 0.4pt,line join=round,line cap=round,fill=fillColor,fill opacity=0.70] ( 83.10, 50.72) --
	( 85.74, 46.15) --
	( 80.45, 46.15) --
	cycle;

\path[draw=drawColor,draw opacity=0.70,line width= 0.4pt,line join=round,line cap=round,fill=fillColor,fill opacity=0.70] ( 98.33, 44.05) --
	(100.98, 39.47) --
	( 95.69, 39.47) --
	cycle;
\definecolor{fillColor}{RGB}{173,216,230}

\path[draw=drawColor,draw opacity=0.70,line width= 0.4pt,line join=round,line cap=round,fill=fillColor,fill opacity=0.70] ( 20.41, 49.05) rectangle ( 23.89, 52.52);

\path[draw=drawColor,draw opacity=0.70,line width= 0.4pt,line join=round,line cap=round,fill=fillColor,fill opacity=0.70] ( 35.65, 42.21) rectangle ( 39.12, 45.69);

\path[draw=drawColor,draw opacity=0.70,line width= 0.4pt,line join=round,line cap=round,fill=fillColor,fill opacity=0.70] ( 50.88, 36.65) rectangle ( 54.36, 40.13);

\path[draw=drawColor,draw opacity=0.70,line width= 0.4pt,line join=round,line cap=round,fill=fillColor,fill opacity=0.70] ( 66.12, 30.52) rectangle ( 69.60, 34.00);

\path[draw=drawColor,draw opacity=0.70,line width= 0.4pt,line join=round,line cap=round,fill=fillColor,fill opacity=0.70] ( 81.36, 24.97) rectangle ( 84.84, 28.45);

\path[draw=drawColor,draw opacity=0.70,line width= 0.4pt,line join=round,line cap=round,fill=fillColor,fill opacity=0.70] ( 96.59, 20.31) rectangle (100.07, 23.79);

\path[draw=drawColor,draw opacity=0.70,line width= 0.4pt,line join=round,line cap=round] ( 20.19, 43.91) rectangle ( 24.11, 47.84);

\path[draw=drawColor,draw opacity=0.70,line width= 0.4pt,line join=round,line cap=round] ( 35.42, 38.67) rectangle ( 39.35, 42.60);

\path[draw=drawColor,draw opacity=0.70,line width= 0.4pt,line join=round,line cap=round] ( 50.66, 33.80) rectangle ( 54.58, 37.73);

\path[draw=drawColor,draw opacity=0.70,line width= 0.4pt,line join=round,line cap=round] ( 65.90, 29.10) rectangle ( 69.82, 33.02);

\path[draw=drawColor,draw opacity=0.70,line width= 0.4pt,line join=round,line cap=round] ( 81.13, 24.70) rectangle ( 85.06, 28.63);

\path[draw=drawColor,draw opacity=0.70,line width= 0.4pt,line join=round,line cap=round] ( 96.37, 20.68) rectangle (100.30, 24.60);
\definecolor{fillColor}{RGB}{255,0,0}

\path[draw=drawColor,draw opacity=0.70,line width= 0.4pt,line join=round,line cap=round,fill=fillColor,fill opacity=0.70] ( 22.15, 44.74) circle (  1.96);

\path[draw=drawColor,draw opacity=0.70,line width= 0.4pt,line join=round,line cap=round,fill=fillColor,fill opacity=0.70] ( 37.39, 39.51) circle (  1.96);

\path[draw=drawColor,draw opacity=0.70,line width= 0.4pt,line join=round,line cap=round,fill=fillColor,fill opacity=0.70] ( 52.62, 35.53) circle (  1.96);

\path[draw=drawColor,draw opacity=0.70,line width= 0.4pt,line join=round,line cap=round,fill=fillColor,fill opacity=0.70] ( 67.86, 31.54) circle (  1.96);

\path[draw=drawColor,draw opacity=0.70,line width= 0.4pt,line join=round,line cap=round,fill=fillColor,fill opacity=0.70] ( 83.10, 26.94) circle (  1.96);

\path[draw=drawColor,draw opacity=0.70,line width= 0.4pt,line join=round,line cap=round,fill=fillColor,fill opacity=0.70] ( 98.33, 23.57) circle (  1.96);
\end{scope}
\begin{scope}
\path[clip] (  0.00,  0.00) rectangle (108.41, 93.95);
\definecolor{drawColor}{gray}{0.10}

\node[text=drawColor,anchor=base east,inner sep=0pt, outer sep=0pt, scale=  0.55] at ( 16.83, 19.24) {20};

\node[text=drawColor,anchor=base east,inner sep=0pt, outer sep=0pt, scale=  0.55] at ( 16.83, 36.17) {35};

\node[text=drawColor,anchor=base east,inner sep=0pt, outer sep=0pt, scale=  0.55] at ( 16.83, 58.74) {55};

\node[text=drawColor,anchor=base east,inner sep=0pt, outer sep=0pt, scale=  0.55] at ( 16.83, 81.31) {75};
\end{scope}
\begin{scope}
\path[clip] (  0.00,  0.00) rectangle (108.41, 93.95);
\definecolor{drawColor}{gray}{0.10}

\node[text=drawColor,anchor=base,inner sep=0pt, outer sep=0pt, scale=  0.55] at ( 37.39, 15.15) {-2.8};

\node[text=drawColor,anchor=base,inner sep=0pt, outer sep=0pt, scale=  0.55] at ( 67.86, 15.15) {-2.6};

\node[text=drawColor,anchor=base,inner sep=0pt, outer sep=0pt, scale=  0.55] at ( 98.33, 15.15) {-2.4};
\end{scope}
\begin{scope}
\path[clip] (  0.00,  0.00) rectangle (108.41, 93.95);
\definecolor{drawColor}{gray}{0.10}

\node[text=drawColor,anchor=base,inner sep=0pt, outer sep=0pt, scale=  0.66] at ( 60.24,  6.78) {$log_{10}(\sigma)$};
\end{scope}
\begin{scope}
\path[clip] (  0.00,  0.00) rectangle (108.41, 93.95);
\definecolor{drawColor}{gray}{0.10}

\node[text=drawColor,rotate= 90.00,anchor=base,inner sep=0pt, outer sep=0pt, scale=  0.66] at ( 10.05, 53.67) {running time (ms)};
\end{scope}
\end{tikzpicture}

%% file: Rfigures/EVDsampson_mean.tex
\begin{tikzpicture}[x=1pt,y=1pt]
\clip (5,3) rectangle (104, 93.95);
\definecolor{fillColor}{RGB}{255,255,255}
\path[use as bounding box,fill=fillColor,fill opacity=0.00] (0,0) rectangle (108.41, 93.95);
\begin{scope}
\path[clip] (  0.00,  0.00) rectangle (108.40, 93.95);
\definecolor{drawColor}{RGB}{255,255,255}
\definecolor{fillColor}{RGB}{255,255,255}

\path[draw=drawColor,line width= 0.6pt,line join=round,line cap=round,fill=fillColor] (  0.00,  0.00) rectangle (108.41, 93.95);
\end{scope}
\begin{scope}
\path[clip] ( 18.58, 18.89) rectangle (102.90, 88.45);
\definecolor{fillColor}{gray}{0.92}

\path[fill=fillColor] ( 18.58, 18.89) rectangle (102.91, 88.45);
\definecolor{drawColor}{RGB}{0,0,0}

\path[draw=drawColor,line width= 0.6pt,line join=round] ( 20.59, 50.93) --
	( 27.28, 48.95) --
	( 33.97, 50.96) --
	( 40.66, 54.21) --
	( 47.36, 59.17) --
	( 54.05, 65.01) --
	( 60.74, 72.13) --
	( 67.43, 79.13) --
	( 72.73, 93.95);

\path[draw=drawColor,line width= 0.6pt,line join=round] ( 20.59, 36.61) --
	( 27.28, 33.96) --
	( 33.97, 32.47) --
	( 40.66, 31.67) --
	( 47.36, 31.23) --
	( 54.05, 31.98) --
	( 60.74, 31.51) --
	( 67.43, 31.43) --
	( 74.13, 31.86) --
	( 80.82, 36.34) --
	( 87.51, 46.56) --
	( 94.20, 51.30) --
	(100.90, 88.58);

\path[draw=drawColor,line width= 0.6pt,line join=round] ( 20.59, 36.10) --
	( 27.28, 33.69) --
	( 33.97, 31.82) --
	( 40.66, 30.81) --
	( 47.36, 29.15) --
	( 54.05, 27.15) --
	( 60.74, 26.39) --
	( 67.43, 25.26) --
	( 74.13, 25.45) --
	( 80.82, 25.71) --
	( 87.51, 26.56) --
	( 94.20, 47.30) --
	(100.90, 58.08);

\path[draw=drawColor,line width= 0.6pt,line join=round] ( 20.59, 34.65) --
	( 27.28, 31.37) --
	( 33.97, 29.01) --
	( 40.66, 27.62) --
	( 47.36, 25.86) --
	( 54.05, 24.21) --
	( 60.74, 23.28) --
	( 67.43, 22.47) --
	( 74.13, 22.39) --
	( 80.82, 22.73) --
	( 87.51, 23.51) --
	( 94.20, 29.05) --
	(100.90, 47.76);
\definecolor{drawColor}{RGB}{0,0,0}
\definecolor{fillColor}{RGB}{190,190,190}

\path[draw=drawColor,draw opacity=0.70,line width= 0.4pt,line join=round,line cap=round,fill=fillColor,fill opacity=0.70] ( 20.59, 53.98) --
	( 23.23, 49.41) --
	( 17.94, 49.41) --
	cycle;

\path[draw=drawColor,draw opacity=0.70,line width= 0.4pt,line join=round,line cap=round,fill=fillColor,fill opacity=0.70] ( 27.28, 52.00) --
	( 29.92, 47.43) --
	( 24.64, 47.43) --
	cycle;

\path[draw=drawColor,draw opacity=0.70,line width= 0.4pt,line join=round,line cap=round,fill=fillColor,fill opacity=0.70] ( 33.97, 54.01) --
	( 36.61, 49.43) --
	( 31.33, 49.43) --
	cycle;

\path[draw=drawColor,draw opacity=0.70,line width= 0.4pt,line join=round,line cap=round,fill=fillColor,fill opacity=0.70] ( 40.66, 57.27) --
	( 43.31, 52.69) --
	( 38.02, 52.69) --
	cycle;

\path[draw=drawColor,draw opacity=0.70,line width= 0.4pt,line join=round,line cap=round,fill=fillColor,fill opacity=0.70] ( 47.36, 62.22) --
	( 50.00, 57.64) --
	( 44.71, 57.64) --
	cycle;

\path[draw=drawColor,draw opacity=0.70,line width= 0.4pt,line join=round,line cap=round,fill=fillColor,fill opacity=0.70] ( 54.05, 68.06) --
	( 56.69, 63.48) --
	( 51.41, 63.48) --
	cycle;

\path[draw=drawColor,draw opacity=0.70,line width= 0.4pt,line join=round,line cap=round,fill=fillColor,fill opacity=0.70] ( 60.74, 75.18) --
	( 63.38, 70.61) --
	( 58.10, 70.61) --
	cycle;

\path[draw=drawColor,draw opacity=0.70,line width= 0.4pt,line join=round,line cap=round,fill=fillColor,fill opacity=0.70] ( 67.43, 82.18) --
	( 70.08, 77.61) --
	( 64.79, 77.61) --
	cycle;
\definecolor{fillColor}{RGB}{173,216,230}

\path[draw=drawColor,draw opacity=0.70,line width= 0.4pt,line join=round,line cap=round,fill=fillColor,fill opacity=0.70] ( 18.85, 34.87) rectangle ( 22.32, 38.35);

\path[draw=drawColor,draw opacity=0.70,line width= 0.4pt,line join=round,line cap=round,fill=fillColor,fill opacity=0.70] ( 25.54, 32.22) rectangle ( 29.02, 35.69);

\path[draw=drawColor,draw opacity=0.70,line width= 0.4pt,line join=round,line cap=round,fill=fillColor,fill opacity=0.70] ( 32.23, 30.73) rectangle ( 35.71, 34.21);

\path[draw=drawColor,draw opacity=0.70,line width= 0.4pt,line join=round,line cap=round,fill=fillColor,fill opacity=0.70] ( 38.92, 29.93) rectangle ( 42.40, 33.40);

\path[draw=drawColor,draw opacity=0.70,line width= 0.4pt,line join=round,line cap=round,fill=fillColor,fill opacity=0.70] ( 45.62, 29.49) rectangle ( 49.09, 32.96);

\path[draw=drawColor,draw opacity=0.70,line width= 0.4pt,line join=round,line cap=round,fill=fillColor,fill opacity=0.70] ( 52.31, 30.24) rectangle ( 55.79, 33.72);

\path[draw=drawColor,draw opacity=0.70,line width= 0.4pt,line join=round,line cap=round,fill=fillColor,fill opacity=0.70] ( 59.00, 29.77) rectangle ( 62.48, 33.25);

\path[draw=drawColor,draw opacity=0.70,line width= 0.4pt,line join=round,line cap=round,fill=fillColor,fill opacity=0.70] ( 65.69, 29.69) rectangle ( 69.17, 33.17);

\path[draw=drawColor,draw opacity=0.70,line width= 0.4pt,line join=round,line cap=round,fill=fillColor,fill opacity=0.70] ( 72.39, 30.12) rectangle ( 75.87, 33.60);

\path[draw=drawColor,draw opacity=0.70,line width= 0.4pt,line join=round,line cap=round,fill=fillColor,fill opacity=0.70] ( 79.08, 34.60) rectangle ( 82.56, 38.08);

\path[draw=drawColor,draw opacity=0.70,line width= 0.4pt,line join=round,line cap=round,fill=fillColor,fill opacity=0.70] ( 85.77, 44.82) rectangle ( 89.25, 48.30);

\path[draw=drawColor,draw opacity=0.70,line width= 0.4pt,line join=round,line cap=round,fill=fillColor,fill opacity=0.70] ( 92.47, 49.56) rectangle ( 95.94, 53.04);

\path[draw=drawColor,draw opacity=0.70,line width= 0.4pt,line join=round,line cap=round,fill=fillColor,fill opacity=0.70] ( 99.16, 86.84) rectangle (102.64, 90.32);

\path[draw=drawColor,draw opacity=0.70,line width= 0.4pt,line join=round,line cap=round] ( 18.62, 34.14) rectangle ( 22.55, 38.06);

\path[draw=drawColor,draw opacity=0.70,line width= 0.4pt,line join=round,line cap=round] ( 25.32, 31.73) rectangle ( 29.24, 35.65);

\path[draw=drawColor,draw opacity=0.70,line width= 0.4pt,line join=round,line cap=round] ( 32.01, 29.85) rectangle ( 35.93, 33.78);

\path[draw=drawColor,draw opacity=0.70,line width= 0.4pt,line join=round,line cap=round] ( 38.70, 28.84) rectangle ( 42.63, 32.77);

\path[draw=drawColor,draw opacity=0.70,line width= 0.4pt,line join=round,line cap=round] ( 45.39, 27.19) rectangle ( 49.32, 31.11);

\path[draw=drawColor,draw opacity=0.70,line width= 0.4pt,line join=round,line cap=round] ( 52.09, 25.19) rectangle ( 56.01, 29.11);

\path[draw=drawColor,draw opacity=0.70,line width= 0.4pt,line join=round,line cap=round] ( 58.78, 24.43) rectangle ( 62.70, 28.36);

\path[draw=drawColor,draw opacity=0.70,line width= 0.4pt,line join=round,line cap=round] ( 65.47, 23.30) rectangle ( 69.40, 27.22);

\path[draw=drawColor,draw opacity=0.70,line width= 0.4pt,line join=round,line cap=round] ( 72.16, 23.49) rectangle ( 76.09, 27.41);

\path[draw=drawColor,draw opacity=0.70,line width= 0.4pt,line join=round,line cap=round] ( 78.86, 23.75) rectangle ( 82.78, 27.67);

\path[draw=drawColor,draw opacity=0.70,line width= 0.4pt,line join=round,line cap=round] ( 85.55, 24.60) rectangle ( 89.47, 28.52);

\path[draw=drawColor,draw opacity=0.70,line width= 0.4pt,line join=round,line cap=round] ( 92.24, 45.34) rectangle ( 96.17, 49.26);

\path[draw=drawColor,draw opacity=0.70,line width= 0.4pt,line join=round,line cap=round] ( 98.94, 56.12) rectangle (102.86, 60.05);
\definecolor{fillColor}{RGB}{255,0,0}

\path[draw=drawColor,draw opacity=0.70,line width= 0.4pt,line join=round,line cap=round,fill=fillColor,fill opacity=0.70] ( 20.59, 34.65) circle (  1.96);

\path[draw=drawColor,draw opacity=0.70,line width= 0.4pt,line join=round,line cap=round,fill=fillColor,fill opacity=0.70] ( 27.28, 31.37) circle (  1.96);

\path[draw=drawColor,draw opacity=0.70,line width= 0.4pt,line join=round,line cap=round,fill=fillColor,fill opacity=0.70] ( 33.97, 29.01) circle (  1.96);

\path[draw=drawColor,draw opacity=0.70,line width= 0.4pt,line join=round,line cap=round,fill=fillColor,fill opacity=0.70] ( 40.66, 27.62) circle (  1.96);

\path[draw=drawColor,draw opacity=0.70,line width= 0.4pt,line join=round,line cap=round,fill=fillColor,fill opacity=0.70] ( 47.36, 25.86) circle (  1.96);

\path[draw=drawColor,draw opacity=0.70,line width= 0.4pt,line join=round,line cap=round,fill=fillColor,fill opacity=0.70] ( 54.05, 24.21) circle (  1.96);

\path[draw=drawColor,draw opacity=0.70,line width= 0.4pt,line join=round,line cap=round,fill=fillColor,fill opacity=0.70] ( 60.74, 23.28) circle (  1.96);

\path[draw=drawColor,draw opacity=0.70,line width= 0.4pt,line join=round,line cap=round,fill=fillColor,fill opacity=0.70] ( 67.43, 22.47) circle (  1.96);

\path[draw=drawColor,draw opacity=0.70,line width= 0.4pt,line join=round,line cap=round,fill=fillColor,fill opacity=0.70] ( 74.13, 22.39) circle (  1.96);

\path[draw=drawColor,draw opacity=0.70,line width= 0.4pt,line join=round,line cap=round,fill=fillColor,fill opacity=0.70] ( 80.82, 22.73) circle (  1.96);

\path[draw=drawColor,draw opacity=0.70,line width= 0.4pt,line join=round,line cap=round,fill=fillColor,fill opacity=0.70] ( 87.51, 23.51) circle (  1.96);

\path[draw=drawColor,draw opacity=0.70,line width= 0.4pt,line join=round,line cap=round,fill=fillColor,fill opacity=0.70] ( 94.20, 29.05) circle (  1.96);

\path[draw=drawColor,draw opacity=0.70,line width= 0.4pt,line join=round,line cap=round,fill=fillColor,fill opacity=0.70] (100.90, 47.76) circle (  1.96);
\end{scope}
\begin{scope}
\path[clip] (  0.00,  0.00) rectangle (108.41, 93.95);
\definecolor{drawColor}{gray}{0.10}

\node[text=drawColor,anchor=base east,inner sep=0pt, outer sep=0pt, scale=  0.55] at ( 17.83, 22.90) {4};

\node[text=drawColor,anchor=base east,inner sep=0pt, outer sep=0pt, scale=  0.55] at ( 17.83, 33.90) {6};

\node[text=drawColor,anchor=base east,inner sep=0pt, outer sep=0pt, scale=  0.55] at ( 17.83, 44.90) {8};

\node[text=drawColor,anchor=base east,inner sep=0pt, outer sep=0pt, scale=  0.55] at ( 17.83, 55.90) {10};

\node[text=drawColor,anchor=base east,inner sep=0pt, outer sep=0pt, scale=  0.55] at ( 17.83, 66.90) {12};

\node[text=drawColor,anchor=base east,inner sep=0pt, outer sep=0pt, scale=  0.55] at ( 17.83, 77.90) {14};
\end{scope}
\begin{scope}
\path[clip] (  0.00,  0.00) rectangle (108.41, 93.95);
\definecolor{drawColor}{gray}{0.10}

\node[text=drawColor,anchor=base,inner sep=0pt, outer sep=0pt, scale=  0.55] at ( 27.28, 15.15) {-2.7};

\node[text=drawColor,anchor=base,inner sep=0pt, outer sep=0pt, scale=  0.55] at ( 40.66, 15.15) {-2.5};

\node[text=drawColor,anchor=base,inner sep=0pt, outer sep=0pt, scale=  0.55] at ( 54.05, 15.15) {-2.3};

\node[text=drawColor,anchor=base,inner sep=0pt, outer sep=0pt, scale=  0.55] at ( 67.43, 15.15) {-2.1};

\node[text=drawColor,anchor=base,inner sep=0pt, outer sep=0pt, scale=  0.55] at ( 80.82, 15.15) {-1.9};

\node[text=drawColor,anchor=base,inner sep=0pt, outer sep=0pt, scale=  0.55] at ( 94.20, 15.15) {-1.7};
\end{scope}
\begin{scope}
\path[clip] (  0.00,  0.00) rectangle (108.41, 93.95);
\definecolor{drawColor}{gray}{0.10}

\node[text=drawColor,anchor=base,inner sep=0pt, outer sep=0pt, scale=  0.66] at ( 60.74,  6.78) {$log_{10}(\sigma)$};
\end{scope}
\begin{scope}
\path[clip] (  0.00,  0.00) rectangle (108.41, 93.95);
\definecolor{drawColor}{gray}{0.10}

\node[text=drawColor,rotate= 90.00,anchor=base,inner sep=0pt, outer sep=0pt, scale=  0.66] at ( 10.05, 53.67) {reprojection error (pixel)};
\end{scope}
\begin{scope}
\path[clip] (  0.00,  0.00) rectangle (108.41, 93.95);
\definecolor{drawColor}{RGB}{0,0,0}
\definecolor{fillColor}{RGB}{190,190,190}

\path[draw=drawColor,draw opacity=0.70,line width= 0.4pt,line join=round,line cap=round,fill=fillColor,fill opacity=0.70] ( 61.47, 60.96) --
	( 64.11, 56.39) --
	( 58.83, 56.39) --
	cycle;
\end{scope}
\begin{scope}
\path[clip] (  0.00,  0.00) rectangle (108.41, 93.95);
\definecolor{drawColor}{RGB}{0,0,0}
\definecolor{fillColor}{RGB}{173,216,230}

\path[draw=drawColor,draw opacity=0.70,line width= 0.4pt,line join=round,line cap=round,fill=fillColor,fill opacity=0.70] ( 59.73, 51.51) rectangle ( 63.21, 54.99);
\end{scope}
\begin{scope}
\path[clip] (  0.00,  0.00) rectangle (108.41, 93.95);
\definecolor{drawColor}{RGB}{0,0,0}

\path[draw=drawColor,draw opacity=0.70,line width= 0.4pt,line join=round,line cap=round] ( 59.51, 46.62) rectangle ( 63.43, 50.55);
\end{scope}
\begin{scope}
\path[clip] (  0.00,  0.00) rectangle (108.41, 93.95);
\definecolor{drawColor}{RGB}{0,0,0}
\definecolor{fillColor}{RGB}{255,0,0}

\path[draw=drawColor,draw opacity=0.70,line width= 0.4pt,line join=round,line cap=round,fill=fillColor,fill opacity=0.70] ( 61.47, 43.92) circle (  1.96);
\end{scope}
\begin{scope}
\path[clip] (  0.00,  0.00) rectangle (108.41, 93.95);
\definecolor{drawColor}{RGB}{0,0,0}

\node[text=drawColor,anchor=base west,inner sep=0pt, outer sep=0pt, scale=  0.53] at ( 65.77, 56.09) {R-GC};
\end{scope}
\begin{scope}
\path[clip] (  0.00,  0.00) rectangle (108.41, 93.95);
\definecolor{drawColor}{RGB}{0,0,0}

\node[text=drawColor,anchor=base west,inner sep=0pt, outer sep=0pt, scale=  0.53] at ( 65.77, 51.43) {R-DLT};
\end{scope}
\begin{scope}
\path[clip] (  0.00,  0.00) rectangle (108.41, 93.95);
\definecolor{drawColor}{RGB}{0,0,0}

\node[text=drawColor,anchor=base west,inner sep=0pt, outer sep=0pt, scale=  0.53] at ( 65.77, 46.77) {R-Huber};
\end{scope}
\begin{scope}
\path[clip] (  0.00,  0.00) rectangle (108.41, 93.95);
\definecolor{drawColor}{RGB}{0,0,0}

\node[text=drawColor,anchor=base west,inner sep=0pt, outer sep=0pt, scale=  0.53] at ( 65.77, 42.11) {R-DPCP};
\end{scope}
\end{tikzpicture}

%% file: Rfigures/HPatchesTrans_mean.tex
\begin{tikzpicture}[x=1pt,y=1pt]
\clip (5,3) rectangle (104, 93.95);
\definecolor{fillColor}{RGB}{255,255,255}
\path[use as bounding box,fill=fillColor,fill opacity=0.00] (0,0) rectangle (108.41, 93.95);
\begin{scope}
\path[clip] (  0.00,  0.00) rectangle (108.41, 93.95);
\definecolor{drawColor}{RGB}{255,255,255}
\definecolor{fillColor}{RGB}{255,255,255}

\path[draw=drawColor,line width= 0.6pt,line join=round,line cap=round,fill=fillColor] ( -0.00,  0.00) rectangle (108.41, 93.95);
\end{scope}
\begin{scope}
\path[clip] ( 19.10, 18.89) rectangle (102.90, 88.45);
\definecolor{fillColor}{gray}{0.92}

\path[fill=fillColor] ( 19.10, 18.89) rectangle (102.90, 88.45);
\definecolor{drawColor}{RGB}{0,0,0}

\path[draw=drawColor,line width= 0.6pt,line join=round] ( 20.46, 67.89) --
	( 24.96, 72.15) --
	( 29.47, 76.43) --
	( 33.97, 79.14) --
	( 38.48, 80.31) --
	( 42.98, 80.15) --
	( 47.49, 79.74) --
	( 51.99, 79.72) --
	( 56.50, 79.96) --
	( 61.00, 80.14) --
	( 65.51, 80.44) --
	( 70.02, 80.82) --
	( 74.52, 81.00) --
	( 79.03, 81.27) --
	( 83.53, 81.63) --
	( 88.04, 82.01) --
	( 92.54, 82.64) --
	( 97.05, 83.48) --
	(101.55, 85.29);

\path[draw=drawColor,line width= 0.6pt,line join=round] ( 20.46, 51.50) --
	( 24.96, 47.79) --
	( 29.47, 43.49) --
	( 33.97, 38.98) --
	( 38.48, 34.74) --
	( 42.98, 31.62) --
	( 47.49, 29.46) --
	( 51.99, 28.40) --
	( 56.50, 27.97) --
	( 61.00, 28.04) --
	( 65.51, 27.98) --
	( 70.02, 28.07) --
	( 74.52, 28.42) --
	( 79.03, 29.04) --
	( 83.53, 29.58) --
	( 88.04, 30.84) --
	( 92.54, 33.03) --
	( 97.05, 35.81) --
	(101.55, 39.41);

\path[draw=drawColor,line width= 0.6pt,line join=round] ( 20.46, 51.49) --
	( 24.96, 47.79) --
	( 29.47, 43.47) --
	( 33.97, 38.95) --
	( 38.48, 34.71) --
	( 42.98, 31.58) --
	( 47.49, 29.41) --
	( 51.99, 28.32) --
	( 56.50, 27.84) --
	( 61.00, 27.82) --
	( 65.51, 27.75) --
	( 70.02, 27.66) --
	( 74.52, 27.73) --
	( 79.03, 27.91) --
	( 83.53, 27.93) --
	( 88.04, 27.91) --
	( 92.54, 28.20) --
	( 97.05, 28.80) --
	(101.55, 30.23);

\path[draw=drawColor,line width= 0.6pt,line join=round] ( 20.46, 46.51) --
	( 24.96, 41.65) --
	( 29.47, 36.72) --
	( 33.97, 32.31) --
	( 38.48, 29.07) --
	( 42.98, 26.21) --
	( 47.49, 23.49) --
	( 51.99, 22.68) --
	( 56.50, 22.19) --
	( 61.00, 22.10) --
	( 65.51, 22.07) --
	( 70.02, 22.05) --
	( 74.52, 22.05) --
	( 79.03, 22.09) --
	( 83.53, 22.09) --
	( 88.04, 22.12) --
	( 92.54, 22.29) --
	( 97.05, 22.90) --
	(101.55, 24.52);
\definecolor{drawColor}{RGB}{0,0,0}
\definecolor{fillColor}{RGB}{190,190,190}

\path[draw=drawColor,draw opacity=0.70,line width= 0.4pt,line join=round,line cap=round,fill=fillColor,fill opacity=0.70] ( 20.46, 70.94) --
	( 23.10, 66.36) --
	( 17.81, 66.36) --
	cycle;

\path[draw=drawColor,draw opacity=0.70,line width= 0.4pt,line join=round,line cap=round,fill=fillColor,fill opacity=0.70] ( 24.96, 75.20) --
	( 27.60, 70.63) --
	( 22.32, 70.63) --
	cycle;

\path[draw=drawColor,draw opacity=0.70,line width= 0.4pt,line join=round,line cap=round,fill=fillColor,fill opacity=0.70] ( 29.47, 79.48) --
	( 32.11, 74.91) --
	( 26.82, 74.91) --
	cycle;

\path[draw=drawColor,draw opacity=0.70,line width= 0.4pt,line join=round,line cap=round,fill=fillColor,fill opacity=0.70] ( 33.97, 82.19) --
	( 36.62, 77.62) --
	( 31.33, 77.62) --
	cycle;

\path[draw=drawColor,draw opacity=0.70,line width= 0.4pt,line join=round,line cap=round,fill=fillColor,fill opacity=0.70] ( 38.48, 83.36) --
	( 41.12, 78.78) --
	( 35.84, 78.78) --
	cycle;

\path[draw=drawColor,draw opacity=0.70,line width= 0.4pt,line join=round,line cap=round,fill=fillColor,fill opacity=0.70] ( 42.98, 83.20) --
	( 45.63, 78.62) --
	( 40.34, 78.62) --
	cycle;

\path[draw=drawColor,draw opacity=0.70,line width= 0.4pt,line join=round,line cap=round,fill=fillColor,fill opacity=0.70] ( 47.49, 82.79) --
	( 50.13, 78.21) --
	( 44.85, 78.21) --
	cycle;

\path[draw=drawColor,draw opacity=0.70,line width= 0.4pt,line join=round,line cap=round,fill=fillColor,fill opacity=0.70] ( 51.99, 82.77) --
	( 54.64, 78.19) --
	( 49.35, 78.19) --
	cycle;

\path[draw=drawColor,draw opacity=0.70,line width= 0.4pt,line join=round,line cap=round,fill=fillColor,fill opacity=0.70] ( 56.50, 83.01) --
	( 59.14, 78.43) --
	( 53.86, 78.43) --
	cycle;

\path[draw=drawColor,draw opacity=0.70,line width= 0.4pt,line join=round,line cap=round,fill=fillColor,fill opacity=0.70] ( 61.00, 83.19) --
	( 63.65, 78.61) --
	( 58.36, 78.61) --
	cycle;

\path[draw=drawColor,draw opacity=0.70,line width= 0.4pt,line join=round,line cap=round,fill=fillColor,fill opacity=0.70] ( 65.51, 83.49) --
	( 68.15, 78.92) --
	( 62.87, 78.92) --
	cycle;

\path[draw=drawColor,draw opacity=0.70,line width= 0.4pt,line join=round,line cap=round,fill=fillColor,fill opacity=0.70] ( 70.02, 83.87) --
	( 72.66, 79.30) --
	( 67.37, 79.30) --
	cycle;

\path[draw=drawColor,draw opacity=0.70,line width= 0.4pt,line join=round,line cap=round,fill=fillColor,fill opacity=0.70] ( 74.52, 84.05) --
	( 77.16, 79.47) --
	( 71.88, 79.47) --
	cycle;

\path[draw=drawColor,draw opacity=0.70,line width= 0.4pt,line join=round,line cap=round,fill=fillColor,fill opacity=0.70] ( 79.03, 84.32) --
	( 81.67, 79.74) --
	( 76.38, 79.74) --
	cycle;

\path[draw=drawColor,draw opacity=0.70,line width= 0.4pt,line join=round,line cap=round,fill=fillColor,fill opacity=0.70] ( 83.53, 84.68) --
	( 86.17, 80.10) --
	( 80.89, 80.10) --
	cycle;

\path[draw=drawColor,draw opacity=0.70,line width= 0.4pt,line join=round,line cap=round,fill=fillColor,fill opacity=0.70] ( 88.04, 85.06) --
	( 90.68, 80.48) --
	( 85.39, 80.48) --
	cycle;

\path[draw=drawColor,draw opacity=0.70,line width= 0.4pt,line join=round,line cap=round,fill=fillColor,fill opacity=0.70] ( 92.54, 85.69) --
	( 95.19, 81.11) --
	( 89.90, 81.11) --
	cycle;

\path[draw=drawColor,draw opacity=0.70,line width= 0.4pt,line join=round,line cap=round,fill=fillColor,fill opacity=0.70] ( 97.05, 86.53) --
	( 99.69, 81.95) --
	( 94.41, 81.95) --
	cycle;

\path[draw=drawColor,draw opacity=0.70,line width= 0.4pt,line join=round,line cap=round,fill=fillColor,fill opacity=0.70] (101.55, 88.34) --
	(104.20, 83.76) --
	( 98.91, 83.76) --
	cycle;
\definecolor{fillColor}{RGB}{173,216,230}

\path[draw=drawColor,draw opacity=0.70,line width= 0.4pt,line join=round,line cap=round,fill=fillColor,fill opacity=0.70] ( 18.72, 49.76) rectangle ( 22.20, 53.24);

\path[draw=drawColor,draw opacity=0.70,line width= 0.4pt,line join=round,line cap=round,fill=fillColor,fill opacity=0.70] ( 23.22, 46.05) rectangle ( 26.70, 49.53);

\path[draw=drawColor,draw opacity=0.70,line width= 0.4pt,line join=round,line cap=round,fill=fillColor,fill opacity=0.70] ( 27.73, 41.75) rectangle ( 31.21, 45.23);

\path[draw=drawColor,draw opacity=0.70,line width= 0.4pt,line join=round,line cap=round,fill=fillColor,fill opacity=0.70] ( 32.23, 37.24) rectangle ( 35.71, 40.72);

\path[draw=drawColor,draw opacity=0.70,line width= 0.4pt,line join=round,line cap=round,fill=fillColor,fill opacity=0.70] ( 36.74, 33.01) rectangle ( 40.22, 36.48);

\path[draw=drawColor,draw opacity=0.70,line width= 0.4pt,line join=round,line cap=round,fill=fillColor,fill opacity=0.70] ( 41.24, 29.88) rectangle ( 44.72, 33.36);

\path[draw=drawColor,draw opacity=0.70,line width= 0.4pt,line join=round,line cap=round,fill=fillColor,fill opacity=0.70] ( 45.75, 27.73) rectangle ( 49.23, 31.20);

\path[draw=drawColor,draw opacity=0.70,line width= 0.4pt,line join=round,line cap=round,fill=fillColor,fill opacity=0.70] ( 50.26, 26.66) rectangle ( 53.73, 30.14);

\path[draw=drawColor,draw opacity=0.70,line width= 0.4pt,line join=round,line cap=round,fill=fillColor,fill opacity=0.70] ( 54.76, 26.23) rectangle ( 58.24, 29.71);

\path[draw=drawColor,draw opacity=0.70,line width= 0.4pt,line join=round,line cap=round,fill=fillColor,fill opacity=0.70] ( 59.27, 26.30) rectangle ( 62.74, 29.78);

\path[draw=drawColor,draw opacity=0.70,line width= 0.4pt,line join=round,line cap=round,fill=fillColor,fill opacity=0.70] ( 63.77, 26.24) rectangle ( 67.25, 29.72);

\path[draw=drawColor,draw opacity=0.70,line width= 0.4pt,line join=round,line cap=round,fill=fillColor,fill opacity=0.70] ( 68.28, 26.33) rectangle ( 71.75, 29.81);

\path[draw=drawColor,draw opacity=0.70,line width= 0.4pt,line join=round,line cap=round,fill=fillColor,fill opacity=0.70] ( 72.78, 26.69) rectangle ( 76.26, 30.16);

\path[draw=drawColor,draw opacity=0.70,line width= 0.4pt,line join=round,line cap=round,fill=fillColor,fill opacity=0.70] ( 77.29, 27.30) rectangle ( 80.77, 30.77);

\path[draw=drawColor,draw opacity=0.70,line width= 0.4pt,line join=round,line cap=round,fill=fillColor,fill opacity=0.70] ( 81.79, 27.84) rectangle ( 85.27, 31.31);

\path[draw=drawColor,draw opacity=0.70,line width= 0.4pt,line join=round,line cap=round,fill=fillColor,fill opacity=0.70] ( 86.30, 29.10) rectangle ( 89.78, 32.58);

\path[draw=drawColor,draw opacity=0.70,line width= 0.4pt,line join=round,line cap=round,fill=fillColor,fill opacity=0.70] ( 90.80, 31.29) rectangle ( 94.28, 34.77);

\path[draw=drawColor,draw opacity=0.70,line width= 0.4pt,line join=round,line cap=round,fill=fillColor,fill opacity=0.70] ( 95.31, 34.07) rectangle ( 98.79, 37.55);

\path[draw=drawColor,draw opacity=0.70,line width= 0.4pt,line join=round,line cap=round,fill=fillColor,fill opacity=0.70] ( 99.81, 37.67) rectangle (103.29, 41.15);

\path[draw=drawColor,draw opacity=0.70,line width= 0.4pt,line join=round,line cap=round] ( 18.49, 49.53) rectangle ( 22.42, 53.45);

\path[draw=drawColor,draw opacity=0.70,line width= 0.4pt,line join=round,line cap=round] ( 23.00, 45.82) rectangle ( 26.92, 49.75);

\path[draw=drawColor,draw opacity=0.70,line width= 0.4pt,line join=round,line cap=round] ( 27.50, 41.50) rectangle ( 31.43, 45.43);

\path[draw=drawColor,draw opacity=0.70,line width= 0.4pt,line join=round,line cap=round] ( 32.01, 36.98) rectangle ( 35.93, 40.91);

\path[draw=drawColor,draw opacity=0.70,line width= 0.4pt,line join=round,line cap=round] ( 36.52, 32.75) rectangle ( 40.44, 36.67);

\path[draw=drawColor,draw opacity=0.70,line width= 0.4pt,line join=round,line cap=round] ( 41.02, 29.62) rectangle ( 44.95, 33.55);

\path[draw=drawColor,draw opacity=0.70,line width= 0.4pt,line join=round,line cap=round] ( 45.53, 27.45) rectangle ( 49.45, 31.37);

\path[draw=drawColor,draw opacity=0.70,line width= 0.4pt,line join=round,line cap=round] ( 50.03, 26.36) rectangle ( 53.96, 30.28);

\path[draw=drawColor,draw opacity=0.70,line width= 0.4pt,line join=round,line cap=round] ( 54.54, 25.88) rectangle ( 58.46, 29.80);

\path[draw=drawColor,draw opacity=0.70,line width= 0.4pt,line join=round,line cap=round] ( 59.04, 25.86) rectangle ( 62.97, 29.79);

\path[draw=drawColor,draw opacity=0.70,line width= 0.4pt,line join=round,line cap=round] ( 63.55, 25.79) rectangle ( 67.47, 29.71);

\path[draw=drawColor,draw opacity=0.70,line width= 0.4pt,line join=round,line cap=round] ( 68.05, 25.69) rectangle ( 71.98, 29.62);

\path[draw=drawColor,draw opacity=0.70,line width= 0.4pt,line join=round,line cap=round] ( 72.56, 25.76) rectangle ( 76.48, 29.69);

\path[draw=drawColor,draw opacity=0.70,line width= 0.4pt,line join=round,line cap=round] ( 77.06, 25.95) rectangle ( 80.99, 29.87);

\path[draw=drawColor,draw opacity=0.70,line width= 0.4pt,line join=round,line cap=round] ( 81.57, 25.97) rectangle ( 85.49, 29.89);

\path[draw=drawColor,draw opacity=0.70,line width= 0.4pt,line join=round,line cap=round] ( 86.08, 25.95) rectangle ( 90.00, 29.87);

\path[draw=drawColor,draw opacity=0.70,line width= 0.4pt,line join=round,line cap=round] ( 90.58, 26.24) rectangle ( 94.50, 30.16);

\path[draw=drawColor,draw opacity=0.70,line width= 0.4pt,line join=round,line cap=round] ( 95.09, 26.84) rectangle ( 99.01, 30.76);

\path[draw=drawColor,draw opacity=0.70,line width= 0.4pt,line join=round,line cap=round] ( 99.59, 28.26) rectangle (103.52, 32.19);
\definecolor{fillColor}{RGB}{255,0,0}

\path[draw=drawColor,draw opacity=0.70,line width= 0.4pt,line join=round,line cap=round,fill=fillColor,fill opacity=0.70] ( 20.46, 46.51) circle (  1.96);

\path[draw=drawColor,draw opacity=0.70,line width= 0.4pt,line join=round,line cap=round,fill=fillColor,fill opacity=0.70] ( 24.96, 41.65) circle (  1.96);

\path[draw=drawColor,draw opacity=0.70,line width= 0.4pt,line join=round,line cap=round,fill=fillColor,fill opacity=0.70] ( 29.47, 36.72) circle (  1.96);

\path[draw=drawColor,draw opacity=0.70,line width= 0.4pt,line join=round,line cap=round,fill=fillColor,fill opacity=0.70] ( 33.97, 32.31) circle (  1.96);

\path[draw=drawColor,draw opacity=0.70,line width= 0.4pt,line join=round,line cap=round,fill=fillColor,fill opacity=0.70] ( 38.48, 29.07) circle (  1.96);

\path[draw=drawColor,draw opacity=0.70,line width= 0.4pt,line join=round,line cap=round,fill=fillColor,fill opacity=0.70] ( 42.98, 26.21) circle (  1.96);

\path[draw=drawColor,draw opacity=0.70,line width= 0.4pt,line join=round,line cap=round,fill=fillColor,fill opacity=0.70] ( 47.49, 23.49) circle (  1.96);

\path[draw=drawColor,draw opacity=0.70,line width= 0.4pt,line join=round,line cap=round,fill=fillColor,fill opacity=0.70] ( 51.99, 22.68) circle (  1.96);

\path[draw=drawColor,draw opacity=0.70,line width= 0.4pt,line join=round,line cap=round,fill=fillColor,fill opacity=0.70] ( 56.50, 22.19) circle (  1.96);

\path[draw=drawColor,draw opacity=0.70,line width= 0.4pt,line join=round,line cap=round,fill=fillColor,fill opacity=0.70] ( 61.00, 22.10) circle (  1.96);

\path[draw=drawColor,draw opacity=0.70,line width= 0.4pt,line join=round,line cap=round,fill=fillColor,fill opacity=0.70] ( 65.51, 22.07) circle (  1.96);

\path[draw=drawColor,draw opacity=0.70,line width= 0.4pt,line join=round,line cap=round,fill=fillColor,fill opacity=0.70] ( 70.02, 22.05) circle (  1.96);

\path[draw=drawColor,draw opacity=0.70,line width= 0.4pt,line join=round,line cap=round,fill=fillColor,fill opacity=0.70] ( 74.52, 22.05) circle (  1.96);

\path[draw=drawColor,draw opacity=0.70,line width= 0.4pt,line join=round,line cap=round,fill=fillColor,fill opacity=0.70] ( 79.03, 22.09) circle (  1.96);

\path[draw=drawColor,draw opacity=0.70,line width= 0.4pt,line join=round,line cap=round,fill=fillColor,fill opacity=0.70] ( 83.53, 22.09) circle (  1.96);

\path[draw=drawColor,draw opacity=0.70,line width= 0.4pt,line join=round,line cap=round,fill=fillColor,fill opacity=0.70] ( 88.04, 22.12) circle (  1.96);

\path[draw=drawColor,draw opacity=0.70,line width= 0.4pt,line join=round,line cap=round,fill=fillColor,fill opacity=0.70] ( 92.54, 22.29) circle (  1.96);

\path[draw=drawColor,draw opacity=0.70,line width= 0.4pt,line join=round,line cap=round,fill=fillColor,fill opacity=0.70] ( 97.05, 22.90) circle (  1.96);

\path[draw=drawColor,draw opacity=0.70,line width= 0.4pt,line join=round,line cap=round,fill=fillColor,fill opacity=0.70] (101.55, 24.52) circle (  1.96);
\end{scope}
\begin{scope}
\path[clip] (  0.00,  0.00) rectangle (108.41, 93.95);
\definecolor{drawColor}{gray}{0.10}

\node[text=drawColor,anchor=base east,inner sep=0pt, outer sep=0pt, scale=  0.55] at ( 18.35, 17.32) {0.5};

\node[text=drawColor,anchor=base east,inner sep=0pt, outer sep=0pt, scale=  0.55] at ( 18.35, 25.69) {0.6};

\node[text=drawColor,anchor=base east,inner sep=0pt, outer sep=0pt, scale=  0.55] at ( 18.35, 34.05) {0.7};

\node[text=drawColor,anchor=base east,inner sep=0pt, outer sep=0pt, scale=  0.55] at ( 18.35, 42.42) {0.8};

\node[text=drawColor,anchor=base east,inner sep=0pt, outer sep=0pt, scale=  0.55] at ( 18.35, 50.78) {0.9};

\node[text=drawColor,anchor=base east,inner sep=0pt, outer sep=0pt, scale=  0.55] at ( 18.35, 59.15) {1.0};

\node[text=drawColor,anchor=base east,inner sep=0pt, outer sep=0pt, scale=  0.55] at ( 18.35, 67.51) {1.1};

\node[text=drawColor,anchor=base east,inner sep=0pt, outer sep=0pt, scale=  0.55] at ( 18.35, 75.87) {1.2};
\end{scope}
\begin{scope}
\path[clip] (  0.00,  0.00) rectangle (108.41, 93.95);
\definecolor{drawColor}{gray}{0.10}

\node[text=drawColor,anchor=base,inner sep=0pt, outer sep=0pt, scale=  0.55] at ( 29.47, 15.15) {-3.2};

\node[text=drawColor,anchor=base,inner sep=0pt, outer sep=0pt, scale=  0.55] at ( 47.49, 15.15) {-2.8};

\node[text=drawColor,anchor=base,inner sep=0pt, outer sep=0pt, scale=  0.55] at ( 65.51, 15.15) {-2.4};

\node[text=drawColor,anchor=base,inner sep=0pt, outer sep=0pt, scale=  0.55] at ( 83.53, 15.15) {-2.0};

\node[text=drawColor,anchor=base,inner sep=0pt, outer sep=0pt, scale=  0.55] at ( 92.54, 15.15) {-1.8};
\end{scope}
\begin{scope}
\path[clip] (  0.00,  0.00) rectangle (108.41, 93.95);
\definecolor{drawColor}{gray}{0.10}

\node[text=drawColor,anchor=base,inner sep=0pt, outer sep=0pt, scale=  0.66] at ( 61.00,  6.78) {$log_{10}(\sigma)$};
\end{scope}
\begin{scope}
\path[clip] (  0.00,  0.00) rectangle (108.41, 93.95);
\definecolor{drawColor}{gray}{0.10}

\node[text=drawColor,rotate= 90.00,anchor=base,inner sep=0pt, outer sep=0pt, scale=  0.66] at ( 10.05, 53.67) {reprojection error (pixel)};
\end{scope}
\end{tikzpicture}

%% file: Rfigures/EVDTime_median.tex
\begin{tikzpicture}[x=1pt,y=1pt]
\clip (5,3) rectangle (104, 93.95);
\definecolor{fillColor}{RGB}{255,255,255}
\path[use as bounding box,fill=fillColor,fill opacity=0.00] (0,0) rectangle (108.41, 93.95);
\begin{scope}
\path[clip] (  0.00,  0.00) rectangle (108.41, 93.95);
\definecolor{drawColor}{RGB}{255,255,255}
\definecolor{fillColor}{RGB}{255,255,255}

\path[draw=drawColor,line width= 0.6pt,line join=round,line cap=round,fill=fillColor] (  0.00,  0.00) rectangle (108.41, 93.95);
\end{scope}
\begin{scope}
\path[clip] ( 18.08, 18.89) rectangle (102.91, 88.45);
\definecolor{fillColor}{gray}{0.92}

\path[fill=fillColor] ( 18.08, 18.89) rectangle (102.91, 88.45);
\definecolor{drawColor}{RGB}{0,0,0}

\path[draw=drawColor,line width= 0.6pt,line join=round] ( 20.10, 85.29) --
	( 26.83, 77.31) --
	( 33.56, 72.47) --
	( 40.29, 61.10) --
	( 47.03, 50.13) --
	( 53.76, 41.27) --
	( 60.49, 37.05) --
	( 67.22, 34.48) --
	( 73.96, 31.97) --
	( 80.69, 30.28) --
	( 87.42, 29.47) --
	( 94.15, 28.40) --
	(100.89, 27.03);

\path[draw=drawColor,line width= 0.6pt,line join=round] ( 20.10, 37.18) --
	( 26.83, 33.75) --
	( 33.56, 30.08) --
	( 40.29, 25.92) --
	( 47.03, 23.85) --
	( 53.76, 22.75) --
	( 60.49, 22.49) --
	( 67.22, 22.23) --
	( 73.96, 22.12) --
	( 80.69, 22.05) --
	( 87.42, 22.16) --
	( 94.15, 22.22) --
	(100.89, 22.47);

\path[draw=drawColor,line width= 0.6pt,line join=round] ( 20.10, 38.72) --
	( 26.83, 34.77) --
	( 33.56, 30.55) --
	( 40.29, 26.19) --
	( 47.03, 24.80) --
	( 53.76, 24.31) --
	( 60.49, 24.18) --
	( 67.22, 24.10) --
	( 73.96, 24.11) --
	( 80.69, 23.96) --
	( 87.42, 23.98) --
	( 94.15, 24.08) --
	(100.89, 24.14);

\path[draw=drawColor,line width= 0.6pt,line join=round] ( 20.10, 48.05) --
	( 26.83, 34.62) --
	( 33.56, 32.66) --
	( 40.29, 29.54) --
	( 47.03, 28.69) --
	( 53.76, 27.52) --
	( 60.49, 26.84) --
	( 67.22, 26.05) --
	( 73.96, 25.61) --
	( 80.69, 25.41) --
	( 87.42, 25.03) --
	( 94.15, 25.03) --
	(100.89, 25.01);
\definecolor{drawColor}{RGB}{0,0,0}
\definecolor{fillColor}{RGB}{190,190,190}

\path[draw=drawColor,draw opacity=0.70,line width= 0.4pt,line join=round,line cap=round,fill=fillColor,fill opacity=0.70] ( 20.10, 88.34) --
	( 22.74, 83.76) --
	( 17.45, 83.76) --
	cycle;

\path[draw=drawColor,draw opacity=0.70,line width= 0.4pt,line join=round,line cap=round,fill=fillColor,fill opacity=0.70] ( 26.83, 80.36) --
	( 29.47, 75.78) --
	( 24.19, 75.78) --
	cycle;

\path[draw=drawColor,draw opacity=0.70,line width= 0.4pt,line join=round,line cap=round,fill=fillColor,fill opacity=0.70] ( 33.56, 75.52) --
	( 36.20, 70.94) --
	( 30.92, 70.94) --
	cycle;

\path[draw=drawColor,draw opacity=0.70,line width= 0.4pt,line join=round,line cap=round,fill=fillColor,fill opacity=0.70] ( 40.29, 64.15) --
	( 42.94, 59.58) --
	( 37.65, 59.58) --
	cycle;

\path[draw=drawColor,draw opacity=0.70,line width= 0.4pt,line join=round,line cap=round,fill=fillColor,fill opacity=0.70] ( 47.03, 53.18) --
	( 49.67, 48.60) --
	( 44.38, 48.60) --
	cycle;

\path[draw=drawColor,draw opacity=0.70,line width= 0.4pt,line join=round,line cap=round,fill=fillColor,fill opacity=0.70] ( 53.76, 44.33) --
	( 56.40, 39.75) --
	( 51.12, 39.75) --
	cycle;

\path[draw=drawColor,draw opacity=0.70,line width= 0.4pt,line join=round,line cap=round,fill=fillColor,fill opacity=0.70] ( 60.49, 40.10) --
	( 63.13, 35.52) --
	( 57.85, 35.52) --
	cycle;

\path[draw=drawColor,draw opacity=0.70,line width= 0.4pt,line join=round,line cap=round,fill=fillColor,fill opacity=0.70] ( 67.22, 37.54) --
	( 69.87, 32.96) --
	( 64.58, 32.96) --
	cycle;

\path[draw=drawColor,draw opacity=0.70,line width= 0.4pt,line join=round,line cap=round,fill=fillColor,fill opacity=0.70] ( 73.96, 35.02) --
	( 76.60, 30.44) --
	( 71.31, 30.44) --
	cycle;

\path[draw=drawColor,draw opacity=0.70,line width= 0.4pt,line join=round,line cap=round,fill=fillColor,fill opacity=0.70] ( 80.69, 33.33) --
	( 83.33, 28.75) --
	( 78.05, 28.75) --
	cycle;

\path[draw=drawColor,draw opacity=0.70,line width= 0.4pt,line join=round,line cap=round,fill=fillColor,fill opacity=0.70] ( 87.42, 32.52) --
	( 90.06, 27.94) --
	( 84.78, 27.94) --
	cycle;

\path[draw=drawColor,draw opacity=0.70,line width= 0.4pt,line join=round,line cap=round,fill=fillColor,fill opacity=0.70] ( 94.15, 31.45) --
	( 96.80, 26.87) --
	( 91.51, 26.87) --
	cycle;

\path[draw=drawColor,draw opacity=0.70,line width= 0.4pt,line join=round,line cap=round,fill=fillColor,fill opacity=0.70] (100.89, 30.08) --
	(103.53, 25.51) --
	( 98.24, 25.51) --
	cycle;
\definecolor{fillColor}{RGB}{173,216,230}

\path[draw=drawColor,draw opacity=0.70,line width= 0.4pt,line join=round,line cap=round,fill=fillColor,fill opacity=0.70] ( 18.36, 35.44) rectangle ( 21.84, 38.92);

\path[draw=drawColor,draw opacity=0.70,line width= 0.4pt,line join=round,line cap=round,fill=fillColor,fill opacity=0.70] ( 25.09, 32.01) rectangle ( 28.57, 35.49);

\path[draw=drawColor,draw opacity=0.70,line width= 0.4pt,line join=round,line cap=round,fill=fillColor,fill opacity=0.70] ( 31.82, 28.34) rectangle ( 35.30, 31.82);

\path[draw=drawColor,draw opacity=0.70,line width= 0.4pt,line join=round,line cap=round,fill=fillColor,fill opacity=0.70] ( 38.56, 24.18) rectangle ( 42.03, 27.66);

\path[draw=drawColor,draw opacity=0.70,line width= 0.4pt,line join=round,line cap=round,fill=fillColor,fill opacity=0.70] ( 45.29, 22.11) rectangle ( 48.77, 25.59);

\path[draw=drawColor,draw opacity=0.70,line width= 0.4pt,line join=round,line cap=round,fill=fillColor,fill opacity=0.70] ( 52.02, 21.02) rectangle ( 55.50, 24.49);

\path[draw=drawColor,draw opacity=0.70,line width= 0.4pt,line join=round,line cap=round,fill=fillColor,fill opacity=0.70] ( 58.75, 20.76) rectangle ( 62.23, 24.23);

\path[draw=drawColor,draw opacity=0.70,line width= 0.4pt,line join=round,line cap=round,fill=fillColor,fill opacity=0.70] ( 65.48, 20.49) rectangle ( 68.96, 23.97);

\path[draw=drawColor,draw opacity=0.70,line width= 0.4pt,line join=round,line cap=round,fill=fillColor,fill opacity=0.70] ( 72.22, 20.38) rectangle ( 75.69, 23.86);

\path[draw=drawColor,draw opacity=0.70,line width= 0.4pt,line join=round,line cap=round,fill=fillColor,fill opacity=0.70] ( 78.95, 20.31) rectangle ( 82.43, 23.79);

\path[draw=drawColor,draw opacity=0.70,line width= 0.4pt,line join=round,line cap=round,fill=fillColor,fill opacity=0.70] ( 85.68, 20.42) rectangle ( 89.16, 23.89);

\path[draw=drawColor,draw opacity=0.70,line width= 0.4pt,line join=round,line cap=round,fill=fillColor,fill opacity=0.70] ( 92.41, 20.48) rectangle ( 95.89, 23.96);

\path[draw=drawColor,draw opacity=0.70,line width= 0.4pt,line join=round,line cap=round,fill=fillColor,fill opacity=0.70] ( 99.15, 20.73) rectangle (102.62, 24.21);

\path[draw=drawColor,draw opacity=0.70,line width= 0.4pt,line join=round,line cap=round] ( 18.13, 36.76) rectangle ( 22.06, 40.68);

\path[draw=drawColor,draw opacity=0.70,line width= 0.4pt,line join=round,line cap=round] ( 24.87, 32.80) rectangle ( 28.79, 36.73);

\path[draw=drawColor,draw opacity=0.70,line width= 0.4pt,line join=round,line cap=round] ( 31.60, 28.59) rectangle ( 35.52, 32.51);

\path[draw=drawColor,draw opacity=0.70,line width= 0.4pt,line join=round,line cap=round] ( 38.33, 24.23) rectangle ( 42.26, 28.15);

\path[draw=drawColor,draw opacity=0.70,line width= 0.4pt,line join=round,line cap=round] ( 45.06, 22.84) rectangle ( 48.99, 26.77);

\path[draw=drawColor,draw opacity=0.70,line width= 0.4pt,line join=round,line cap=round] ( 51.80, 22.35) rectangle ( 55.72, 26.28);

\path[draw=drawColor,draw opacity=0.70,line width= 0.4pt,line join=round,line cap=round] ( 58.53, 22.22) rectangle ( 62.45, 26.14);

\path[draw=drawColor,draw opacity=0.70,line width= 0.4pt,line join=round,line cap=round] ( 65.26, 22.14) rectangle ( 69.19, 26.06);

\path[draw=drawColor,draw opacity=0.70,line width= 0.4pt,line join=round,line cap=round] ( 71.99, 22.15) rectangle ( 75.92, 26.07);

\path[draw=drawColor,draw opacity=0.70,line width= 0.4pt,line join=round,line cap=round] ( 78.73, 21.99) rectangle ( 82.65, 25.92);

\path[draw=drawColor,draw opacity=0.70,line width= 0.4pt,line join=round,line cap=round] ( 85.46, 22.02) rectangle ( 89.38, 25.94);

\path[draw=drawColor,draw opacity=0.70,line width= 0.4pt,line join=round,line cap=round] ( 92.19, 22.12) rectangle ( 96.12, 26.04);

\path[draw=drawColor,draw opacity=0.70,line width= 0.4pt,line join=round,line cap=round] ( 98.92, 22.18) rectangle (102.85, 26.10);
\definecolor{fillColor}{RGB}{255,0,0}

\path[draw=drawColor,draw opacity=0.70,line width= 0.4pt,line join=round,line cap=round,fill=fillColor,fill opacity=0.70] ( 20.10, 48.05) circle (  1.96);

\path[draw=drawColor,draw opacity=0.70,line width= 0.4pt,line join=round,line cap=round,fill=fillColor,fill opacity=0.70] ( 26.83, 34.62) circle (  1.96);

\path[draw=drawColor,draw opacity=0.70,line width= 0.4pt,line join=round,line cap=round,fill=fillColor,fill opacity=0.70] ( 33.56, 32.66) circle (  1.96);

\path[draw=drawColor,draw opacity=0.70,line width= 0.4pt,line join=round,line cap=round,fill=fillColor,fill opacity=0.70] ( 40.29, 29.54) circle (  1.96);

\path[draw=drawColor,draw opacity=0.70,line width= 0.4pt,line join=round,line cap=round,fill=fillColor,fill opacity=0.70] ( 47.03, 28.69) circle (  1.96);

\path[draw=drawColor,draw opacity=0.70,line width= 0.4pt,line join=round,line cap=round,fill=fillColor,fill opacity=0.70] ( 53.76, 27.52) circle (  1.96);

\path[draw=drawColor,draw opacity=0.70,line width= 0.4pt,line join=round,line cap=round,fill=fillColor,fill opacity=0.70] ( 60.49, 26.84) circle (  1.96);

\path[draw=drawColor,draw opacity=0.70,line width= 0.4pt,line join=round,line cap=round,fill=fillColor,fill opacity=0.70] ( 67.22, 26.05) circle (  1.96);

\path[draw=drawColor,draw opacity=0.70,line width= 0.4pt,line join=round,line cap=round,fill=fillColor,fill opacity=0.70] ( 73.96, 25.61) circle (  1.96);

\path[draw=drawColor,draw opacity=0.70,line width= 0.4pt,line join=round,line cap=round,fill=fillColor,fill opacity=0.70] ( 80.69, 25.41) circle (  1.96);

\path[draw=drawColor,draw opacity=0.70,line width= 0.4pt,line join=round,line cap=round,fill=fillColor,fill opacity=0.70] ( 87.42, 25.03) circle (  1.96);

\path[draw=drawColor,draw opacity=0.70,line width= 0.4pt,line join=round,line cap=round,fill=fillColor,fill opacity=0.70] ( 94.15, 25.03) circle (  1.96);

\path[draw=drawColor,draw opacity=0.70,line width= 0.4pt,line join=round,line cap=round,fill=fillColor,fill opacity=0.70] (100.89, 25.01) circle (  1.96);
\end{scope}
\begin{scope}
\path[clip] (  0.00,  0.00) rectangle (108.41, 93.95);
\definecolor{drawColor}{gray}{0.10}

\node[text=drawColor,anchor=base east,inner sep=0pt, outer sep=0pt, scale=  0.55] at ( 17.33, 33.68) {5};

\node[text=drawColor,anchor=base east,inner sep=0pt, outer sep=0pt, scale=  0.55] at ( 17.33, 48.31) {10};

\node[text=drawColor,anchor=base east,inner sep=0pt, outer sep=0pt, scale=  0.55] at ( 17.33, 62.94) {15};

\node[text=drawColor,anchor=base east,inner sep=0pt, outer sep=0pt, scale=  0.55] at ( 17.33, 77.57) {20};
\end{scope}
\begin{scope}
\path[clip] (  0.00,  0.00) rectangle (108.41, 93.95);
\definecolor{drawColor}{gray}{0.10}

\node[text=drawColor,anchor=base,inner sep=0pt, outer sep=0pt, scale=  0.55] at ( 26.83, 15.15) {-2.7};

\node[text=drawColor,anchor=base,inner sep=0pt, outer sep=0pt, scale=  0.55] at ( 40.29, 15.15) {-2.5};

\node[text=drawColor,anchor=base,inner sep=0pt, outer sep=0pt, scale=  0.55] at ( 53.76, 15.15) {-2.3};

\node[text=drawColor,anchor=base,inner sep=0pt, outer sep=0pt, scale=  0.55] at ( 67.22, 15.15) {-2.1};

\node[text=drawColor,anchor=base,inner sep=0pt, outer sep=0pt, scale=  0.55] at ( 80.69, 15.15) {-1.9};

\node[text=drawColor,anchor=base,inner sep=0pt, outer sep=0pt, scale=  0.55] at ( 94.15, 15.15) {-1.7};
\end{scope}
\begin{scope}
\path[clip] (  0.00,  0.00) rectangle (108.41, 93.95);
\definecolor{drawColor}{gray}{0.10}

\node[text=drawColor,anchor=base,inner sep=0pt, outer sep=0pt, scale=  0.66] at ( 60.49,  6.78) {$log_{10}(\sigma)$};
\end{scope}
\begin{scope}
\path[clip] (  0.00,  0.00) rectangle (108.41, 93.95);
\definecolor{drawColor}{gray}{0.10}

\node[text=drawColor,rotate= 90.00,anchor=base,inner sep=0pt, outer sep=0pt, scale=  0.66] at ( 10.05, 53.67) {running time (ms)};
\end{scope}
\end{tikzpicture}

%% file: Rfigures/HPatchesTime_mean.tex
\begin{tikzpicture}[x=1pt,y=1pt]
\clip (5,3) rectangle (104, 93.95);
\definecolor{fillColor}{RGB}{255,255,255}
\path[use as bounding box,fill=fillColor,fill opacity=0.00] (0,0) rectangle (108.41, 93.95);
\begin{scope}
\path[clip] (  0.00,  0.00) rectangle (108.41, 93.95);
\definecolor{drawColor}{RGB}{255,255,255}
\definecolor{fillColor}{RGB}{255,255,255}

\path[draw=drawColor,line width= 0.6pt,line join=round,line cap=round,fill=fillColor] (  0.00,  0.00) rectangle (108.41, 93.95);
\end{scope}
\begin{scope}
\path[clip] ( 18.08, 18.89) rectangle (102.91, 88.45);
\definecolor{fillColor}{gray}{0.92}

\path[fill=fillColor] ( 18.08, 18.89) rectangle (102.91, 88.45);
\definecolor{drawColor}{RGB}{0,0,0}

\path[draw=drawColor,line width= 0.6pt,line join=round] ( 19.45, 51.05) --
	( 24.01, 46.43) --
	( 28.57, 44.78) --
	( 33.13, 43.79) --
	( 37.69, 43.55) --
	( 42.25, 43.50) --
	( 46.81, 43.66) --
	( 51.37, 43.72) --
	( 55.93, 43.48) --
	( 60.49, 43.31) --
	( 65.05, 43.13) --
	( 69.61, 42.88) --
	( 74.17, 42.73) --
	( 78.73, 42.49) --
	( 83.29, 42.11) --
	( 87.85, 41.82) --
	( 92.42, 41.50) --
	( 96.98, 41.20) --
	(101.54, 41.24);

\path[draw=drawColor,line width= 0.6pt,line join=round] ( 19.45, 31.39) --
	( 24.01, 34.27) --
	( 28.57, 34.69) --
	( 33.13, 34.40) --
	( 37.69, 31.21) --
	( 42.25, 30.52) --
	( 46.81, 28.60) --
	( 51.37, 30.11) --
	( 55.93, 28.79) --
	( 60.49, 28.59) --
	( 65.05, 27.05) --
	( 69.61, 26.57) --
	( 74.17, 27.62) --
	( 78.73, 30.38) --
	( 83.29, 31.93) --
	( 87.85, 30.34) --
	( 92.42, 31.18) --
	( 96.98, 31.86) --
	(101.54, 31.68);

\path[draw=drawColor,line width= 0.6pt,line join=round] ( 19.45, 42.01) --
	( 24.01, 37.96) --
	( 28.57, 39.28) --
	( 33.13, 41.81) --
	( 37.69, 39.82) --
	( 42.25, 37.64) --
	( 46.81, 38.79) --
	( 51.37, 49.37) --
	( 55.93, 46.80) --
	( 60.49, 49.17) --
	( 65.05, 36.86) --
	( 69.61, 25.24) --
	( 74.17, 25.49) --
	( 78.73, 25.19) --
	( 83.29, 25.11) --
	( 87.85, 25.54) --
	( 92.42, 24.88) --
	( 96.98, 22.21) --
	(101.54, 22.05);

\path[draw=drawColor,line width= 0.6pt,line join=round] ( 19.45, 85.29) --
	( 24.01, 55.64) --
	( 28.57, 48.05) --
	( 33.13, 48.10) --
	( 37.69, 40.55) --
	( 42.25, 33.41) --
	( 46.81, 31.84) --
	( 51.37, 30.53) --
	( 55.93, 29.23) --
	( 60.49, 28.52) --
	( 65.05, 25.98) --
	( 69.61, 25.55) --
	( 74.17, 25.17) --
	( 78.73, 25.09) --
	( 83.29, 24.98) --
	( 87.85, 24.71) --
	( 92.42, 24.74) --
	( 96.98, 24.93) --
	(101.54, 25.15);
\definecolor{drawColor}{RGB}{0,0,0}
\definecolor{fillColor}{RGB}{190,190,190}

\path[draw=drawColor,draw opacity=0.70,line width= 0.4pt,line join=round,line cap=round,fill=fillColor,fill opacity=0.70] ( 19.45, 54.11) --
	( 22.09, 49.53) --
	( 16.80, 49.53) --
	cycle;

\path[draw=drawColor,draw opacity=0.70,line width= 0.4pt,line join=round,line cap=round,fill=fillColor,fill opacity=0.70] ( 24.01, 49.48) --
	( 26.65, 44.91) --
	( 21.36, 44.91) --
	cycle;

\path[draw=drawColor,draw opacity=0.70,line width= 0.4pt,line join=round,line cap=round,fill=fillColor,fill opacity=0.70] ( 28.57, 47.83) --
	( 31.21, 43.25) --
	( 25.92, 43.25) --
	cycle;

\path[draw=drawColor,draw opacity=0.70,line width= 0.4pt,line join=round,line cap=round,fill=fillColor,fill opacity=0.70] ( 33.13, 46.84) --
	( 35.77, 42.26) --
	( 30.48, 42.26) --
	cycle;

\path[draw=drawColor,draw opacity=0.70,line width= 0.4pt,line join=round,line cap=round,fill=fillColor,fill opacity=0.70] ( 37.69, 46.60) --
	( 40.33, 42.02) --
	( 35.05, 42.02) --
	cycle;

\path[draw=drawColor,draw opacity=0.70,line width= 0.4pt,line join=round,line cap=round,fill=fillColor,fill opacity=0.70] ( 42.25, 46.55) --
	( 44.89, 41.97) --
	( 39.61, 41.97) --
	cycle;

\path[draw=drawColor,draw opacity=0.70,line width= 0.4pt,line join=round,line cap=round,fill=fillColor,fill opacity=0.70] ( 46.81, 46.72) --
	( 49.45, 42.14) --
	( 44.17, 42.14) --
	cycle;

\path[draw=drawColor,draw opacity=0.70,line width= 0.4pt,line join=round,line cap=round,fill=fillColor,fill opacity=0.70] ( 51.37, 46.77) --
	( 54.01, 42.20) --
	( 48.73, 42.20) --
	cycle;

\path[draw=drawColor,draw opacity=0.70,line width= 0.4pt,line join=round,line cap=round,fill=fillColor,fill opacity=0.70] ( 55.93, 46.53) --
	( 58.57, 41.95) --
	( 53.29, 41.95) --
	cycle;

\path[draw=drawColor,draw opacity=0.70,line width= 0.4pt,line join=round,line cap=round,fill=fillColor,fill opacity=0.70] ( 60.49, 46.36) --
	( 63.13, 41.78) --
	( 57.85, 41.78) --
	cycle;

\path[draw=drawColor,draw opacity=0.70,line width= 0.4pt,line join=round,line cap=round,fill=fillColor,fill opacity=0.70] ( 65.05, 46.19) --
	( 67.69, 41.61) --
	( 62.41, 41.61) --
	cycle;

\path[draw=drawColor,draw opacity=0.70,line width= 0.4pt,line join=round,line cap=round,fill=fillColor,fill opacity=0.70] ( 69.61, 45.94) --
	( 72.25, 41.36) --
	( 66.97, 41.36) --
	cycle;

\path[draw=drawColor,draw opacity=0.70,line width= 0.4pt,line join=round,line cap=round,fill=fillColor,fill opacity=0.70] ( 74.17, 45.78) --
	( 76.82, 41.20) --
	( 71.53, 41.20) --
	cycle;

\path[draw=drawColor,draw opacity=0.70,line width= 0.4pt,line join=round,line cap=round,fill=fillColor,fill opacity=0.70] ( 78.73, 45.54) --
	( 81.38, 40.96) --
	( 76.09, 40.96) --
	cycle;

\path[draw=drawColor,draw opacity=0.70,line width= 0.4pt,line join=round,line cap=round,fill=fillColor,fill opacity=0.70] ( 83.29, 45.16) --
	( 85.94, 40.59) --
	( 80.65, 40.59) --
	cycle;

\path[draw=drawColor,draw opacity=0.70,line width= 0.4pt,line join=round,line cap=round,fill=fillColor,fill opacity=0.70] ( 87.85, 44.87) --
	( 90.50, 40.29) --
	( 85.21, 40.29) --
	cycle;

\path[draw=drawColor,draw opacity=0.70,line width= 0.4pt,line join=round,line cap=round,fill=fillColor,fill opacity=0.70] ( 92.42, 44.55) --
	( 95.06, 39.97) --
	( 89.77, 39.97) --
	cycle;

\path[draw=drawColor,draw opacity=0.70,line width= 0.4pt,line join=round,line cap=round,fill=fillColor,fill opacity=0.70] ( 96.98, 44.25) --
	( 99.62, 39.67) --
	( 94.33, 39.67) --
	cycle;

\path[draw=drawColor,draw opacity=0.70,line width= 0.4pt,line join=round,line cap=round,fill=fillColor,fill opacity=0.70] (101.54, 44.29) --
	(104.18, 39.72) --
	( 98.89, 39.72) --
	cycle;
\definecolor{fillColor}{RGB}{173,216,230}

\path[draw=drawColor,draw opacity=0.70,line width= 0.4pt,line join=round,line cap=round,fill=fillColor,fill opacity=0.70] ( 17.71, 29.65) rectangle ( 21.18, 33.13);

\path[draw=drawColor,draw opacity=0.70,line width= 0.4pt,line join=round,line cap=round,fill=fillColor,fill opacity=0.70] ( 22.27, 32.53) rectangle ( 25.74, 36.01);

\path[draw=drawColor,draw opacity=0.70,line width= 0.4pt,line join=round,line cap=round,fill=fillColor,fill opacity=0.70] ( 26.83, 32.95) rectangle ( 30.31, 36.43);

\path[draw=drawColor,draw opacity=0.70,line width= 0.4pt,line join=round,line cap=round,fill=fillColor,fill opacity=0.70] ( 31.39, 32.66) rectangle ( 34.87, 36.13);

\path[draw=drawColor,draw opacity=0.70,line width= 0.4pt,line join=round,line cap=round,fill=fillColor,fill opacity=0.70] ( 35.95, 29.48) rectangle ( 39.43, 32.95);

\path[draw=drawColor,draw opacity=0.70,line width= 0.4pt,line join=round,line cap=round,fill=fillColor,fill opacity=0.70] ( 40.51, 28.78) rectangle ( 43.99, 32.26);

\path[draw=drawColor,draw opacity=0.70,line width= 0.4pt,line join=round,line cap=round,fill=fillColor,fill opacity=0.70] ( 45.07, 26.86) rectangle ( 48.55, 30.34);

\path[draw=drawColor,draw opacity=0.70,line width= 0.4pt,line join=round,line cap=round,fill=fillColor,fill opacity=0.70] ( 49.63, 28.37) rectangle ( 53.11, 31.85);

\path[draw=drawColor,draw opacity=0.70,line width= 0.4pt,line join=round,line cap=round,fill=fillColor,fill opacity=0.70] ( 54.19, 27.05) rectangle ( 57.67, 30.53);

\path[draw=drawColor,draw opacity=0.70,line width= 0.4pt,line join=round,line cap=round,fill=fillColor,fill opacity=0.70] ( 58.75, 26.85) rectangle ( 62.23, 30.33);

\path[draw=drawColor,draw opacity=0.70,line width= 0.4pt,line join=round,line cap=round,fill=fillColor,fill opacity=0.70] ( 63.31, 25.31) rectangle ( 66.79, 28.79);

\path[draw=drawColor,draw opacity=0.70,line width= 0.4pt,line join=round,line cap=round,fill=fillColor,fill opacity=0.70] ( 67.87, 24.83) rectangle ( 71.35, 28.31);

\path[draw=drawColor,draw opacity=0.70,line width= 0.4pt,line join=round,line cap=round,fill=fillColor,fill opacity=0.70] ( 72.43, 25.88) rectangle ( 75.91, 29.36);

\path[draw=drawColor,draw opacity=0.70,line width= 0.4pt,line join=round,line cap=round,fill=fillColor,fill opacity=0.70] ( 76.99, 28.64) rectangle ( 80.47, 32.12);

\path[draw=drawColor,draw opacity=0.70,line width= 0.4pt,line join=round,line cap=round,fill=fillColor,fill opacity=0.70] ( 81.56, 30.19) rectangle ( 85.03, 33.67);

\path[draw=drawColor,draw opacity=0.70,line width= 0.4pt,line join=round,line cap=round,fill=fillColor,fill opacity=0.70] ( 86.12, 28.61) rectangle ( 89.59, 32.08);

\path[draw=drawColor,draw opacity=0.70,line width= 0.4pt,line join=round,line cap=round,fill=fillColor,fill opacity=0.70] ( 90.68, 29.44) rectangle ( 94.15, 32.92);

\path[draw=drawColor,draw opacity=0.70,line width= 0.4pt,line join=round,line cap=round,fill=fillColor,fill opacity=0.70] ( 95.24, 30.12) rectangle ( 98.72, 33.60);

\path[draw=drawColor,draw opacity=0.70,line width= 0.4pt,line join=round,line cap=round,fill=fillColor,fill opacity=0.70] ( 99.80, 29.94) rectangle (103.28, 33.42);

\path[draw=drawColor,draw opacity=0.70,line width= 0.4pt,line join=round,line cap=round] ( 17.48, 40.05) rectangle ( 21.41, 43.97);

\path[draw=drawColor,draw opacity=0.70,line width= 0.4pt,line join=round,line cap=round] ( 22.04, 36.00) rectangle ( 25.97, 39.92);

\path[draw=drawColor,draw opacity=0.70,line width= 0.4pt,line join=round,line cap=round] ( 26.60, 37.31) rectangle ( 30.53, 41.24);

\path[draw=drawColor,draw opacity=0.70,line width= 0.4pt,line join=round,line cap=round] ( 31.17, 39.85) rectangle ( 35.09, 43.77);

\path[draw=drawColor,draw opacity=0.70,line width= 0.4pt,line join=round,line cap=round] ( 35.73, 37.85) rectangle ( 39.65, 41.78);

\path[draw=drawColor,draw opacity=0.70,line width= 0.4pt,line join=round,line cap=round] ( 40.29, 35.68) rectangle ( 44.21, 39.61);

\path[draw=drawColor,draw opacity=0.70,line width= 0.4pt,line join=round,line cap=round] ( 44.85, 36.83) rectangle ( 48.77, 40.75);

\path[draw=drawColor,draw opacity=0.70,line width= 0.4pt,line join=round,line cap=round] ( 49.41, 47.41) rectangle ( 53.33, 51.34);

\path[draw=drawColor,draw opacity=0.70,line width= 0.4pt,line join=round,line cap=round] ( 53.97, 44.84) rectangle ( 57.89, 48.77);

\path[draw=drawColor,draw opacity=0.70,line width= 0.4pt,line join=round,line cap=round] ( 58.53, 47.21) rectangle ( 62.45, 51.13);

\path[draw=drawColor,draw opacity=0.70,line width= 0.4pt,line join=round,line cap=round] ( 63.09, 34.90) rectangle ( 67.01, 38.82);

\path[draw=drawColor,draw opacity=0.70,line width= 0.4pt,line join=round,line cap=round] ( 67.65, 23.27) rectangle ( 71.57, 27.20);

\path[draw=drawColor,draw opacity=0.70,line width= 0.4pt,line join=round,line cap=round] ( 72.21, 23.53) rectangle ( 76.14, 27.45);

\path[draw=drawColor,draw opacity=0.70,line width= 0.4pt,line join=round,line cap=round] ( 76.77, 23.23) rectangle ( 80.70, 27.15);

\path[draw=drawColor,draw opacity=0.70,line width= 0.4pt,line join=round,line cap=round] ( 81.33, 23.15) rectangle ( 85.26, 27.08);

\path[draw=drawColor,draw opacity=0.70,line width= 0.4pt,line join=round,line cap=round] ( 85.89, 23.57) rectangle ( 89.82, 27.50);

\path[draw=drawColor,draw opacity=0.70,line width= 0.4pt,line join=round,line cap=round] ( 90.45, 22.91) rectangle ( 94.38, 26.84);

\path[draw=drawColor,draw opacity=0.70,line width= 0.4pt,line join=round,line cap=round] ( 95.01, 20.25) rectangle ( 98.94, 24.17);

\path[draw=drawColor,draw opacity=0.70,line width= 0.4pt,line join=round,line cap=round] ( 99.57, 20.09) rectangle (103.50, 24.01);
\definecolor{fillColor}{RGB}{255,0,0}

\path[draw=drawColor,draw opacity=0.70,line width= 0.4pt,line join=round,line cap=round,fill=fillColor,fill opacity=0.70] ( 19.45, 85.29) circle (  1.96);

\path[draw=drawColor,draw opacity=0.70,line width= 0.4pt,line join=round,line cap=round,fill=fillColor,fill opacity=0.70] ( 24.01, 55.64) circle (  1.96);

\path[draw=drawColor,draw opacity=0.70,line width= 0.4pt,line join=round,line cap=round,fill=fillColor,fill opacity=0.70] ( 28.57, 48.05) circle (  1.96);

\path[draw=drawColor,draw opacity=0.70,line width= 0.4pt,line join=round,line cap=round,fill=fillColor,fill opacity=0.70] ( 33.13, 48.10) circle (  1.96);

\path[draw=drawColor,draw opacity=0.70,line width= 0.4pt,line join=round,line cap=round,fill=fillColor,fill opacity=0.70] ( 37.69, 40.55) circle (  1.96);

\path[draw=drawColor,draw opacity=0.70,line width= 0.4pt,line join=round,line cap=round,fill=fillColor,fill opacity=0.70] ( 42.25, 33.41) circle (  1.96);

\path[draw=drawColor,draw opacity=0.70,line width= 0.4pt,line join=round,line cap=round,fill=fillColor,fill opacity=0.70] ( 46.81, 31.84) circle (  1.96);

\path[draw=drawColor,draw opacity=0.70,line width= 0.4pt,line join=round,line cap=round,fill=fillColor,fill opacity=0.70] ( 51.37, 30.53) circle (  1.96);

\path[draw=drawColor,draw opacity=0.70,line width= 0.4pt,line join=round,line cap=round,fill=fillColor,fill opacity=0.70] ( 55.93, 29.23) circle (  1.96);

\path[draw=drawColor,draw opacity=0.70,line width= 0.4pt,line join=round,line cap=round,fill=fillColor,fill opacity=0.70] ( 60.49, 28.52) circle (  1.96);

\path[draw=drawColor,draw opacity=0.70,line width= 0.4pt,line join=round,line cap=round,fill=fillColor,fill opacity=0.70] ( 65.05, 25.98) circle (  1.96);

\path[draw=drawColor,draw opacity=0.70,line width= 0.4pt,line join=round,line cap=round,fill=fillColor,fill opacity=0.70] ( 69.61, 25.55) circle (  1.96);

\path[draw=drawColor,draw opacity=0.70,line width= 0.4pt,line join=round,line cap=round,fill=fillColor,fill opacity=0.70] ( 74.17, 25.17) circle (  1.96);

\path[draw=drawColor,draw opacity=0.70,line width= 0.4pt,line join=round,line cap=round,fill=fillColor,fill opacity=0.70] ( 78.73, 25.09) circle (  1.96);

\path[draw=drawColor,draw opacity=0.70,line width= 0.4pt,line join=round,line cap=round,fill=fillColor,fill opacity=0.70] ( 83.29, 24.98) circle (  1.96);

\path[draw=drawColor,draw opacity=0.70,line width= 0.4pt,line join=round,line cap=round,fill=fillColor,fill opacity=0.70] ( 87.85, 24.71) circle (  1.96);

\path[draw=drawColor,draw opacity=0.70,line width= 0.4pt,line join=round,line cap=round,fill=fillColor,fill opacity=0.70] ( 92.42, 24.74) circle (  1.96);

\path[draw=drawColor,draw opacity=0.70,line width= 0.4pt,line join=round,line cap=round,fill=fillColor,fill opacity=0.70] ( 96.98, 24.93) circle (  1.96);

\path[draw=drawColor,draw opacity=0.70,line width= 0.4pt,line join=round,line cap=round,fill=fillColor,fill opacity=0.70] (101.54, 25.15) circle (  1.96);
\end{scope}
\begin{scope}
\path[clip] (  0.00,  0.00) rectangle (108.41, 93.95);
\definecolor{drawColor}{gray}{0.10}

\node[text=drawColor,anchor=base east,inner sep=0pt, outer sep=0pt, scale=  0.55] at ( 17.33, 35.65) {5};

\node[text=drawColor,anchor=base east,inner sep=0pt, outer sep=0pt, scale=  0.55] at ( 17.33, 58.20) {10};

\node[text=drawColor,anchor=base east,inner sep=0pt, outer sep=0pt, scale=  0.55] at ( 17.33, 80.76) {15};
\end{scope}
\begin{scope}
\path[clip] (  0.00,  0.00) rectangle (108.41, 93.95);
\definecolor{drawColor}{gray}{0.10}

\node[text=drawColor,anchor=base,inner sep=0pt, outer sep=0pt, scale=  0.55] at ( 28.57, 15.15) {-3.2};

\node[text=drawColor,anchor=base,inner sep=0pt, outer sep=0pt, scale=  0.55] at ( 46.81, 15.15) {-2.8};

\node[text=drawColor,anchor=base,inner sep=0pt, outer sep=0pt, scale=  0.55] at ( 65.05, 15.15) {-2.4};

\node[text=drawColor,anchor=base,inner sep=0pt, outer sep=0pt, scale=  0.55] at ( 83.29, 15.15) {-2.0};

\node[text=drawColor,anchor=base,inner sep=0pt, outer sep=0pt, scale=  0.55] at ( 92.42, 15.15) {-1.8};
\end{scope}
\begin{scope}
\path[clip] (  0.00,  0.00) rectangle (108.41, 93.95);
\definecolor{drawColor}{gray}{0.10}

\node[text=drawColor,anchor=base,inner sep=0pt, outer sep=0pt, scale=  0.66] at ( 60.49,  6.78) {$log_{10}(\sigma)$};
\end{scope}
\begin{scope}
\path[clip] (  0.00,  0.00) rectangle (108.41, 93.95);
\definecolor{drawColor}{gray}{0.10}

\node[text=drawColor,rotate= 90.00,anchor=base,inner sep=0pt, outer sep=0pt, scale=  0.66] at ( 10.05, 53.67) {running time (ms)};
\end{scope}
\end{tikzpicture}

%% file: Rfigures/NearRot_mean.tex
\begin{tikzpicture}[x=1pt,y=1pt]
\clip (5,3) rectangle (104, 93.95);
\definecolor{fillColor}{RGB}{255,255,255}
\path[use as bounding box,fill=fillColor,fill opacity=0.00] (0,0) rectangle (108.41, 93.95);
\begin{scope}
\path[clip] (  0.00,  0.00) rectangle (108.41, 93.95);
\definecolor{drawColor}{RGB}{255,255,255}
\definecolor{fillColor}{RGB}{255,255,255}

\path[draw=drawColor,line width= 0.6pt,line join=round,line cap=round,fill=fillColor] ( -0.00,  0.00) rectangle (108.41, 93.95);
\end{scope}
\begin{scope}
\path[clip] ( 21.85, 18.89) rectangle (102.91, 88.45);
\definecolor{fillColor}{gray}{0.92}

\path[fill=fillColor] ( 21.85, 18.89) rectangle (102.91, 88.45);
\definecolor{drawColor}{RGB}{0,0,0}

\path[draw=drawColor,line width= 0.6pt,line join=round] ( 23.52, 89.25) --
	( 29.07, 64.42) --
	( 34.62, 60.29) --
	( 40.17, 66.40) --
	( 45.73, 70.30) --
	( 51.28, 64.80) --
	( 56.83, 60.19) --
	( 62.38, 58.70) --
	( 67.93, 58.14) --
	( 73.48, 58.62) --
	( 79.03, 57.97) --
	( 84.59, 60.25) --
	( 90.14, 60.70) --
	( 95.69, 64.59) --
	(101.24, 70.68);

\path[draw=drawColor,line width= 0.6pt,line join=round] ( 23.52, 83.30) --
	( 29.07, 60.14) --
	( 34.62, 56.54) --
	( 40.17, 59.49) --
	( 45.73, 62.35) --
	( 51.28, 57.36) --
	( 56.83, 52.70) --
	( 62.38, 51.66) --
	( 67.93, 51.10) --
	( 73.48, 51.36) --
	( 79.03, 50.95) --
	( 84.59, 52.94) --
	( 90.14, 53.93) --
	( 95.69, 57.01) --
	(101.24, 60.90);

\path[draw=drawColor,line width= 0.6pt,line join=round] ( 23.52, 57.78) --
	( 29.07, 40.87) --
	( 34.62, 34.46) --
	( 40.17, 33.67) --
	( 45.73, 32.94) --
	( 51.28, 28.47) --
	( 56.83, 26.77) --
	( 62.38, 25.58) --
	( 67.93, 28.62) --
	( 73.48, 25.71) --
	( 79.03, 26.16) --
	( 84.59, 27.55) --
	( 90.14, 29.31) --
	( 95.69, 31.57) --
	(101.24, 34.46);
\definecolor{drawColor}{RGB}{0,0,0}
\definecolor{fillColor}{RGB}{173,216,230}

\path[draw=drawColor,draw opacity=0.70,line width= 0.4pt,line join=round,line cap=round,fill=fillColor,fill opacity=0.70] ( 21.78, 87.51) rectangle ( 25.26, 90.98);

\path[draw=drawColor,draw opacity=0.70,line width= 0.4pt,line join=round,line cap=round,fill=fillColor,fill opacity=0.70] ( 27.33, 62.68) rectangle ( 30.81, 66.16);

\path[draw=drawColor,draw opacity=0.70,line width= 0.4pt,line join=round,line cap=round,fill=fillColor,fill opacity=0.70] ( 32.88, 58.55) rectangle ( 36.36, 62.03);

\path[draw=drawColor,draw opacity=0.70,line width= 0.4pt,line join=round,line cap=round,fill=fillColor,fill opacity=0.70] ( 38.43, 64.67) rectangle ( 41.91, 68.14);

\path[draw=drawColor,draw opacity=0.70,line width= 0.4pt,line join=round,line cap=round,fill=fillColor,fill opacity=0.70] ( 43.99, 68.56) rectangle ( 47.46, 72.04);

\path[draw=drawColor,draw opacity=0.70,line width= 0.4pt,line join=round,line cap=round,fill=fillColor,fill opacity=0.70] ( 49.54, 63.06) rectangle ( 53.02, 66.54);

\path[draw=drawColor,draw opacity=0.70,line width= 0.4pt,line join=round,line cap=round,fill=fillColor,fill opacity=0.70] ( 55.09, 58.45) rectangle ( 58.57, 61.92);

\path[draw=drawColor,draw opacity=0.70,line width= 0.4pt,line join=round,line cap=round,fill=fillColor,fill opacity=0.70] ( 60.64, 56.96) rectangle ( 64.12, 60.44);

\path[draw=drawColor,draw opacity=0.70,line width= 0.4pt,line join=round,line cap=round,fill=fillColor,fill opacity=0.70] ( 66.19, 56.40) rectangle ( 69.67, 59.88);

\path[draw=drawColor,draw opacity=0.70,line width= 0.4pt,line join=round,line cap=round,fill=fillColor,fill opacity=0.70] ( 71.74, 56.88) rectangle ( 75.22, 60.36);

\path[draw=drawColor,draw opacity=0.70,line width= 0.4pt,line join=round,line cap=round,fill=fillColor,fill opacity=0.70] ( 77.29, 56.23) rectangle ( 80.77, 59.71);

\path[draw=drawColor,draw opacity=0.70,line width= 0.4pt,line join=round,line cap=round,fill=fillColor,fill opacity=0.70] ( 82.85, 58.51) rectangle ( 86.32, 61.99);

\path[draw=drawColor,draw opacity=0.70,line width= 0.4pt,line join=round,line cap=round,fill=fillColor,fill opacity=0.70] ( 88.40, 58.96) rectangle ( 91.88, 62.44);

\path[draw=drawColor,draw opacity=0.70,line width= 0.4pt,line join=round,line cap=round,fill=fillColor,fill opacity=0.70] ( 93.95, 62.85) rectangle ( 97.43, 66.32);

\path[draw=drawColor,draw opacity=0.70,line width= 0.4pt,line join=round,line cap=round,fill=fillColor,fill opacity=0.70] ( 99.50, 68.95) rectangle (102.98, 72.42);

\path[draw=drawColor,draw opacity=0.70,line width= 0.4pt,line join=round,line cap=round] ( 21.56, 81.34) rectangle ( 25.48, 85.26);

\path[draw=drawColor,draw opacity=0.70,line width= 0.4pt,line join=round,line cap=round] ( 27.11, 58.18) rectangle ( 31.03, 62.10);

\path[draw=drawColor,draw opacity=0.70,line width= 0.4pt,line join=round,line cap=round] ( 32.66, 54.58) rectangle ( 36.58, 58.50);

\path[draw=drawColor,draw opacity=0.70,line width= 0.4pt,line join=round,line cap=round] ( 38.21, 57.53) rectangle ( 42.14, 61.45);

\path[draw=drawColor,draw opacity=0.70,line width= 0.4pt,line join=round,line cap=round] ( 43.76, 60.39) rectangle ( 47.69, 64.31);

\path[draw=drawColor,draw opacity=0.70,line width= 0.4pt,line join=round,line cap=round] ( 49.31, 55.40) rectangle ( 53.24, 59.32);

\path[draw=drawColor,draw opacity=0.70,line width= 0.4pt,line join=round,line cap=round] ( 54.87, 50.73) rectangle ( 58.79, 54.66);

\path[draw=drawColor,draw opacity=0.70,line width= 0.4pt,line join=round,line cap=round] ( 60.42, 49.70) rectangle ( 64.34, 53.63);

\path[draw=drawColor,draw opacity=0.70,line width= 0.4pt,line join=round,line cap=round] ( 65.97, 49.14) rectangle ( 69.89, 53.06);

\path[draw=drawColor,draw opacity=0.70,line width= 0.4pt,line join=round,line cap=round] ( 71.52, 49.40) rectangle ( 75.44, 53.33);

\path[draw=drawColor,draw opacity=0.70,line width= 0.4pt,line join=round,line cap=round] ( 77.07, 48.99) rectangle ( 81.00, 52.91);

\path[draw=drawColor,draw opacity=0.70,line width= 0.4pt,line join=round,line cap=round] ( 82.62, 50.98) rectangle ( 86.55, 54.90);

\path[draw=drawColor,draw opacity=0.70,line width= 0.4pt,line join=round,line cap=round] ( 88.17, 51.97) rectangle ( 92.10, 55.90);

\path[draw=drawColor,draw opacity=0.70,line width= 0.4pt,line join=round,line cap=round] ( 93.73, 55.05) rectangle ( 97.65, 58.98);

\path[draw=drawColor,draw opacity=0.70,line width= 0.4pt,line join=round,line cap=round] ( 99.28, 58.94) rectangle (103.20, 62.87);
\definecolor{fillColor}{RGB}{255,0,0}

\path[draw=drawColor,draw opacity=0.70,line width= 0.4pt,line join=round,line cap=round,fill=fillColor,fill opacity=0.70] ( 23.52, 57.78) circle (  1.96);

\path[draw=drawColor,draw opacity=0.70,line width= 0.4pt,line join=round,line cap=round,fill=fillColor,fill opacity=0.70] ( 29.07, 40.87) circle (  1.96);

\path[draw=drawColor,draw opacity=0.70,line width= 0.4pt,line join=round,line cap=round,fill=fillColor,fill opacity=0.70] ( 34.62, 34.46) circle (  1.96);

\path[draw=drawColor,draw opacity=0.70,line width= 0.4pt,line join=round,line cap=round,fill=fillColor,fill opacity=0.70] ( 40.17, 33.67) circle (  1.96);

\path[draw=drawColor,draw opacity=0.70,line width= 0.4pt,line join=round,line cap=round,fill=fillColor,fill opacity=0.70] ( 45.73, 32.94) circle (  1.96);

\path[draw=drawColor,draw opacity=0.70,line width= 0.4pt,line join=round,line cap=round,fill=fillColor,fill opacity=0.70] ( 51.28, 28.47) circle (  1.96);

\path[draw=drawColor,draw opacity=0.70,line width= 0.4pt,line join=round,line cap=round,fill=fillColor,fill opacity=0.70] ( 56.83, 26.77) circle (  1.96);

\path[draw=drawColor,draw opacity=0.70,line width= 0.4pt,line join=round,line cap=round,fill=fillColor,fill opacity=0.70] ( 62.38, 25.58) circle (  1.96);

\path[draw=drawColor,draw opacity=0.70,line width= 0.4pt,line join=round,line cap=round,fill=fillColor,fill opacity=0.70] ( 67.93, 28.62) circle (  1.96);

\path[draw=drawColor,draw opacity=0.70,line width= 0.4pt,line join=round,line cap=round,fill=fillColor,fill opacity=0.70] ( 73.48, 25.71) circle (  1.96);

\path[draw=drawColor,draw opacity=0.70,line width= 0.4pt,line join=round,line cap=round,fill=fillColor,fill opacity=0.70] ( 79.03, 26.16) circle (  1.96);

\path[draw=drawColor,draw opacity=0.70,line width= 0.4pt,line join=round,line cap=round,fill=fillColor,fill opacity=0.70] ( 84.59, 27.55) circle (  1.96);

\path[draw=drawColor,draw opacity=0.70,line width= 0.4pt,line join=round,line cap=round,fill=fillColor,fill opacity=0.70] ( 90.14, 29.31) circle (  1.96);

\path[draw=drawColor,draw opacity=0.70,line width= 0.4pt,line join=round,line cap=round,fill=fillColor,fill opacity=0.70] ( 95.69, 31.57) circle (  1.96);

\path[draw=drawColor,draw opacity=0.70,line width= 0.4pt,line join=round,line cap=round,fill=fillColor,fill opacity=0.70] (101.24, 34.46) circle (  1.96);
\end{scope}
\begin{scope}
\path[clip] (  0.00,  0.00) rectangle (108.41, 93.95);
\definecolor{drawColor}{gray}{0.10}

\node[text=drawColor,anchor=base east,inner sep=0pt, outer sep=0pt, scale=  0.55] at ( 21.10, 20.15) {0.48};

\node[text=drawColor,anchor=base east,inner sep=0pt, outer sep=0pt, scale=  0.55] at ( 21.10, 32.67) {0.50};

\node[text=drawColor,anchor=base east,inner sep=0pt, outer sep=0pt, scale=  0.55] at ( 21.10, 50.52) {0.53};

\node[text=drawColor,anchor=base east,inner sep=0pt, outer sep=0pt, scale=  0.55] at ( 21.10, 67.40) {0.56};

\node[text=drawColor,anchor=base east,inner sep=0pt, outer sep=0pt, scale=  0.55] at ( 21.10, 78.16) {0.58};
\end{scope}
\begin{scope}
\path[clip] (  0.00,  0.00) rectangle (108.41, 93.95);
\definecolor{drawColor}{gray}{0.10}

\node[text=drawColor,anchor=base,inner sep=0pt, outer sep=0pt, scale=  0.55] at ( 23.52, 15.15) {-2.5};

\node[text=drawColor,anchor=base,inner sep=0pt, outer sep=0pt, scale=  0.55] at ( 45.73, 15.15) {-2.1};

\node[text=drawColor,anchor=base,inner sep=0pt, outer sep=0pt, scale=  0.55] at ( 67.93, 15.15) {-1.7};

\node[text=drawColor,anchor=base,inner sep=0pt, outer sep=0pt, scale=  0.55] at ( 90.14, 15.15) {-1.3};
\end{scope}
\begin{scope}
\path[clip] (  0.00,  0.00) rectangle (108.41, 93.95);
\definecolor{drawColor}{gray}{0.10}

\node[text=drawColor,anchor=base,inner sep=0pt, outer sep=0pt, scale=  0.66] at ( 62.38,  6.78) {$log_{10}(\sigma)$};
\end{scope}
\begin{scope}
\path[clip] (  0.00,  0.00) rectangle (108.41, 93.95);
\definecolor{drawColor}{gray}{0.10}

\node[text=drawColor,rotate= 90.00,anchor=base,inner sep=0pt, outer sep=0pt, scale=  0.66] at ( 10.05, 53.67) {rotation error (degree)};
\end{scope}
\begin{scope}
\path[clip] (  0.00,  0.00) rectangle (108.41, 93.95);
\definecolor{drawColor}{RGB}{0,0,0}
\definecolor{fillColor}{RGB}{173,216,230}

\path[draw=drawColor,draw opacity=0.70,line width= 0.4pt,line join=round,line cap=round,fill=fillColor,fill opacity=0.70] ( 48.72, 81.67) rectangle ( 52.20, 85.15);
\end{scope}
\begin{scope}
\path[clip] (  0.00,  0.00) rectangle (108.41, 93.95);
\definecolor{drawColor}{RGB}{0,0,0}

\path[draw=drawColor,draw opacity=0.70,line width= 0.4pt,line join=round,line cap=round] ( 48.50, 76.78) rectangle ( 52.42, 80.71);
\end{scope}
\begin{scope}
\path[clip] (  0.00,  0.00) rectangle (108.41, 93.95);
\definecolor{drawColor}{RGB}{0,0,0}
\definecolor{fillColor}{RGB}{255,0,0}

\path[draw=drawColor,draw opacity=0.70,line width= 0.4pt,line join=round,line cap=round,fill=fillColor,fill opacity=0.70] ( 50.46, 74.08) circle (  1.96);
\end{scope}
\begin{scope}
\path[clip] (  0.00,  0.00) rectangle (108.41, 93.95);
\definecolor{drawColor}{RGB}{0,0,0}

\node[text=drawColor,anchor=base west,inner sep=0pt, outer sep=0pt, scale=  0.53] at ( 54.76, 81.59) {R-DLT};
\end{scope}
\begin{scope}
\path[clip] (  0.00,  0.00) rectangle (108.41, 93.95);
\definecolor{drawColor}{RGB}{0,0,0}

\node[text=drawColor,anchor=base west,inner sep=0pt, outer sep=0pt, scale=  0.53] at ( 54.76, 76.93) {R-Huber};
\end{scope}
\begin{scope}
\path[clip] (  0.00,  0.00) rectangle (108.41, 93.95);
\definecolor{drawColor}{RGB}{0,0,0}

\node[text=drawColor,anchor=base west,inner sep=0pt, outer sep=0pt, scale=  0.53] at ( 54.76, 72.26) {R-DPCP};
\end{scope}
\end{tikzpicture}

%% file: Rfigures/HemiRot_mean.tex
\begin{tikzpicture}[x=1pt,y=1pt]
\clip (5,3) rectangle (104, 93.95);
\definecolor{fillColor}{RGB}{255,255,255}
\path[use as bounding box,fill=fillColor,fill opacity=0.00] (0,0) rectangle (108.41, 93.95);
\begin{scope}
\path[clip] (  0.00,  0.00) rectangle (108.41, 93.95);
\definecolor{drawColor}{RGB}{255,255,255}
\definecolor{fillColor}{RGB}{255,255,255}

\path[draw=drawColor,line width= 0.6pt,line join=round,line cap=round,fill=fillColor] ( -0.00,  0.00) rectangle (108.41, 93.95);
\end{scope}
\begin{scope}
\path[clip] ( 21.85, 18.89) rectangle (102.91, 88.45);
\definecolor{fillColor}{gray}{0.92}

\path[fill=fillColor] ( 21.85, 18.89) rectangle (102.91, 88.45);
\definecolor{drawColor}{RGB}{0,0,0}

\path[draw=drawColor,line width= 0.6pt,line join=round] ( 23.52, 83.55) --
	( 29.07, 72.60) --
	( 34.62, 69.35) --
	( 40.17, 69.32) --
	( 45.73, 63.76) --
	( 51.28, 55.60) --
	( 56.83, 51.45) --
	( 62.38, 45.67) --
	( 67.93, 47.67) --
	( 73.48, 50.20) --
	( 79.03, 55.10) --
	( 84.59, 69.57) --
	( 90.14, 66.62) --
	( 95.69, 68.14) --
	(101.24, 66.34);

\path[draw=drawColor,line width= 0.6pt,line join=round] ( 23.52, 73.74) --
	( 29.07, 59.35) --
	( 34.62, 56.87) --
	( 40.17, 55.89) --
	( 45.73, 50.86) --
	( 51.28, 44.17) --
	( 56.83, 43.61) --
	( 62.38, 38.09) --
	( 67.93, 38.47) --
	( 73.48, 46.61) --
	( 79.03, 46.86) --
	( 84.59, 53.65) --
	( 90.14, 51.29) --
	( 95.69, 52.13) --
	(101.24, 51.19);

\path[draw=drawColor,line width= 0.6pt,line join=round] ( 23.52, 53.59) --
	( 29.07, 41.96) --
	( 34.62, 53.02) --
	( 40.17, 40.37) --
	( 45.73, 35.52) --
	( 51.28, 30.54) --
	( 56.83, 33.26) --
	( 62.38, 27.56) --
	( 67.93, 33.83) --
	( 73.48, 36.58) --
	( 79.03, 33.94) --
	( 84.59, 32.69) --
	( 90.14, 30.86) --
	( 95.69, 28.35) --
	(101.24, 27.71);
\definecolor{drawColor}{RGB}{0,0,0}
\definecolor{fillColor}{RGB}{173,216,230}

\path[draw=drawColor,draw opacity=0.70,line width= 0.4pt,line join=round,line cap=round,fill=fillColor,fill opacity=0.70] ( 21.78, 81.82) rectangle ( 25.26, 85.29);

\path[draw=drawColor,draw opacity=0.70,line width= 0.4pt,line join=round,line cap=round,fill=fillColor,fill opacity=0.70] ( 27.33, 70.86) rectangle ( 30.81, 74.34);

\path[draw=drawColor,draw opacity=0.70,line width= 0.4pt,line join=round,line cap=round,fill=fillColor,fill opacity=0.70] ( 32.88, 67.61) rectangle ( 36.36, 71.09);

\path[draw=drawColor,draw opacity=0.70,line width= 0.4pt,line join=round,line cap=round,fill=fillColor,fill opacity=0.70] ( 38.43, 67.58) rectangle ( 41.91, 71.06);

\path[draw=drawColor,draw opacity=0.70,line width= 0.4pt,line join=round,line cap=round,fill=fillColor,fill opacity=0.70] ( 43.99, 62.02) rectangle ( 47.46, 65.50);

\path[draw=drawColor,draw opacity=0.70,line width= 0.4pt,line join=round,line cap=round,fill=fillColor,fill opacity=0.70] ( 49.54, 53.86) rectangle ( 53.02, 57.34);

\path[draw=drawColor,draw opacity=0.70,line width= 0.4pt,line join=round,line cap=round,fill=fillColor,fill opacity=0.70] ( 55.09, 49.72) rectangle ( 58.57, 53.19);

\path[draw=drawColor,draw opacity=0.70,line width= 0.4pt,line join=round,line cap=round,fill=fillColor,fill opacity=0.70] ( 60.64, 43.93) rectangle ( 64.12, 47.41);

\path[draw=drawColor,draw opacity=0.70,line width= 0.4pt,line join=round,line cap=round,fill=fillColor,fill opacity=0.70] ( 66.19, 45.93) rectangle ( 69.67, 49.41);

\path[draw=drawColor,draw opacity=0.70,line width= 0.4pt,line join=round,line cap=round,fill=fillColor,fill opacity=0.70] ( 71.74, 48.46) rectangle ( 75.22, 51.93);

\path[draw=drawColor,draw opacity=0.70,line width= 0.4pt,line join=round,line cap=round,fill=fillColor,fill opacity=0.70] ( 77.29, 53.36) rectangle ( 80.77, 56.84);

\path[draw=drawColor,draw opacity=0.70,line width= 0.4pt,line join=round,line cap=round,fill=fillColor,fill opacity=0.70] ( 82.85, 67.84) rectangle ( 86.32, 71.31);

\path[draw=drawColor,draw opacity=0.70,line width= 0.4pt,line join=round,line cap=round,fill=fillColor,fill opacity=0.70] ( 88.40, 64.89) rectangle ( 91.88, 68.36);

\path[draw=drawColor,draw opacity=0.70,line width= 0.4pt,line join=round,line cap=round,fill=fillColor,fill opacity=0.70] ( 93.95, 66.40) rectangle ( 97.43, 69.88);

\path[draw=drawColor,draw opacity=0.70,line width= 0.4pt,line join=round,line cap=round,fill=fillColor,fill opacity=0.70] ( 99.50, 64.60) rectangle (102.98, 68.08);

\path[draw=drawColor,draw opacity=0.70,line width= 0.4pt,line join=round,line cap=round] ( 21.56, 71.78) rectangle ( 25.48, 75.71);

\path[draw=drawColor,draw opacity=0.70,line width= 0.4pt,line join=round,line cap=round] ( 27.11, 57.39) rectangle ( 31.03, 61.31);

\path[draw=drawColor,draw opacity=0.70,line width= 0.4pt,line join=round,line cap=round] ( 32.66, 54.90) rectangle ( 36.58, 58.83);

\path[draw=drawColor,draw opacity=0.70,line width= 0.4pt,line join=round,line cap=round] ( 38.21, 53.92) rectangle ( 42.14, 57.85);

\path[draw=drawColor,draw opacity=0.70,line width= 0.4pt,line join=round,line cap=round] ( 43.76, 48.90) rectangle ( 47.69, 52.82);

\path[draw=drawColor,draw opacity=0.70,line width= 0.4pt,line join=round,line cap=round] ( 49.31, 42.21) rectangle ( 53.24, 46.14);

\path[draw=drawColor,draw opacity=0.70,line width= 0.4pt,line join=round,line cap=round] ( 54.87, 41.65) rectangle ( 58.79, 45.57);

\path[draw=drawColor,draw opacity=0.70,line width= 0.4pt,line join=round,line cap=round] ( 60.42, 36.13) rectangle ( 64.34, 40.06);

\path[draw=drawColor,draw opacity=0.70,line width= 0.4pt,line join=round,line cap=round] ( 65.97, 36.51) rectangle ( 69.89, 40.43);

\path[draw=drawColor,draw opacity=0.70,line width= 0.4pt,line join=round,line cap=round] ( 71.52, 44.65) rectangle ( 75.44, 48.57);

\path[draw=drawColor,draw opacity=0.70,line width= 0.4pt,line join=round,line cap=round] ( 77.07, 44.90) rectangle ( 81.00, 48.82);

\path[draw=drawColor,draw opacity=0.70,line width= 0.4pt,line join=round,line cap=round] ( 82.62, 51.69) rectangle ( 86.55, 55.61);

\path[draw=drawColor,draw opacity=0.70,line width= 0.4pt,line join=round,line cap=round] ( 88.17, 49.32) rectangle ( 92.10, 53.25);

\path[draw=drawColor,draw opacity=0.70,line width= 0.4pt,line join=round,line cap=round] ( 93.73, 50.17) rectangle ( 97.65, 54.09);

\path[draw=drawColor,draw opacity=0.70,line width= 0.4pt,line join=round,line cap=round] ( 99.28, 49.23) rectangle (103.20, 53.15);
\definecolor{fillColor}{RGB}{255,0,0}

\path[draw=drawColor,draw opacity=0.70,line width= 0.4pt,line join=round,line cap=round,fill=fillColor,fill opacity=0.70] ( 23.52, 53.59) circle (  1.96);

\path[draw=drawColor,draw opacity=0.70,line width= 0.4pt,line join=round,line cap=round,fill=fillColor,fill opacity=0.70] ( 29.07, 41.96) circle (  1.96);

\path[draw=drawColor,draw opacity=0.70,line width= 0.4pt,line join=round,line cap=round,fill=fillColor,fill opacity=0.70] ( 34.62, 53.02) circle (  1.96);

\path[draw=drawColor,draw opacity=0.70,line width= 0.4pt,line join=round,line cap=round,fill=fillColor,fill opacity=0.70] ( 40.17, 40.37) circle (  1.96);

\path[draw=drawColor,draw opacity=0.70,line width= 0.4pt,line join=round,line cap=round,fill=fillColor,fill opacity=0.70] ( 45.73, 35.52) circle (  1.96);

\path[draw=drawColor,draw opacity=0.70,line width= 0.4pt,line join=round,line cap=round,fill=fillColor,fill opacity=0.70] ( 51.28, 30.54) circle (  1.96);

\path[draw=drawColor,draw opacity=0.70,line width= 0.4pt,line join=round,line cap=round,fill=fillColor,fill opacity=0.70] ( 56.83, 33.26) circle (  1.96);

\path[draw=drawColor,draw opacity=0.70,line width= 0.4pt,line join=round,line cap=round,fill=fillColor,fill opacity=0.70] ( 62.38, 27.56) circle (  1.96);

\path[draw=drawColor,draw opacity=0.70,line width= 0.4pt,line join=round,line cap=round,fill=fillColor,fill opacity=0.70] ( 67.93, 33.83) circle (  1.96);

\path[draw=drawColor,draw opacity=0.70,line width= 0.4pt,line join=round,line cap=round,fill=fillColor,fill opacity=0.70] ( 73.48, 36.58) circle (  1.96);

\path[draw=drawColor,draw opacity=0.70,line width= 0.4pt,line join=round,line cap=round,fill=fillColor,fill opacity=0.70] ( 79.03, 33.94) circle (  1.96);

\path[draw=drawColor,draw opacity=0.70,line width= 0.4pt,line join=round,line cap=round,fill=fillColor,fill opacity=0.70] ( 84.59, 32.69) circle (  1.96);

\path[draw=drawColor,draw opacity=0.70,line width= 0.4pt,line join=round,line cap=round,fill=fillColor,fill opacity=0.70] ( 90.14, 30.86) circle (  1.96);

\path[draw=drawColor,draw opacity=0.70,line width= 0.4pt,line join=round,line cap=round,fill=fillColor,fill opacity=0.70] ( 95.69, 28.35) circle (  1.96);

\path[draw=drawColor,draw opacity=0.70,line width= 0.4pt,line join=round,line cap=round,fill=fillColor,fill opacity=0.70] (101.24, 27.71) circle (  1.96);
\end{scope}
\begin{scope}
\path[clip] (  0.00,  0.00) rectangle (108.41, 93.95);
\definecolor{drawColor}{gray}{0.10}

\node[text=drawColor,anchor=base east,inner sep=0pt, outer sep=0pt, scale=  0.55] at ( 21.10, 20.15) {1.10};

\node[text=drawColor,anchor=base east,inner sep=0pt, outer sep=0pt, scale=  0.55] at ( 21.10, 31.81) {1.15};

\node[text=drawColor,anchor=base east,inner sep=0pt, outer sep=0pt, scale=  0.55] at ( 21.10, 42.97) {1.20};

\node[text=drawColor,anchor=base east,inner sep=0pt, outer sep=0pt, scale=  0.55] at ( 21.10, 53.68) {1.25};

\node[text=drawColor,anchor=base east,inner sep=0pt, outer sep=0pt, scale=  0.55] at ( 21.10, 63.96) {1.30};

\node[text=drawColor,anchor=base east,inner sep=0pt, outer sep=0pt, scale=  0.55] at ( 21.10, 73.86) {1.35};
\end{scope}
\begin{scope}
\path[clip] (  0.00,  0.00) rectangle (108.41, 93.95);
\definecolor{drawColor}{gray}{0.10}

\node[text=drawColor,anchor=base,inner sep=0pt, outer sep=0pt, scale=  0.55] at ( 23.52, 15.15) {-2.5};

\node[text=drawColor,anchor=base,inner sep=0pt, outer sep=0pt, scale=  0.55] at ( 45.73, 15.15) {-2.1};

\node[text=drawColor,anchor=base,inner sep=0pt, outer sep=0pt, scale=  0.55] at ( 67.93, 15.15) {-1.7};

\node[text=drawColor,anchor=base,inner sep=0pt, outer sep=0pt, scale=  0.55] at ( 90.14, 15.15) {-1.3};
\end{scope}
\begin{scope}
\path[clip] (  0.00,  0.00) rectangle (108.41, 93.95);
\definecolor{drawColor}{gray}{0.10}

\node[text=drawColor,anchor=base,inner sep=0pt, outer sep=0pt, scale=  0.66] at ( 62.38,  6.78) {$log_{10}(\sigma)$};
\end{scope}
\begin{scope}
\path[clip] (  0.00,  0.00) rectangle (108.41, 93.95);
\definecolor{drawColor}{gray}{0.10}

\node[text=drawColor,rotate= 90.00,anchor=base,inner sep=0pt, outer sep=0pt, scale=  0.66] at ( 10.05, 53.67) {rotation error (degree)};
\end{scope}
\end{tikzpicture}

%% file: Rfigures/NearTrans_mean.tex
\begin{tikzpicture}[x=1pt,y=1pt]
\clip (5,3) rectangle (104, 93.95);
\definecolor{fillColor}{RGB}{255,255,255}
\path[use as bounding box,fill=fillColor,fill opacity=0.00] (0,0) rectangle (108.41, 93.95);
\begin{scope}
\path[clip] (  0.00,  0.00) rectangle (108.41, 93.95);
\definecolor{drawColor}{RGB}{255,255,255}
\definecolor{fillColor}{RGB}{255,255,255}

\path[draw=drawColor,line width= 0.6pt,line join=round,line cap=round,fill=fillColor] (  0.00,  0.00) rectangle (108.41, 93.95);
\end{scope}
\begin{scope}
\path[clip] ( 21.10, 18.89) rectangle (102.90, 88.45);
\definecolor{fillColor}{gray}{0.92}

\path[fill=fillColor] ( 21.10, 18.89) rectangle (102.90, 88.45);
\definecolor{drawColor}{RGB}{0,0,0}

\path[draw=drawColor,line width= 0.6pt,line join=round] ( 22.79, 91.13) --
	( 28.39, 73.84) --
	( 33.99, 60.15) --
	( 39.59, 61.52) --
	( 45.20, 63.88) --
	( 50.80, 62.40) --
	( 56.40, 51.66) --
	( 62.00, 55.88) --
	( 67.61, 57.95) --
	( 73.21, 62.05) --
	( 78.81, 64.85) --
	( 84.42, 69.84) --
	( 90.02, 73.14) --
	( 95.62, 76.77) --
	(101.22, 83.09);

\path[draw=drawColor,line width= 0.6pt,line join=round] ( 22.79, 82.93) --
	( 28.39, 66.01) --
	( 33.99, 54.39) --
	( 39.59, 54.31) --
	( 45.20, 56.36) --
	( 50.80, 54.09) --
	( 56.40, 44.65) --
	( 62.00, 47.71) --
	( 67.61, 49.40) --
	( 73.21, 52.64) --
	( 78.81, 54.33) --
	( 84.42, 57.62) --
	( 90.02, 60.32) --
	( 95.62, 63.73) --
	(101.22, 67.33);

\path[draw=drawColor,line width= 0.6pt,line join=round] ( 22.79, 65.17) --
	( 28.39, 50.84) --
	( 33.99, 42.41) --
	( 39.59, 37.66) --
	( 45.20, 34.46) --
	( 50.80, 31.47) --
	( 56.40, 26.82) --
	( 62.00, 25.53) --
	( 67.61, 27.24) --
	( 73.21, 29.29) --
	( 78.81, 30.01) --
	( 84.42, 30.92) --
	( 90.02, 33.13) --
	( 95.62, 34.73) --
	(101.22, 39.22);
\definecolor{drawColor}{RGB}{0,0,0}
\definecolor{fillColor}{RGB}{173,216,230}

\path[draw=drawColor,draw opacity=0.70,line width= 0.4pt,line join=round,line cap=round,fill=fillColor,fill opacity=0.70] ( 21.05, 89.40) rectangle ( 24.52, 92.87);

\path[draw=drawColor,draw opacity=0.70,line width= 0.4pt,line join=round,line cap=round,fill=fillColor,fill opacity=0.70] ( 26.65, 72.10) rectangle ( 30.13, 75.58);

\path[draw=drawColor,draw opacity=0.70,line width= 0.4pt,line join=round,line cap=round,fill=fillColor,fill opacity=0.70] ( 32.25, 58.42) rectangle ( 35.73, 61.89);

\path[draw=drawColor,draw opacity=0.70,line width= 0.4pt,line join=round,line cap=round,fill=fillColor,fill opacity=0.70] ( 37.85, 59.78) rectangle ( 41.33, 63.26);

\path[draw=drawColor,draw opacity=0.70,line width= 0.4pt,line join=round,line cap=round,fill=fillColor,fill opacity=0.70] ( 43.46, 62.14) rectangle ( 46.94, 65.62);

\path[draw=drawColor,draw opacity=0.70,line width= 0.4pt,line join=round,line cap=round,fill=fillColor,fill opacity=0.70] ( 49.06, 60.66) rectangle ( 52.54, 64.14);

\path[draw=drawColor,draw opacity=0.70,line width= 0.4pt,line join=round,line cap=round,fill=fillColor,fill opacity=0.70] ( 54.66, 49.92) rectangle ( 58.14, 53.40);

\path[draw=drawColor,draw opacity=0.70,line width= 0.4pt,line join=round,line cap=round,fill=fillColor,fill opacity=0.70] ( 60.27, 54.14) rectangle ( 63.74, 57.62);

\path[draw=drawColor,draw opacity=0.70,line width= 0.4pt,line join=round,line cap=round,fill=fillColor,fill opacity=0.70] ( 65.87, 56.21) rectangle ( 69.35, 59.69);

\path[draw=drawColor,draw opacity=0.70,line width= 0.4pt,line join=round,line cap=round,fill=fillColor,fill opacity=0.70] ( 71.47, 60.31) rectangle ( 74.95, 63.79);

\path[draw=drawColor,draw opacity=0.70,line width= 0.4pt,line join=round,line cap=round,fill=fillColor,fill opacity=0.70] ( 77.07, 63.11) rectangle ( 80.55, 66.59);

\path[draw=drawColor,draw opacity=0.70,line width= 0.4pt,line join=round,line cap=round,fill=fillColor,fill opacity=0.70] ( 82.68, 68.10) rectangle ( 86.15, 71.58);

\path[draw=drawColor,draw opacity=0.70,line width= 0.4pt,line join=round,line cap=round,fill=fillColor,fill opacity=0.70] ( 88.28, 71.40) rectangle ( 91.76, 74.88);

\path[draw=drawColor,draw opacity=0.70,line width= 0.4pt,line join=round,line cap=round,fill=fillColor,fill opacity=0.70] ( 93.88, 75.03) rectangle ( 97.36, 78.51);

\path[draw=drawColor,draw opacity=0.70,line width= 0.4pt,line join=round,line cap=round,fill=fillColor,fill opacity=0.70] ( 99.49, 81.35) rectangle (102.96, 84.82);

\path[draw=drawColor,draw opacity=0.70,line width= 0.4pt,line join=round,line cap=round] ( 20.82, 80.97) rectangle ( 24.75, 84.89);

\path[draw=drawColor,draw opacity=0.70,line width= 0.4pt,line join=round,line cap=round] ( 26.43, 64.05) rectangle ( 30.35, 67.97);

\path[draw=drawColor,draw opacity=0.70,line width= 0.4pt,line join=round,line cap=round] ( 32.03, 52.42) rectangle ( 35.95, 56.35);

\path[draw=drawColor,draw opacity=0.70,line width= 0.4pt,line join=round,line cap=round] ( 37.63, 52.35) rectangle ( 41.56, 56.28);

\path[draw=drawColor,draw opacity=0.70,line width= 0.4pt,line join=round,line cap=round] ( 43.23, 54.40) rectangle ( 47.16, 58.33);

\path[draw=drawColor,draw opacity=0.70,line width= 0.4pt,line join=round,line cap=round] ( 48.84, 52.12) rectangle ( 52.76, 56.05);

\path[draw=drawColor,draw opacity=0.70,line width= 0.4pt,line join=round,line cap=round] ( 54.44, 42.68) rectangle ( 58.36, 46.61);

\path[draw=drawColor,draw opacity=0.70,line width= 0.4pt,line join=round,line cap=round] ( 60.04, 45.75) rectangle ( 63.97, 49.67);

\path[draw=drawColor,draw opacity=0.70,line width= 0.4pt,line join=round,line cap=round] ( 65.65, 47.44) rectangle ( 69.57, 51.37);

\path[draw=drawColor,draw opacity=0.70,line width= 0.4pt,line join=round,line cap=round] ( 71.25, 50.68) rectangle ( 75.17, 54.60);

\path[draw=drawColor,draw opacity=0.70,line width= 0.4pt,line join=round,line cap=round] ( 76.85, 52.36) rectangle ( 80.78, 56.29);

\path[draw=drawColor,draw opacity=0.70,line width= 0.4pt,line join=round,line cap=round] ( 82.45, 55.66) rectangle ( 86.38, 59.58);

\path[draw=drawColor,draw opacity=0.70,line width= 0.4pt,line join=round,line cap=round] ( 88.06, 58.36) rectangle ( 91.98, 62.28);

\path[draw=drawColor,draw opacity=0.70,line width= 0.4pt,line join=round,line cap=round] ( 93.66, 61.76) rectangle ( 97.58, 65.69);

\path[draw=drawColor,draw opacity=0.70,line width= 0.4pt,line join=round,line cap=round] ( 99.26, 65.36) rectangle (103.19, 69.29);
\definecolor{fillColor}{RGB}{255,0,0}

\path[draw=drawColor,draw opacity=0.70,line width= 0.4pt,line join=round,line cap=round,fill=fillColor,fill opacity=0.70] ( 22.79, 65.17) circle (  1.96);

\path[draw=drawColor,draw opacity=0.70,line width= 0.4pt,line join=round,line cap=round,fill=fillColor,fill opacity=0.70] ( 28.39, 50.84) circle (  1.96);

\path[draw=drawColor,draw opacity=0.70,line width= 0.4pt,line join=round,line cap=round,fill=fillColor,fill opacity=0.70] ( 33.99, 42.41) circle (  1.96);

\path[draw=drawColor,draw opacity=0.70,line width= 0.4pt,line join=round,line cap=round,fill=fillColor,fill opacity=0.70] ( 39.59, 37.66) circle (  1.96);

\path[draw=drawColor,draw opacity=0.70,line width= 0.4pt,line join=round,line cap=round,fill=fillColor,fill opacity=0.70] ( 45.20, 34.46) circle (  1.96);

\path[draw=drawColor,draw opacity=0.70,line width= 0.4pt,line join=round,line cap=round,fill=fillColor,fill opacity=0.70] ( 50.80, 31.47) circle (  1.96);

\path[draw=drawColor,draw opacity=0.70,line width= 0.4pt,line join=round,line cap=round,fill=fillColor,fill opacity=0.70] ( 56.40, 26.82) circle (  1.96);

\path[draw=drawColor,draw opacity=0.70,line width= 0.4pt,line join=round,line cap=round,fill=fillColor,fill opacity=0.70] ( 62.00, 25.53) circle (  1.96);

\path[draw=drawColor,draw opacity=0.70,line width= 0.4pt,line join=round,line cap=round,fill=fillColor,fill opacity=0.70] ( 67.61, 27.24) circle (  1.96);

\path[draw=drawColor,draw opacity=0.70,line width= 0.4pt,line join=round,line cap=round,fill=fillColor,fill opacity=0.70] ( 73.21, 29.29) circle (  1.96);

\path[draw=drawColor,draw opacity=0.70,line width= 0.4pt,line join=round,line cap=round,fill=fillColor,fill opacity=0.70] ( 78.81, 30.01) circle (  1.96);

\path[draw=drawColor,draw opacity=0.70,line width= 0.4pt,line join=round,line cap=round,fill=fillColor,fill opacity=0.70] ( 84.42, 30.92) circle (  1.96);

\path[draw=drawColor,draw opacity=0.70,line width= 0.4pt,line join=round,line cap=round,fill=fillColor,fill opacity=0.70] ( 90.02, 33.13) circle (  1.96);

\path[draw=drawColor,draw opacity=0.70,line width= 0.4pt,line join=round,line cap=round,fill=fillColor,fill opacity=0.70] ( 95.62, 34.73) circle (  1.96);

\path[draw=drawColor,draw opacity=0.70,line width= 0.4pt,line join=round,line cap=round,fill=fillColor,fill opacity=0.70] (101.22, 39.22) circle (  1.96);
\end{scope}
\begin{scope}
\path[clip] (  0.00,  0.00) rectangle (108.41, 93.95);
\definecolor{drawColor}{gray}{0.10}

\node[text=drawColor,anchor=base east,inner sep=0pt, outer sep=0pt, scale=  0.55] at ( 20.35, 27.90) {3.8};

\node[text=drawColor,anchor=base east,inner sep=0pt, outer sep=0pt, scale=  0.55] at ( 20.35, 42.80) {4.0};

\node[text=drawColor,anchor=base east,inner sep=0pt, outer sep=0pt, scale=  0.55] at ( 20.35, 56.97) {4.2};

\node[text=drawColor,anchor=base east,inner sep=0pt, outer sep=0pt, scale=  0.55] at ( 20.35, 70.48) {4.4};
\end{scope}
\begin{scope}
\path[clip] (  0.00,  0.00) rectangle (108.41, 93.95);
\definecolor{drawColor}{gray}{0.10}

\node[text=drawColor,anchor=base,inner sep=0pt, outer sep=0pt, scale=  0.55] at ( 22.79, 15.15) {-2.5};

\node[text=drawColor,anchor=base,inner sep=0pt, outer sep=0pt, scale=  0.55] at ( 45.20, 15.15) {-2.1};

\node[text=drawColor,anchor=base,inner sep=0pt, outer sep=0pt, scale=  0.55] at ( 67.61, 15.15) {-1.7};

\node[text=drawColor,anchor=base,inner sep=0pt, outer sep=0pt, scale=  0.55] at ( 90.02, 15.15) {-1.3};
\end{scope}
\begin{scope}
\path[clip] (  0.00,  0.00) rectangle (108.41, 93.95);
\definecolor{drawColor}{gray}{0.10}

\node[text=drawColor,anchor=base,inner sep=0pt, outer sep=0pt, scale=  0.66] at ( 62.00,  6.78) {$log_{10}(\sigma)$};
\end{scope}
\begin{scope}
\path[clip] (  0.00,  0.00) rectangle (108.41, 93.95);
\definecolor{drawColor}{gray}{0.10}

\node[text=drawColor,rotate= 90.00,anchor=base,inner sep=0pt, outer sep=0pt, scale=  0.66] at ( 10.05, 53.67) {translation error (degree)};
\end{scope}
\end{tikzpicture}

%% file: Rfigures/HemiTrans_mean.tex
\begin{tikzpicture}[x=1pt,y=1pt]
\clip (5,3) rectangle (104, 93.95);
\definecolor{fillColor}{RGB}{255,255,255}
\path[use as bounding box,fill=fillColor,fill opacity=0.00] (0,0) rectangle (108.41, 93.95);
\begin{scope}
\path[clip] (  0.00,  0.00) rectangle (108.41, 93.95);
\definecolor{drawColor}{RGB}{255,255,255}
\definecolor{fillColor}{RGB}{255,255,255}

\path[draw=drawColor,line width= 0.6pt,line join=round,line cap=round,fill=fillColor] ( -0.00,  0.00) rectangle (108.41, 93.95);
\end{scope}
\begin{scope}
\path[clip] ( 21.85, 18.89) rectangle (102.91, 88.45);
\definecolor{fillColor}{gray}{0.92}

\path[fill=fillColor] ( 21.85, 18.89) rectangle (102.91, 88.45);
\definecolor{drawColor}{RGB}{0,0,0}

\path[draw=drawColor,line width= 0.6pt,line join=round] ( 23.52, 85.18) --
	( 29.07, 68.74) --
	( 34.62, 64.02) --
	( 40.17, 43.49) --
	( 45.73, 33.41) --
	( 51.28, 38.61) --
	( 56.83, 43.53) --
	( 62.38, 43.07) --
	( 67.93, 42.12) --
	( 73.48, 37.44) --
	( 79.03, 30.34) --
	( 84.59, 30.71) --
	( 90.14, 38.15) --
	( 95.69, 41.32) --
	(101.24, 44.92);

\path[draw=drawColor,line width= 0.6pt,line join=round] ( 23.52, 82.80) --
	( 29.07, 74.22) --
	( 34.62, 66.13) --
	( 40.17, 46.99) --
	( 45.73, 34.94) --
	( 51.28, 36.69) --
	( 56.83, 36.25) --
	( 62.38, 36.16) --
	( 67.93, 37.58) --
	( 73.48, 32.87) --
	( 79.03, 29.41) --
	( 84.59, 30.44) --
	( 90.14, 34.28) --
	( 95.69, 35.50) --
	(101.24, 37.21);

\path[draw=drawColor,line width= 0.6pt,line join=round] ( 23.52, 79.48) --
	( 29.07, 77.61) --
	( 34.62, 72.77) --
	( 40.17, 56.94) --
	( 45.73, 36.13) --
	( 51.28, 32.98) --
	( 56.83, 32.58) --
	( 62.38, 25.50) --
	( 67.93, 32.68) --
	( 73.48, 31.67) --
	( 79.03, 29.76) --
	( 84.59, 33.49) --
	( 90.14, 33.65) --
	( 95.69, 34.24) --
	(101.24, 34.10);
\definecolor{drawColor}{RGB}{0,0,0}
\definecolor{fillColor}{RGB}{173,216,230}

\path[draw=drawColor,draw opacity=0.70,line width= 0.4pt,line join=round,line cap=round,fill=fillColor,fill opacity=0.70] ( 21.78, 83.44) rectangle ( 25.26, 86.92);

\path[draw=drawColor,draw opacity=0.70,line width= 0.4pt,line join=round,line cap=round,fill=fillColor,fill opacity=0.70] ( 27.33, 67.00) rectangle ( 30.81, 70.48);

\path[draw=drawColor,draw opacity=0.70,line width= 0.4pt,line join=round,line cap=round,fill=fillColor,fill opacity=0.70] ( 32.88, 62.28) rectangle ( 36.36, 65.76);

\path[draw=drawColor,draw opacity=0.70,line width= 0.4pt,line join=round,line cap=round,fill=fillColor,fill opacity=0.70] ( 38.43, 41.75) rectangle ( 41.91, 45.23);

\path[draw=drawColor,draw opacity=0.70,line width= 0.4pt,line join=round,line cap=round,fill=fillColor,fill opacity=0.70] ( 43.99, 31.67) rectangle ( 47.46, 35.15);

\path[draw=drawColor,draw opacity=0.70,line width= 0.4pt,line join=round,line cap=round,fill=fillColor,fill opacity=0.70] ( 49.54, 36.87) rectangle ( 53.02, 40.35);

\path[draw=drawColor,draw opacity=0.70,line width= 0.4pt,line join=round,line cap=round,fill=fillColor,fill opacity=0.70] ( 55.09, 41.79) rectangle ( 58.57, 45.27);

\path[draw=drawColor,draw opacity=0.70,line width= 0.4pt,line join=round,line cap=round,fill=fillColor,fill opacity=0.70] ( 60.64, 41.33) rectangle ( 64.12, 44.81);

\path[draw=drawColor,draw opacity=0.70,line width= 0.4pt,line join=round,line cap=round,fill=fillColor,fill opacity=0.70] ( 66.19, 40.38) rectangle ( 69.67, 43.86);

\path[draw=drawColor,draw opacity=0.70,line width= 0.4pt,line join=round,line cap=round,fill=fillColor,fill opacity=0.70] ( 71.74, 35.70) rectangle ( 75.22, 39.18);

\path[draw=drawColor,draw opacity=0.70,line width= 0.4pt,line join=round,line cap=round,fill=fillColor,fill opacity=0.70] ( 77.29, 28.60) rectangle ( 80.77, 32.08);

\path[draw=drawColor,draw opacity=0.70,line width= 0.4pt,line join=round,line cap=round,fill=fillColor,fill opacity=0.70] ( 82.85, 28.97) rectangle ( 86.32, 32.45);

\path[draw=drawColor,draw opacity=0.70,line width= 0.4pt,line join=round,line cap=round,fill=fillColor,fill opacity=0.70] ( 88.40, 36.41) rectangle ( 91.88, 39.89);

\path[draw=drawColor,draw opacity=0.70,line width= 0.4pt,line join=round,line cap=round,fill=fillColor,fill opacity=0.70] ( 93.95, 39.58) rectangle ( 97.43, 43.06);

\path[draw=drawColor,draw opacity=0.70,line width= 0.4pt,line join=round,line cap=round,fill=fillColor,fill opacity=0.70] ( 99.50, 43.18) rectangle (102.98, 46.66);

\path[draw=drawColor,draw opacity=0.70,line width= 0.4pt,line join=round,line cap=round] ( 21.56, 80.83) rectangle ( 25.48, 84.76);

\path[draw=drawColor,draw opacity=0.70,line width= 0.4pt,line join=round,line cap=round] ( 27.11, 72.26) rectangle ( 31.03, 76.18);

\path[draw=drawColor,draw opacity=0.70,line width= 0.4pt,line join=round,line cap=round] ( 32.66, 64.17) rectangle ( 36.58, 68.09);

\path[draw=drawColor,draw opacity=0.70,line width= 0.4pt,line join=round,line cap=round] ( 38.21, 45.03) rectangle ( 42.14, 48.95);

\path[draw=drawColor,draw opacity=0.70,line width= 0.4pt,line join=round,line cap=round] ( 43.76, 32.98) rectangle ( 47.69, 36.90);

\path[draw=drawColor,draw opacity=0.70,line width= 0.4pt,line join=round,line cap=round] ( 49.31, 34.73) rectangle ( 53.24, 38.65);

\path[draw=drawColor,draw opacity=0.70,line width= 0.4pt,line join=round,line cap=round] ( 54.87, 34.29) rectangle ( 58.79, 38.22);

\path[draw=drawColor,draw opacity=0.70,line width= 0.4pt,line join=round,line cap=round] ( 60.42, 34.20) rectangle ( 64.34, 38.12);

\path[draw=drawColor,draw opacity=0.70,line width= 0.4pt,line join=round,line cap=round] ( 65.97, 35.62) rectangle ( 69.89, 39.54);

\path[draw=drawColor,draw opacity=0.70,line width= 0.4pt,line join=round,line cap=round] ( 71.52, 30.91) rectangle ( 75.44, 34.83);

\path[draw=drawColor,draw opacity=0.70,line width= 0.4pt,line join=round,line cap=round] ( 77.07, 27.45) rectangle ( 81.00, 31.37);

\path[draw=drawColor,draw opacity=0.70,line width= 0.4pt,line join=round,line cap=round] ( 82.62, 28.48) rectangle ( 86.55, 32.40);

\path[draw=drawColor,draw opacity=0.70,line width= 0.4pt,line join=round,line cap=round] ( 88.17, 32.32) rectangle ( 92.10, 36.24);

\path[draw=drawColor,draw opacity=0.70,line width= 0.4pt,line join=round,line cap=round] ( 93.73, 33.53) rectangle ( 97.65, 37.46);

\path[draw=drawColor,draw opacity=0.70,line width= 0.4pt,line join=round,line cap=round] ( 99.28, 35.24) rectangle (103.20, 39.17);
\definecolor{fillColor}{RGB}{255,0,0}

\path[draw=drawColor,draw opacity=0.70,line width= 0.4pt,line join=round,line cap=round,fill=fillColor,fill opacity=0.70] ( 23.52, 79.48) circle (  1.96);

\path[draw=drawColor,draw opacity=0.70,line width= 0.4pt,line join=round,line cap=round,fill=fillColor,fill opacity=0.70] ( 29.07, 77.61) circle (  1.96);

\path[draw=drawColor,draw opacity=0.70,line width= 0.4pt,line join=round,line cap=round,fill=fillColor,fill opacity=0.70] ( 34.62, 72.77) circle (  1.96);

\path[draw=drawColor,draw opacity=0.70,line width= 0.4pt,line join=round,line cap=round,fill=fillColor,fill opacity=0.70] ( 40.17, 56.94) circle (  1.96);

\path[draw=drawColor,draw opacity=0.70,line width= 0.4pt,line join=round,line cap=round,fill=fillColor,fill opacity=0.70] ( 45.73, 36.13) circle (  1.96);

\path[draw=drawColor,draw opacity=0.70,line width= 0.4pt,line join=round,line cap=round,fill=fillColor,fill opacity=0.70] ( 51.28, 32.98) circle (  1.96);

\path[draw=drawColor,draw opacity=0.70,line width= 0.4pt,line join=round,line cap=round,fill=fillColor,fill opacity=0.70] ( 56.83, 32.58) circle (  1.96);

\path[draw=drawColor,draw opacity=0.70,line width= 0.4pt,line join=round,line cap=round,fill=fillColor,fill opacity=0.70] ( 62.38, 25.50) circle (  1.96);

\path[draw=drawColor,draw opacity=0.70,line width= 0.4pt,line join=round,line cap=round,fill=fillColor,fill opacity=0.70] ( 67.93, 32.68) circle (  1.96);

\path[draw=drawColor,draw opacity=0.70,line width= 0.4pt,line join=round,line cap=round,fill=fillColor,fill opacity=0.70] ( 73.48, 31.67) circle (  1.96);

\path[draw=drawColor,draw opacity=0.70,line width= 0.4pt,line join=round,line cap=round,fill=fillColor,fill opacity=0.70] ( 79.03, 29.76) circle (  1.96);

\path[draw=drawColor,draw opacity=0.70,line width= 0.4pt,line join=round,line cap=round,fill=fillColor,fill opacity=0.70] ( 84.59, 33.49) circle (  1.96);

\path[draw=drawColor,draw opacity=0.70,line width= 0.4pt,line join=round,line cap=round,fill=fillColor,fill opacity=0.70] ( 90.14, 33.65) circle (  1.96);

\path[draw=drawColor,draw opacity=0.70,line width= 0.4pt,line join=round,line cap=round,fill=fillColor,fill opacity=0.70] ( 95.69, 34.24) circle (  1.96);

\path[draw=drawColor,draw opacity=0.70,line width= 0.4pt,line join=round,line cap=round,fill=fillColor,fill opacity=0.70] (101.24, 34.10) circle (  1.96);
\end{scope}
\begin{scope}
\path[clip] (  0.00,  0.00) rectangle (108.41, 93.95);
\definecolor{drawColor}{gray}{0.10}

\node[text=drawColor,anchor=base east,inner sep=0pt, outer sep=0pt, scale=  0.55] at ( 21.10, 34.86) {36.0};

\node[text=drawColor,anchor=base east,inner sep=0pt, outer sep=0pt, scale=  0.55] at ( 21.10, 49.37) {36.5};

\node[text=drawColor,anchor=base east,inner sep=0pt, outer sep=0pt, scale=  0.55] at ( 21.10, 63.68) {37.0};

\node[text=drawColor,anchor=base east,inner sep=0pt, outer sep=0pt, scale=  0.55] at ( 21.10, 77.80) {37.5};
\end{scope}
\begin{scope}
\path[clip] (  0.00,  0.00) rectangle (108.41, 93.95);
\definecolor{drawColor}{gray}{0.10}

\node[text=drawColor,anchor=base,inner sep=0pt, outer sep=0pt, scale=  0.55] at ( 23.52, 15.15) {-2.5};

\node[text=drawColor,anchor=base,inner sep=0pt, outer sep=0pt, scale=  0.55] at ( 45.73, 15.15) {-2.1};

\node[text=drawColor,anchor=base,inner sep=0pt, outer sep=0pt, scale=  0.55] at ( 67.93, 15.15) {-1.7};

\node[text=drawColor,anchor=base,inner sep=0pt, outer sep=0pt, scale=  0.55] at ( 90.14, 15.15) {-1.3};
\end{scope}
\begin{scope}
\path[clip] (  0.00,  0.00) rectangle (108.41, 93.95);
\definecolor{drawColor}{gray}{0.10}

\node[text=drawColor,anchor=base,inner sep=0pt, outer sep=0pt, scale=  0.66] at ( 62.38,  6.78) {$log_{10}(\sigma)$};
\end{scope}
\begin{scope}
\path[clip] (  0.00,  0.00) rectangle (108.41, 93.95);
\definecolor{drawColor}{gray}{0.10}

\node[text=drawColor,rotate= 90.00,anchor=base,inner sep=0pt, outer sep=0pt, scale=  0.66] at ( 10.05, 53.67) {translation error (degree)};
\end{scope}
\end{tikzpicture}